\documentclass[twoside]{IEEEtran}
\usepackage{cite}
\usepackage{amsmath}
\usepackage{amssymb}
\usepackage{algorithm}
\usepackage{algpseudocode}
\usepackage{graphicx} 
\usepackage{multirow}
\usepackage{multicol}
\usepackage[caption=false]{subfig}

\DeclareMathOperator*{\argmin}{arg\,min}

\makeatletter
\def\munderbar#1{\underline{\sbox\tw@{$#1$}\dp\tw@\z@\box\tw@}}
\makeatother
\graphicspath{{./figs/}}

\begin{document}

\title{Identity-Preserving Pose-Robust Face Hallucination Through Face Subspace Prior}%

\author{Ali~Abbasi
        and~Mohammad~Rahmati,~\IEEEmembership{Member,~IEEE}%
\thanks{Ali Abbasi is with the Pattern Recognition and Image Processing Lab, Department of Computer Engineering, Amirkabir University of Technology (Tehran Polytechnic), 424 Hafez Ave., Tehran, Iran. e-mail: ali.abbasi@aut.ac.ir}}%

\markboth{IEEE Transactions on Image Processing,~Vol.~XX, No.~XX, November~20XX}%
{Abbasi and Rahmati: Face Hallucination Through Face Subspace Prior}

\maketitle

\begin{abstract}
	Over the past few decades, numerous attempts have been made to address the problem of recovering a high-resolution (HR) facial image from its corresponding low-resolution (LR) counterpart, a task commonly referred to as face hallucination. Despite the impressive performance achieved by position-patch and deep learning-based methods, most of these techniques are still unable to recover identity-specific features of faces. The former group of algorithms often produces blurry and oversmoothed outputs particularly in the presence of higher levels of degradation, whereas the latter generates faces which sometimes by no means resemble the individuals in the input images. In this paper, a novel face super-resolution approach will be introduced, in which the hallucinated face is forced to lie in a subspace spanned by the available training faces. Therefore, in contrast to the majority of existing face hallucination techniques and thanks to this \textit{face subspace prior}, the reconstruction is performed in favor of recovering person-specific facial features, rather than merely increasing image quantitative scores. Furthermore, inspired by recent advances in the area of 3D face reconstruction, an efficient 3D dictionary alignment scheme is also presented, through which the algorithm becomes capable of dealing with low-resolution faces taken in uncontrolled conditions. In extensive experiments carried out on several well-known face datasets, the proposed algorithm shows remarkable performance by generating detailed and close to ground truth results which outperform the state-of-the-art face hallucination algorithms by significant margins both in quantitative and qualitative evaluations.
\end{abstract}

\IEEEpeerreviewmaketitle

\section{Introduction}
\label{sec:intro}
\IEEEPARstart{O}{ur} desire to enhance the resolution of an already-recorded image is arguably as old as the time when the early photographs were taken. With the emergence of digital images, the idea also started to attract the attention of many researchers, leading to the introduction of a popular field in the area of image processing, known as image super-resolution \cite{refs:Nasrollahi2014}. Even today, despite cameras having ever-increasing resolution, there is still a huge demand in increasing the resolution of the existing images, particularly in specific applications such as law enforcement, surveillance, and monitoring, where images are taken under uncontrolled conditions and are required to be further processed before being used for a particular purpose. More importantly, most computer vision algorithms are designed to work with high quality images, which means their performance would severely affected when given a low-resolution input \cite{refs:Zou2012}.

Among different variations of image super-resolution applications, those which deal with super-resolving face images have always been of special interest to researchers, and are often categorized under the name face hallucination. The term was first coined by Baker and Kanade \cite{refs:Baker2000} in 2000, and since then has gained huge popularity due to its wide range of applications, with dozens of algorithms proposed so far.

One can hardly offer an explicit classification of the algorithms presented in the literature, as in many cases, the distinction among different categories of methods is not clear. Consequently, different studies have suggested different criteria to classify face hallucination algorithms; including operating domain (spatial vs. frequency), number of input images (single vs. multiple), and reconstruction method (reconstruction-based vs. learning-based). Early researchers relied more on statistical approaches to predict the HR face image given the LR observation. In their pioneering effort \cite{refs:Baker2000}, Baker and Kanade used a Bayesian framework with gradient priors to estimate high-frequency components of a face image. Inspired by their work, Su \textit{et al.} \cite{refs:Su2005a} proposed a similar formulation in which the prior was estimated by matching local low-level facial features from the input LR and the training HR face images. Meanwhile, Markov random fields (MRF) also started to draw the attention of researchers, after a two-step method was suggested by Liu \textit{et al.} \cite{refs:Liu2001}.

Another major group of researchers focused on making use of training samples to learn a projection matrix which could be later used to project the LR input into high dimensional space and obtain the reconstructed HR output. They based their idea on the fact that face images share structural similarities, and therefore can be synthesized from a linear combination of other samples. Wang and Tang \cite{refs:Wang2005} addressed the problem by applying PCA to fit the input face image as a linear combination of the training low-resolution face images, and then reconstructing the HR output by using the combination weights for the training high-resolution images. Despite considerable performance, their method failed to recover fine details of face images as it only focused on global face information. To alleviate this, various methods has been suggested in the literature. In \cite{refs:Zhuang2007}, authors adopted locality preserving projection (LPP) to learn the projection weights. Yang \textit{et al.} \cite{refs:Yang2008} employed non-negative matrix factorization (NMF) to find the face subspace, along with a patch-based sparse representation method using coupled overcomplete dictionaries to generate final hallucinated image. Also in \cite{refs:Park2008}, the coefficient vector was obtained through a recursive error back-projection method.

To find the aforementioned subspace, many studies have utilized the idea of manifold learning by assuming that LR face images and their HR counterparts are sampled from two manifolds which have similar local neighborhood structures. Liu \textit{et al.} \cite{refs:Liu2005} maintained the local features by developing a multilinear patch-based reconstruction method. Fan and Yeung \cite{refs:Fan2007} addressed the problem through a two-step approach using neighbor embedding over visual primitive features. Huang \textit{et al.} \cite{refs:Huang2010} applied canonical correlation analysis (CCA) to determine this subspace. Authors in \cite{refs:Hu2011} managed to learn pixel-wise structure prior represented as embedding coefficients to estimate the final result. Because of the one-to-many mapping relation between LR and HR samples, some researchers cast doubt on the above manifold assumption and suggested different alternative strategies. Li \textit{et al.} \cite{refs:Li2009} presented a manifold alignment approach which projected the two manifolds to a common hidden manifold. In another study \cite{refs:Farrugia2017} a strategy was devised to learn linear models based on the local geometrical structure on the high-resolution manifold. To avoid dealing with the difficulties of preserving local geometry in various resolutions, \cite{refs:Shi2018} directly regularized the relationship between target patch and training patches in the HR space. Later, Shi \textit{et al.} \cite{refs:Shi2019} addressed this challenge by training a series of adaptive kernel regression mappings for predicting the missing details from LR patches.

Position-patch based face hallucination methods have also gained wide popularity during the last decade. The main intuition behind these algorithms is that the HR counterpart of a given input LR image patch can be reconstructed by applying neighbor embedding to those patches located in the same position as the test patch. Ma \textit{et al.} \cite{refs:Ma2010} was first to suggest this method, by computing the reconstruction weights through solving a least square problem. To obtain a more suitable solution, Jung \textit{et al.} \cite{refs:Jung2011} decided to replace the least square estimation with a convex constrained optimization. Various attempts have been made recently to use the idea of locality-constrained representation (LcR) in order to impose a locality constraint on the least square inversion problem to encourage sparsity and locality simultaneously \cite{refs:Jiang2014a, refs:Jiang2014b, refs:Jiang2018}.

In recent years, with the advancement of neural networks, deep learning-based face hallucination algorithms have become increasingly prevalent in the literature. Motivated by powerful representation abilities of CNNs, Zhou \textit{et al.} \cite{refs:Zhou2015} designed a network architecture to learn the mapping between the raw input image and the face representations extracted by a deep convolutional network. To avoid oversmoothing problem and preserve more textural details, WaveletSRNet \cite{refs:Huang2017} reconstructed HR images in wavelet coefficient domain. Chen \textit{et al.} \cite{refs:Chen2021}, extracted multi-scale features by incorporating multiple encoders and decoders in bottom-up and top-bottom patterns. In \cite{refs:Hu2019}, a super-resolution technique was suggested which decomposed faces and recovered different components. Jiang \textit{et al.} \cite{refs:Jiang2020a} also developed a network with two individual branches to learn global facial shape and local facial components.

The emergence of generative adversarial networks had also a great impact on face super-resolution studies. Yu \textit{et al.} \cite{refs:Yu2016} pioneered in GAN-based face hallucination algorithms, by considering an architecture which consisted of two discriminative and generative networks. They further extended their work \cite{refs:Yu2017a} by incorporating multiple spatial transformer networks (STN) in their network to improve the alignment and upsampling performance. Later, they enhance robustness against noisy inputs and inputs with non-fixed resolution in \cite{refs:Yu2017b} and \cite{refs:Yu2020}, respectively. Bulat \textit{et al.} \cite{refs:Bulat2018} discussed the idea of learning the degradation before super-resolution in a two-stage process. In \cite{refs:Yang2020}, the problem was formulated with a collaborative suppression and replenishment framework, whereas the algorithm of \cite{refs:Chen2020a} made the training phase more effective and efficient by introducing spatial attention into the generator. A self-supervised method in which the problem of face super-resolution is expressed as generation problem was developed in \cite{refs:Menon2020}. \cite{refs:Jiang2020b} also took a different approach by integrating multiple deep learning networks of different types.

\begin{figure}[t]
	\captionsetup[subfloat]{farskip=0pt,captionskip=1pt}
	\centering
	\def\twidth{0.5}
	\includegraphics[width=\twidth\textwidth]{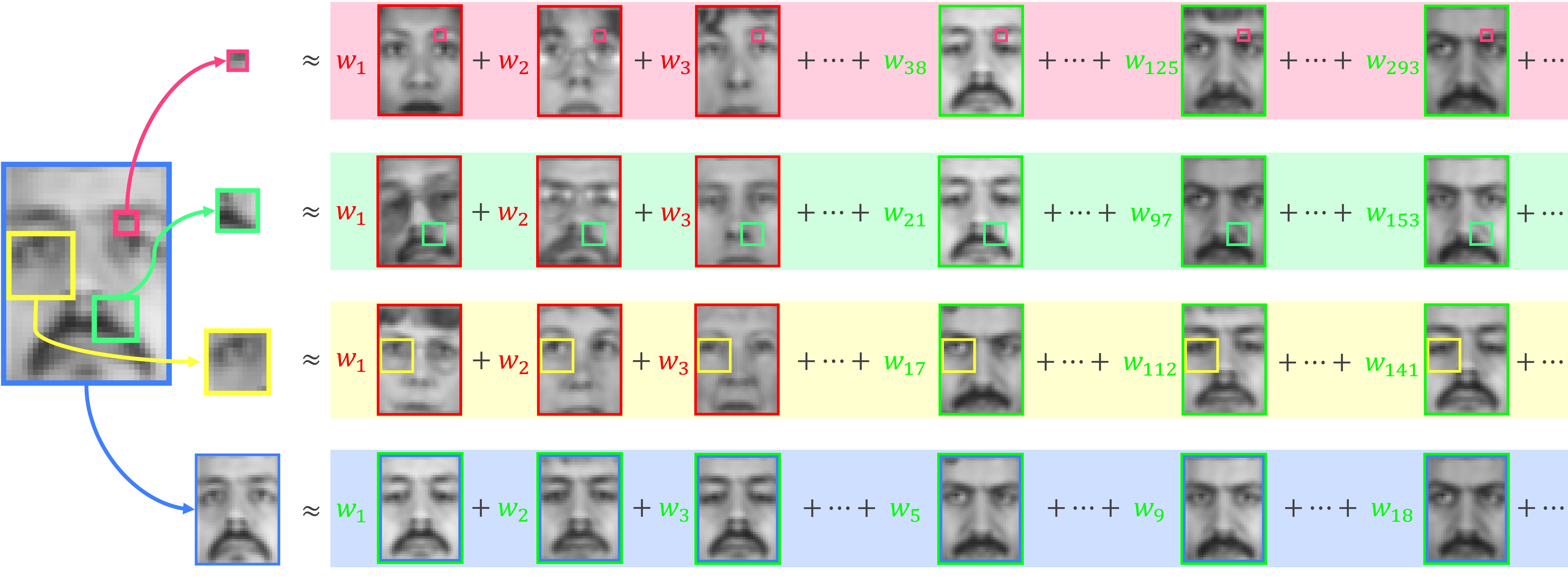}
	\caption{Effects of the patch size on the subject of the most similar training patches. Four different sized LR patches (denoted by different colors) are written as the linear combinations of the training patches in the same corresponding positions, whereas training images with green and red boundaries indicate identical and different subjects with reference to the test image subject, respectively. The top three most similar patches as well as the position of three patches belonging to the test image subject are displayed. When the whole image is taken into account, the top three most similar patches all belong to the subject of the LR test image, whereas the reverse is the case with smaller patches. Furthermore, considering the position of the training patches in each case (i.e., the coefficients), one can notice that the similarity between the test patch and those patches which belong to the test image subject is directly correlated with the patch size.}
	\label{fig:patchsize}
\end{figure}

\subsection{Motivation and Contribution}
Recent face hallucination studies have been dominated by two approaches: position-patch and deep learning-based methods. The first relies on the basic assumption that small patches in LR and HR spaces create manifolds with similar local geometry, hence they consider the reconstruction weights in both spaces equal. However, it has been shown \cite{refs:Li2009} that, due to the nonisometric one-to-multiple mappings from LR patches to HR ones, this assumption is not always met in practice. Therefore, face images of two entirely different individuals may have similar LR patches, whereas HR/LR patch pairs of a specific person may bear no similarity at all \cite{refs:Chen2019}. This becomes more severe as the LR input degradation level increases \cite{refs:Jiang2014b}, or patches with smaller sizes are considered. Fig. \ref{fig:patchsize} illustrates four patches of different sizes extracted from the same LR image. As shown, training patches belonging to the test subject tend to be more similar to the LR test patch when they increase in size. To further demonstrate this point, the average neighborhood preservation rates (NPR) \cite{refs:Su2005b} between the LR and HR image manifolds according to different patch size and based on three levels of degradation is presented in Fig. \ref{fig:npr}. As the graph suggests, selecting patches with bigger sizes gradually increases the average NPR, hence the reliability of the selected patches. However, the rising trend gradually weakens before the case when the whole image is taken into consideration, in which a sudden jump in the values of NPR is observable. The figure also reveals that, as image becomes more degraded, more invalid patches will be selected as the neighboring patches. Still, the case when the entire image is considered is relatively less affected by this change.

\begin{figure}[t]
	\captionsetup[subfloat]{farskip=0pt,captionskip=1pt}
	\centering
	\def\theight{0.22}
	\includegraphics[height=\theight\textwidth]{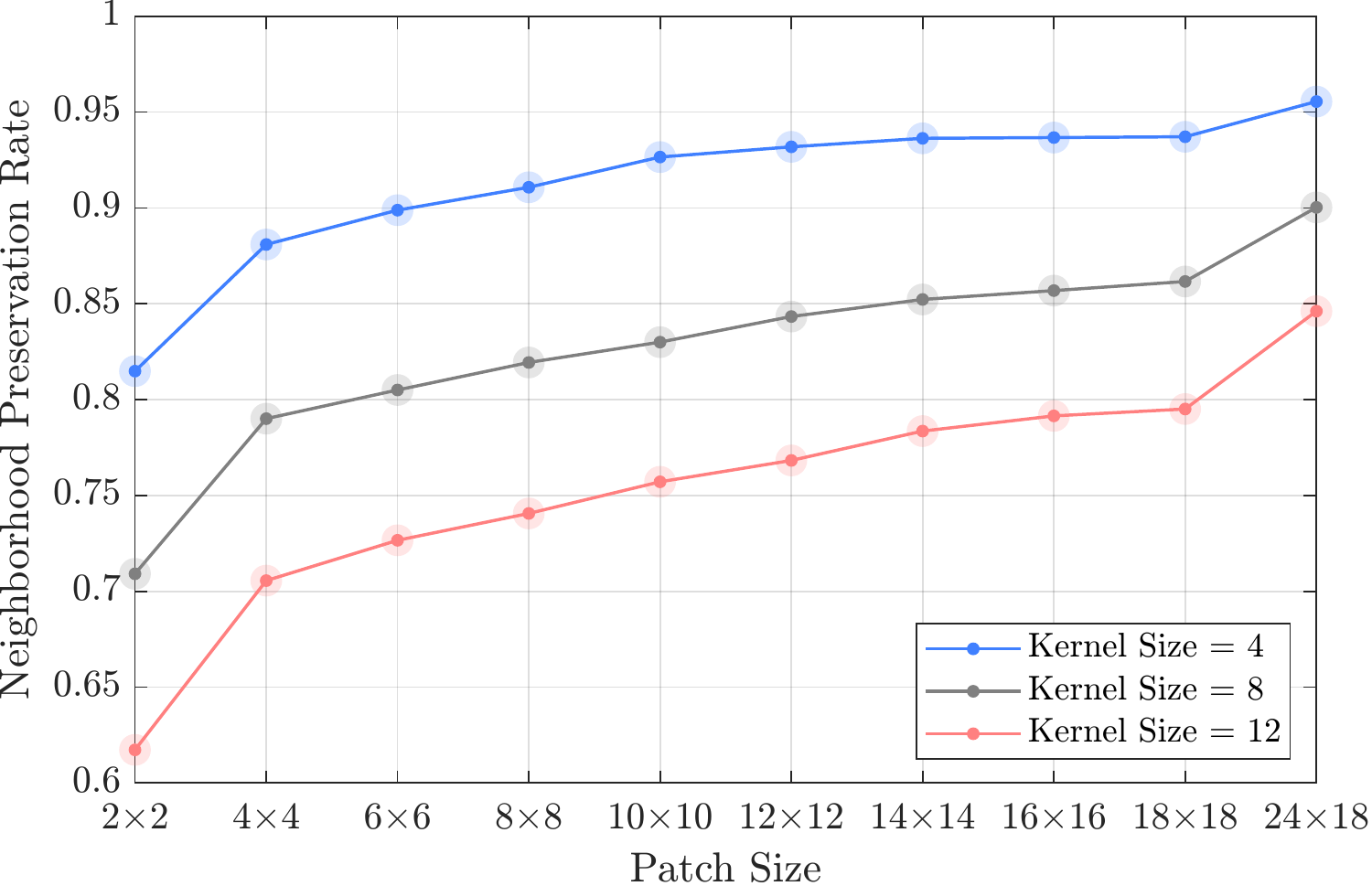}
	\caption{Influence of the patch size on the average neighborhood preservation rates between the LR and HR image manifolds, based on different levels of degradation (i.e., blur kernel size) and when the number of neighbors \textit{K} is set to 200. Patch-based strategies tend to be more erroneous as the patches decrease in size, and this becomes even worse in the presence of higher levels of degradation. Considering the entire face image as a patch leads to a sharp increase in the level of NPR and more robustness against facial degradation.}
	\label{fig:npr}
\end{figure}

Generally, patch-based methods often face the dilemma of selecting appropriate patch size. On one hand, to capture the global nature of faces and finding more meaningful and accurate neighbors, larger patches are preferred. On the other hand, as a consequence of the curse of dimensionality \cite{refs:Duda2001}, the size of the training set should grow exponentially with the patch size to guarantee valid matches \cite{refs:Romano2016} and avoid ghosting effects \cite{refs:Jiang2018}. Although several approaches have been suggested to alleviate this problem \cite{refs:Jiang2014b, refs:Jiang2018, refs:Chen2019}, existing patch-based methods still fail to recover person-specific facial features due to the above-mentioned limitations.

A similar argument can also be made in the case of deep learning-based face super-resolution algorithms. In spite of their great ability to add visually pleasing details to LR images, these algorithms often neglect how much beneficial the added information is for the task of recognizing the identity of the face \cite{refs:Hsu2019}. Most of the loss functions that have been considered in the literature are designed to minimize the mean square error (MSE) between the HR image and its corresponding reconstructed one, which, although can sometimes achieve high MSE-oriented quality metrics, in most cases produce blurry and over-smoothed results \cite{refs:Zhang2021a}. In order for deep learning-based methods to be able to learn identity-aware representations, they are required to be trained with a large well-labeled face dataset, which tends to be very costly \cite{refs:Hsu2019}. In recent years, several network architectures and loss functions have been suggested to incorporate the identity prior into the learning procedure \cite{refs:Hsu2019, refs:Zhang2021a}, however, still in many cases the hallucinated face hardly resembles the person in the test image, as shown in Fig. \ref{fig:vspulse}.

In this paper, a novel face hallucination approach is presented in which super-resolution is performed in the subspace spanned by the available training faces, often referred to as \textit{face subspace} \cite{refs:Wright2009}. To accomplish this, face subspace prior has been incorporated as a regularization term, and through a simple, yet vastly effective MAP-based formulation, the benefits of global hallucination are achieved whereas the drawbacks of patch-based methods are avoided. Additionally, although the proposed algorithm can be considered as a global reconstruction scheme, however, the hallucinated faces are artifact-free and robust to ghosting effects, and there is no constraints on dataset size either. The optimization process of the proposed objective function is also addressed through a highly effective recently introduced closed-form solution \cite{refs:Zhao2016}.

Furthermore, to better deal with the cases where there is a significant misalignment between the input LR face and the training faces, an effective 3D dictionary alignment technique which allows us to perform face super-resolution on unconstrained face images will be suggested. The proposed alignment procedure can also be used in the pipeline of other similar algorithms, and since the majority of current face hallucination approaches only produce satisfactory results when given frontal LR face images, this method can substantially increase their robustness against pose variations in LR inputs.

Therefore, the major contributions of the paper can be summarized as follows:
\begin{itemize}
	\item We force the reconstructed face to lie in the linear span of the training faces, hence, unlike most existing face hallucination algorithms, the reconstruction is performed in favor of both recovering identity-specific face attributes as well as enhancing image quantitative measures.
	\item We will show that not more than three samples per subject are required for our algorithm to guarantee an identity-preserving result and outperform the existing methods in both tasks of face super-resolution and face recognition.
	\item By incorporating an efficient 3D alignment procedure, the algorithm can extend its superior performance to the case where LR face pose is significantly different from the ones in the training set. More importantly, the proposed alignment scheme allows us to deal with face hallucination problems in which face images from both the training and testing sets are unconstrained.
	\item In contrast to the patch-based algorithms, the proposed method is barely affected by increasing the level of degradation, and shows outstanding robustness when given very low-resolution (VLR) face images.
	\item By utilizing an efficient closed-form solution for the proposed objective function, our method is considered to be a very fast face hallucination algorithm in which the computational time is comparatively less affected by increasing the size of the HR image or the number of samples in the dataset.
	\item The proposed method achieves superior performance over the competitive state-of-the-art algorithms in various frontal, non-frontal, and in-the-wild face hallucination experiments conducted on several well-known face datasets, and surpasses the position-patch and deep learning-based methods both quantitatively and qualitatively.
\end{itemize}
\begin{figure}
	\captionsetup[subfloat]{farskip=0pt,captionskip=1pt}
	\centering
	\def\imw{0.97}
	\def\imh{1.61} 
	\def\hdis{0.085}
	\def\vdiss{0.04}
	\def\vdisb{0.1}
	\begin{minipage}{\hdis\textwidth} 
		\centering
		\stepcounter{figure}\addtocounter{figure}{-1}
		\addtocounter{subfigure}{0}
		\subfloat[]{\includegraphics[width=\imw\textwidth]{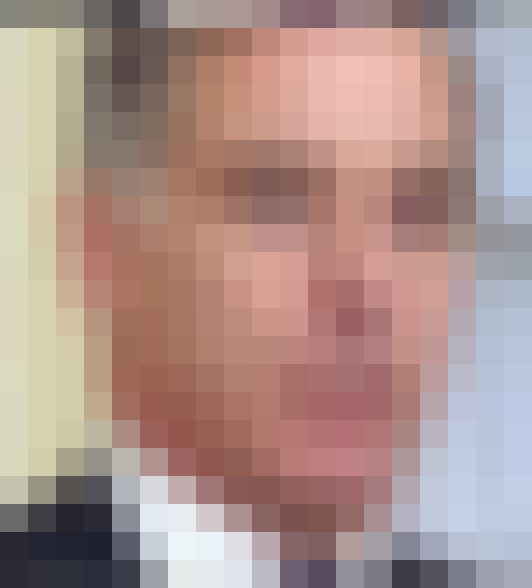}}
	\end{minipage}%
	\begin{minipage}{\hdis\textwidth} 
		\centering
		\stepcounter{figure}\addtocounter{figure}{-1}
	    \addtocounter{subfigure}{1}
		\subfloat[]{\includegraphics[width=\imw\textwidth]{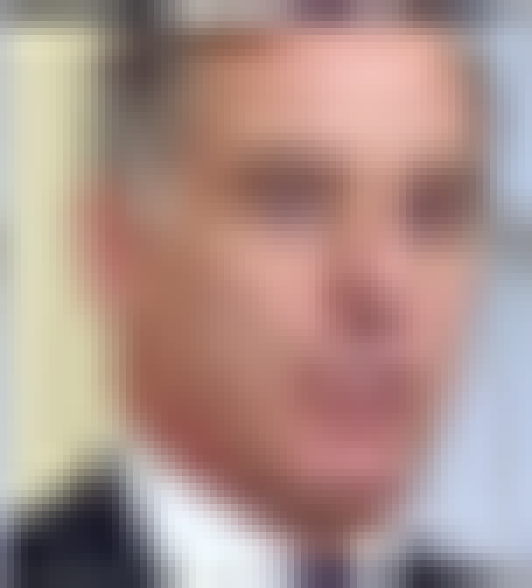}}
	\end{minipage}%
	\begin{minipage}{\hdis\textwidth} 
		\centering		
		\stepcounter{figure}\addtocounter{figure}{-1}
		\addtocounter{subfigure}{2}
		\subfloat[]{\includegraphics[width=\imw\textwidth]{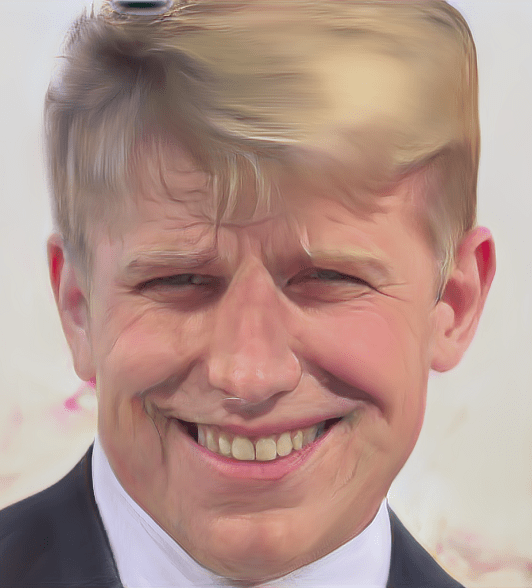}}
	\end{minipage}%
	\begin{minipage}{\hdis\textwidth} 
		\centering
		\stepcounter{figure}\addtocounter{figure}{-1}
		\addtocounter{subfigure}{3}
		\subfloat[]{\includegraphics[width=\imw\textwidth]{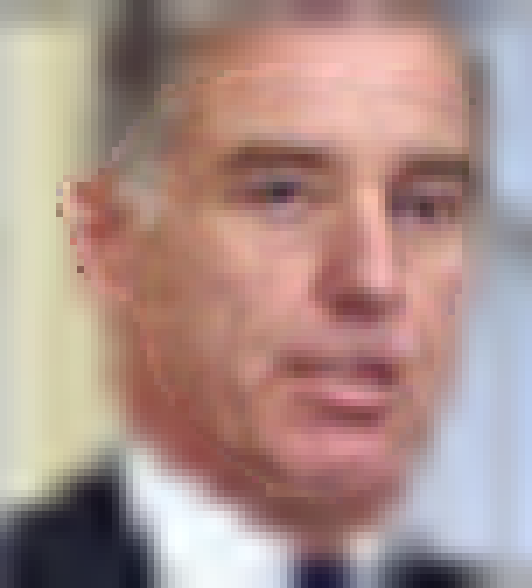}}
	\end{minipage}%
	\begin{minipage}{\hdis\textwidth} 
		\centering
		\stepcounter{figure}\addtocounter{figure}{-1}
		\addtocounter{subfigure}{4}
		\subfloat[]{\includegraphics[width=\imw\textwidth]{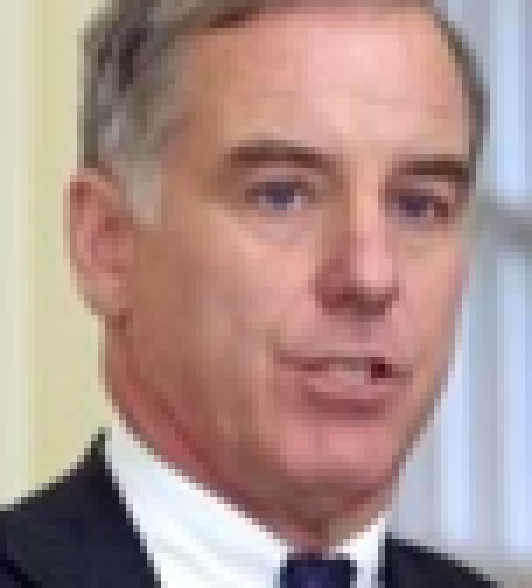}}
	\end{minipage}%
	\caption{The results obtained by a recently introduced GAN-based face hallucination technique \cite{refs:Menon2020} and the proposed algorithm in a real-world problem. Despite reconstructing aesthetically pleasing faces, the facial details recovered by deep learning methods may sometimes be vastly different from the ones in the ground truth face image. (a) Low-resolution input. (b) Bicubic interpolation. (c) PULSE \cite{refs:Menon2020}. (d) Proposed method. (e) Ground truth image.}
	\label{fig:vspulse}
\end{figure}

The organization of the rest of the paper is as follows: in section \ref{sec:proposed}, the face subspace prior along with our main MAP-based model are explained. Later, the details of the optimization procedure of the proposed algorithm is scrutinized, before the introduction of our 3D dictionary alignment pipeline which is detailed in the same section. The experimental evaluations and comparison with other competitive algorithms are the subject of section \ref{sec:results}, followed by conclusion and possible future works which are presented in section \ref{sec:conclusion}.

\section{Proposed Method}
\label{sec:proposed}
A common assumption in the problem of single image super-resolution is that the low-resolution input is a noisy, blurred, and decimated counterpart of the unknown high-resolution image. Consequently, the following forward degradation model is often taken into consideration:
\begin{equation} \label{eq:sr_model}
	\textbf{y} = \textbf{SHx} + \boldsymbol\epsilon
\end{equation}
in which $\textbf{y} \in{\rm I\!R}^{m_l \times 1}$ is the low-resolution input with $\textbf{x} \in {\rm I\!R}^{m_h \times 1}$ as its high-resolution equivalent, where $m_l = w_l \times h_l$ and $m_h = w_h \times h_h$. In addition, $\textbf{H} \in {\rm I\!R}^{m_h \times m_h}$ represents blurring filter, $\textbf{S} \in {\rm I\!R}^{m_l \times m_h}$ denotes decimation operator with scaling factor $d$, hence $m_h = m_l \times d^2$, and $\boldsymbol\epsilon \in{\rm I\!R}^{m_l \times 1}$ indicates additive white Gaussian noise (AWGN) encountered through the image acquisition process. The problem of super-resolution can therefore be written as an optimization problem derived from maximum likelihood (ML) estimator of the high-resolution image $\textbf{x}$ as below:
\begin{equation} \label{eq:sr_ml}
	\hat{\textbf{x}} = \argmin_\textbf{x}{\| \textbf{y} - \textbf{SHx} \|_2 ^ 2}
\end{equation}
which leads to the following solution:
\begin{equation} \label{eq:sr_ml_closed}
	\hat{\textbf{x}} = (\textbf{H}^T\textbf{S}^T\textbf{SH}) ^ {-1} (\textbf{H}^T\textbf{S}^T\textbf{y})
\end{equation}
This solution, which is equivalent to the least-square solution of the inverse problem of \eqref{eq:sr_model}, is known to be an ill-conditioned problem due to its sensitivity to small noise and measurement errors \cite{refs:Nasrollahi2014}. On condition that $\textbf{H}^T\textbf{S}^T\textbf{SH}$ is singular, the problem is also ill-posed with infinite space of possible solutions. Moreover, solving \eqref{eq:sr_ml_closed} requires inverting the matrix $\textbf{H}^T\textbf{S}^T\textbf{SH}$ with a computational complexity of the order $\mathcal{O}(m_h^3)$, which makes it practically inefficient in many real scenarios \cite{refs:Zhao2016}.

To overcome these problems, some additional information is needed to constrain the space of solutions and stabilize the problem. This is often accomplished by introducing a new term to \eqref{eq:sr_ml}, converting the maximum likelihood problem to a maximum \textit{a posteriori} (MAP) problem:
\begin{equation} \label{eq:sr_map}
	\hat{\textbf{x}} = \argmin_\textbf{x}{\| \textbf{y} - \textbf{SHx} \|_2 ^ 2 + \xi\;\Phi(\textbf{R}\textbf{x})}
\end{equation}
which consists of a fidelity term corresponding to model likelihood and a regularization term which represents \textit{a priori} knowledge about the original image, with the regularization parameter $\xi$ which determines the contribution of each term. $\textbf{R}$ is also a matrix which can be defined according to the application. Various priors for natural images have been suggested in the literature, with Tikhonov regularization and Total Variation as the most notable ones \cite{refs:Nasrollahi2014}. However, to the best of our knowledge, there are few, if any, such priors introduced for the purpose of addressing single-frame global face hallucination problem, and the studies mostly include multi-image \cite{refs:Chakrabarti2007} or patch-based \cite{refs:Jia2008} approaches.

\subsection{Identity-Preserving Face Prior}
In the area of pattern recognition, a well-established assumption is that patterns from a specific object class lie on a linear subspace \cite{refs:Duda2001}. In regard to facial recognition problem, it has been verified that face images belonging to a certain subject create a low-dimensional subspace, and the idea has been the cornerstone of various successful face recognition algorithms \cite{refs:Wright2009}. Let $\textbf{D}_{h,i} = [\textbf{x}_{i,1}, \textbf{x}_{i,2} ,\dots, \textbf{x}_{i,n_i}]$ be the HR training faces of the \textit{i}th subject. On condition that $n_i$ is sufficiently large, the above assumption implies that if the hallucinated face image $\textbf{x}$ belongs to the subject \textit{i}, for some scalar coefficients $\boldsymbol\alpha_{i,j} \in {\rm I\!R}$, $j = 1,2,\dots, n_i$, it can be represented as $\textbf{x} = \boldsymbol\alpha_{i,1}\textbf{x}_{i,1} + \boldsymbol\alpha_{i,2}\textbf{x}_{i,2} +\dots+ \boldsymbol\alpha_{i,n_i}\textbf{x}_{i,n_i}$. Therefore, in case the subject of the input LR face is given beforehand, a suitable prior term for \eqref{eq:sr_map} would be
\begin{equation} \label{eq:prior_single}
	\Phi(\textbf{x}) =  \| \textbf{x} - \boldsymbol\alpha_{i,1}\textbf{x}_{i,1} - \boldsymbol\alpha_{i,2}\textbf{x}_{i,2} -\dots- \boldsymbol\alpha_{i,n_i}\textbf{x}_{i,n_i} \|^2_2
\end{equation}
To generalize the above term, a dictionary matrix $\textbf{D}_h$ is defined which contains the whole $n$ training faces of all $c$ subjects, in which, to facilitate classification task, face images of the same subject are arranged beside each other, that is, $\textbf{D}_h = [\textbf{D}_{h,1}, \textbf{D}_{h,2},\dots, \textbf{D}_{h,c}] = [\textbf{x}_{1,1}, \textbf{x}_{1,2},\dots, \textbf{x}_{c,n_c}]$. Therefore, \eqref{eq:prior_single} changes to
\begin{equation} \label{eq:prior_all}
	\Phi(\textbf{x}) =  \| \textbf{x} - \textbf{D}_h\boldsymbol\alpha \|^2_2
\end{equation}
where $\boldsymbol\alpha$ is the coefficient vector with non-zero entries for those elements associated with subject \textit{i}, and zero elsewhere. Incorporating this prior term into \eqref{eq:sr_map} gives
\begin{equation} \label{eq:sr_map_dic}
	\{ \hat{\textbf{x}}, \hat{\boldsymbol\alpha} \} = \argmin_{\textbf{x}, \boldsymbol\alpha}{\| \textbf{y} - \textbf{SHx} \|_2 ^ 2 + \mu \| \textbf{x} - \textbf{D}_h\boldsymbol\alpha \|^2_2}
\end{equation}
Provided that the number of subjects is sufficiently large, $\boldsymbol\alpha$ is expected to have a sparse representation, since only few of its elements will have nonzero values. Therefore, a new regularization term $\|\boldsymbol\alpha\|_0$ should be included in \eqref{eq:sr_map_dic}, and since this leads to an NP-hard problem, the $l_0$-norm is replaced by $l_1$-norm \cite{refs:Wright2009}: 
\begin{equation} \label{eq:sr_map_dic_sparse}
	\{ \hat{\textbf{x}}, \hat{\boldsymbol\alpha} \}  = \argmin_{\textbf{x}, \boldsymbol\alpha}{\| \textbf{y} - \textbf{SHx} \|_2 ^ 2 + \mu \| \textbf{x} - \textbf{D}_h\boldsymbol\alpha \|^2_2 + \lambda \| \boldsymbol\alpha \|_1}
\end{equation}
With the above formulation, the hallucination is constrained to the subspace spanned by the subject which gives the sparsest coefficient vector with respect to the input image. Thus, if there are enough samples of the subject to which the input face belongs, the super-resolution is performed for the benefit of recovering true facial attributes. In section \ref{sec:results} we will show that not more than three samples per subject are required to satisfy this condition.

\subsection{Optimization}
The optimization problem \eqref{eq:sr_map_dic_sparse} can be divided into two subproblems associated with each of the variables $\textbf{x}$ and $\boldsymbol\alpha$, before being solved iteratively for one while fixing the other. The following two optimization steps can therefore be defined: 

\subsubsection{Optimizing for $x$}
The intermediate HR estimate of the LR input $\textbf{y}$ in a given iteration $t$ can be obtained through the following $l_2$-regularized optimization problem:
\begin{equation} \label{eq:subproblem_x}
	{\textbf{x}}_{t+1} = \argmin_{\textbf{x}}{\| \textbf{y} - \textbf{SHx} \|_2 ^ 2 + \mu \| \textbf{x} - \textbf{D}_h\boldsymbol\alpha_t \|^2_2}
\end{equation}
whose closed-form solution is given by
\begin{equation} \label{eq:subproblem_x_closed}
	{\textbf{x}}_{t+1} = (\textbf{H}^H \textbf{S}^H \textbf{SH} + 2 \mu\textbf{I}_{m_h}) ^ {-1} (\textbf{H}^H \textbf{S}^H \textbf{y} + 2 \mu \textbf{D}_h\boldsymbol\alpha_t)
\end{equation}
Unlike the optimization procedure of other similar inverse problems (e.g., image deblurring \cite{refs:Zhang2011b}) which can be solved efficiently in the frequency domain, here, the presence of the decimation operator $\textbf{S}$ in the fidelity term, and the fact that the product matrix $\textbf{SH}$ does not have a block-circulant structure and cannot be diagonalized in the frequency domain, makes the problem impossible to be solved using the Fourier transform.
However, \cite{refs:Zhao2016} showed that under certain assumptions on the decimation operator $\textbf{S}$ and the blurring matrix $\textbf{H}$, the optimization problem admits a closed-form solution in the frequency domain. 

\begin{figure*}
	\captionsetup[subfloat]{farskip=0pt,captionskip=1pt}
	\centering
	\def\theight{0.3}
	\includegraphics[height=\theight\textwidth]{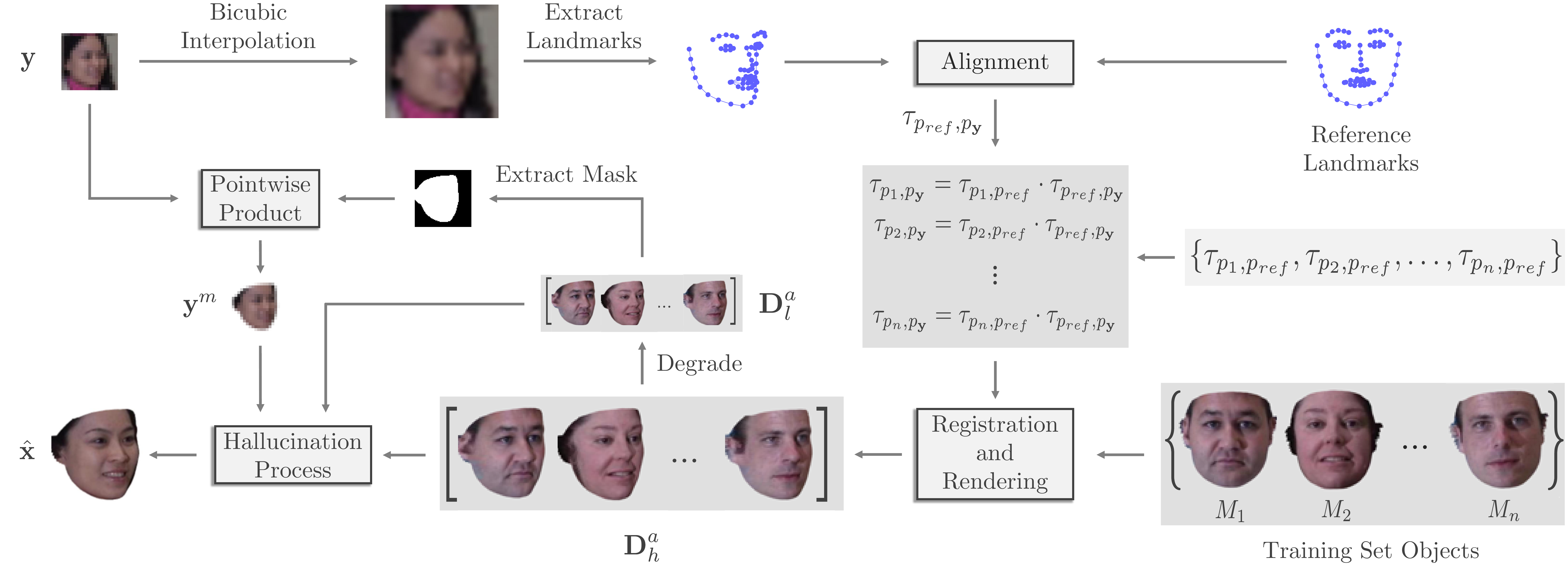}
	\caption{Flowchart of the 3D dictionary alignment framework. The 3D facial landmarks are first extracted from the upscaled version of the LR input face, before being used to align it with respect to a set of landmark points defined as reference and obtain the associated transformation matrix. Having previously calculated the transformations between the dictionary samples and the reference landmarks, the transformation matrices between the training samples and the input face image can therefore be efficiently obtained. After applying these transformation matrices to their corresponding training face objects and performing 3D rendering, the aligned HR and LR dictionaries are obtained which will be later used along with the masked LR input in the process of face hallucination.}
	\label{fig:flowchart}
\end{figure*}

More precisely, assuming $\textbf{H}$ as the matrix representation of the cyclic convolution operator, one can decompose the blurring operator $\textbf{H}$ and its conjugate transpose $\textbf{H}^H$ as
\begin{equation} \label{eq:blur_mat_dec}
	\textbf{H} = \textbf{F}^H \boldsymbol\Lambda \textbf{F}, \quad \textbf{H}^H = \textbf{F}^H \boldsymbol\Lambda^H \textbf{F}
\end{equation}
where $\textbf{F} \in \mathbb{C} ^ {m_h \times m_h}$ is the discrete Fourier transform matrix with the property $\textbf{F} ^ H = \textbf{F} ^ {-1}$, hence $\textbf{F}\textbf{F}^H = \textbf{F}^H\textbf{F} = \textbf{I}_{m_h}$, and $\boldsymbol\Lambda \in \mathbb{C} ^ {m_h \times m_h}$ is a diagonal matrix whose elements are the Fourier transform of the zero-padded PSF, that is, the first column of the blurring matrix $\textbf{H}$.

 Additionally, $\textbf{S}$ is assumed to be a downsampling operator whose conjugate transpose $\textbf{S}^H$ interpolates the decimated image with zeros, and satisfies the relationship $\textbf{S}\textbf{S}^H = \textbf{I}_{m_l}$. By considering $ \underbar{\textbf{S}} \triangleq  \textbf{S}^{H}\textbf{S} $, which operates as an element-wise multiplication by an $m_h \times m_h$ matrix with ones at the sampled positions and zeros elsewhere,  \cite{refs:Wei2015} showed that:
\begin{equation} \label{eq:kronecker}
	\textbf{F} \underbar{\textbf{S}} \textbf{F}^H = \frac{1}{d^2} \textbf{J}_{d^2} \otimes \textbf{I}_{m_l}
\end{equation}
in which $\otimes$ denotes the Kronecker product, $\textbf{J}_{d^2} \in {\rm I\!R} ^ {{d^2} \times {d^2}}$ is a matrix of ones, and $\textbf{I}_{m_l} \in {\rm I\!R} ^ {m_l \times m_l}$ is an identity matrix. Bearing in mind \eqref{eq:kronecker} as well as the previously mentioned assumptions, one can rewrite the analytical solution \eqref{eq:subproblem_x_closed} as
\begin{equation} \label{eq:subproblem_x_closed_fourier1}
	{\textbf{x}}_{t+1} = \textbf{F}^{H} \left(\frac{1}{{d^2}} \munderbar{\boldsymbol\Lambda}^{H} \munderbar{\boldsymbol\Lambda} + 2\mu \textbf{F}\textbf{F}^H\right) ^ {-1} \textbf{F}(\textbf{H}^H \textbf{S}^H \textbf{y} + 2 \mu \textbf{D}_h\boldsymbol\alpha_t)
\end{equation}
where $\munderbar{\boldsymbol\Lambda} \in \mathbb{C} ^ {m_l \times m_h} $ is defined as $\munderbar{\boldsymbol\Lambda} = [\boldsymbol\Lambda_1, \boldsymbol\Lambda_2,\dots, \boldsymbol\Lambda_{d^2}]$, with the blocks $\boldsymbol\Lambda_i \in \mathbb{C} ^ {m_l \times m_l} (i = 1,\dots, {d^2})$ such that $diag\{\boldsymbol\Lambda_1,\dots, \boldsymbol\Lambda_{d^2}\} = \boldsymbol\Lambda$. This can be further simplified by incorporating the Woodbury matrix identity \cite{refs:Hager1989} into \eqref{eq:subproblem_x_closed_fourier1} and obtaining the following closed-form solution:
\begin{equation} \label{eq:subproblem_x_closed_fourier2}
	{\textbf{x}}_{t+1} = \frac{1}{2 \mu} \left(\textbf{r} - \textbf{F}^H \munderbar{\boldsymbol\Lambda}^H (2\mu {d^2} \textbf{I}_{m_l} + \munderbar{\boldsymbol\Lambda}\munderbar{\boldsymbol\Lambda}^H) ^ {-1} \munderbar{\boldsymbol\Lambda}\textbf{Fr}\right) 
\end{equation}
where $\textbf{r} = \textbf{H}^H \textbf{S}^H \textbf{y} + 2 \mu \textbf{D}_h\boldsymbol\alpha_t$.

\subsubsection{Optimizing for $\alpha$}
The coefficient vector $\boldsymbol\alpha$ is updated using the following well-known $l_1$-minimization problem:
\begin{equation} \label{eq:subproblem_a}
	{\boldsymbol\alpha}_{t+1} = \argmin_{\boldsymbol\alpha}{\| \textbf{x}_t - \textbf{D}_h\boldsymbol\alpha \|^2_2 + \lambda \| \boldsymbol\alpha \|_1}
\end{equation}
which can be efficiently solved using various $l_1$-minimization algorithms. It should also be noted that the initial $\boldsymbol\alpha$ vector is obtained through solving the above minimization problem for the input face $\textbf{y}$ and the low-resolution training dictionary $\textbf{D}_l$, i.e., ${\boldsymbol\alpha}_0 = \argmin_{\boldsymbol\alpha}{\| \textbf{y} - \textbf{D}_l\boldsymbol\alpha \|^2_2 + \lambda \| \boldsymbol\alpha \|_1}$.

Algorithm \ref{alg:main} summarizes the entire procedure of the proposed face hallucination approach.

\begin{algorithm}[t]
	\fontsize{9}{10}\selectfont
	\caption{Identity-Preserving Face Hallucination}
	\textbf{Input:} HR and LR dictionaries $\textbf{D}_h$ and $\textbf{D}_l$, low-resolution image $\textbf{y}$, blurring matrix $\textbf{H}$, scaling factor $d$, regularization parameters $\mu$ and $\lambda$, number of iterations $T$.
	\begin{algorithmic}[1]
		\State Initialize $\boldsymbol\alpha$: ${\boldsymbol\alpha}_0 = \argmin_{\boldsymbol\alpha}{\| \textbf{y} - \textbf{D}_l\boldsymbol\alpha \|^2_2 + \lambda \| \boldsymbol\alpha \|_1}$.
		\State Factorize the blurring matrix: $\textbf{H} = \textbf{F}^H \boldsymbol\Lambda \textbf{F}$.
		\State Compute $\munderbar{\boldsymbol\Lambda}$: $\munderbar{\boldsymbol\Lambda} = [(\textbf{1}^{T}_{d} \otimes \textbf{I}_{w_l}) \otimes (\textbf{1}^{T}_{d} \otimes \textbf{I}_{h_l})] \boldsymbol\Lambda$.
    	\For{$t = 0 : T - 1$}
			\State Find the \textit{FFT} of $\textbf{r}$: $\textbf{Fr} = \textbf{F}(\textbf{H}^H \textbf{S}^H \textbf{y} + 2 \mu \textbf{D}_h\boldsymbol\alpha_{t})$.
			\State Find entrywise product in the frequency domain: 
			\\\qquad\quad $\textbf{x}_f = \left(\munderbar{\boldsymbol\Lambda}^H (2\mu {d^2} \textbf{I}_{m_l} + \munderbar{\boldsymbol\Lambda}\munderbar{\boldsymbol\Lambda}^H) ^ {-1} \munderbar{\boldsymbol\Lambda}\right)\textbf{Fr}$.
			\State Update $\textbf{x}$: ${\textbf{x}}_{t+1} = \frac{1}{2 \mu}(\textbf{r} - \textbf{F}^H\textbf{x}_f)$.
			\State Update $\boldsymbol{\alpha}$: ${\boldsymbol\alpha}_{t+1} = \argmin_{\boldsymbol\alpha}{\| \textbf{x}_{t} - \textbf{D}_h\boldsymbol\alpha \|^2_2 + \lambda \| \boldsymbol\alpha \|_1}$.
		\EndFor
	\end{algorithmic}
	\textbf{Output:} Hallucinated face $\hat{\textbf{x}} = {\textbf{x}}_{t+1}$.
	\label{alg:main}
\end{algorithm}

\subsection{3D Dictionary Alignment}
As face images of different poses are distributed on a highly nonlinear manifold \cite{refs:Tenenbaum2000}, the majority of dictionary-based face hallucination algorithms fail to achieve satisfactory results when given non-frontal input faces. Inspired by recent advances in 3D face reconstruction and alignment studies, in this section an efficient dictionary alignment procedure will be presented, by which the training faces are registered with respect to the LR face pose before being used in the hallucination process. This additional step gives significant flexibility to the main algorithm and boosts its performance in reconstructing LR faces in the presence of high pose variations, even when the training faces are also non-frontal and unconstrained. The proposed alignment procedure, which is visually presented in Fig. \ref{fig:flowchart}, contains the following steps:

\subsubsection{Training Faces 3D Reconstruction}
In order to perform 3D alignment, we first use \cite{refs:Feng2018} to generate 3D geometries associated with each of the training faces, and obtain two sets $\{M_i\}^{n}_{i=1}$ and $\{C_i\}^{n}_{i=1}$, such that $M_i = (V_i, T_i)$ is the 3D mesh associated with the \textit{i}th training face, with $V_i \in {\rm I\!R}^{n_v \times 3} $, $T_i \in {\rm I\!N}^{n_t \times 3} $, and $C_i \in {\rm I\!R}^{n_v \times 3} $ as its vertices, triangles, and color attributes, where $n_v$ and $n_t$ denote the number of vertices and triangles, respectively. As might be expected, the whole process of face reconstruction is performed offline, hence does not affect the runtime of the main hallucination process. 

\subsubsection{LR Face Landmark Detection}
The most crucial part of the alignment pipeline is to locate facial landmarks on the LR input face. Since the proposed hallucination algorithm will accept degraded facial images with high variations in pose, the landmark detection method is expected to be highly robust and perform well in uncontrolled conditions. Fortunately, recent advances in deep neural networks has allowed researchers to propose powerful facial landmark detectors with considerable speed, accuracy, and stability. According to our experiments, most of the current state-of-the-art approaches fully satisfy our desired level of robustness, and therefore are eligible to be used in the alignment procedure. Fig. \ref{fig:nme} illustrates the average normalized mean error (NME) between the landmarks of a set of upscaled LR face images detected by \cite{refs:Guo2020} and their corresponding ground truth points, based on different levels of degradation. According to the figure, in most cases the difference between landmarks detected in the LR face images and their ground truth points is fairly negligible, even when face images as small as $15\times15$ pixels with $5\times5$ blur kernel are considered. Since conventional degradation settings are often far more lenient, we can rest assured that the detected landmarks $p_{y}$ will not affect the main alignment procedure.

\begin{figure}
	\captionsetup[subfloat]{farskip=0pt,captionskip=1pt}
	\centering
	\def\twidth{0.48}
	\subfloat{\includegraphics[width=\twidth\textwidth]{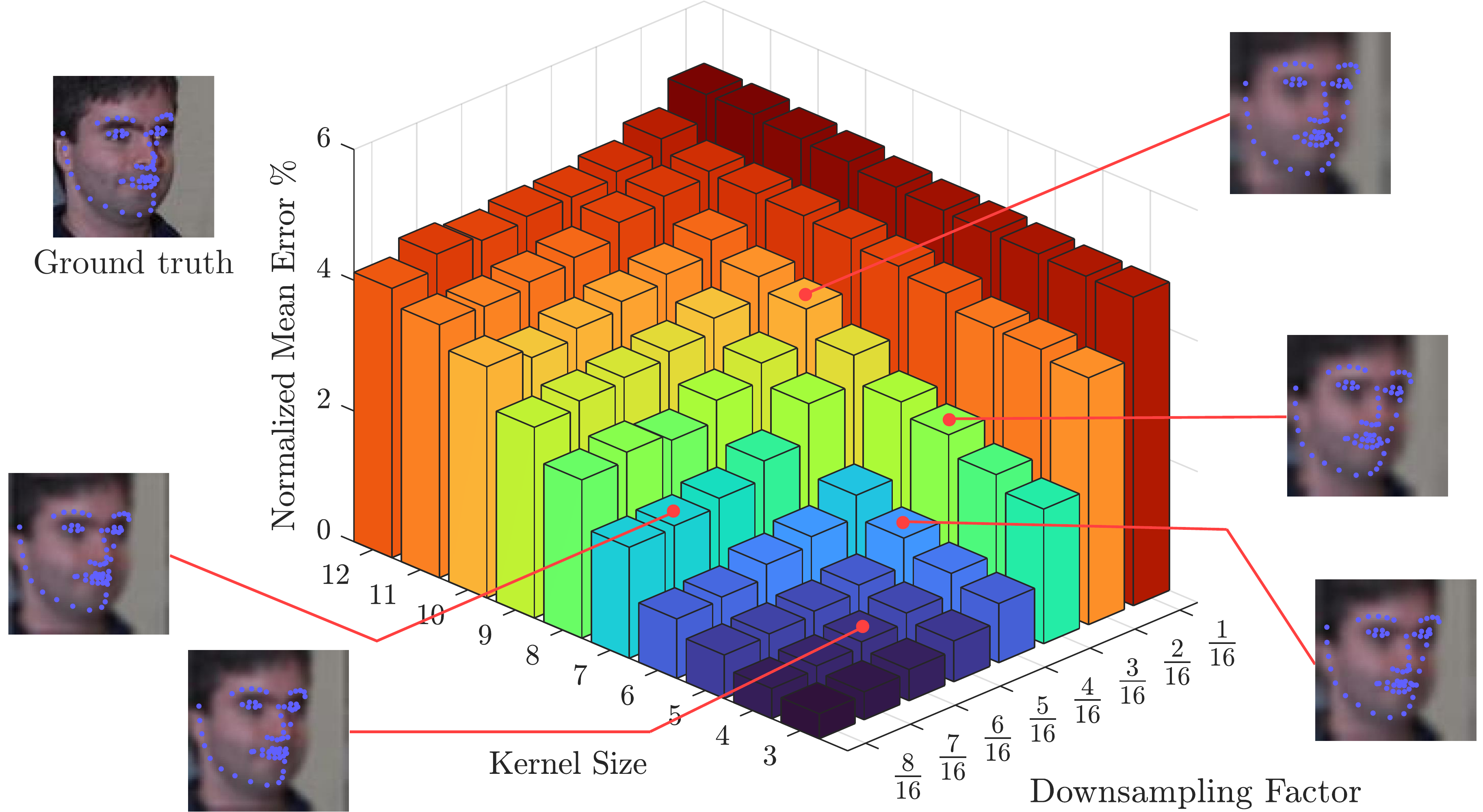}}\hfil
	\caption{The average normalized mean error achieved by a state-of-the-art landmark detector \cite{refs:Guo2020} on 20 randomly selected faces from the Multi-PIE dataset with pose variations 04-1, 05-0, 13-0, and 14-0, degraded by average filter of different sizes with various downsampling factors from their $80\times60$ HR images. The results of landmark detection on a test face with different levels of degradation as well as its ground truth landmark points are also presented. The figure suggests that the detector robustness is more than enough to be used in our dictionary alignment pipeline.}
	\label{fig:nme}
\end{figure}

\subsubsection{Registering Training Faces}
Having the 3D landmark points associated with both the training HR faces and the input LR face, one can easily estimate the transformation matrices needed for registering each training face with respect to the LR face pose. To speed up the process, a set of landmark points $p_{ref}$ are considered as the reference, and the transformation matrices $\{ \tau_{p_1,p_{ref}}, \tau_{p_2,p_{ref}},\dots, \tau_{p_n,p_{ref}}\}$ are calculated in the training phase, in which $\tau_{p_i,p_{ref}}$ denotes the transformation between the landmarks associated with the \textit{i}th training sample and the reference. In the test phase, it is only required to find the transformation matrix $\tau_{p_{ref},p_\textbf{y}}$ between the LR input face and the reference landmarks, and obtain $\{ \tau_{p_1,p_\textbf{y}}, \tau_{p_2,p_\textbf{y}},\dots, \tau_{p_n,p_\textbf{y}}\}$ simply by $ \tau_{p_i,p_\textbf{y}} = \tau_{p_i,p_{ref}} \cdot \tau_{p_{ref}, p_\textbf{y}} $. Performing the registrations and finding the transformed face objects is also efficiently implemented by applying each transformation to the corresponding object vertices.

\subsubsection{3D Face Rendering}
Finally, the 3D face objects are converted to 2D images using the available object rendering algorithms. This step is the most time-consuming part of the whole process, and extra care must be taken to preserve the details and information of the face object.

The above steps are performed on each of the training samples to obtain the aligned HR and LR dictionaries $\textbf{D}^{a}_h$ and $\textbf{D}^{a}_l$. To reduce the error caused by non-face regions in the process of hallucination, we also remove the excess pixels in the LR input face using a mask extracted from the average of the registered LR faces to obtain $\textbf{y}^m$, which will be later used along with $\textbf{D}^{a}_h$ and $\textbf{D}^{a}_l$ in Algorithm \ref{alg:main} to perform pose-robust face hallucination. As will be shown in section \ref{sec:results}, the proposed 3D dictionary alignment significantly improves the face hallucination performance even in the presence of high pose variations. Moreover, the whole process takes roughly 0.04 seconds for a $20\times20$ training face image and its corresponding object with $n_v = 43,867$ and $n_t = 86,906$, which is indeed a reasonable time considering the amazing benefits it offers. A summary of the 3D dictionary alignment procedure is presented in Algorithm \ref{alg:align} as pseudo-code.

\begin{algorithm}
	\fontsize{9}{10}\selectfont
	\caption{3D Dictionary Alignment}
	\textbf{Input:} Training face objects $\{M_i\}^{n}_{i=1}$ and their color attributes $\{C_i\}^{n}_{i=1}$, reference landmark points $p_{ref}$, transformations between the training faces and the reference landmarks $\{\tau_{p_1,p_{ref}}, \tau_{p_2,p_{ref}},\dots, \tau_{p_n,p_{ref}}\}$, low-resolution image $\textbf{y}$.
	\begin{algorithmic}[1]
		\State $\textbf{y}_{bic} \gets upscale(\textbf{y})$
		\State Extract 3D facial landmark points from $\textbf{y}_{bic}$.
		\State Align $\textbf{y}_{bic}$ with respect to the reference landmarks $p_{ref}$ and obtain the transformation matrix $\tau_{p_{ref},p_\textbf{y}}$.
		\For{$i = 1 : n$}
			\State $\tau_{p_i,p_\textbf{y}} = \tau_{p_i,p_{ref}} \cdot \tau_{p_{ref}, p_\textbf{y}}$
			\State $V^{a}_i = V_i \circ \tau_{p_i,p_\textbf{y}}$
			\State $\textbf{x}^{a}_i \gets render(V^{a}_i, T_i, C_i)$
			\State Add $\textbf{x}^{a}_i$ to the HR dictionary $\textbf{D}^{a}_h$.
		\EndFor
		\State $\textbf{D}^{a}_l \gets degrade(\textbf{D}^{a}_h)$
		\State Obtain $\textbf{y}^{m}$ by applying the mask extracted from the average of the LR aligned faces to $\textbf{y}$.
	\end{algorithmic}
	\textbf{Output:} Aligned dictionaries $\textbf{D}^{a}_h$ and $\textbf{D}^{a}_l$, masked LR input $\textbf{y}^{m}$.
	\label{alg:align}
\end{algorithm}

\section{Experimental Results}
\label{sec:results}
In this section, several experiments have been carried out to evaluate the performance of the proposed algorithm and demonstrate its efficiency and applicability. For this purpose, a number of recently published state-of-the-art face hallucination methods have been selected\footnote{All the implementations used in the evaluations are from the official source codes released by the authors.} with their parameters tuned so that they produce their optimal results. The experiments on frontal face hallucination are performed on the FERET \cite{refs:Phillips2000}, the CMU Multi-PIE \cite{refs:Gross2010}, and the AR \cite{refs:Martinez1998} public face datasets, whereas the pose-robust face super-resolution algorithm is tested on the CMU Multi-PIE and the LFW \cite{refs:Huang2008} databases.

\subsection{Experiments on Frontal Faces}
The proposed method is first assessed on faces taken in controlled condition. Unless otherwise specified, all the LR face images are obtained after applying downsampling and blurring (by a $4 \times 4$ average smoothing filter) to their HR counterparts. Samples from all databases were aligned based on the location of the eye corners. For all the experiments, one random image per each subject was selected as the test sample and the remaining were used in the training phase. The regularization parameters $\mu$ and $\lambda$ are chosen to be $10^{-8}$ and $2700$, respectively, whereas the number of iterations $T$ is set to 30. In the patch-based approaches, the patch size and the overlapping parameters are chosen according to the LR and HR image sizes. In \cite{refs:Wang2005}, the eigenvalues accumulation contribution rate is set to 0.99. \cite{refs:Farrugia2017} is modified so that it includes the blur information of the LR inputs. The implementations of \cite{refs:Jiang2014a} and \cite{refs:Jiang2014b} were slightly changed to prevent errors in recovering very low-resolution face images. 

\begin{table}[t]
	\caption{Evaluation of quantitative performance of different methods on the $15\times10$ samples of the FERET database by different scaling factors}
	\centering
	\begin{tabular}{c||cc|cc|cc} 
		\hline\hline
		\multirow{2}{*}{Algorithm} & \multicolumn{2}{c|}{$15\times10$ \textbar{} $\times2$} & \multicolumn{2}{c|}{$15\times10$ \textbar{} $\times4$} & \multicolumn{2}{c}{$15\times10$ \textbar{} $\times8$}  \\ 
		\cline{2-7}
		& PSNR           & SSIM                                  & PSNR           & SSIM                                  & PSNR           & SSIM                                  \\ 
		\hline
		Bicubic                    & 24.11          & 0.7272                                & 23.12          & 0.6327                                & 20.89          & 0.5468                                \\
		Wang \cite{refs:Wang2005}                  & 26.29          & 0.8648                                & 25.93          & 0.7799                                & 25.36          & 0.7122                                \\
		LSR \cite{refs:Ma2010}                       & 31.71          & 0.9568                                & 27.33          & 0.8223                                & 24.84          & 0.6739                                \\
		LcR \cite{refs:Jiang2014a}                       & 31.12          & 0.9485                                & 28.24          & 0.8494                                & 23.05          & 0.5969                                \\
		LINE \cite{refs:Jiang2014b}                      & 32.69          & 0.9669                                & 28.67          & 0.8656                                & 23.41          & 0.6225                                \\
		SSR \cite{refs:Jiang2017}                       & 30.51          & 0.9466                                & 28.91          & 0.8752                                & 26.76          & 0.7619                                \\
		LM-CSS \cite{refs:Farrugia2017}                    & 29.89          & 0.9416                                & 27.75          & 0.8466                                & 26.49          & 0.7482                                \\
		TRNR \cite{refs:Jiang2016}                      & 32.99          & 0.9690                                & 29.86          & 0.8976                                & 25.91          & 0.7323                                \\
		TLcR-RL \cite{refs:Jiang2018}                   & 33.25          & 0.9700                                & 29.69          & 0.8955                                & 27.58          & 0.7998                                \\
		Proposed                   & \textbf{34.43} & \textbf{0.9795}                       & \textbf{30.86} & \textbf{0.9175}                       & \textbf{28.58} & \textbf{0.8351}                       \\
		\hline\hline
	\end{tabular}
	\label{tab:feretres}
\end{table}

\begin{figure*}[!t]
	\captionsetup[subfloat]{farskip=0pt,captionskip=1pt}
	\centering
	\def\imw{1.45}
	\def\imh{2.03}
	\def\hdis{0.082}
	\def\vdiss{0.04}
	\def\vdisb{0.1}
	\begin{minipage}{\hdis\textwidth} 
		\centering
		\subfloat {\includegraphics[width=\imw cm, height=\imh cm]{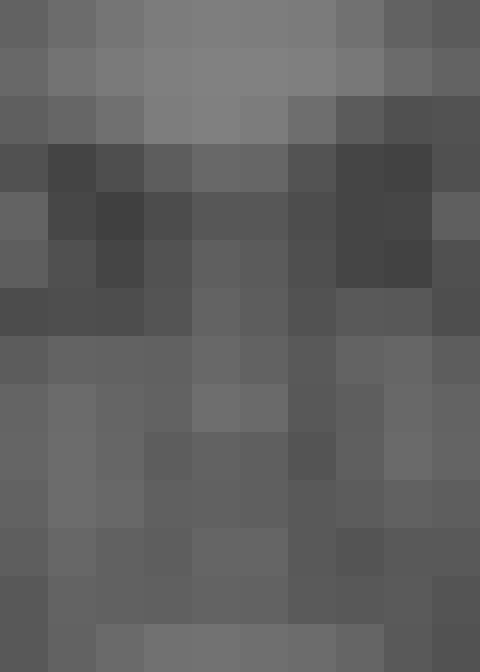}}\par\vspace{\vdiss cm}		
		\subfloat {\includegraphics[width=\imw cm, height=\imh cm]{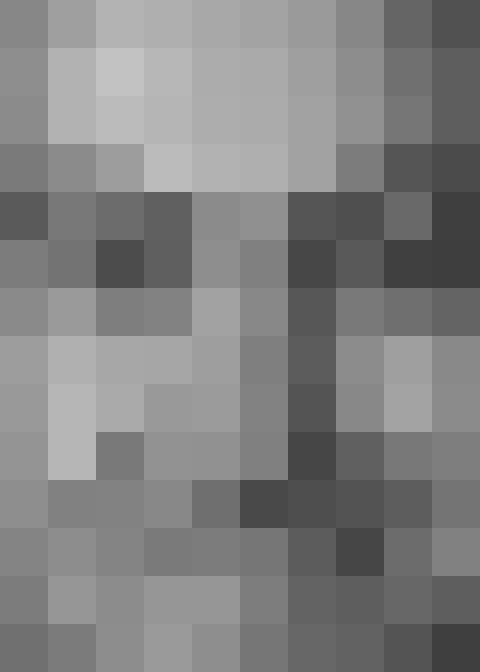}}\par\vspace{\vdiss cm}
		\subfloat {\includegraphics[width=\imw cm, height=\imh cm]{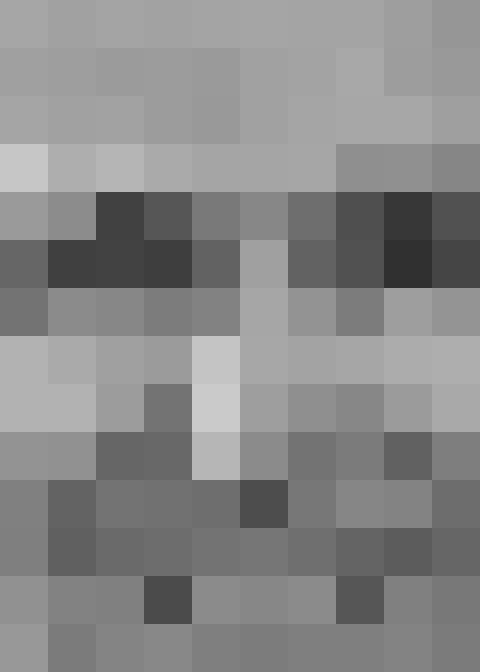}}\par\vspace{\vdisb cm}
		\subfloat {\includegraphics[width=\imw cm, height=\imh cm]{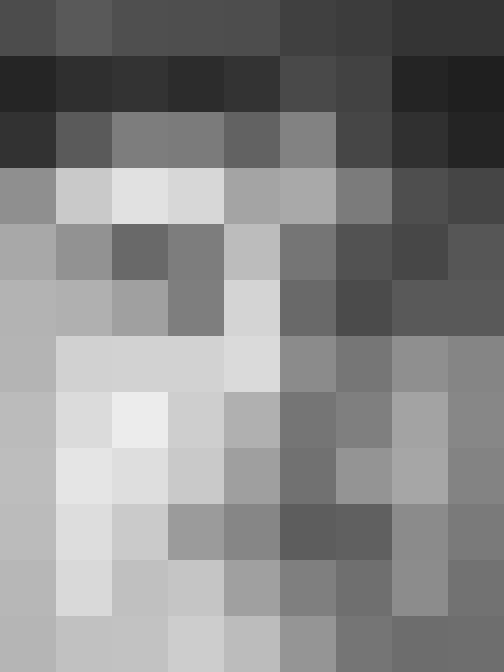}}\par\vspace{\vdiss cm}
		\subfloat {\includegraphics[width=\imw cm, height=\imh cm]{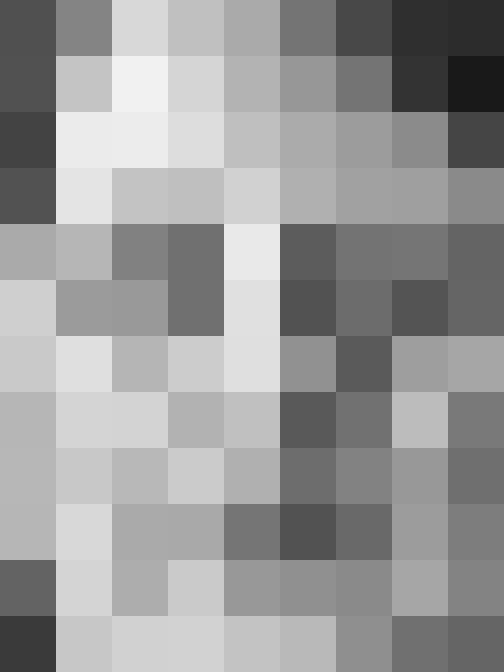}}\par\vspace{\vdiss cm}
		\subfloat {\includegraphics[width=\imw cm, height=\imh cm]{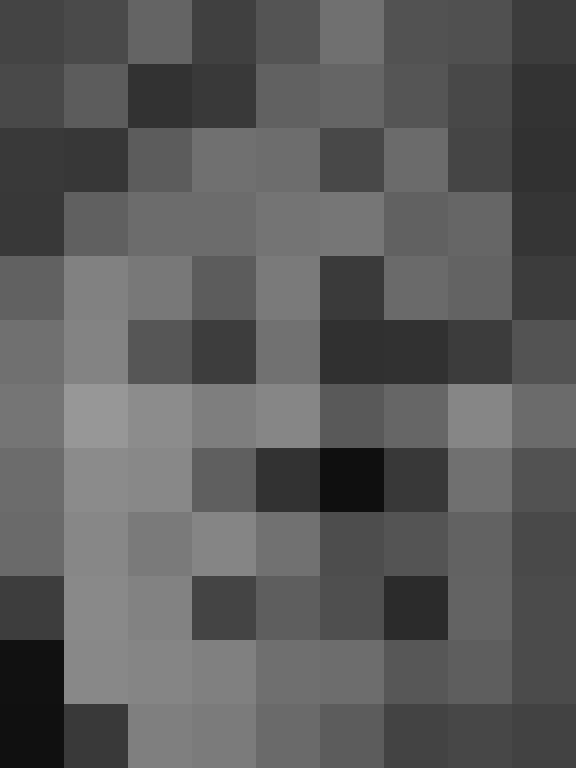}}\par\vspace{\vdisb cm}
		\subfloat {\includegraphics[width=\imw cm, height=\imh cm]{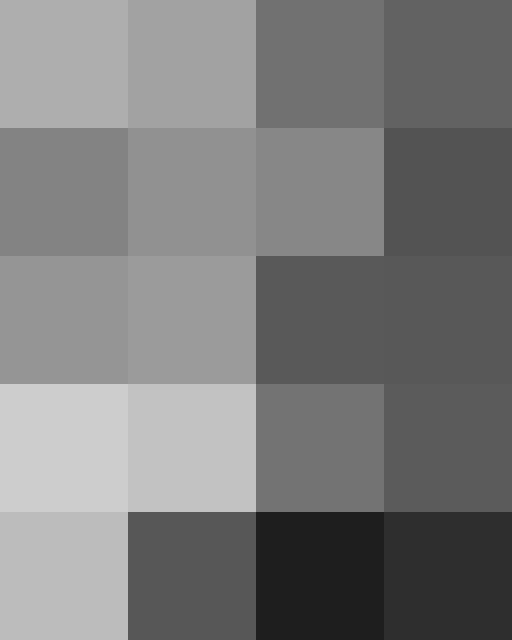}}\par\vspace{\vdiss cm}
		\subfloat {\includegraphics[width=\imw cm, height=\imh cm]{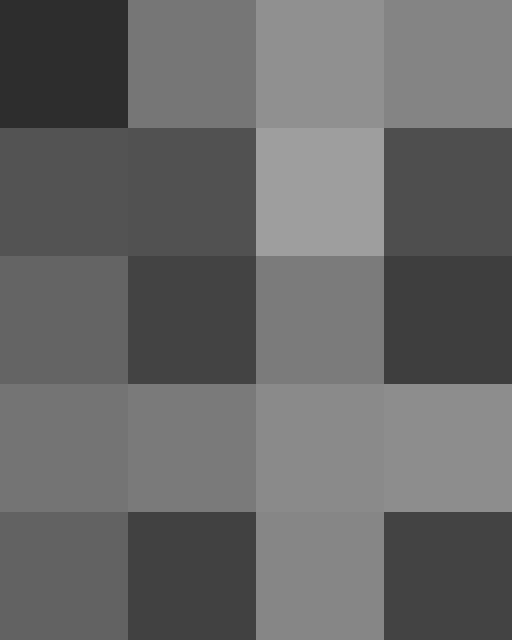}}\par\vspace{\vdiss cm}
		\stepcounter{figure}\addtocounter{figure}{-1}
		\addtocounter{subfigure}{0}
		\subfloat[]{\includegraphics[width=\imw cm, height=\imh cm]{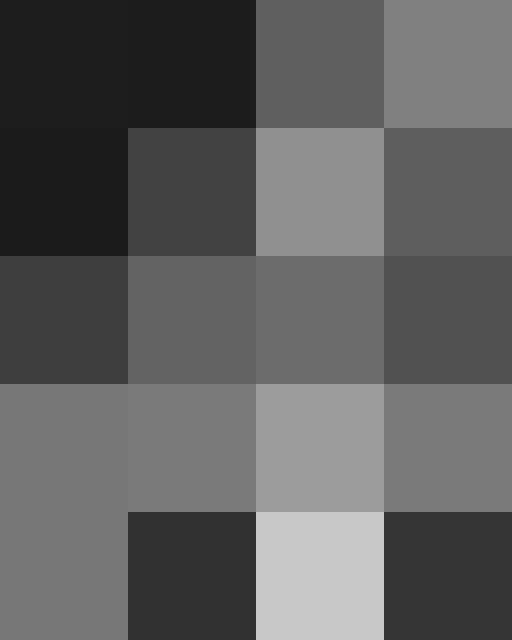}}
	\end{minipage}%
	\begin{minipage}{\hdis\textwidth} 
		\centering
		\subfloat {\includegraphics[width=\imw cm, height=\imh cm]{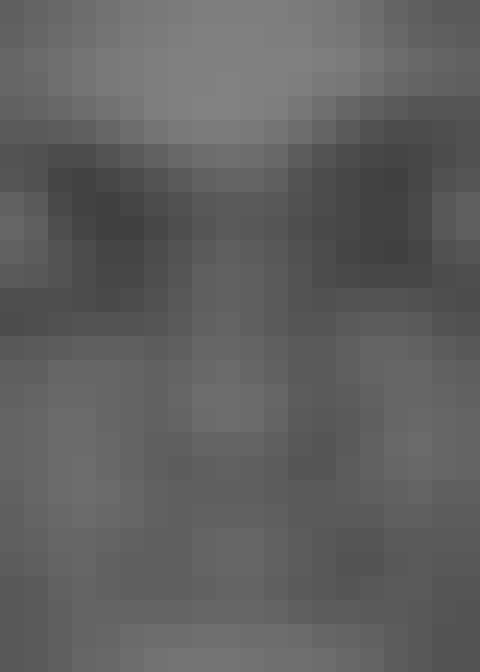}}\par\vspace{\vdiss cm}		
		\subfloat {\includegraphics[width=\imw cm, height=\imh cm]{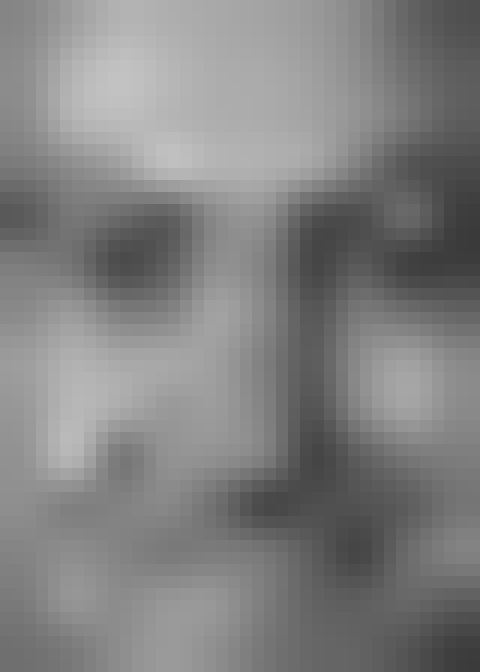}}\par\vspace{\vdiss cm}
		\subfloat {\includegraphics[width=\imw cm, height=\imh cm]{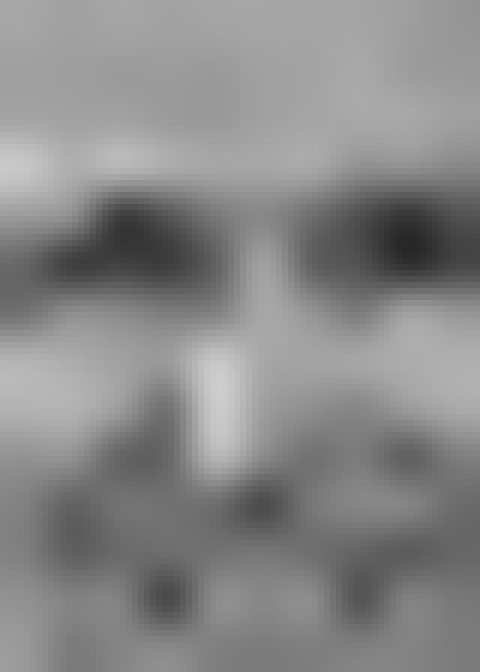}}\par\vspace{\vdisb cm}
		\subfloat {\includegraphics[width=\imw cm, height=\imh cm]{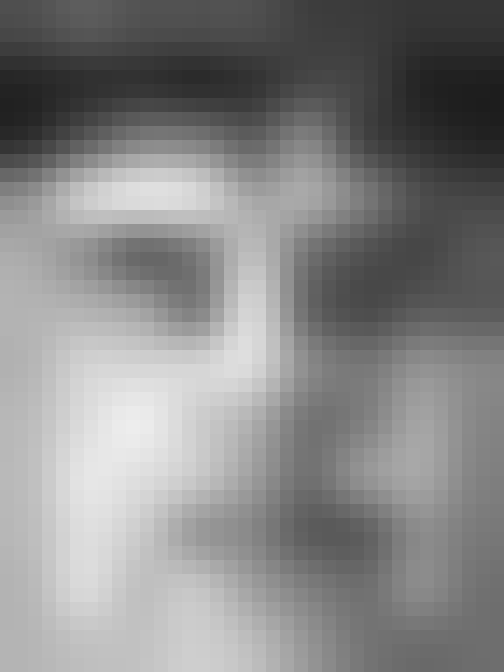}}\par\vspace{\vdiss cm}
		\subfloat {\includegraphics[width=\imw cm, height=\imh cm]{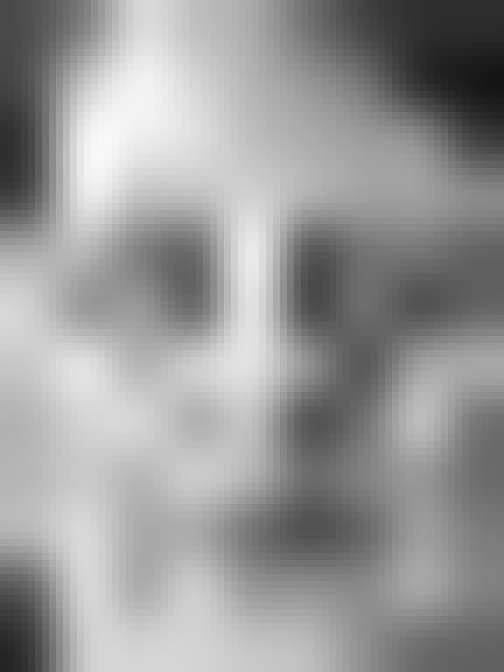}}\par\vspace{\vdiss cm}
		\subfloat {\includegraphics[width=\imw cm, height=\imh cm]{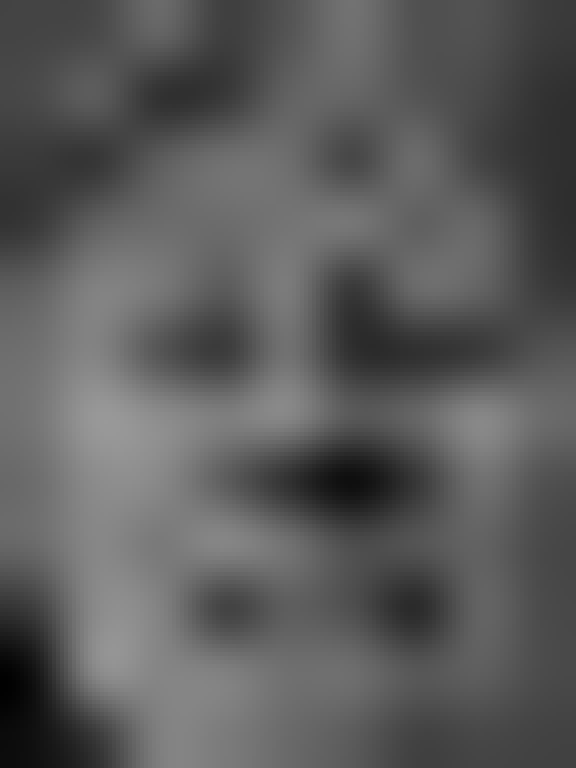}}\par\vspace{\vdisb cm}
		\subfloat {\includegraphics[width=\imw cm, height=\imh cm]{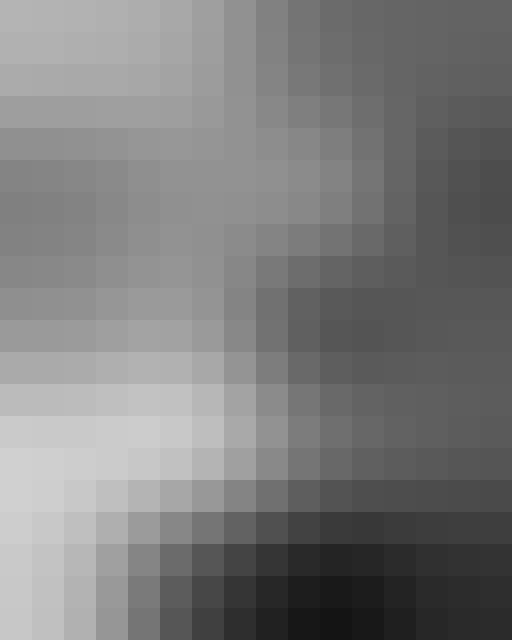}}\par\vspace{\vdiss cm}
		\subfloat {\includegraphics[width=\imw cm, height=\imh cm]{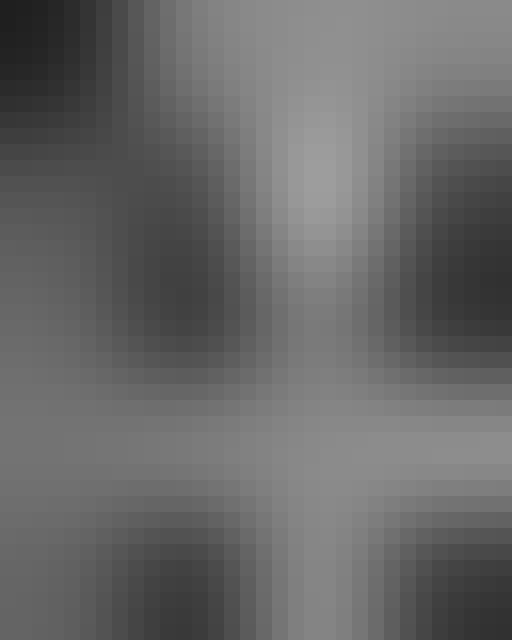}}\par\vspace{\vdiss cm}
		\stepcounter{figure}\addtocounter{figure}{-1}
		\addtocounter{subfigure}{1}
		\subfloat[]{\includegraphics[width=\imw cm, height=\imh cm]{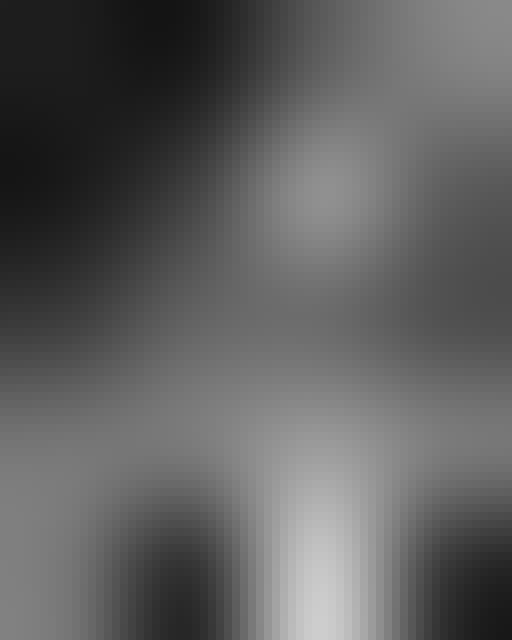}}
	\end{minipage}%
	\begin{minipage}{\hdis\textwidth} 
		\centering
		\subfloat {\includegraphics[width=\imw cm, height=\imh cm]{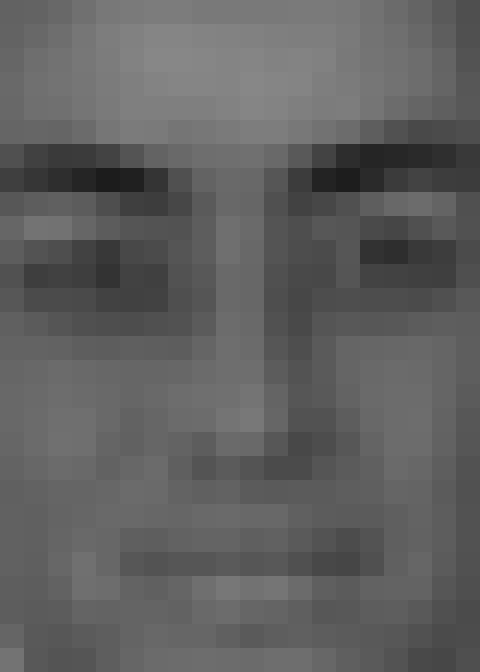}}\par\vspace{\vdiss cm}		
		\subfloat {\includegraphics[width=\imw cm, height=\imh cm]{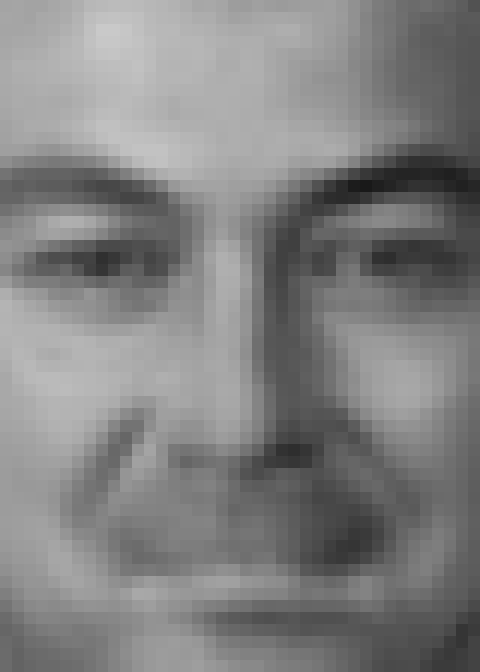}}\par\vspace{\vdiss cm}
		\subfloat {\includegraphics[width=\imw cm, height=\imh cm]{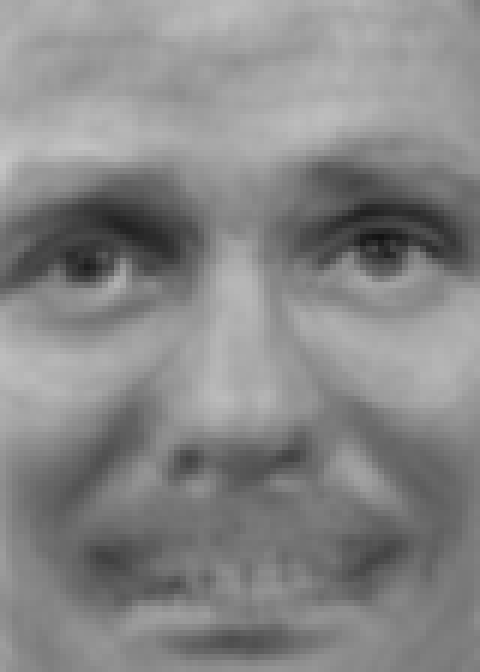}}\par\vspace{\vdisb cm}
		\subfloat {\includegraphics[width=\imw cm, height=\imh cm]{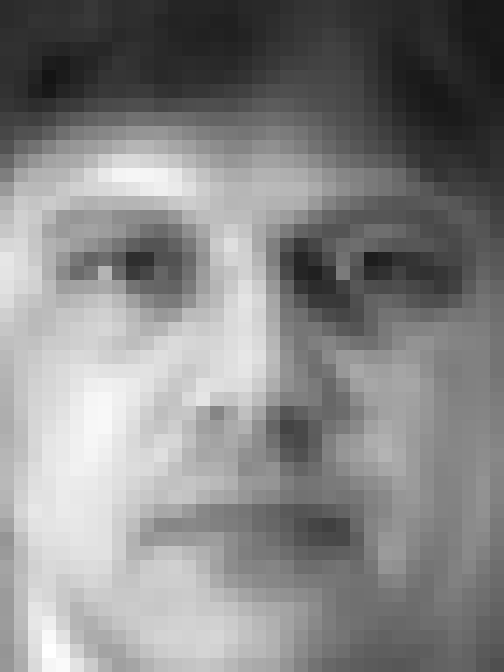}}\par\vspace{\vdiss cm}
		\subfloat {\includegraphics[width=\imw cm, height=\imh cm]{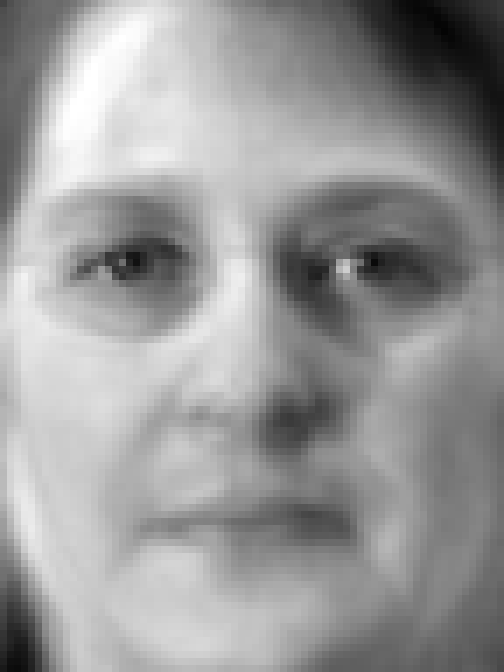}}\par\vspace{\vdiss cm}
		\subfloat {\includegraphics[width=\imw cm, height=\imh cm]{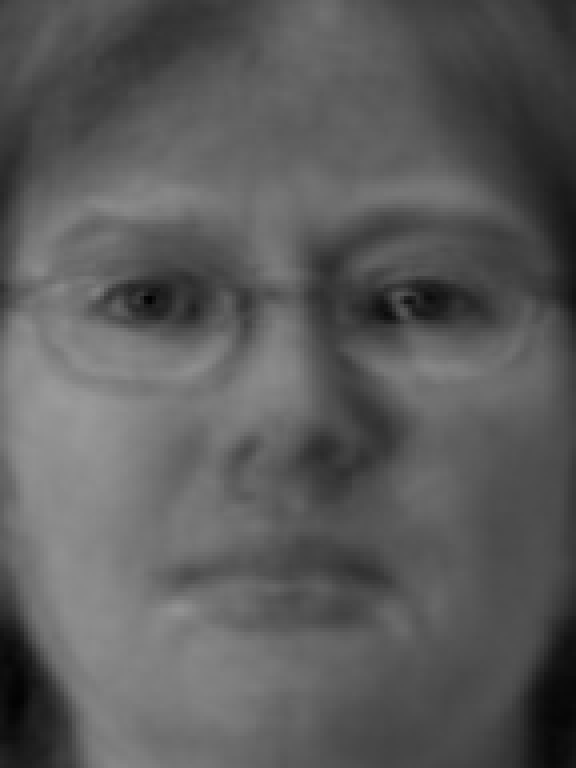}}\par\vspace{\vdisb cm}
		\subfloat {\includegraphics[width=\imw cm, height=\imh cm]{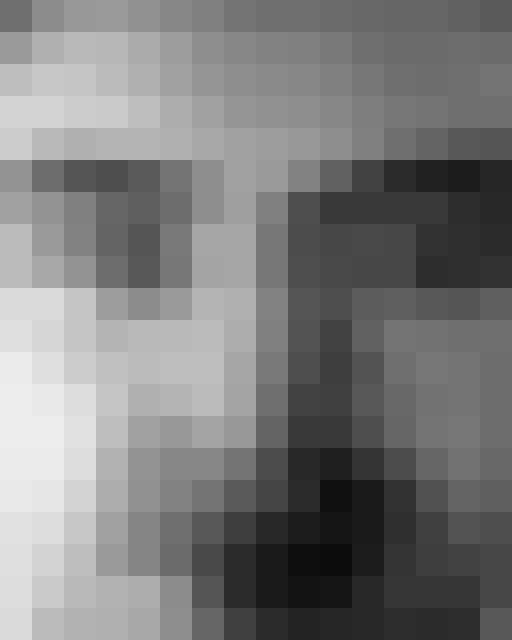}}\par\vspace{\vdiss cm}
		\subfloat {\includegraphics[width=\imw cm, height=\imh cm]{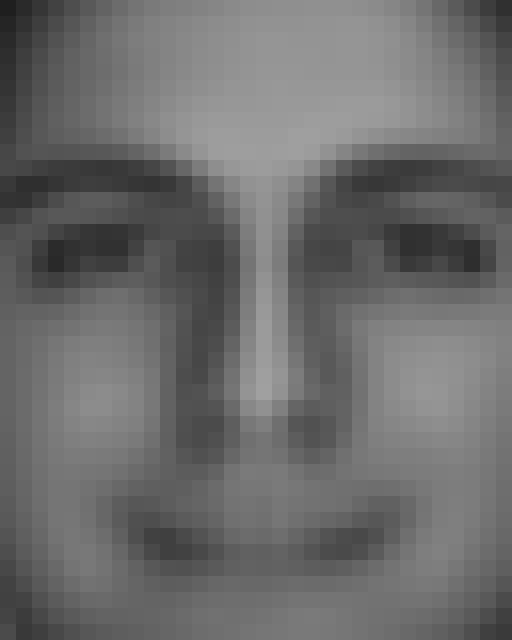}}\par\vspace{\vdiss cm}
		\stepcounter{figure}\addtocounter{figure}{-1}
		\addtocounter{subfigure}{2}
		\subfloat[]{\includegraphics[width=\imw cm, height=\imh cm]{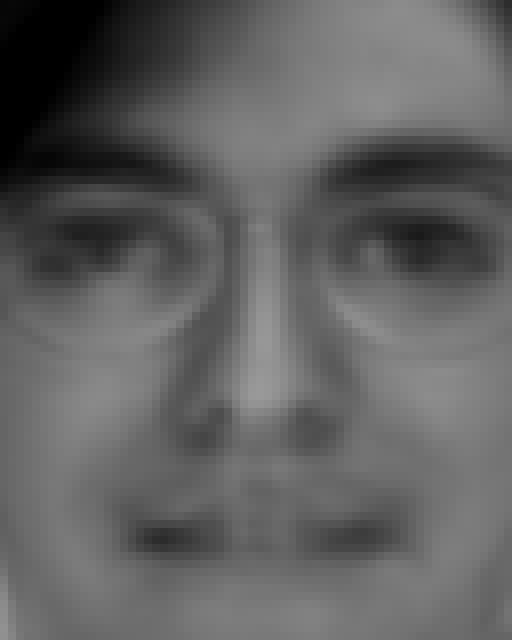}}
	\end{minipage}%
	\begin{minipage}{\hdis\textwidth} 
		\centering
		\subfloat {\includegraphics[width=\imw cm, height=\imh cm]{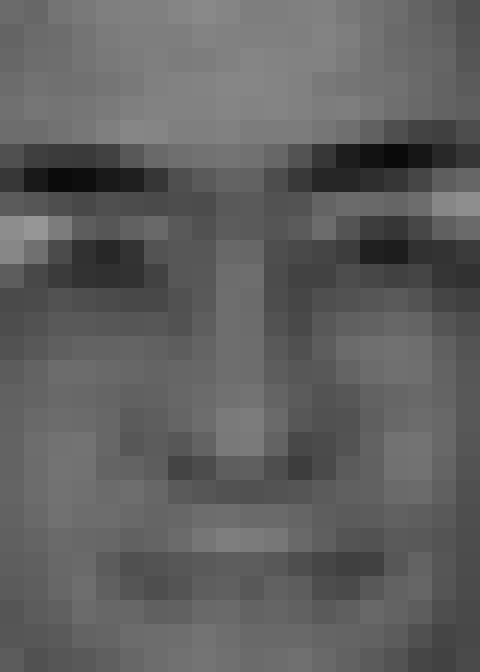}}\par\vspace{\vdiss cm}		
		\subfloat {\includegraphics[width=\imw cm, height=\imh cm]{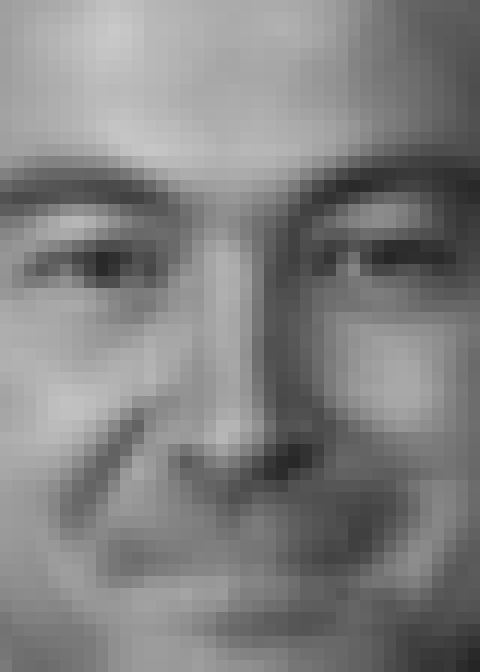}}\par\vspace{\vdiss cm}
		\subfloat {\includegraphics[width=\imw cm, height=\imh cm]{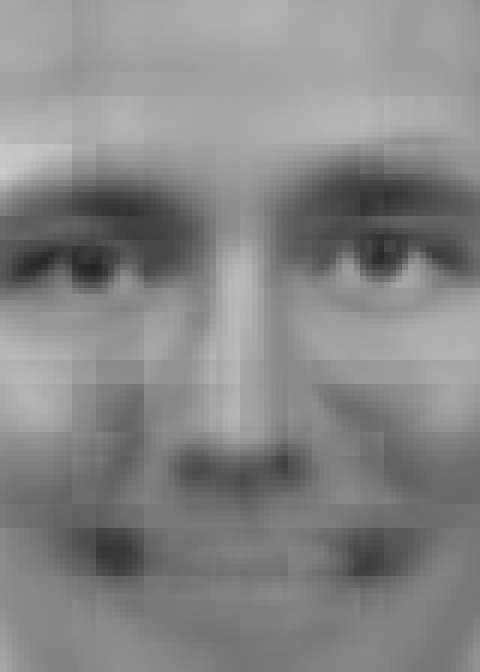}}\par\vspace{\vdisb cm}
		\subfloat {\includegraphics[width=\imw cm, height=\imh cm]{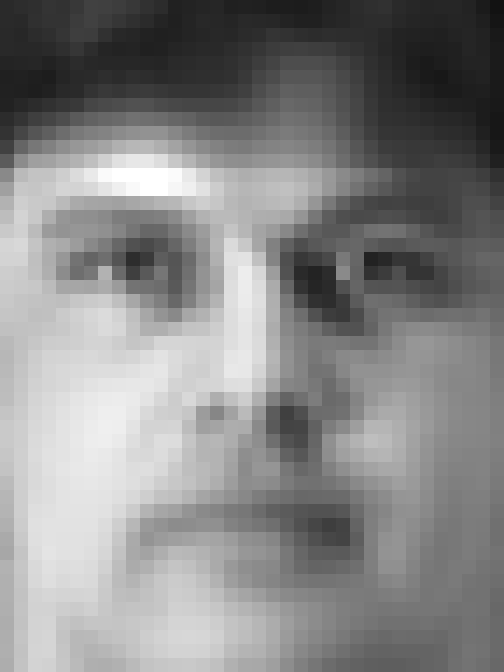}}\par\vspace{\vdiss cm}
		\subfloat {\includegraphics[width=\imw cm, height=\imh cm]{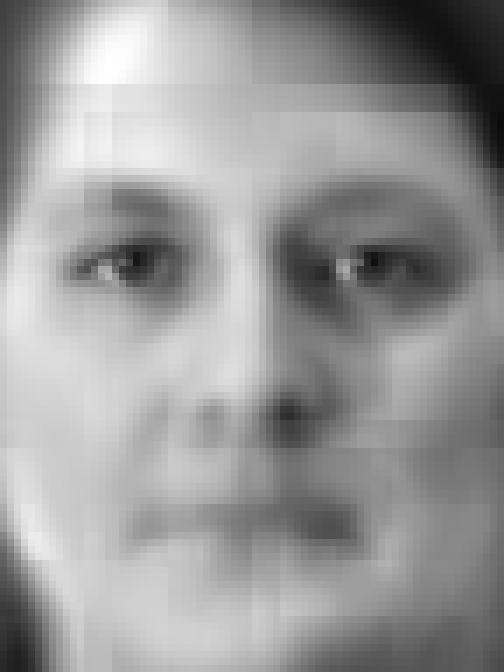}}\par\vspace{\vdiss cm}
		\subfloat {\includegraphics[width=\imw cm, height=\imh cm]{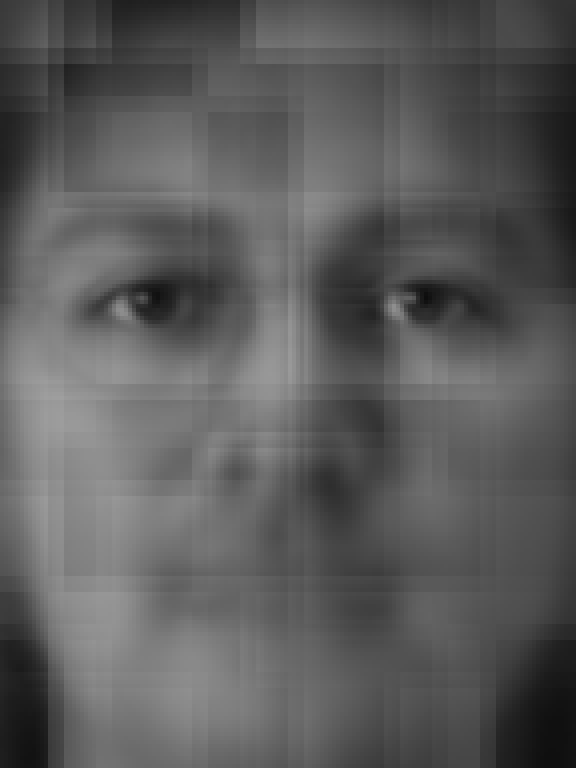}}\par\vspace{\vdisb cm}
		\subfloat {\includegraphics[width=\imw cm, height=\imh cm]{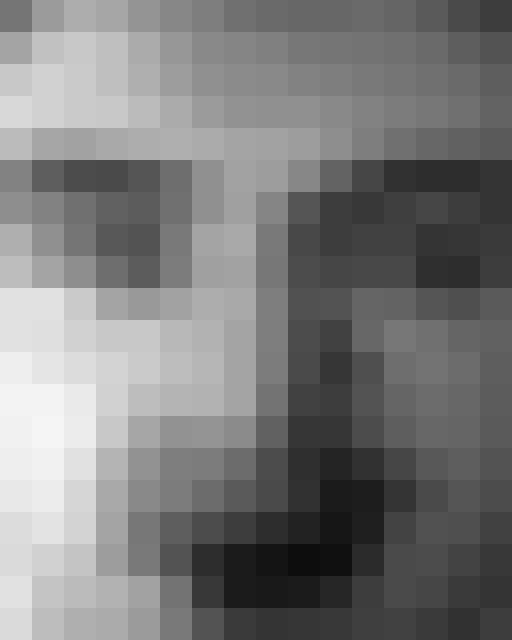}}\par\vspace{\vdiss cm}
		\subfloat {\includegraphics[width=\imw cm, height=\imh cm]{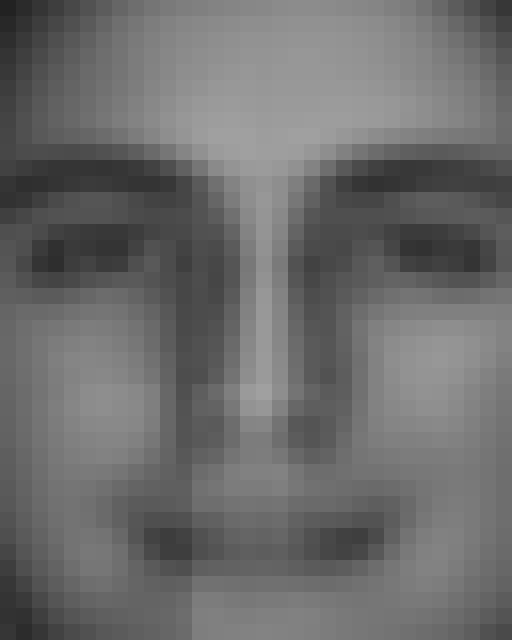}}\par\vspace{\vdiss cm}
		\stepcounter{figure}\addtocounter{figure}{-1}
		\addtocounter{subfigure}{3}
		\subfloat[]{\includegraphics[width=\imw cm, height=\imh cm]{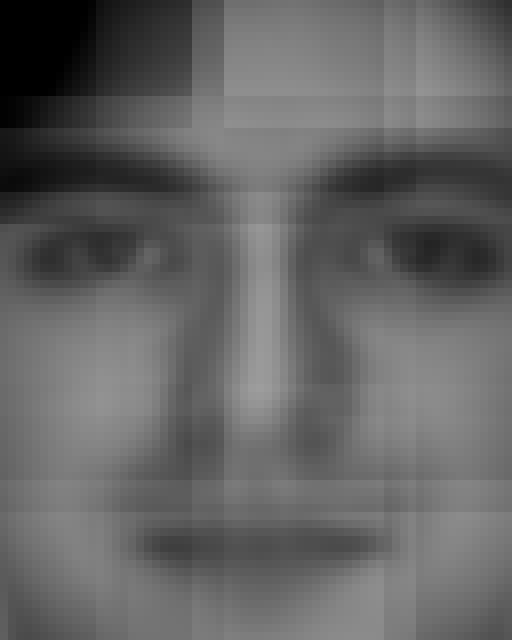}}
	\end{minipage}%
	\begin{minipage}{\hdis\textwidth} 
		\centering
		\subfloat {\includegraphics[width=\imw cm, height=\imh cm]{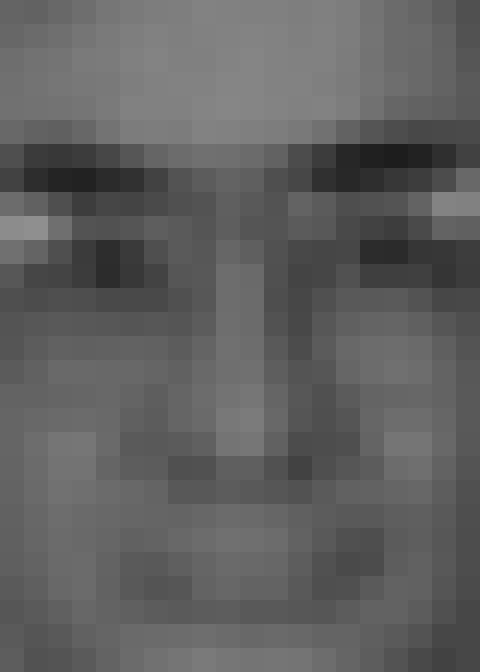}}\par\vspace{\vdiss cm}		
		\subfloat {\includegraphics[width=\imw cm, height=\imh cm]{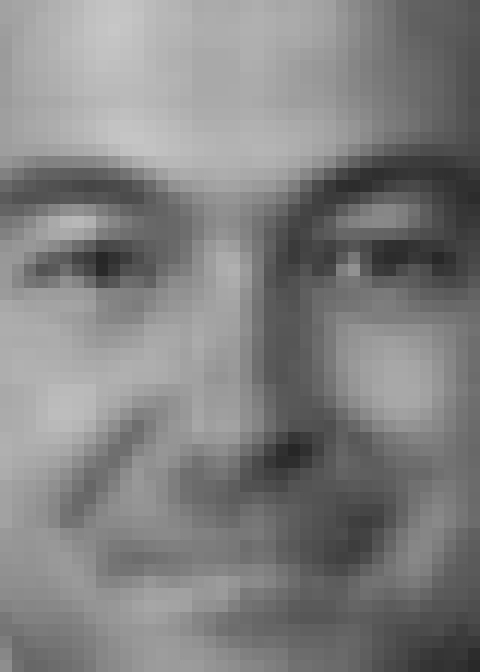}}\par\vspace{\vdiss cm}
		\subfloat {\includegraphics[width=\imw cm, height=\imh cm]{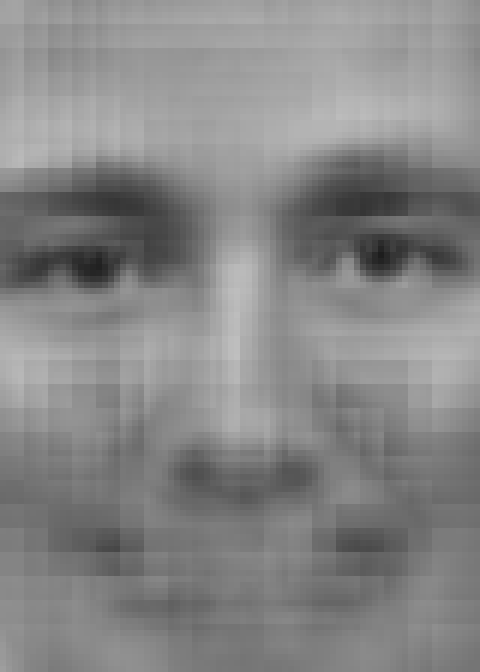}}\par\vspace{\vdisb cm}
		\subfloat {\includegraphics[width=\imw cm, height=\imh cm]{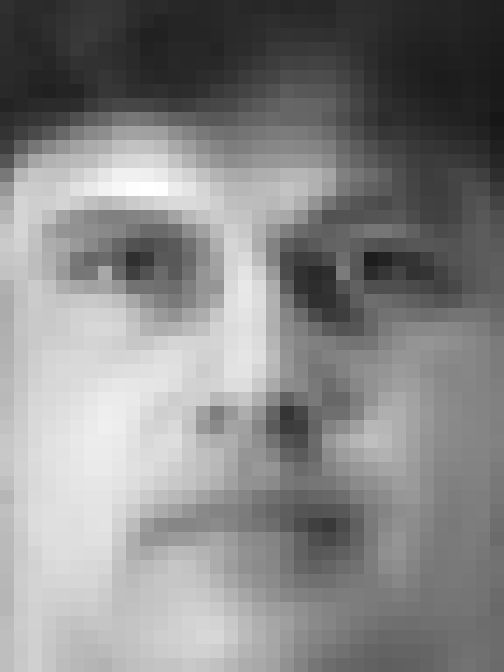}}\par\vspace{\vdiss cm}
		\subfloat {\includegraphics[width=\imw cm, height=\imh cm]{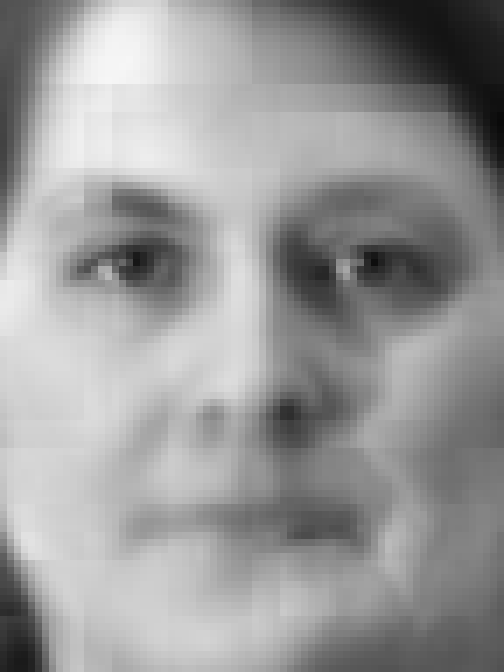}}\par\vspace{\vdiss cm}
		\subfloat {\includegraphics[width=\imw cm, height=\imh cm]{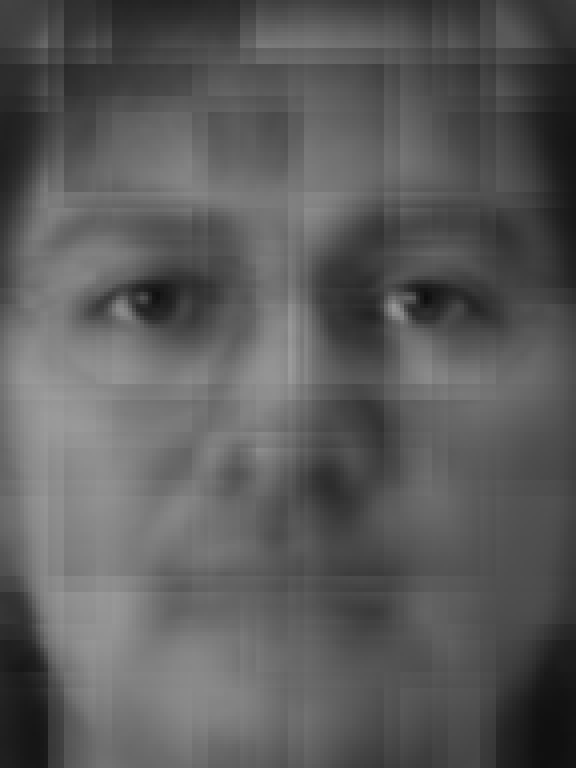}}\par\vspace{\vdisb cm}
		\subfloat {\includegraphics[width=\imw cm, height=\imh cm]{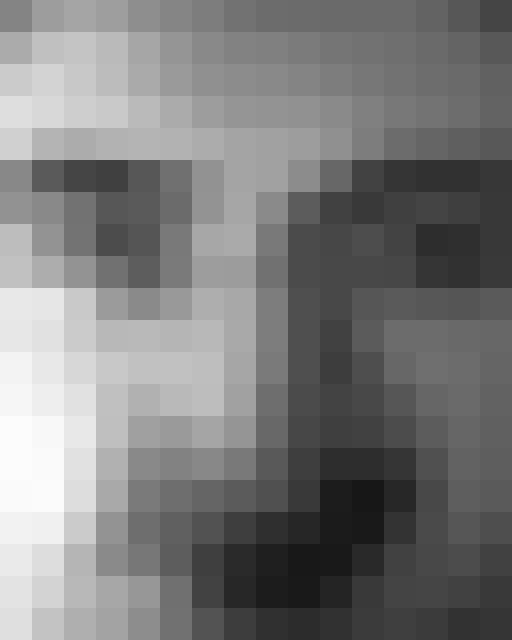}}\par\vspace{\vdiss cm}
		\subfloat {\includegraphics[width=\imw cm, height=\imh cm]{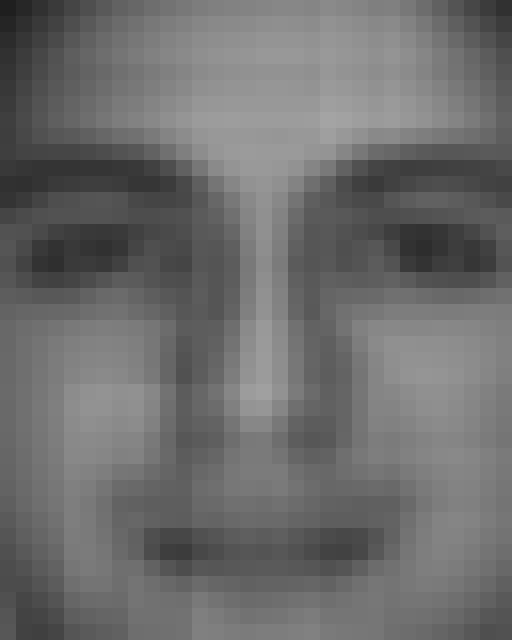}}\par\vspace{\vdiss cm}
		\stepcounter{figure}\addtocounter{figure}{-1}
		\addtocounter{subfigure}{4}
		\subfloat[]{\includegraphics[width=\imw cm, height=\imh cm]{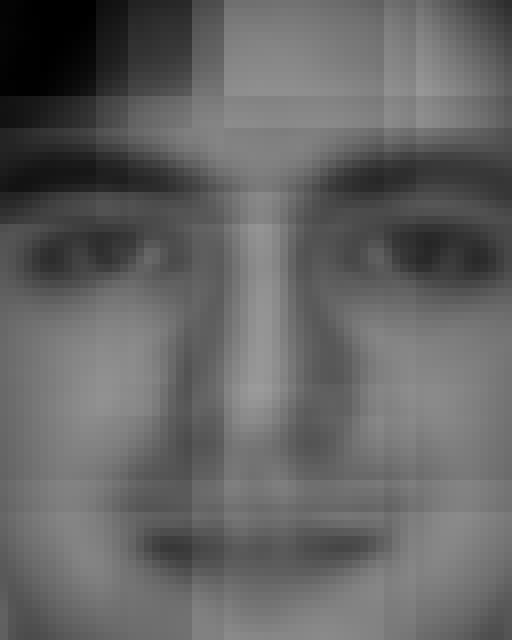}}
	\end{minipage}%
	\begin{minipage}{\hdis\textwidth} 
		\centering
		\subfloat {\includegraphics[width=\imw cm, height=\imh cm]{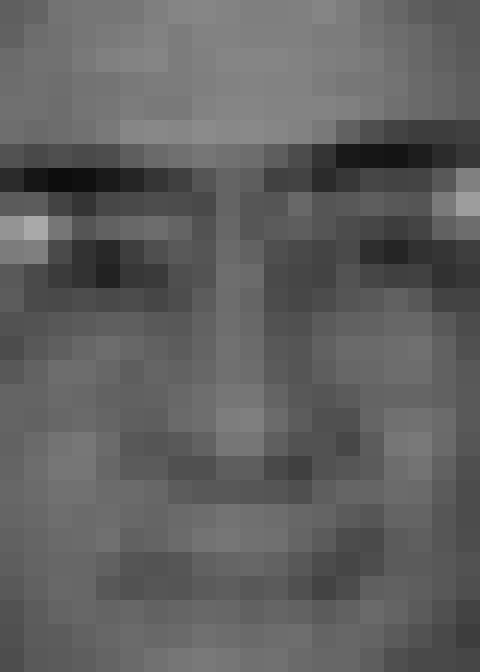}}\par\vspace{\vdiss cm}		
		\subfloat {\includegraphics[width=\imw cm, height=\imh cm]{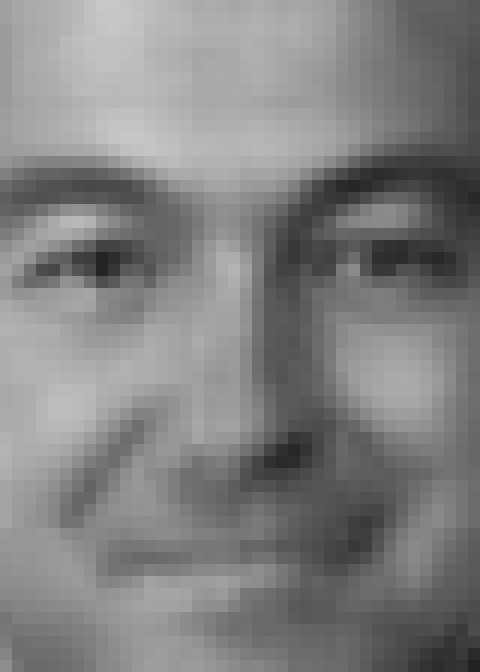}}\par\vspace{\vdiss cm}
		\subfloat {\includegraphics[width=\imw cm, height=\imh cm]{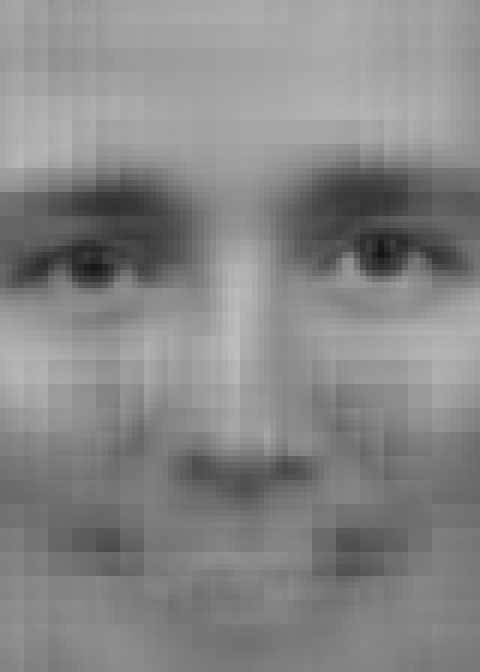}}\par\vspace{\vdisb cm}
		\subfloat {\includegraphics[width=\imw cm, height=\imh cm]{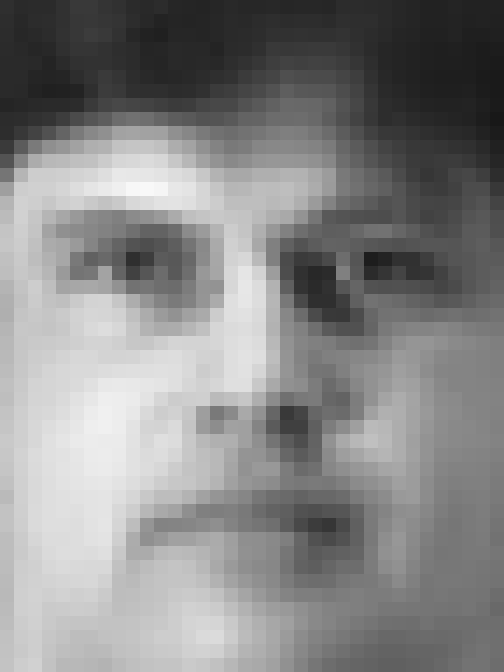}}\par\vspace{\vdiss cm}
		\subfloat {\includegraphics[width=\imw cm, height=\imh cm]{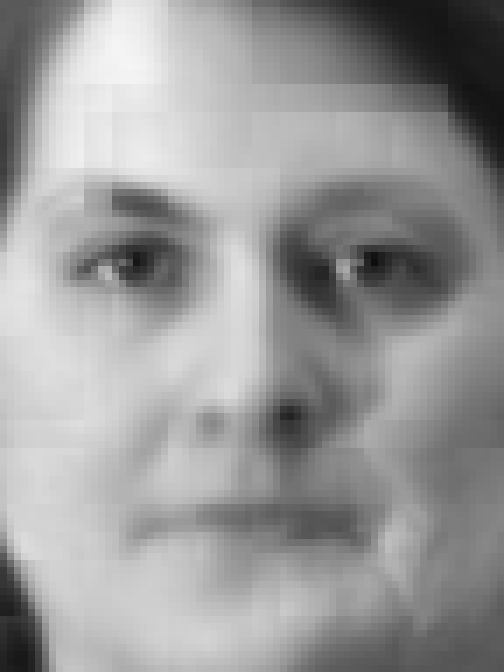}}\par\vspace{\vdiss cm}
		\subfloat {\includegraphics[width=\imw cm, height=\imh cm]{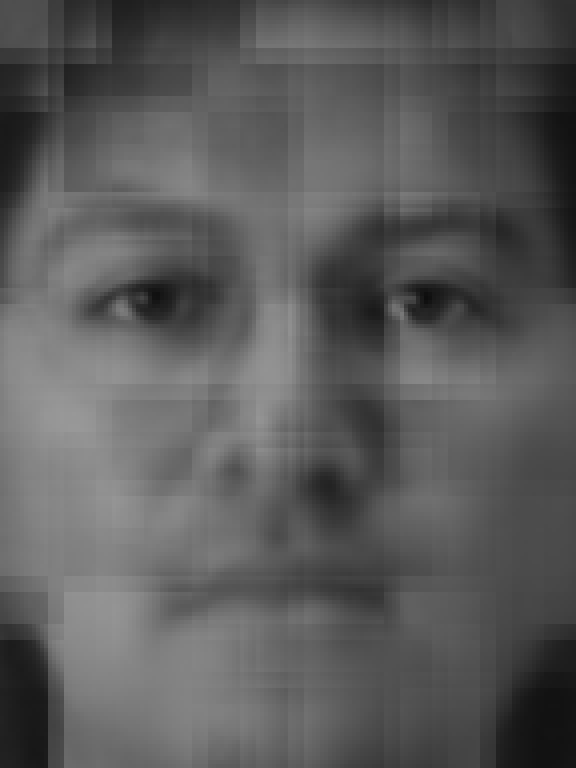}}\par\vspace{\vdisb cm}
		\subfloat {\includegraphics[width=\imw cm, height=\imh cm]{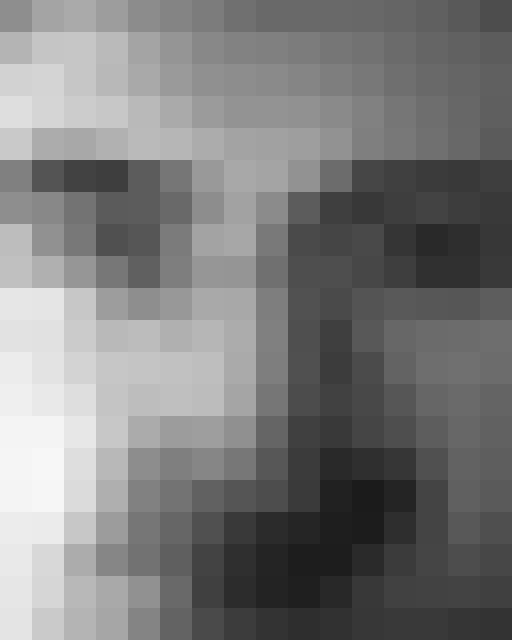}}\par\vspace{\vdiss cm}
		\subfloat {\includegraphics[width=\imw cm, height=\imh cm]{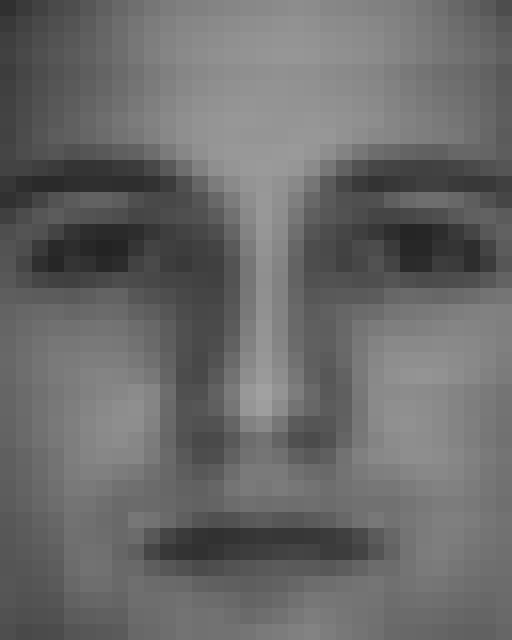}}\par\vspace{\vdiss cm}
		\stepcounter{figure}\addtocounter{figure}{-1}
		\addtocounter{subfigure}{5}
		\subfloat[]{\includegraphics[width=\imw cm, height=\imh cm]{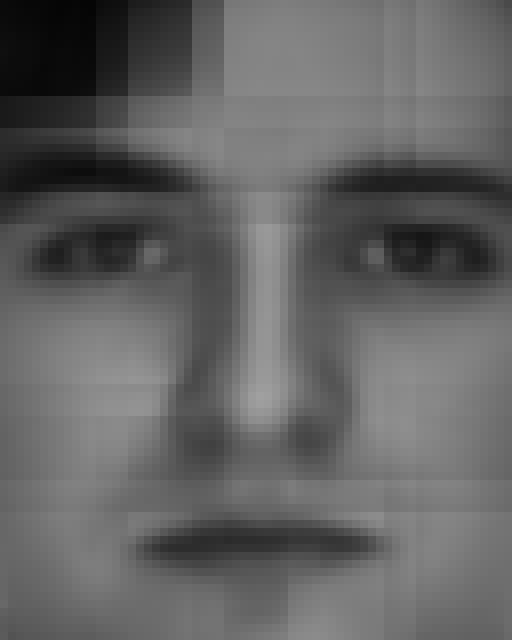}}
	\end{minipage}%
	\begin{minipage}{\hdis\textwidth} 
		\centering
		\subfloat {\includegraphics[width=\imw cm, height=\imh cm]{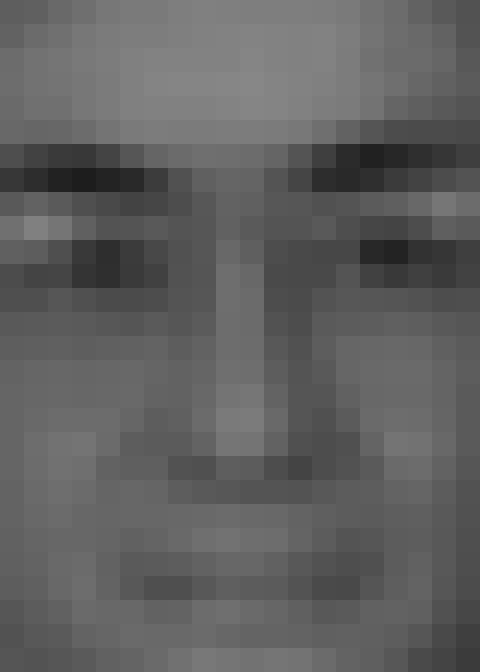}}\par\vspace{\vdiss cm}		
		\subfloat {\includegraphics[width=\imw cm, height=\imh cm]{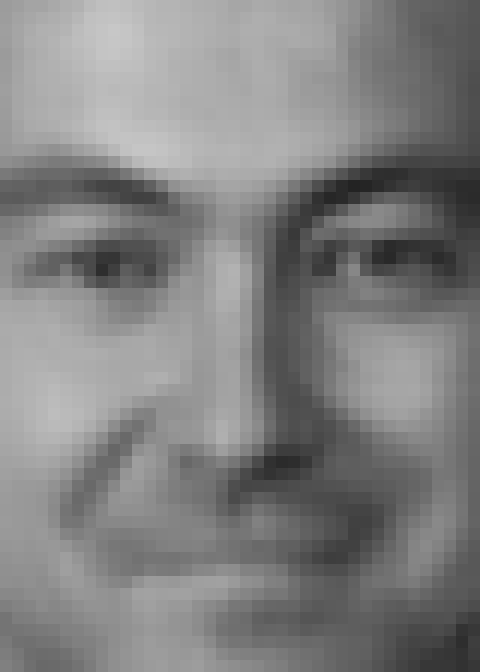}}\par\vspace{\vdiss cm}
		\subfloat {\includegraphics[width=\imw cm, height=\imh cm]{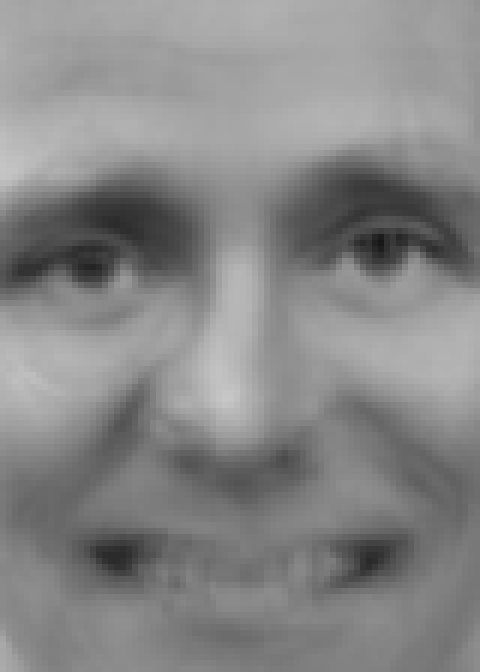}}\par\vspace{\vdisb cm}
		\subfloat {\includegraphics[width=\imw cm, height=\imh cm]{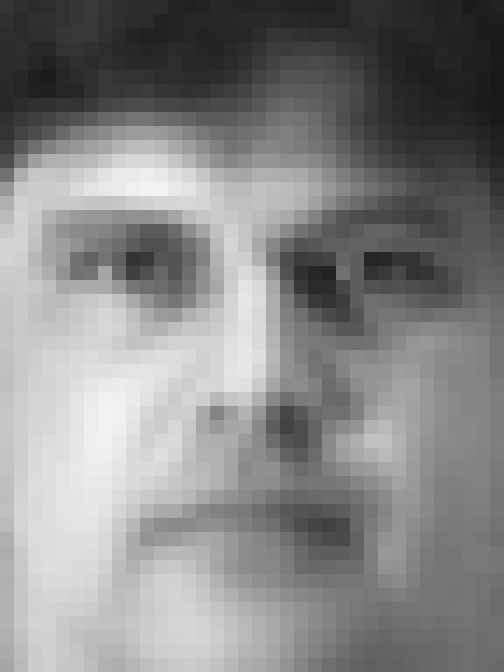}}\par\vspace{\vdiss cm}
		\subfloat {\includegraphics[width=\imw cm, height=\imh cm]{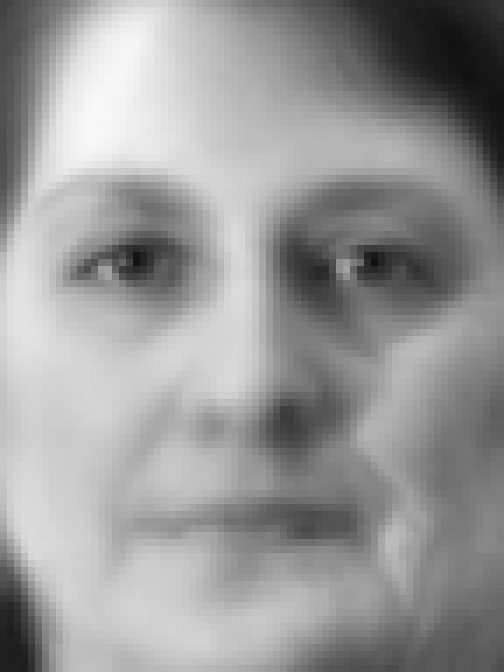}}\par\vspace{\vdiss cm}
		\subfloat {\includegraphics[width=\imw cm, height=\imh cm]{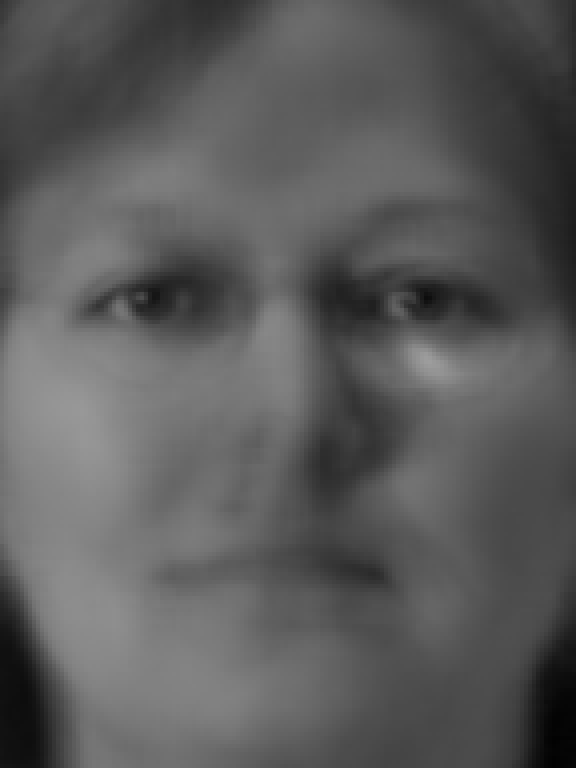}}\par\vspace{\vdisb cm}
		\subfloat {\includegraphics[width=\imw cm, height=\imh cm]{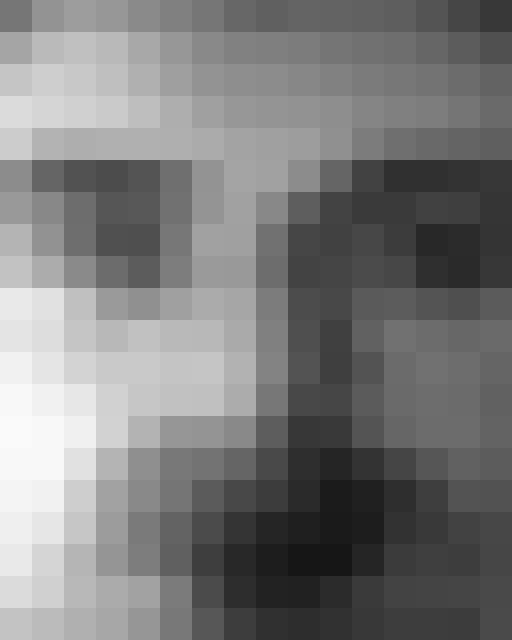}}\par\vspace{\vdiss cm}
		\subfloat {\includegraphics[width=\imw cm, height=\imh cm]{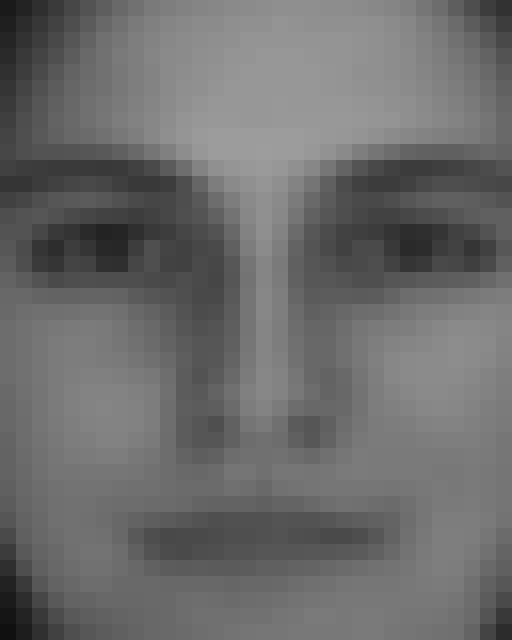}}\par\vspace{\vdiss cm}
		\stepcounter{figure}\addtocounter{figure}{-1}
		\addtocounter{subfigure}{6}
		\subfloat[]{\includegraphics[width=\imw cm, height=\imh cm]{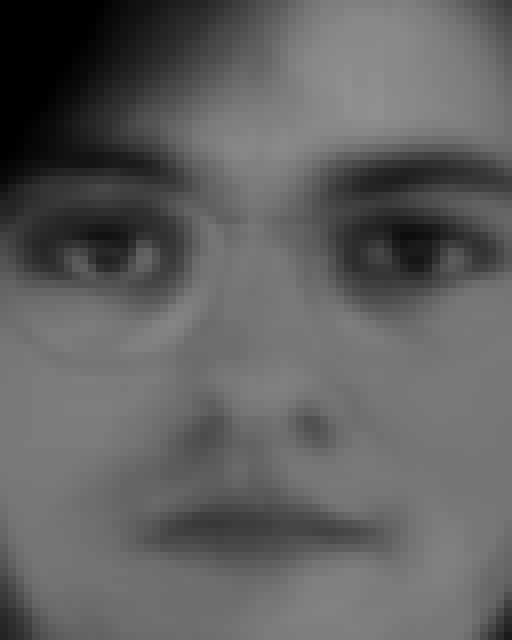}}
	\end{minipage}%
	\begin{minipage}{\hdis\textwidth} 
		\centering
		\subfloat {\includegraphics[width=\imw cm, height=\imh cm]{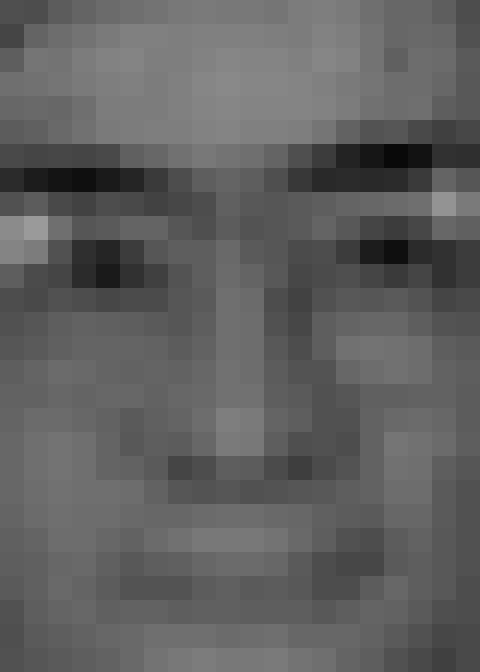}}\par\vspace{\vdiss cm}		
		\subfloat {\includegraphics[width=\imw cm, height=\imh cm]{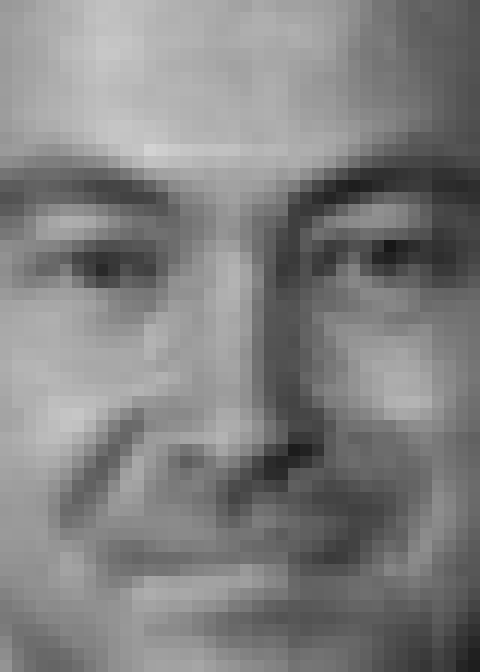}}\par\vspace{\vdiss cm}
		\subfloat {\includegraphics[width=\imw cm, height=\imh cm]{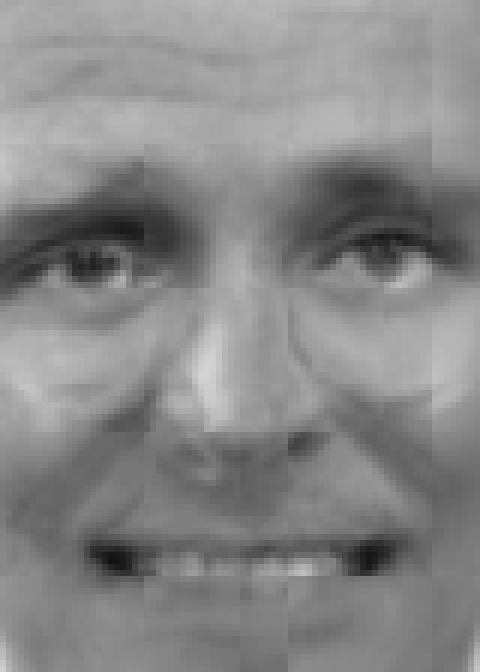}}\par\vspace{\vdisb cm}
		\subfloat {\includegraphics[width=\imw cm, height=\imh cm]{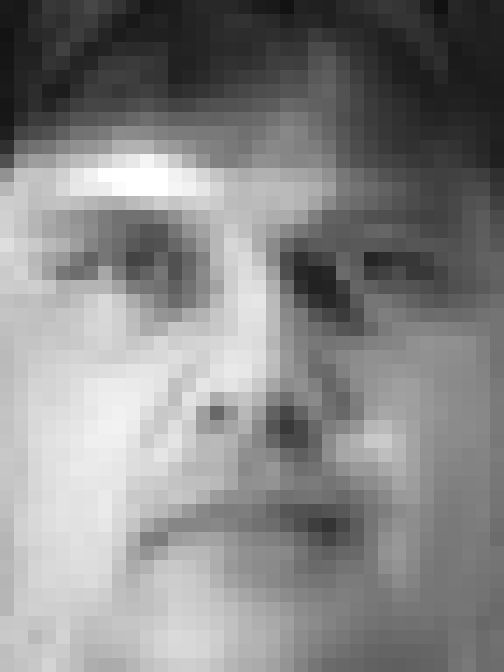}}\par\vspace{\vdiss cm}
		\subfloat {\includegraphics[width=\imw cm, height=\imh cm]{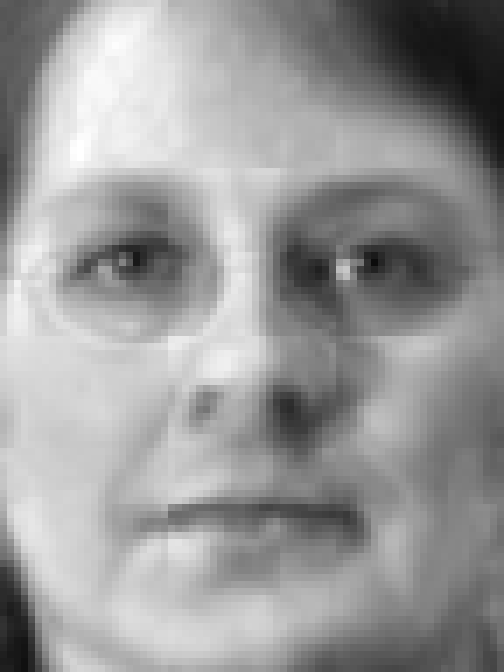}}\par\vspace{\vdiss cm}
		\subfloat {\includegraphics[width=\imw cm, height=\imh cm]{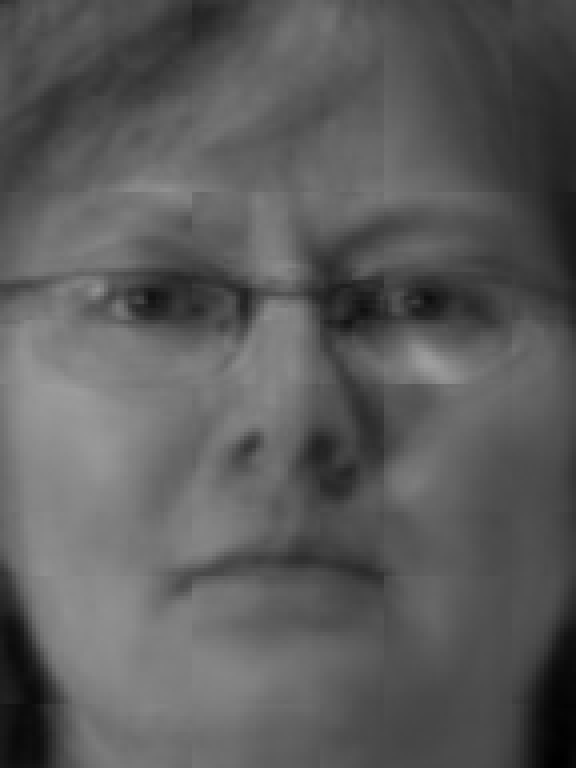}}\par\vspace{\vdisb cm}
		\subfloat {\includegraphics[width=\imw cm, height=\imh cm]{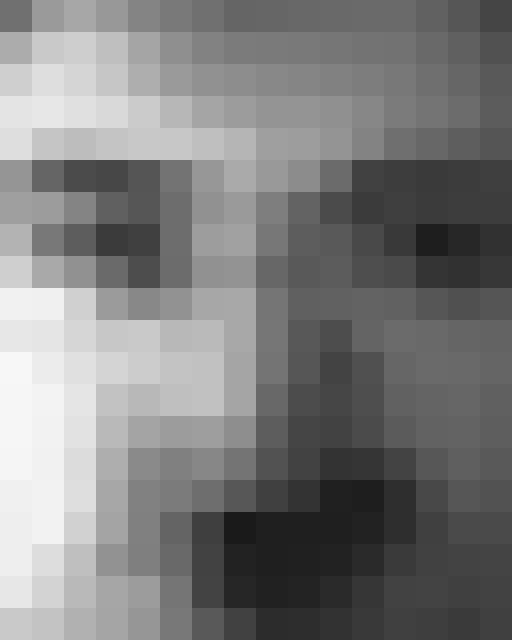}}\par\vspace{\vdiss cm}
		\subfloat {\includegraphics[width=\imw cm, height=\imh cm]{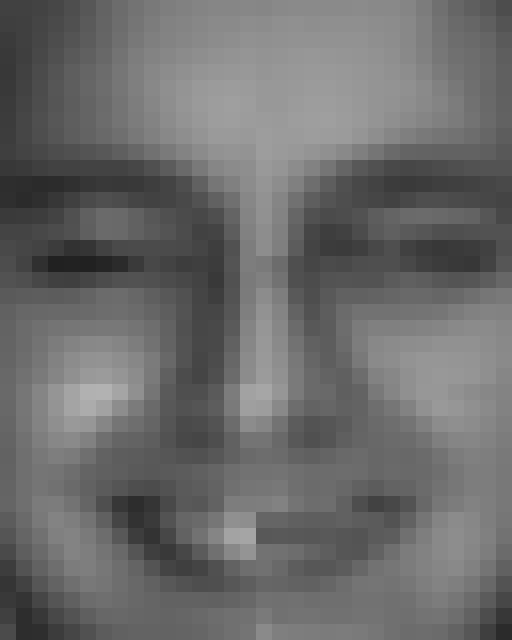}}\par\vspace{\vdiss cm}
		\stepcounter{figure}\addtocounter{figure}{-1}
		\addtocounter{subfigure}{7}
		\subfloat[]{\includegraphics[width=\imw cm, height=\imh cm]{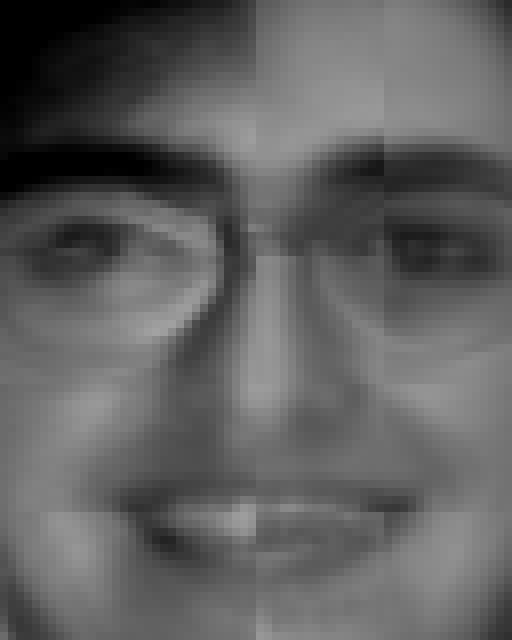}}
	\end{minipage}%
	\begin{minipage}{\hdis\textwidth} 
		\centering		
		\subfloat {\includegraphics[width=\imw cm, height=\imh cm]{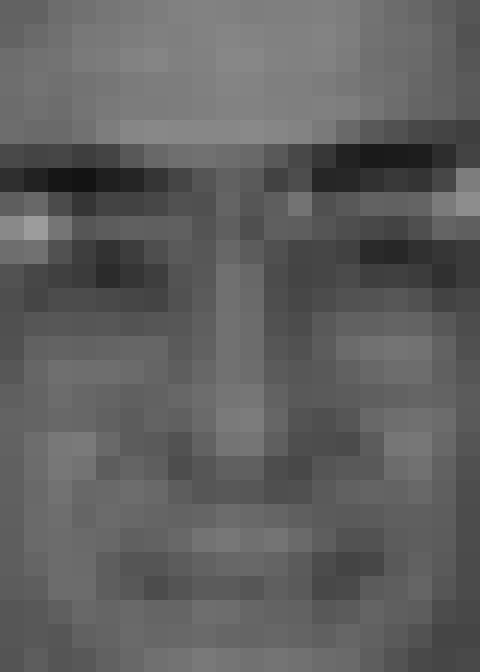}}\par\vspace{\vdiss cm}		
		\subfloat {\includegraphics[width=\imw cm, height=\imh cm]{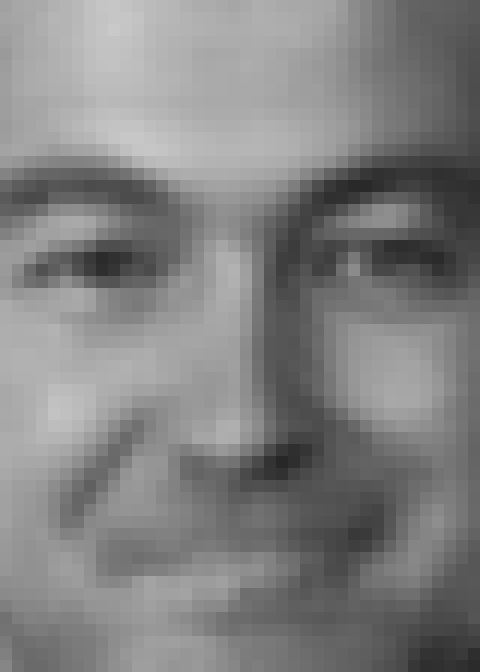}}\par\vspace{\vdiss cm}
		\subfloat {\includegraphics[width=\imw cm, height=\imh cm]{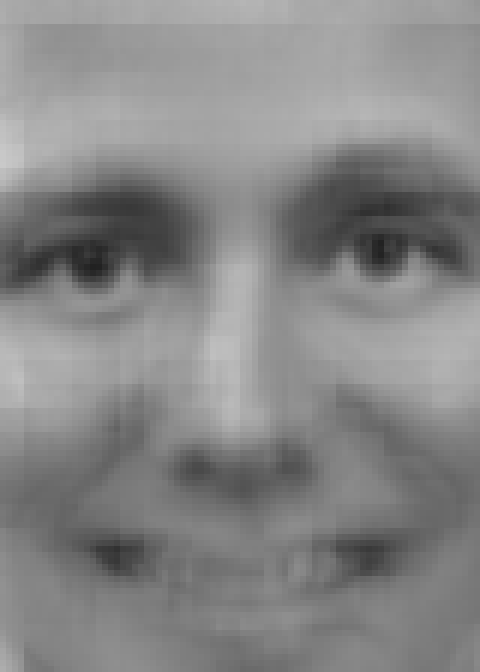}}\par\vspace{\vdisb cm}
		\subfloat {\includegraphics[width=\imw cm, height=\imh cm]{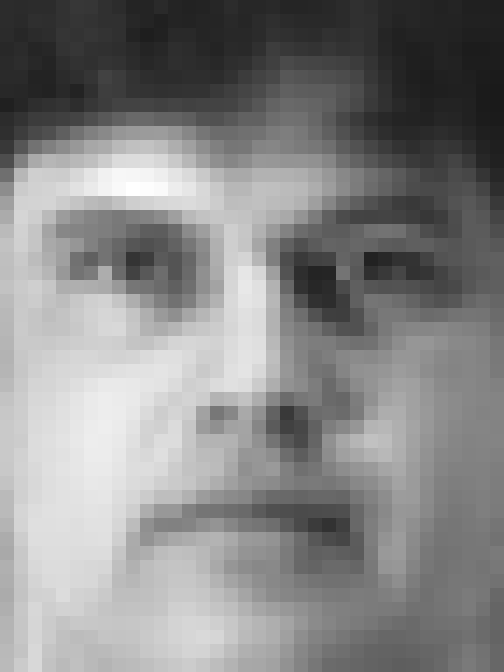}}\par\vspace{\vdiss cm}
		\subfloat {\includegraphics[width=\imw cm, height=\imh cm]{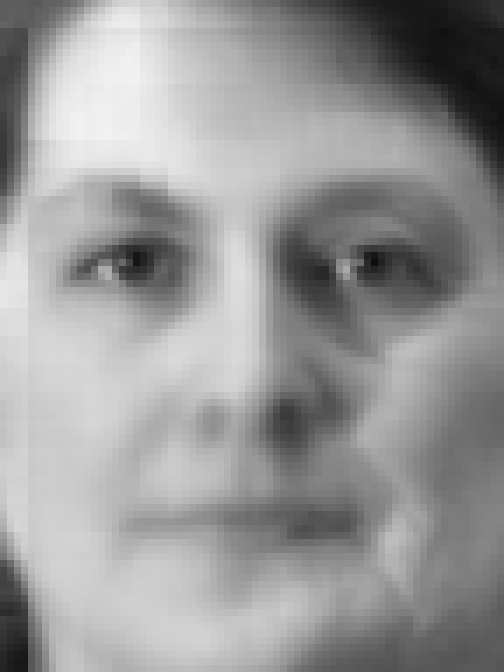}}\par\vspace{\vdiss cm}
		\subfloat {\includegraphics[width=\imw cm, height=\imh cm]{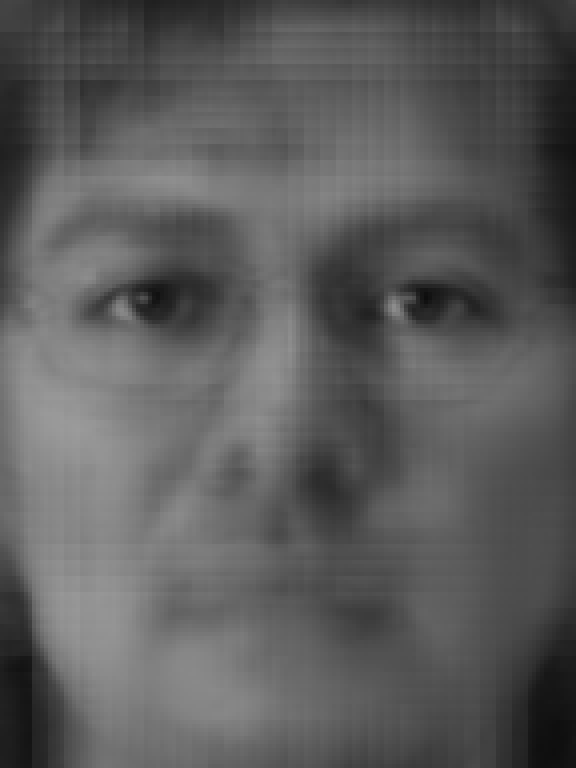}}\par\vspace{\vdisb cm}
		\subfloat {\includegraphics[width=\imw cm, height=\imh cm]{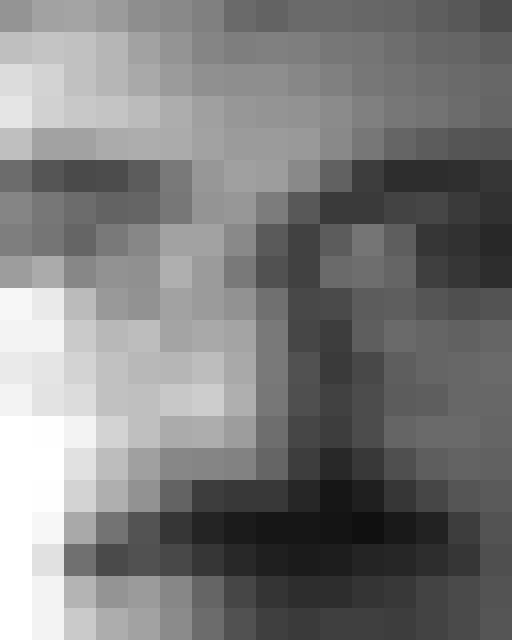}}\par\vspace{\vdiss cm}
		\subfloat {\includegraphics[width=\imw cm, height=\imh cm]{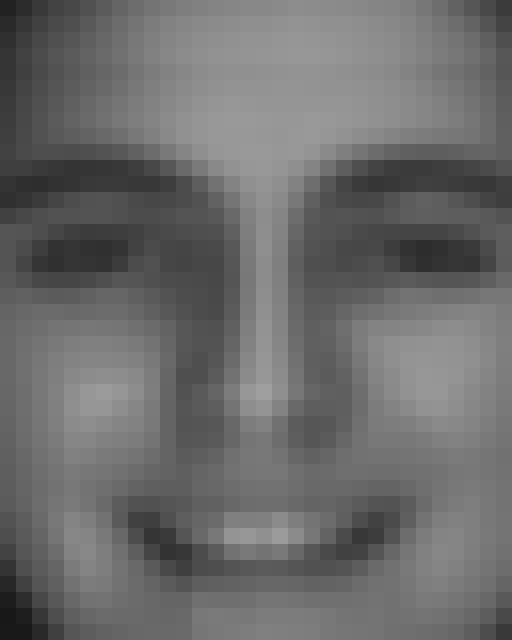}}\par\vspace{\vdiss cm}
		\stepcounter{figure}\addtocounter{figure}{-1}
		\addtocounter{subfigure}{8}
		\subfloat[]{\includegraphics[width=\imw cm, height=\imh cm]{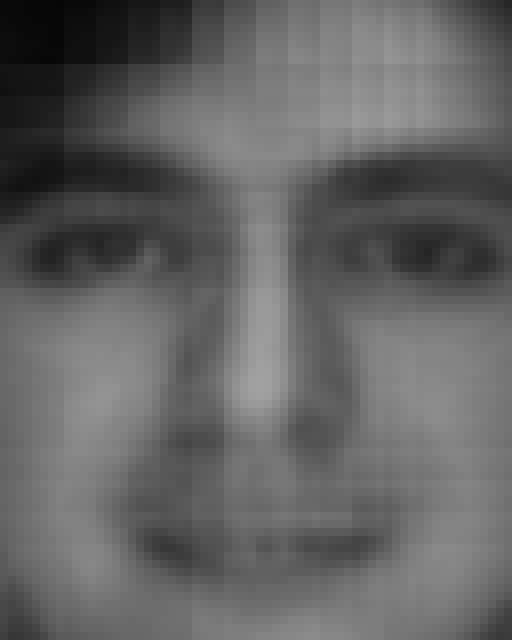}}
	\end{minipage}%
	\begin{minipage}{\hdis\textwidth} 
		\centering
		\subfloat {\includegraphics[width=\imw cm, height=\imh cm]{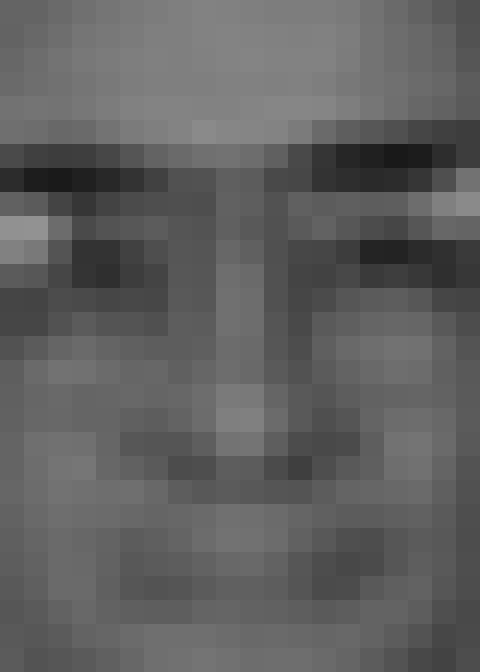}}\par\vspace{\vdiss cm}		
		\subfloat {\includegraphics[width=\imw cm, height=\imh cm]{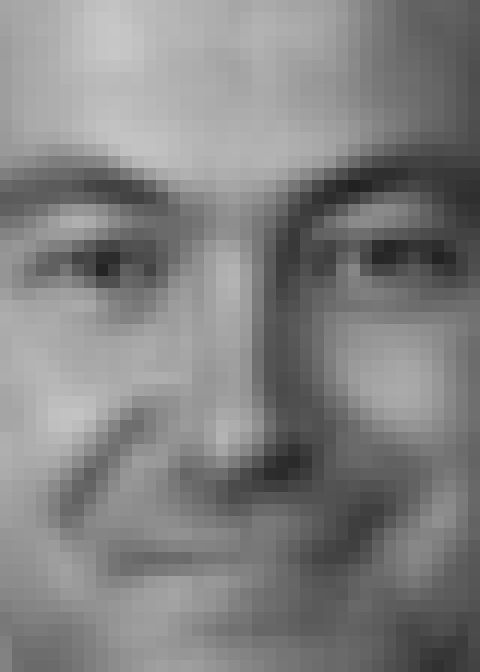}}\par\vspace{\vdiss cm}
		\subfloat {\includegraphics[width=\imw cm, height=\imh cm]{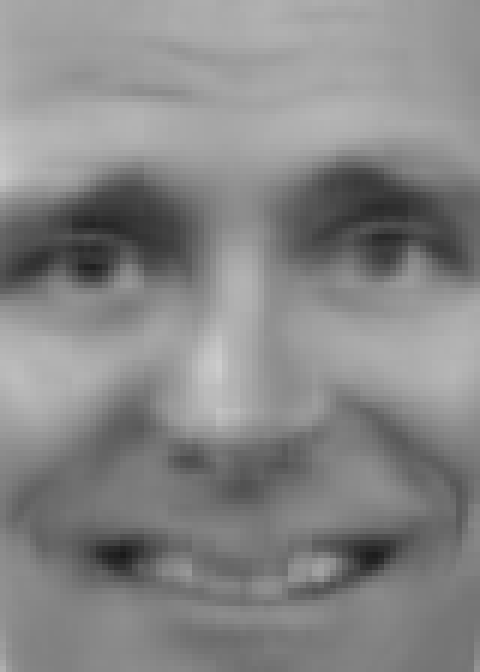}}\par\vspace{\vdisb cm}
		\subfloat {\includegraphics[width=\imw cm, height=\imh cm]{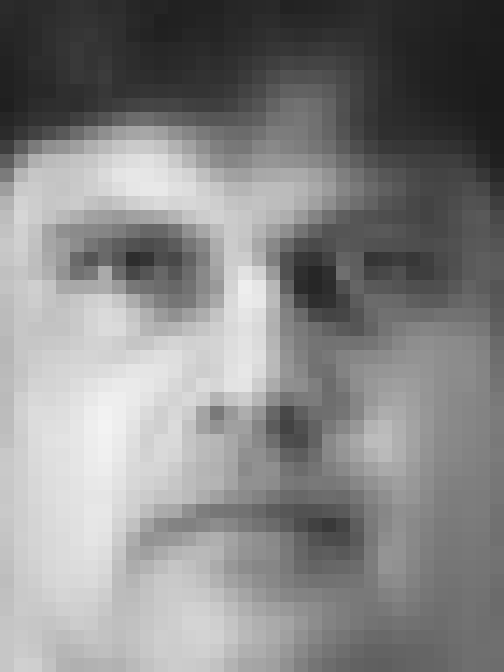}}\par\vspace{\vdiss cm}
		\subfloat {\includegraphics[width=\imw cm, height=\imh cm]{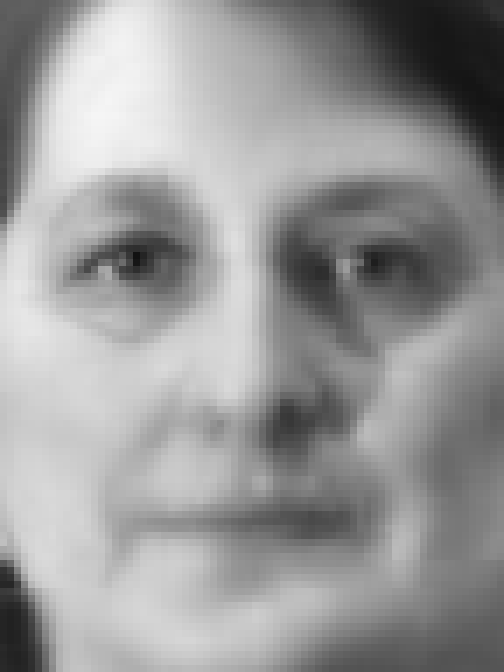}}\par\vspace{\vdiss cm}
		\subfloat {\includegraphics[width=\imw cm, height=\imh cm]{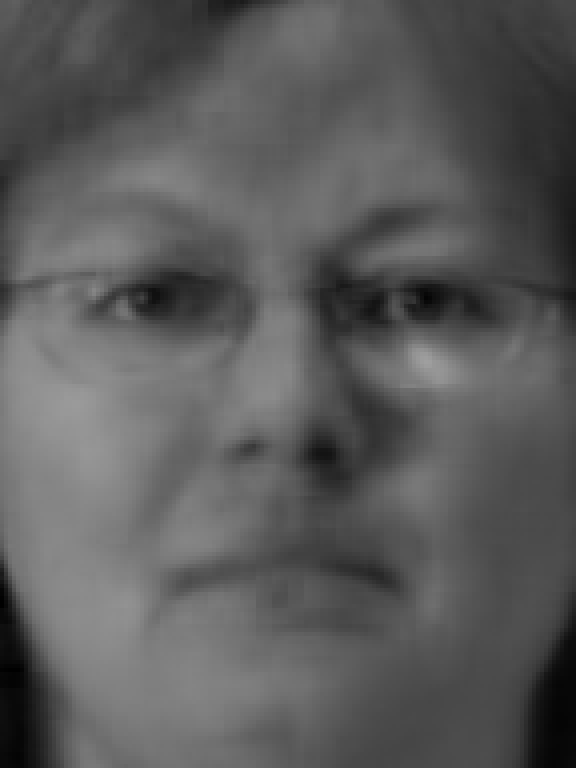}}\par\vspace{\vdisb cm}
		\subfloat {\includegraphics[width=\imw cm, height=\imh cm]{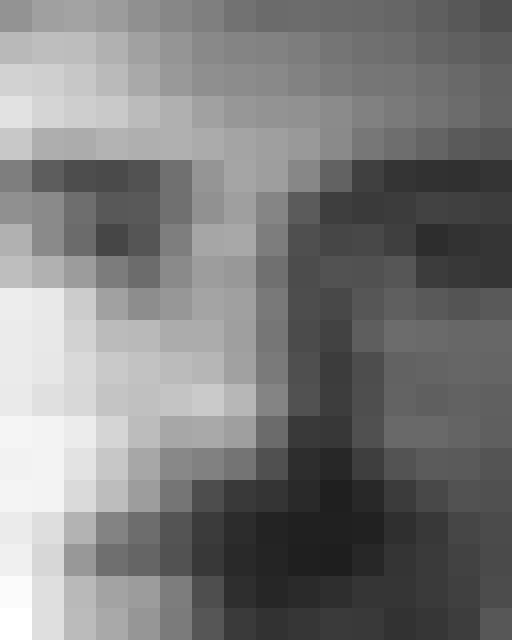}}\par\vspace{\vdiss cm}
		\subfloat {\includegraphics[width=\imw cm, height=\imh cm]{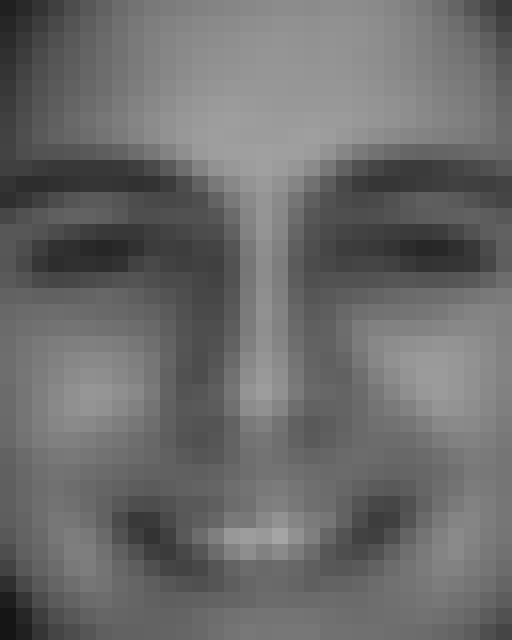}}\par\vspace{\vdiss cm}
		\stepcounter{figure}\addtocounter{figure}{-1}
		\addtocounter{subfigure}{9}
		\subfloat[]{\includegraphics[width=\imw cm, height=\imh cm]{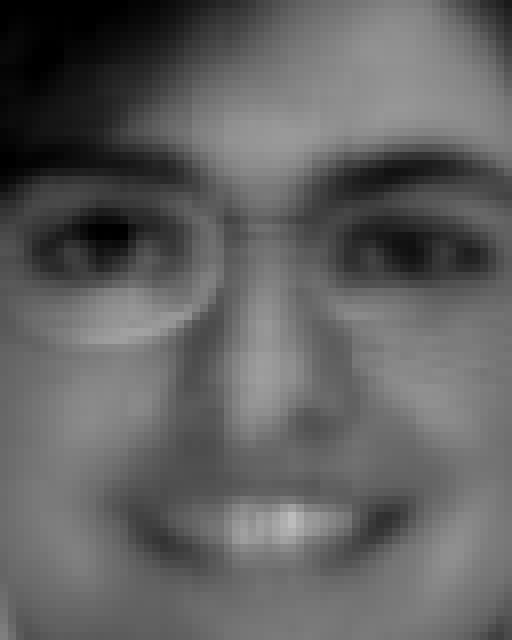}}
	\end{minipage}%
	\begin{minipage}{\hdis\textwidth} 
		\centering
		\subfloat {\includegraphics[width=\imw cm, height=\imh cm]{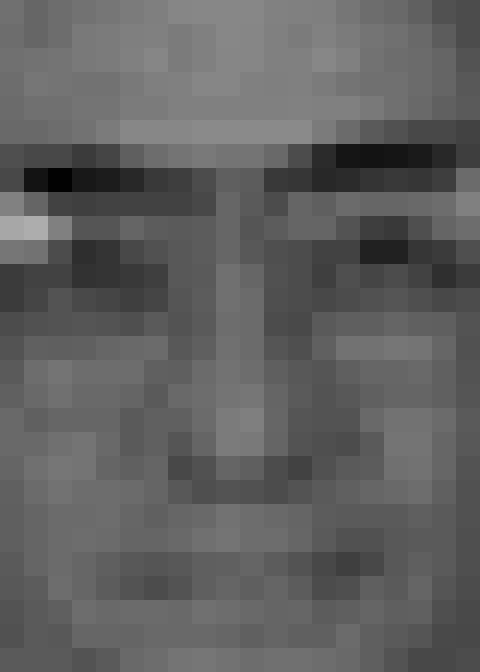}}\par\vspace{\vdiss cm}		
		\subfloat {\includegraphics[width=\imw cm, height=\imh cm]{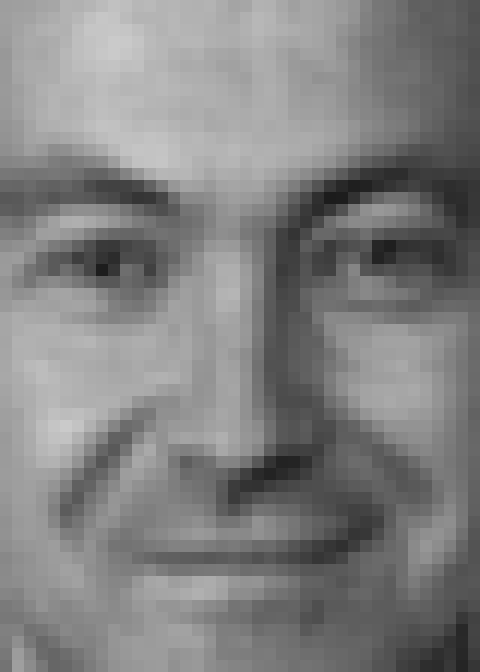}}\par\vspace{\vdiss cm}
		\subfloat {\includegraphics[width=\imw cm, height=\imh cm]{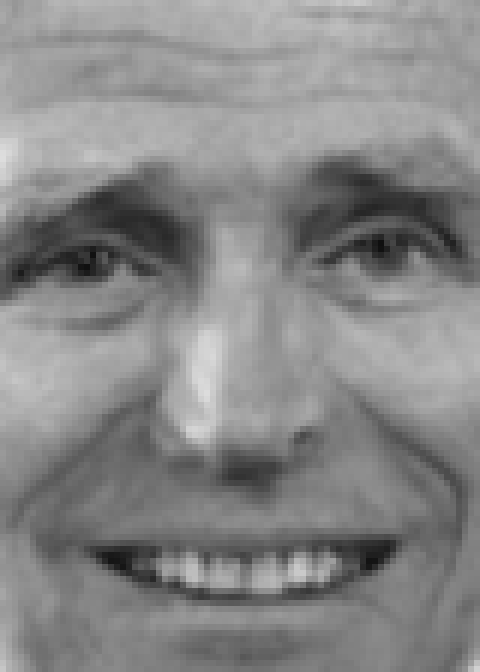}}\par\vspace{\vdisb cm}
		\subfloat {\includegraphics[width=\imw cm, height=\imh cm]{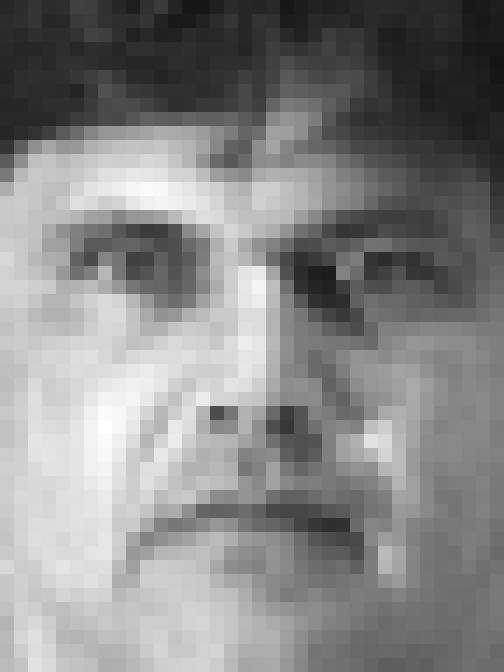}}\par\vspace{\vdiss cm}
		\subfloat {\includegraphics[width=\imw cm, height=\imh cm]{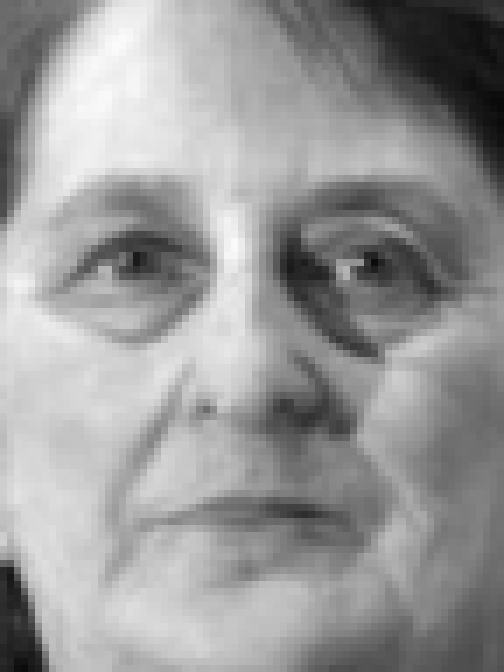}}\par\vspace{\vdiss cm}
		\subfloat {\includegraphics[width=\imw cm, height=\imh cm]{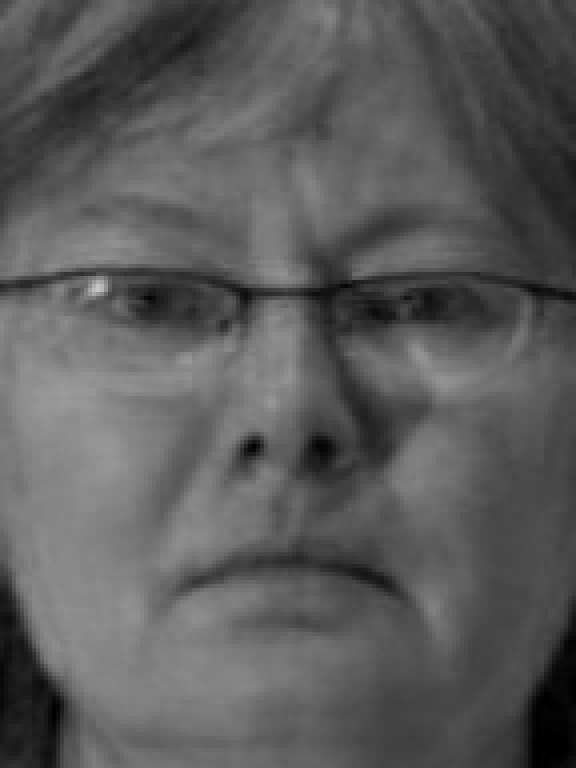}}\par\vspace{\vdisb cm}
		\subfloat {\includegraphics[width=\imw cm, height=\imh cm]{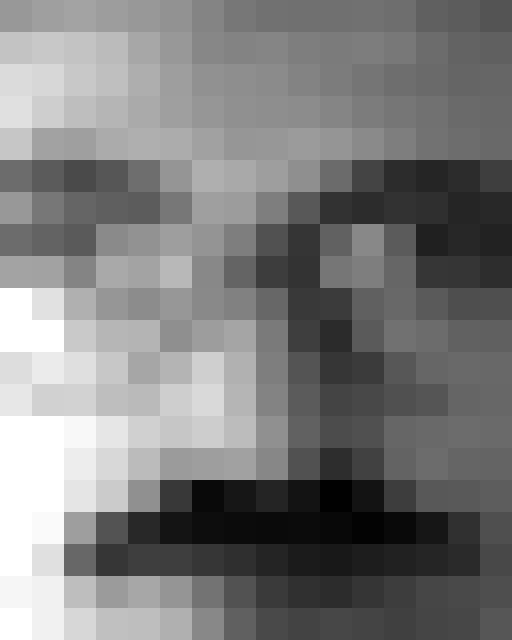}}\par\vspace{\vdiss cm}
		\subfloat {\includegraphics[width=\imw cm, height=\imh cm]{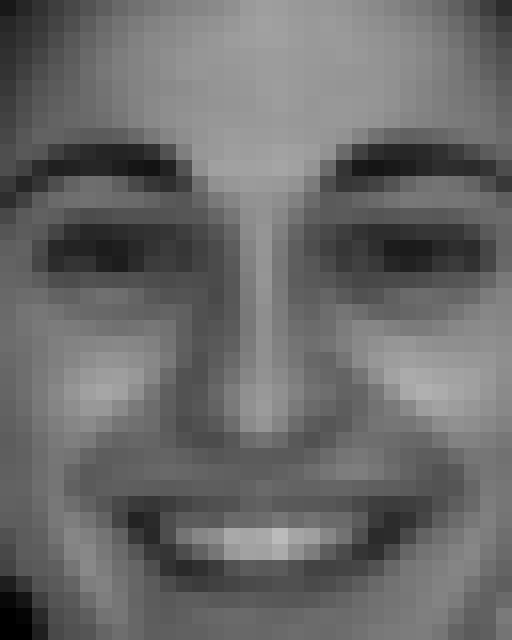}}\par\vspace{\vdiss cm}
		\stepcounter{figure}\addtocounter{figure}{-1}
		\addtocounter{subfigure}{10}
		\subfloat[]{\includegraphics[width=\imw cm, height=\imh cm]{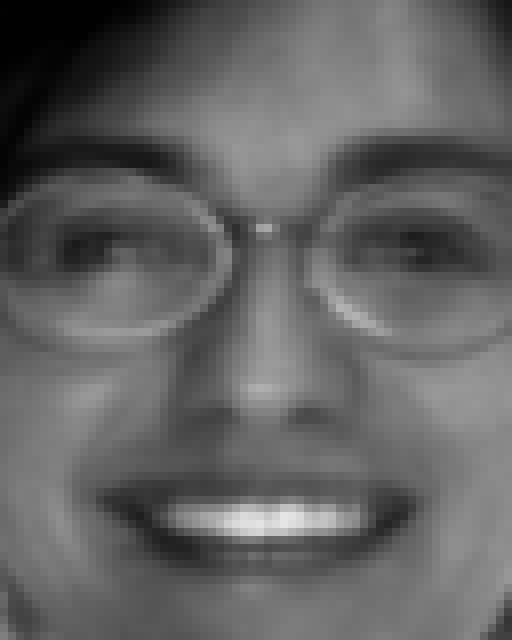}}
	\end{minipage}%
	\begin{minipage}{\hdis\textwidth} 
		\centering
		\subfloat {\includegraphics[width=\imw cm, height=\imh cm]{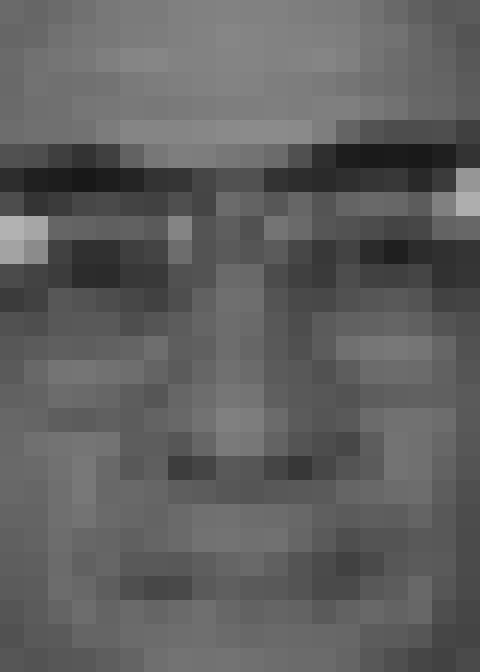}}\par\vspace{\vdiss cm}		
		\subfloat {\includegraphics[width=\imw cm, height=\imh cm]{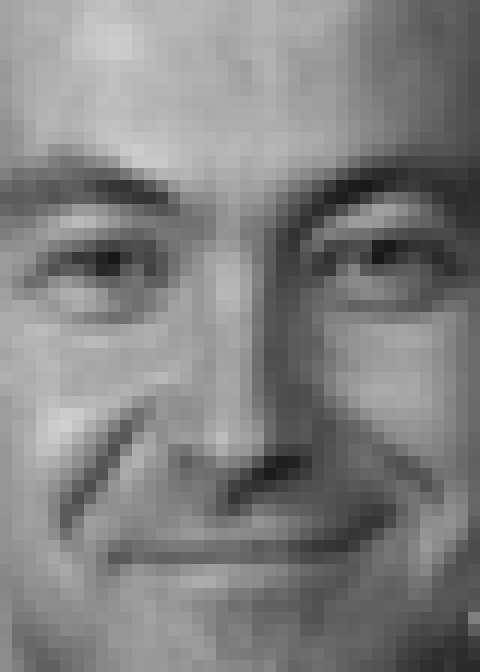}}\par\vspace{\vdiss cm}
		\subfloat {\includegraphics[width=\imw cm, height=\imh cm]{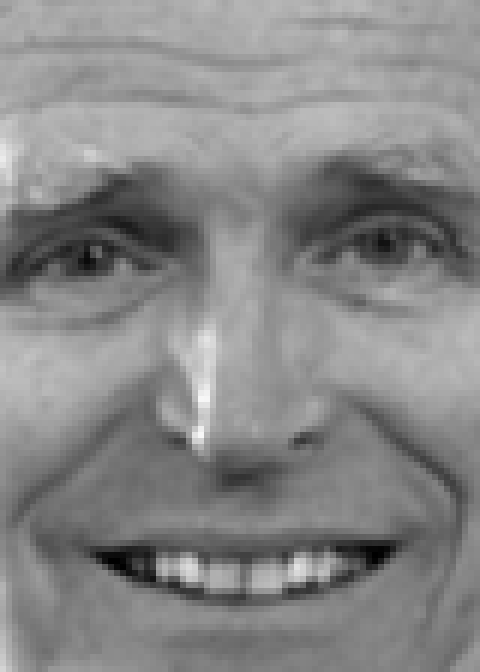}}\par\vspace{\vdisb cm}
		\subfloat {\includegraphics[width=\imw cm, height=\imh cm]{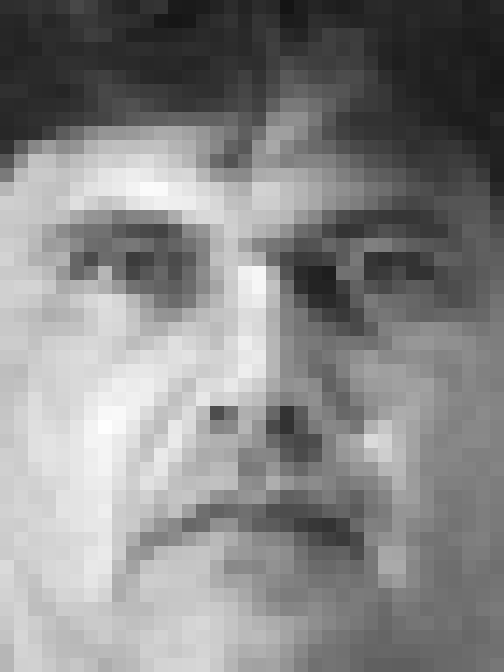}}\par\vspace{\vdiss cm}
		\subfloat {\includegraphics[width=\imw cm, height=\imh cm]{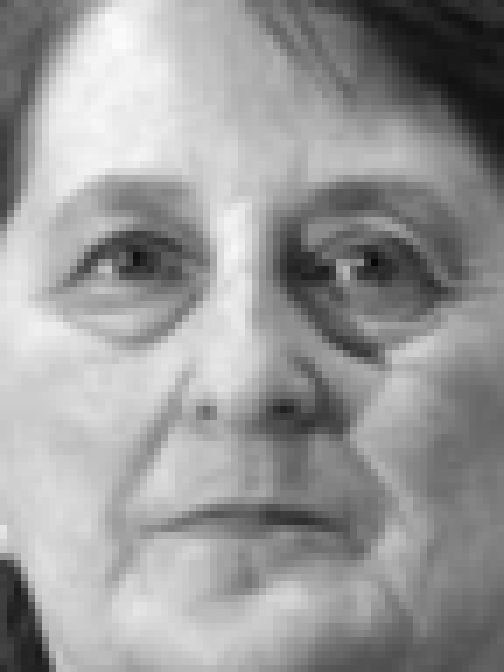}}\par\vspace{\vdiss cm}
		\subfloat {\includegraphics[width=\imw cm, height=\imh cm]{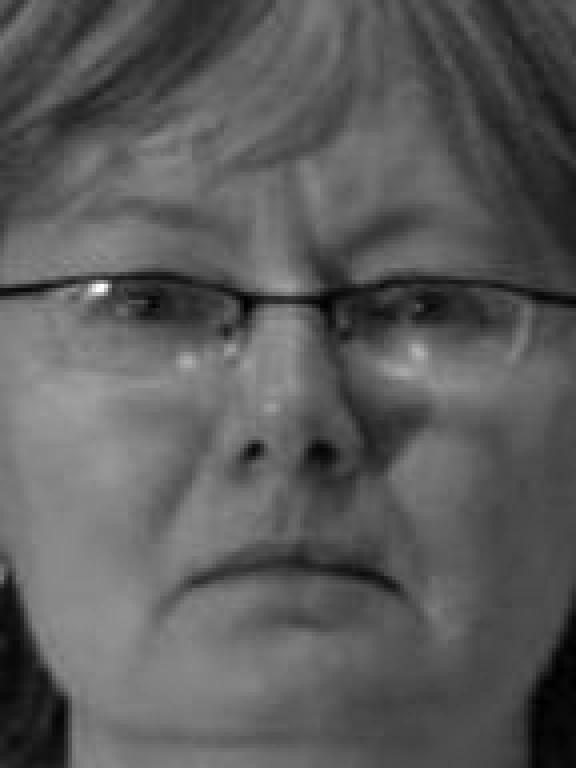}}\par\vspace{\vdisb cm}
		\subfloat {\includegraphics[width=\imw cm, height=\imh cm]{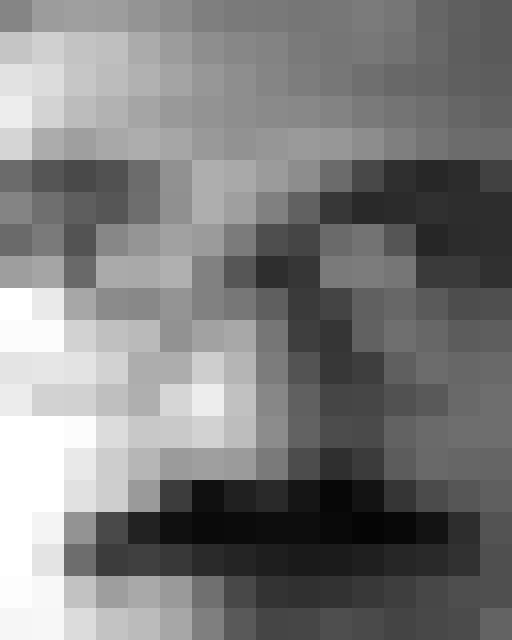}}\par\vspace{\vdiss cm}
		\subfloat {\includegraphics[width=\imw cm, height=\imh cm]{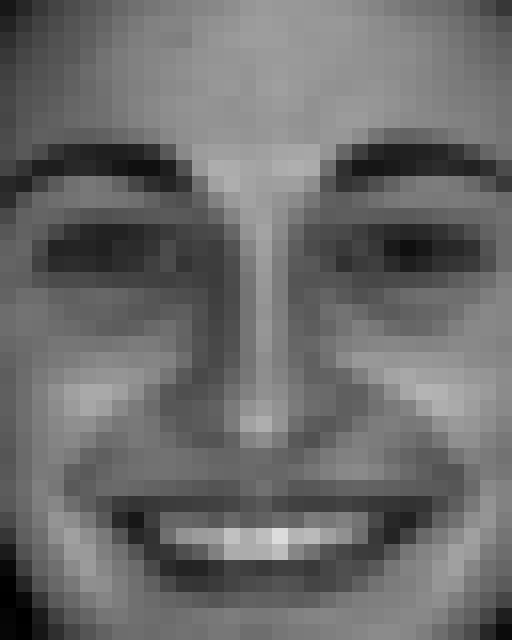}}\par\vspace{\vdiss cm}
		\stepcounter{figure}\addtocounter{figure}{-1}
		\addtocounter{subfigure}{11}
		\subfloat[]{\includegraphics[width=\imw cm, height=\imh cm]{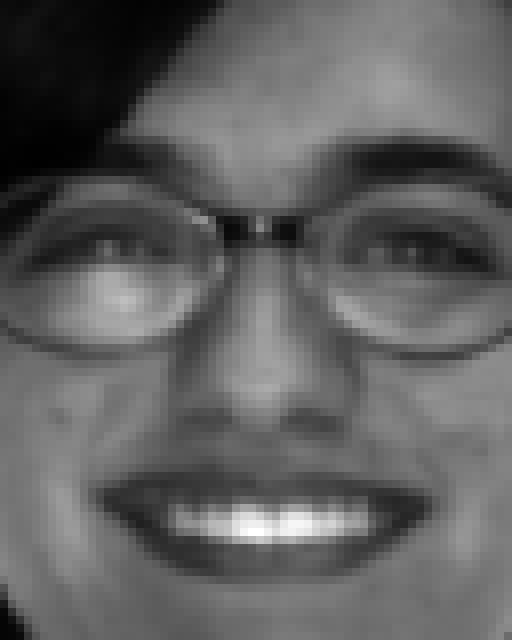}}
	\end{minipage}\par\medskip
	\caption{Face hallucination results obtained by various methods on different databases. The first three rows indicate the results on the FERET database (scaling factors 2, 4, and 8, respectively), the second three rows are the ones conducted on the Multi-PIE dataset (scaling factors 4, 8, and 16, respectively), and the last three rows represent the results associated with the AR face database (scaling factors 4, 8, and 16, respectively). (a) LR input. (b) Bicubic interpolation. (c) Wang \cite{refs:Wang2005}. (d) LSR \cite{refs:Ma2010}. (e) LcR \cite{refs:Jiang2014a}. (f) LINE \cite{refs:Jiang2014b}. (g) SSR \cite{refs:Jiang2017}. (h) LM-CSS \cite{refs:Farrugia2017}. (i) TRNR \cite{refs:Jiang2016}. (j) TLcR-RL \cite{refs:Jiang2018}. (k) Proposed. (l) Ground truth.}
	\label{fig:frontres}
\end{figure*}

\subsubsection{The FERET dataset}
\begin{figure}
	\captionsetup[subfloat]{farskip=0pt,captionskip=1pt}
	\centering
	\def\theight{0.18}
	\subfloat{\includegraphics[height=\theight\textwidth]{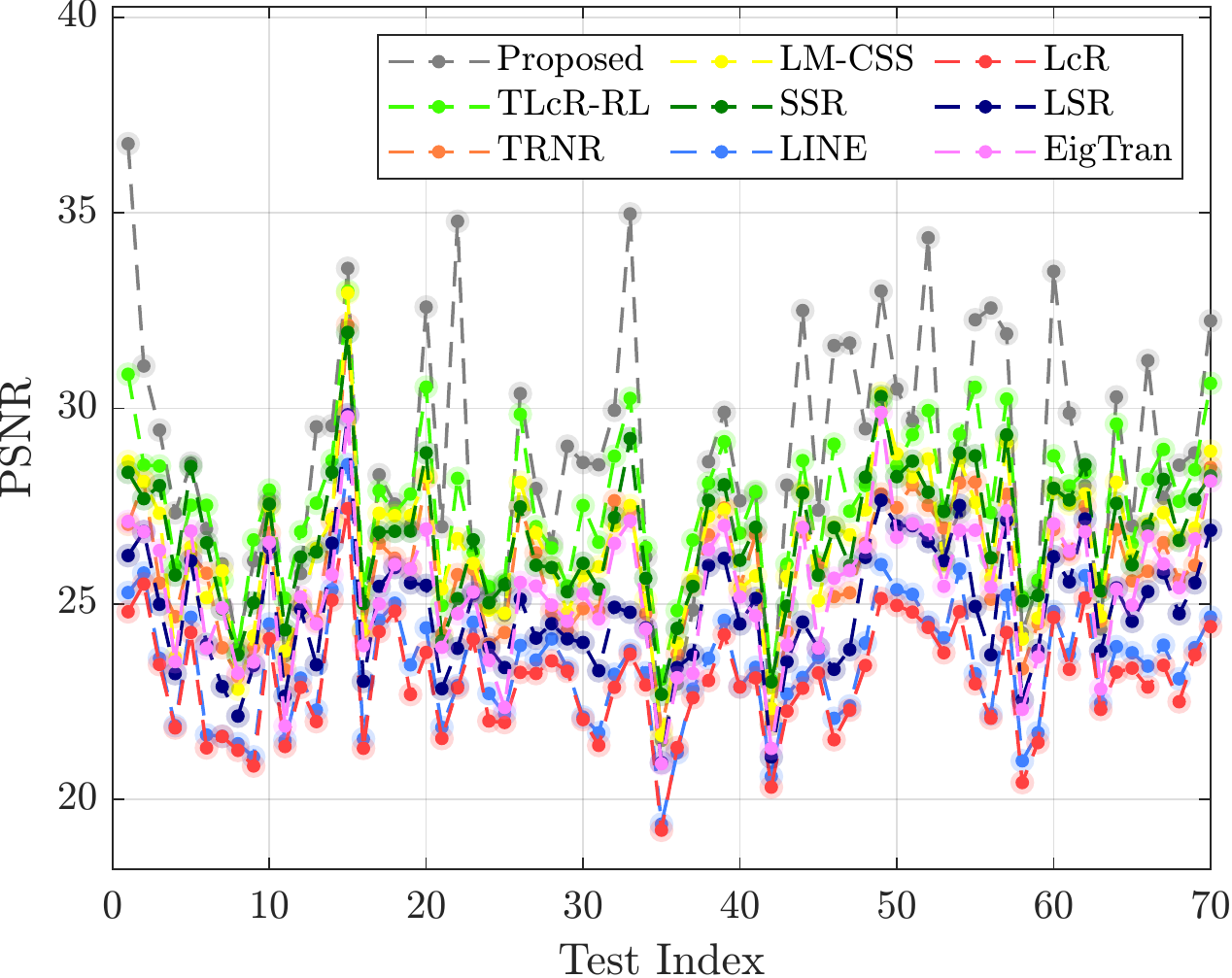}}\hfil
	\subfloat{\includegraphics[height=\theight\textwidth]{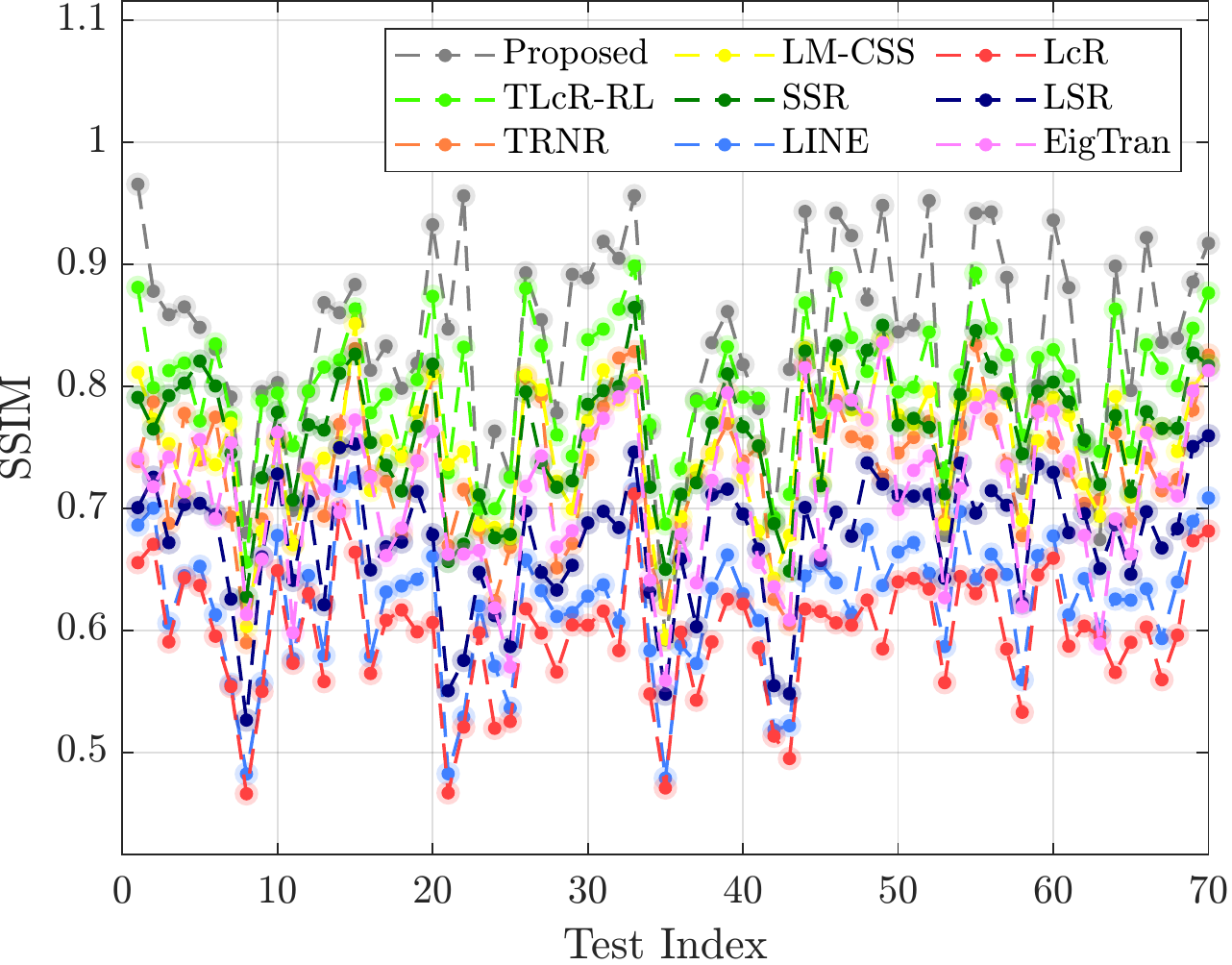}}
	\caption{Quantitative metrics achieved by different methods on all the 70 test samples of the FERET database when scaling factor is set to 4.}
	\label{fig:feretall}
\end{figure}

We first evaluate the performance of the proposed method on the frontal facial images from the FERET database. We only select subjects with five or more samples in the database, which leads to a subset of 519 images from 70 individuals, each with unequal number of samples. The LR input faces are of size $15\times10$, and scaling factors are set to 2, 4, and 8. The PSNR and SSIM performance of different algorithms is summarized in Table \ref{tab:feretres}. In all three experiments and with different scaling factors, the proposed method outperforms the second best algorithm by 1.18 dB, 1.00 dB, and 1.00 dB in PSNR and 0.0095, 0.0199, and 0.0353 in SSIM, respectively. In Fig. \ref{fig:feretall}, the performance of the algorithms on each one of the test samples when scaling factor is 4 is displayed, which demonstrates the dominance of the proposed method over the competitive ones in almost all the available test images. Fig. \ref{fig:frontres} (top three rows) also qualitatively compares the methods on three testing images with different scaling factors. It is observable that the competitive methods were unable to recover facial details including eyeglasses and wrinkles. In the presence of facial expressions, \cite{refs:Wang2005} produced undesirable artifacts in the recovered face images. The methods based on position-patch also produced blurry and oversmoothed images, particularly around the mouth regions. The results of LM-CSS \cite{refs:Farrugia2017} appear to be more similar to the original HR faces than those of the remaining approaches, however, the noise and artifacts added to the resultant faces have made this method quantitatively unsatisfactory. In general, the results obtained by the proposed algorithm are obviously more detailed, clear, and artifact-free compared to those produced by the others.

\subsubsection{The Multi-PIE dataset}
We next focus on the Multi-PIE database frontal images (camera 05-1) taken in neutral expression and under normal illumination (illumination condition 10) from all four sessions and two recordings. hence there will be 650 images representing 130 subjects, each with five different samples. In this experiment, the size of the LR input images are $12\times9$, and three different scaling factors 4, 8, and 16 are considered. The quantitative results of different methods can be seen in Table \ref{tab:multipieres}. One can notice that the proposed method shows superior performance over the other algorithms by considerable margins (1.09 dB, 1.46 dB, and 1.28 dB in PSNR and 0.0167, 0.0445, and 0.0308 in SSIM, in scaling factors 4, 8, and 16, respectively). Visually speaking, as depicted in the second three rows of Fig. \ref{fig:frontres}, the results of the proposed algorithm is by far more similar to the ground truth images, particularly when individual-specific face attributes (such as hair, mouth, wrinkles, and eyeglasses) are taken into consideration. Due to their inability to recover edges, almost all the other methods were unable to reconstruct age-specific features of the faces, leading to faces which appear younger than the person's actual age. Owing to its locality constrained approach, TLcR-LR \cite{refs:Jiang2018} has done slightly better than the other methods in recovering some facial attributes (e.g., eyeglasses), however, its results still suffer from blurriness and lack of details.

\subsubsection{The AR dataset (VLR Face Hallucination)}
To demonstrate the efficiency of the proposed algorithm in hallucinating very low-resolution face images \cite{refs:Zou2012}, extensive experiments were conducted on the AR face database. From the cropped version of the database \cite{refs:Martinez2001}, we discard the images with occlusion, and select a subset of 1400 images associated with each of the 100 subjects, each with 14 samples. Low-resolution input faces are chosen to be $5\times4$, making it an extreme case of VLR face super-resolution task with only 20 LR pixels available. The dataset also contains facial images with significant expression variations, which cause the problem to be even more challenging. We upscale the existing LR faces by the factors of 4, 8, and 16. As can be seen in Table \ref{tab:arres}, our approach shows its capability in recovering very low-resolution inputs in all three experiments by improving the PSNR by 1.18 dB, 2.37 dB, and 1.5 dB, and the SSIM by 0.0186, 0.1235, and 0.1166 units, respectively. More importantly, as the visual comparison in the last three rows of Fig. \ref{fig:frontres} suggests, the performance of the position-patch based methods falls dramatically when given VLR inputs, with their results being blurry and mostly irrelevant. Conversely, despite being a classic algorithm, \cite{refs:Wang2005} manages to outperform several recently introduced patch-based techniques due to its global reconstruction approach. All in all, the results achieved by the proposed algorithm bear much more visual resemblance to the ground truth faces compared to those of the others.

\begin{table}
	\caption{Quantitative evaluation scores of different algorithms for the $12\times9$ test faces of the Multi-PIE database based on different scaling factors}
	\centering
	\begin{tabular}{c||cc|cc|cc} 
		\hline\hline
		\multirow{2}{*}{Algorithm} & \multicolumn{2}{c|}{$12\times9$ \textbar{} $\times4$} & \multicolumn{2}{c|}{$12\times9$ \textbar{} $\times8$} & \multicolumn{2}{c}{$12\times9$ \textbar{} $\times16$}  \\ 
		\cline{2-7}
		& PSNR           & SSIM                                  & PSNR           & SSIM                                  & PSNR           & SSIM                                  \\ 
		\hline
		Bicubic                    & 21.20          & 0.5967                                & 18.83          & 0.4920                                & 17.69          & 0.5564                                \\
		Wang \cite{refs:Wang2005}                  & 25.85          & 0.8338                                & 25.43          & 0.7588                                & 25.21          & 0.7429                                \\
		LSR \cite{refs:Ma2010}                       & 26.88          & 0.8548                                & 24.57          & 0.7102                                & 22.72          & 0.6498                                \\
		LcR \cite{refs:Jiang2014a}                       & 28.08          & 0.8841                                & 25.43          & 0.7493                                & 22.93          & 0.6602                                \\
		LINE \cite{refs:Jiang2014b}                      & 28.97          & 0.9042                                & 25.53          & 0.7585                                & 23.02          & 0.6727                                \\
		SSR \cite{refs:Jiang2017}                       & 28.39          & 0.8963                                & 26.52          & 0.7908                                & 24.97          & 0.7118                                \\
		LM-CSS \cite{refs:Farrugia2017}                    & 27.48          & 0.8748                                & 26.46          & 0.7861                                & 26.18          & 0.7867                                \\
		TRNR \cite{refs:Jiang2016}                      & 30.25          & 0.9278                                & 26.01          & 0.7790                                & 23.36          & 0.6860                                \\
		TLcR-RL \cite{refs:Jiang2018}                   & 29.88          & 0.9211                                & 27.66          & 0.8324                                & 26.78          & 0.7920                                \\
		Proposed                   & \textbf{31.34} & \textbf{0.9445}                       & \textbf{29.12} & \textbf{0.8769}                       & \textbf{28.06} & \textbf{0.8228}                       \\
		\hline\hline
	\end{tabular}
	\label{tab:multipieres}
\end{table}

\begin{table}
	\caption{Objective comparison of the proposed method with the competitive algorithms on the $5\times4$ test samples of the AR database by different scaling factors}
	\centering
	\begin{tabular}{c||cc|cc|cc} 
		\hline\hline
		\multirow{2}{*}{Algorithm} & \multicolumn{2}{c|}{$5\times4$ \textbar{} $\times4$} & \multicolumn{2}{c|}{$5\times4$ \textbar{} $\times8$} & \multicolumn{2}{c}{$5\times4$ \textbar{} $\times16$}  \\ 
		\cline{2-7}
		& PSNR           & SSIM                                  & PSNR           & SSIM                                  & PSNR           & SSIM                                  \\ 
		\hline
		Bicubic                    & 17.14          & 0.3444                                & 14.92          & 0.2041                                & 13.70          & 0.2674                                \\
		Wang \cite{refs:Wang2005}                  & 22.42          & 0.8223                                & 20.91          & 0.6834                                & 20.27          & 0.6091                                \\
		LSR \cite{refs:Ma2010}                       & 22.98          & 0.8407                                & 20.37          & 0.6359                                & 18.69          & 0.5127                                \\
		LcR \cite{refs:Jiang2014a}                       & 23.62          & 0.8580                                & 19.44          & 0.6052                                & 18.98          & 0.5344                                \\
		LINE \cite{refs:Jiang2014b}                      & 24.61          & 0.8877                                & 19.58          & 0.6367                                & 19.11          & 0.5664                                \\
		SSR \cite{refs:Jiang2017}                       & 23.79          & 0.8562                                & 21.31          & 0.6876                                & 20.17          & 0.5844                                \\
		LM-CSS \cite{refs:Farrugia2017}                    & 22.19          & 0.8121                                & 20.52          & 0.6503                                & 19.83          & 0.5683                                \\
		TRNR \cite{refs:Jiang2016}                      & 25.34          & 0.9042                                & 21.61          & 0.7274                                & 19.45          & 0.5625                                \\
		TLcR-RL \cite{refs:Jiang2018}                   & 24.59          & 0.8859                                & 21.86          & 0.7225                                & 21.04          & 0.6455                                \\
		Proposed                   & \textbf{26.52} & \textbf{0.9228}                       & \textbf{24.23} & \textbf{0.8509}                       & \textbf{22.54} & \textbf{0.7621}                       \\
		\hline\hline
	\end{tabular}
	\label{tab:arres}
\end{table}

\subsection{Parameters Analysis}
\subsubsection{Regularization Parameters}
\begin{figure}[t]
	\captionsetup[subfloat]{farskip=0pt}
	\centering
	\def\theight{0.18}
	\subfloat{\includegraphics[height=\theight\textwidth]{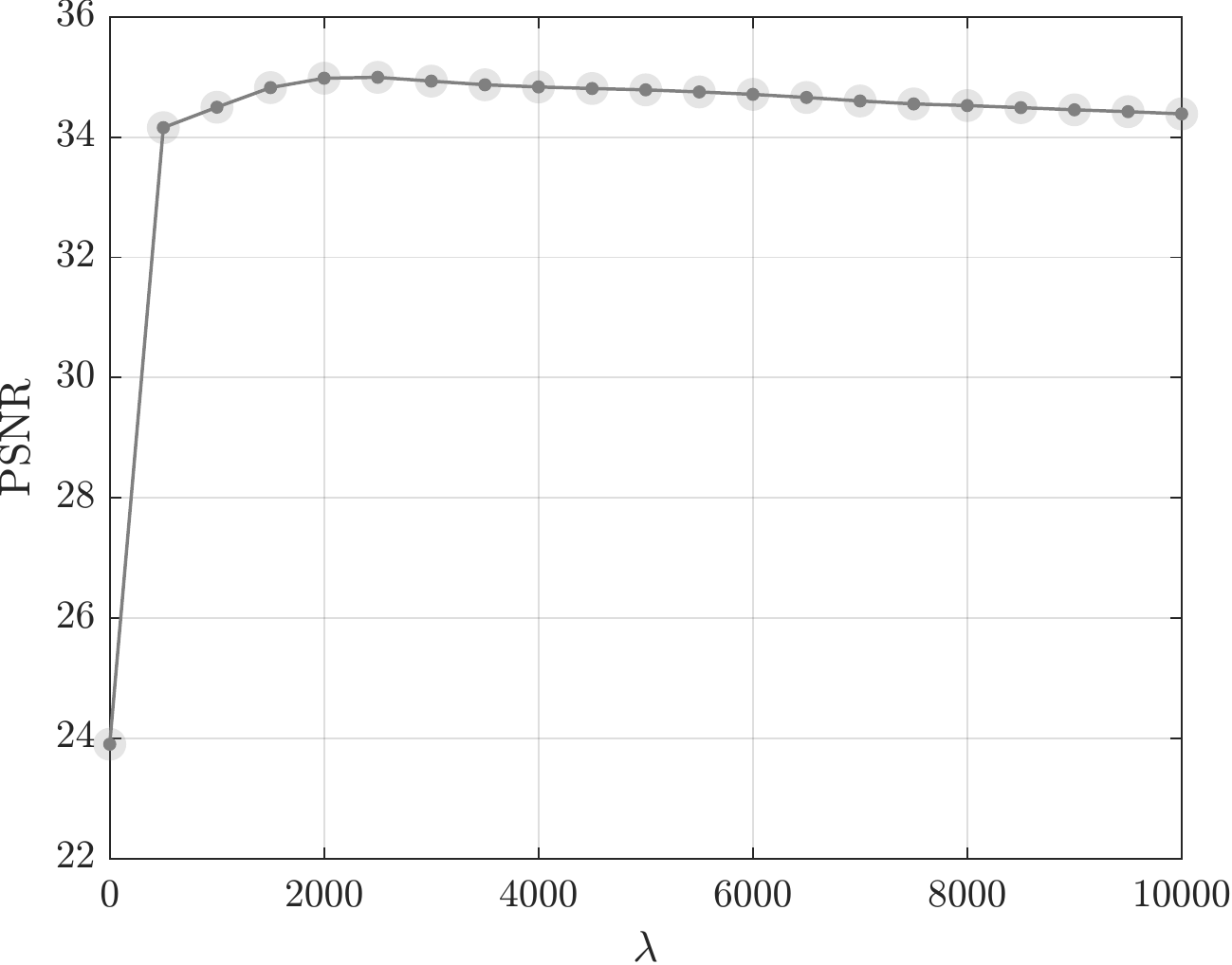}}\hfil
	\subfloat{\includegraphics[height=\theight\textwidth]{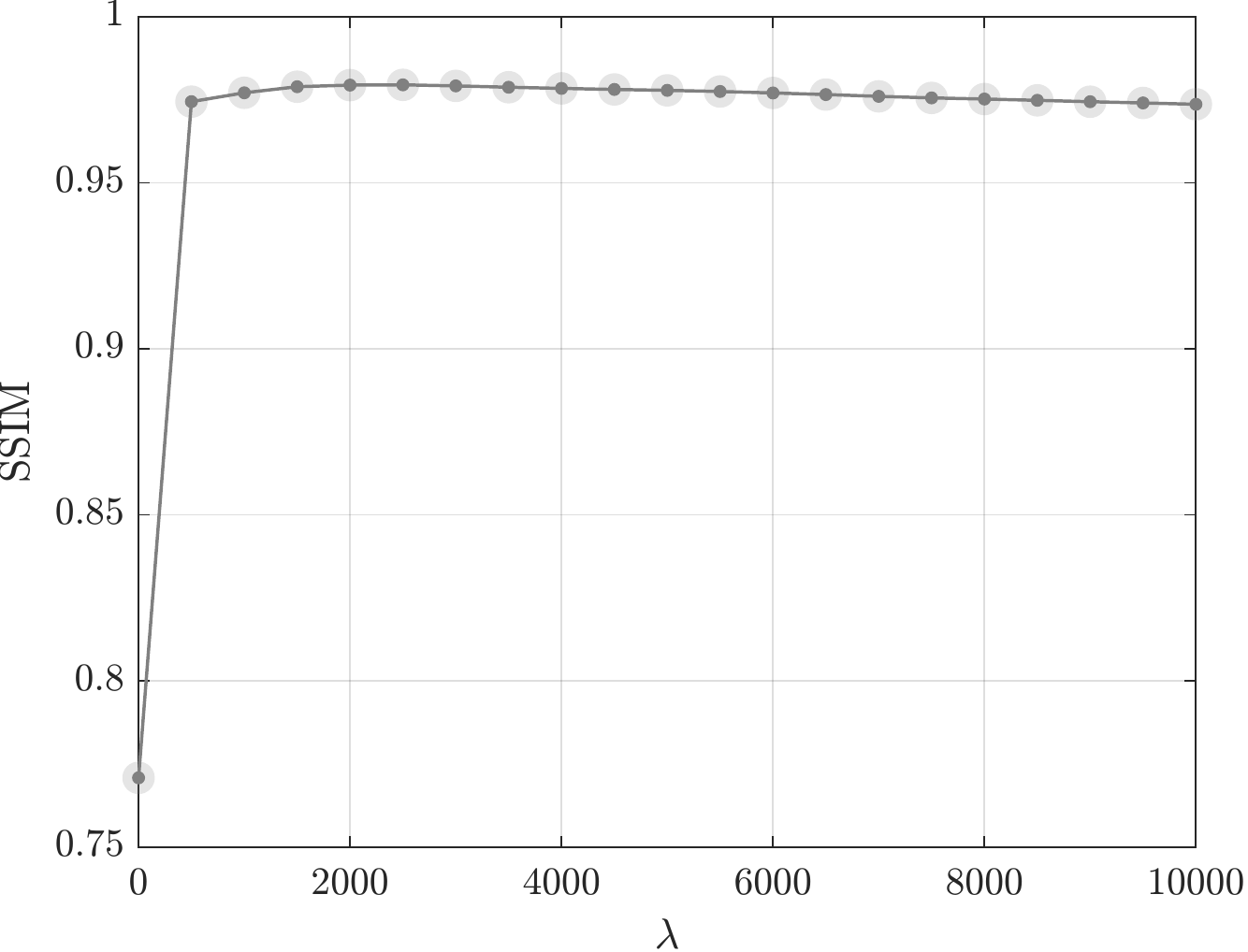}}
	\caption{Influence of the parameter $\lambda$ on the face hallucination performance.}
	\label{fig:lambda}
\end{figure}

\begin{figure}[t]
	\captionsetup[subfloat]{farskip=0pt}
	\centering
	\def\theight{0.18}
	\subfloat{\includegraphics[height=\theight\textwidth]{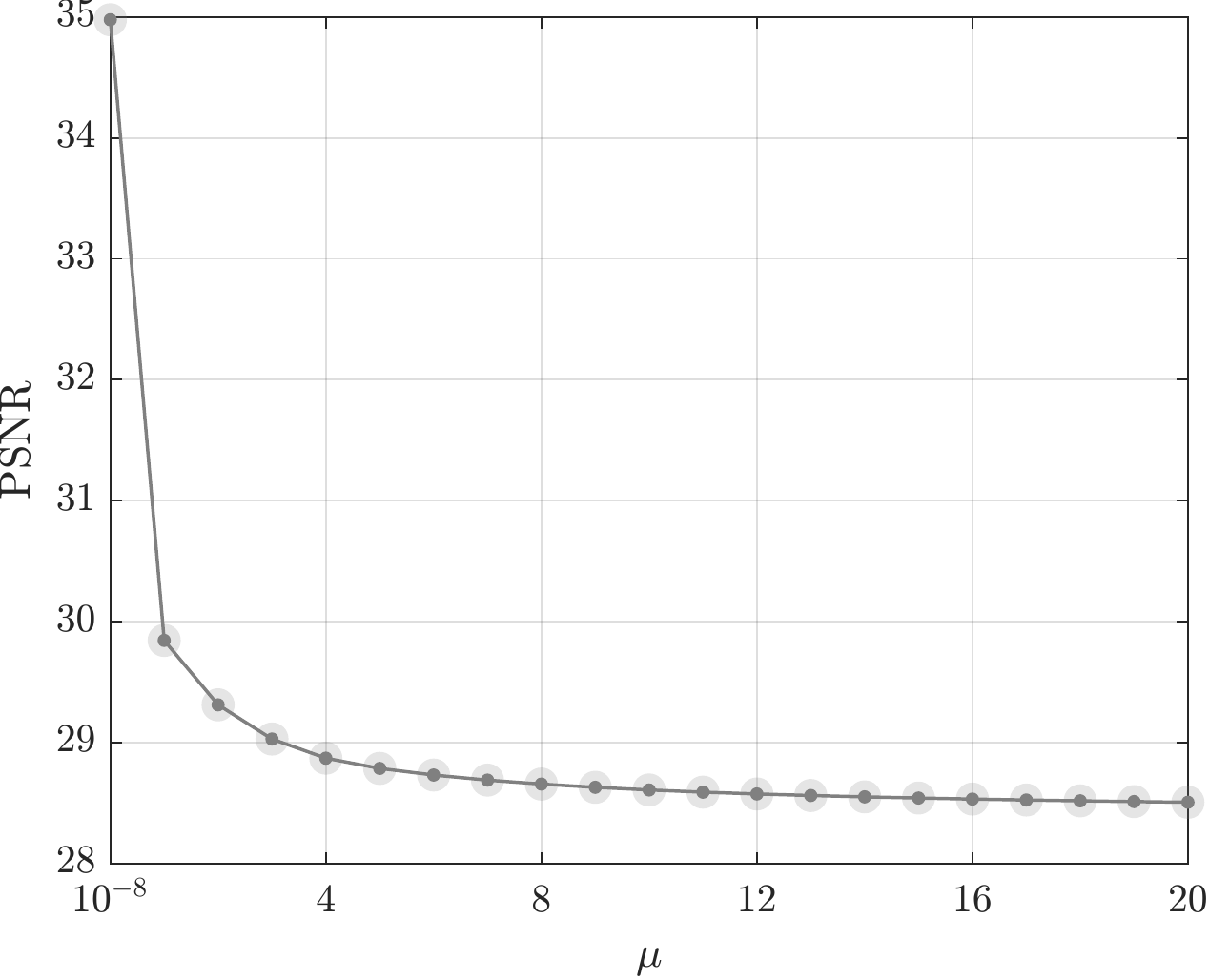}}\hfil
	\subfloat{\includegraphics[height=\theight\textwidth]{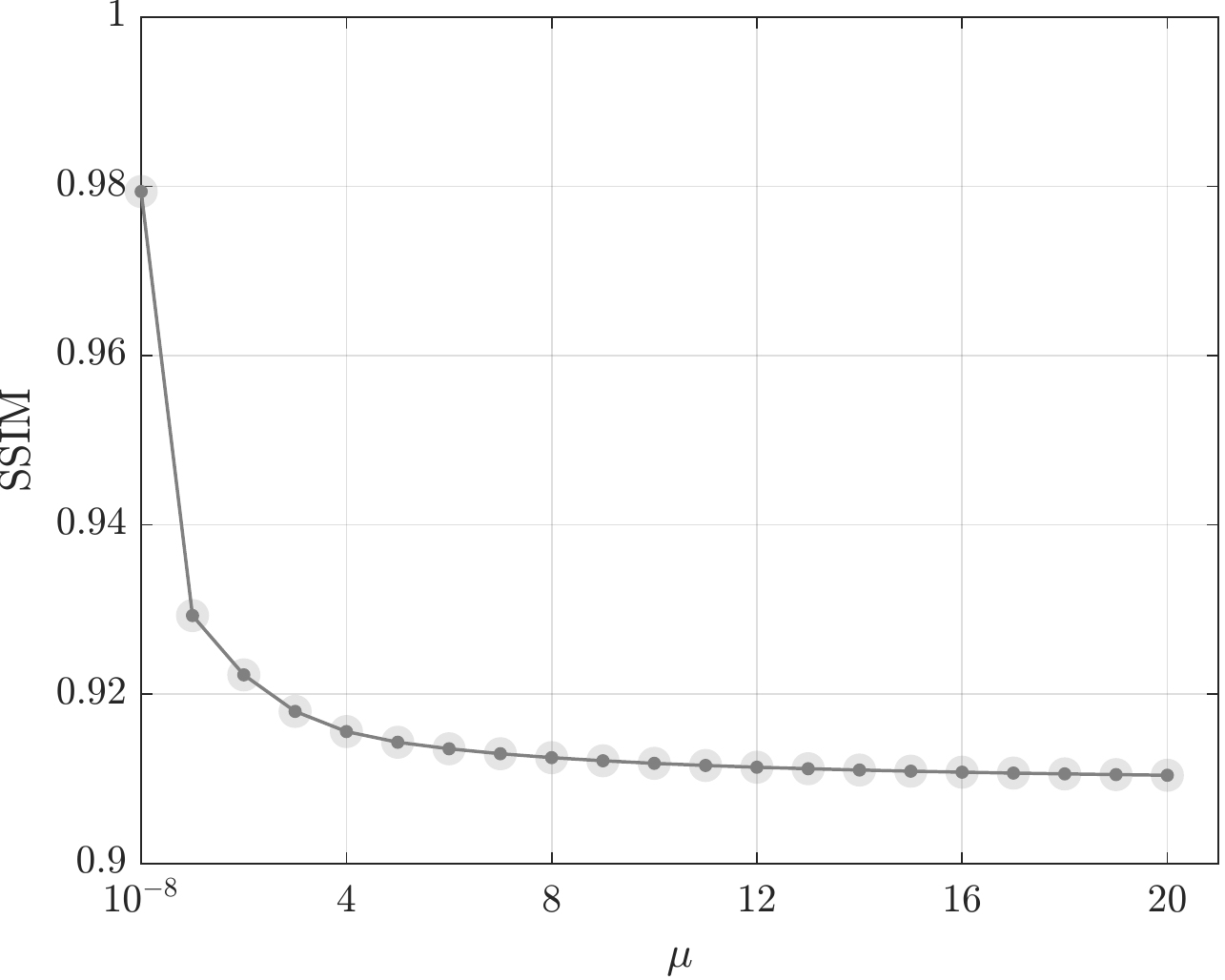}}
	\caption{Objective results based on different values of $\mu$.}
	\label{fig:tau}
\end{figure}

There are two regularization parameters in the proposed formulation, namely $\mu$ and $\lambda$. The first determines the closeness of the reconstructed face to the subspace spanned by the available faces, whereas the second decides how strictly this face subspace should be estimated. In this subsection, we perform experiments on a randomly selected face image and tune each parameter separately while keeping the other fixed. By fixing $\mu$ and changing $\lambda$ values over the range $[0,10^4]$ with an interval of 500, as plotted in Fig. \ref{fig:lambda}, one can notice that the best performance is achieved when $\lambda$ is roughly set to $10^3$. We next fix the value of $\lambda$ and chose 20 different values for $\mu$ from the range $[0,20]$. The variations of PSNR and SSIM (Fig. \ref{fig:tau}) suggest that values closer to zero are more desirable for this parameter. When $\mu$ is set to zero, however, the hallucinated face image will be incalculable (NaN) due to the absence of regularization term in the optimization function. We therefore set $\lambda=2700$ and $\mu=10^{-8}$ in our experiments.


\subsubsection{Number of Training Samples per Subject}
The proposed algorithm utilizes the face subspace as a prior knowledge, thus it is desirable to see how the quality of the subspace spanned by the training face images affects its performance. In this regard, the diversity among the faces (i.e., the availability of faces with different variations), which is related to the number of training faces per subject, is expected to be decisive. To justify this, we create a test set by randomly selecting one sample from each subject of the AR dataset, and perform a series of experiments on the selected set, each time with different number of training samples per subject. Fig. \ref{fig:samppersub} shows the quantitative metrics obtained by different methods in each experiment. It is observable that when there is only one sample per subject, TLcR-RL \cite{refs:Jiang2018} and TRNR \cite{refs:Jiang2016} achieve better performance compared to the proposed method. This can be justified by the fact that in this case, no subject-specific face subspace is formed and subsequently the prior term used in the formulation will be of no benefit. However, the performance of the proposed method improves significantly when another training sample is added for each subject, with only 0.1 dB lower PSNR and 0.0246 higher SSIM compared to the pioneering method. When there are three samples per each subject, the proposed method clearly outperforms the remaining algorithms, and by increasing the number of samples continues to improve its dominance whereas the performance of the other methods remains relatively unchanged. In Fig. \ref{fig:spsexample}, the influence of the number of samples per subject on the visual appearance of a face image reconstructed by the proposed method (bottom) and \cite{refs:Jiang2018} (top) is displayed. As the number of samples per subject increases, more details appear in the hallucinated face, and the undesired effects (for example, on forehead region) diminish. With four samples per subject, a clean face image close to the ground truth is obtained, whereas the true facial expression is reconstructed when seven samples per subject are used. The figure also reveals that the results generated by TLcR-RL are less affected by the addition of extra samples to the training set, and are still oversmoothed and blurry even when there are 13 samples per subject available. To summarize, the experiment demonstrates that the proposed algorithm requires only two to three images per each subject to outperform the competitive methods both quantitatively and qualitatively.

\begin{figure}
	\captionsetup[subfloat]{farskip=0pt,captionskip=1pt}
	\centering
	\def\theight{0.18}
	\subfloat{\includegraphics[height=\theight\textwidth]{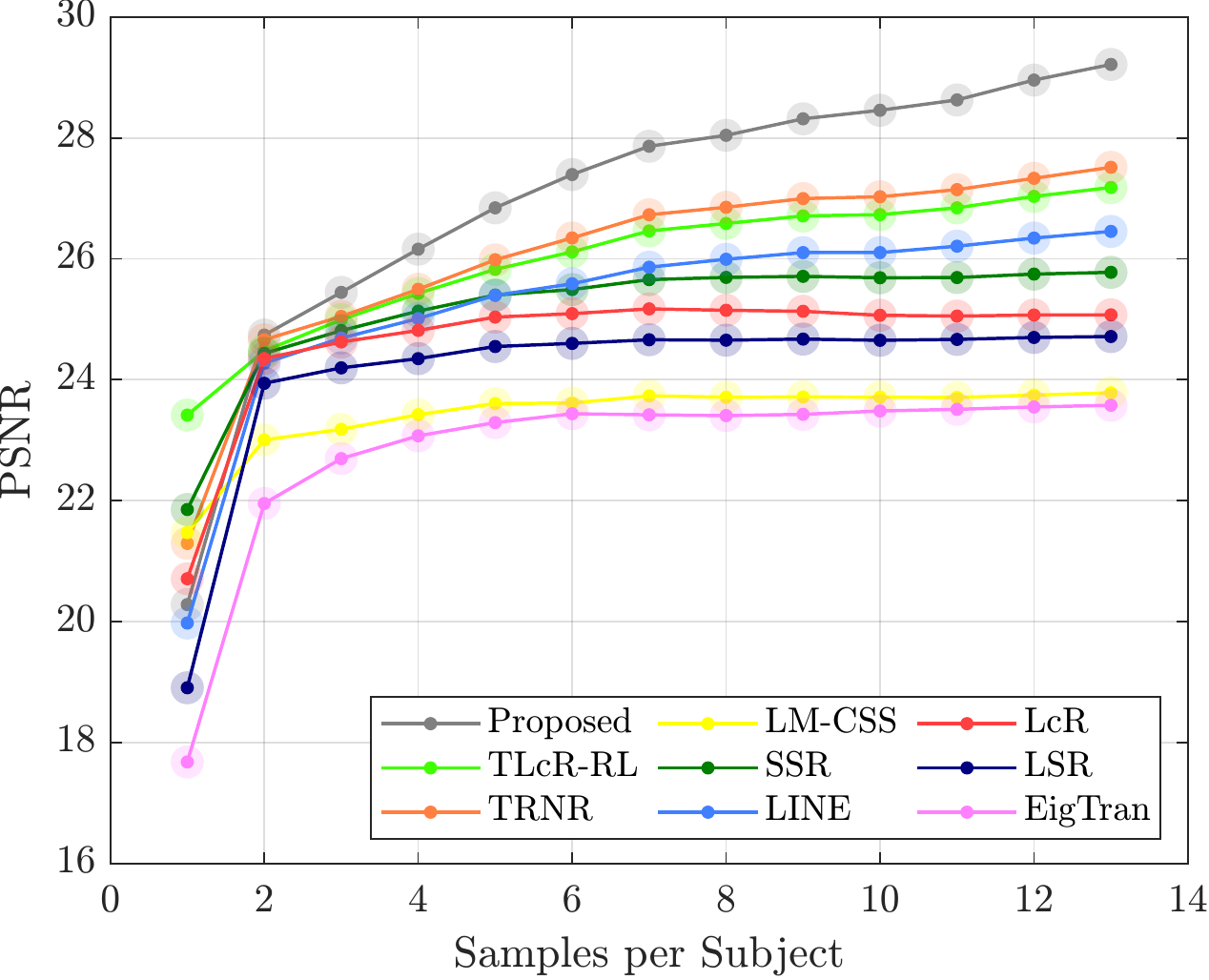}}\hfil
	\subfloat{\includegraphics[height=\theight\textwidth]{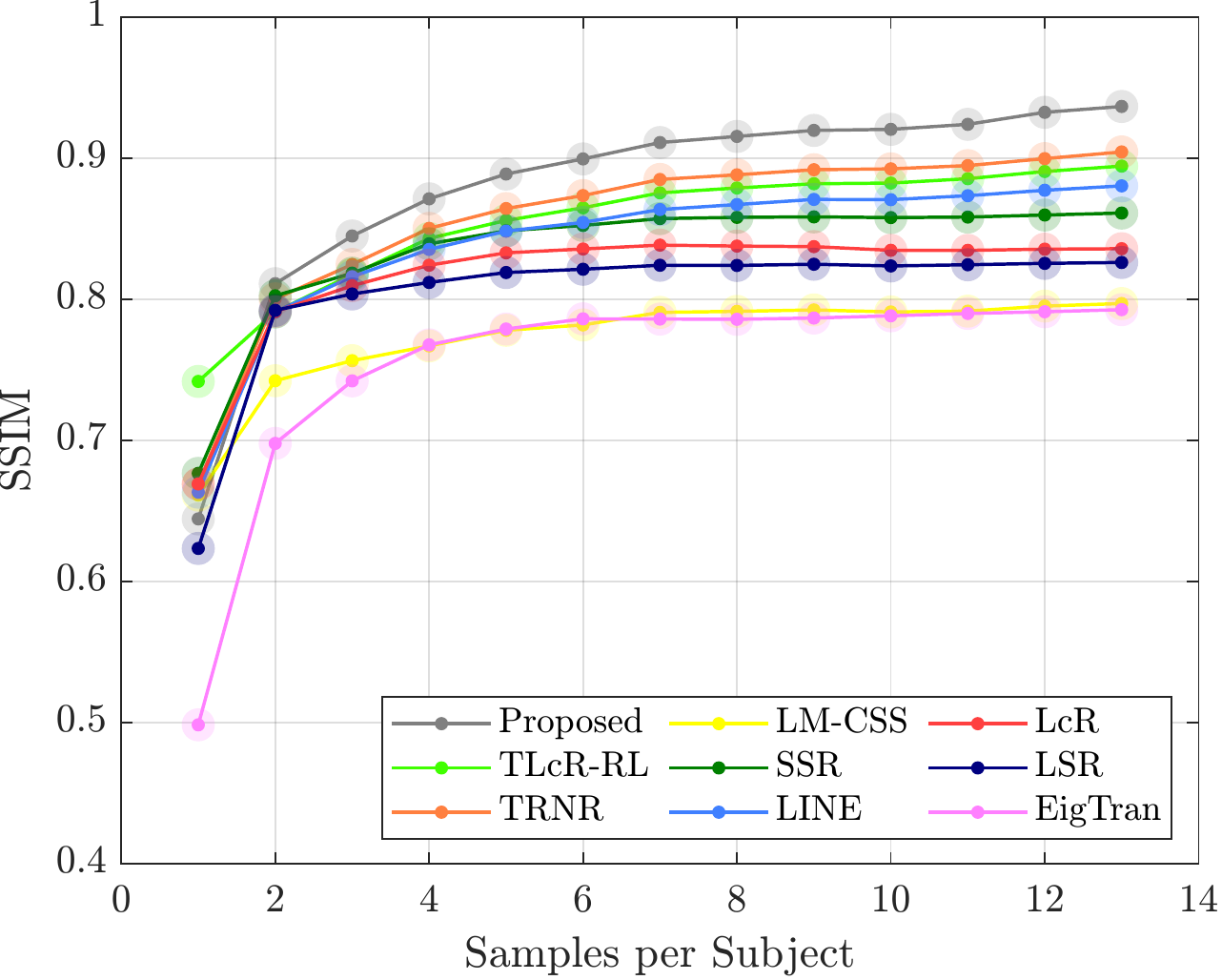}}
	\caption{Effect of the number of samples per subject on the performance of different algorithms.}
	\label{fig:samppersub}
\end{figure}

\begin{figure}
	\captionsetup[subfloat]{farskip=0pt,captionskip=1pt}
	\centering
	\def\imw{1.15}
	\def\imh{1.61} 
	\def\hdis{0.066}
	\def\vdiss{0.04}
	\def\vdisb{0.1}
	\begin{minipage}{\hdis\textwidth} 
		\centering
		\subfloat {\includegraphics[width=\imw cm, height=\imh cm]{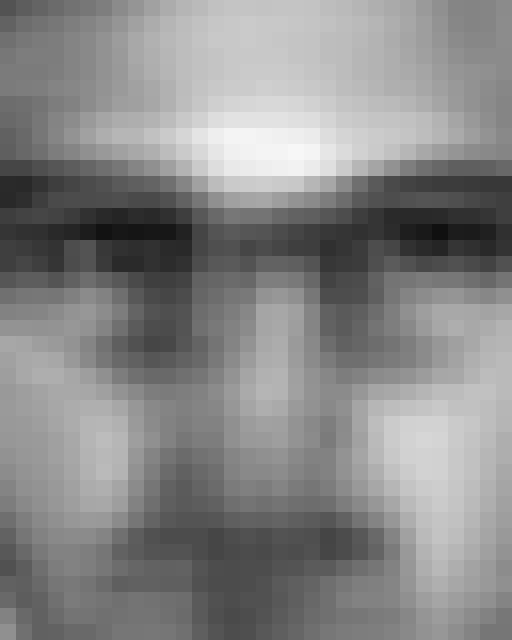}}\par\vspace{\vdiss cm}		
		\stepcounter{figure}\addtocounter{figure}{-1}
		\addtocounter{subfigure}{0}
		\subfloat[]{\includegraphics[width=\imw cm, height=\imh cm]{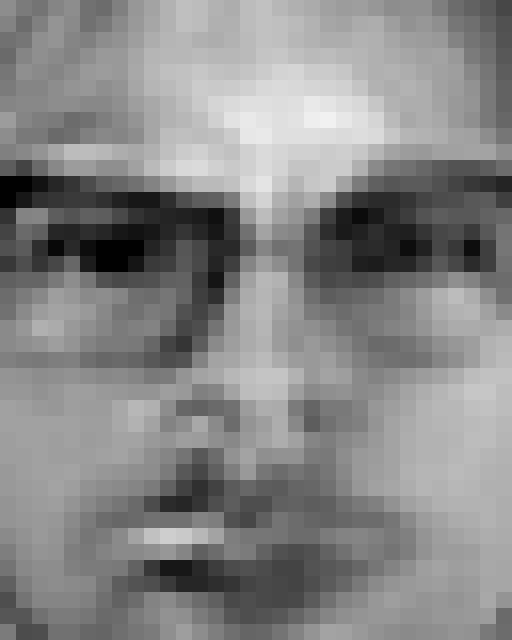}}
	\end{minipage}%
	\begin{minipage}{\hdis\textwidth} 
		\centering
		\subfloat {\includegraphics[width=\imw cm, height=\imh cm]{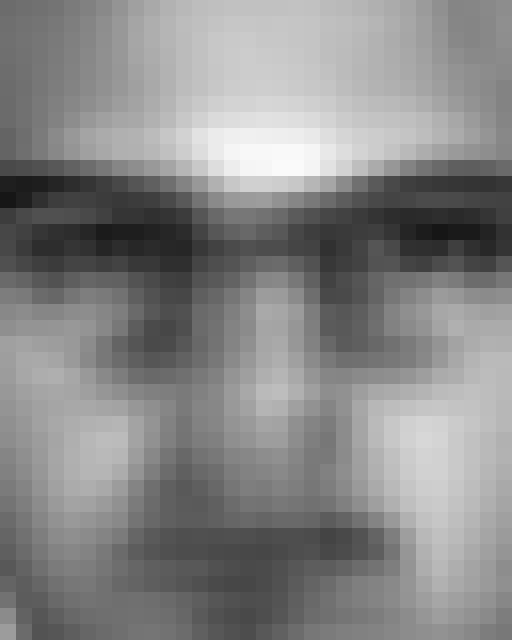}}\par\vspace{\vdiss cm}		
		\stepcounter{figure}\addtocounter{figure}{-1}
		\addtocounter{subfigure}{1}
		\subfloat[]{\includegraphics[width=\imw cm, height=\imh cm]{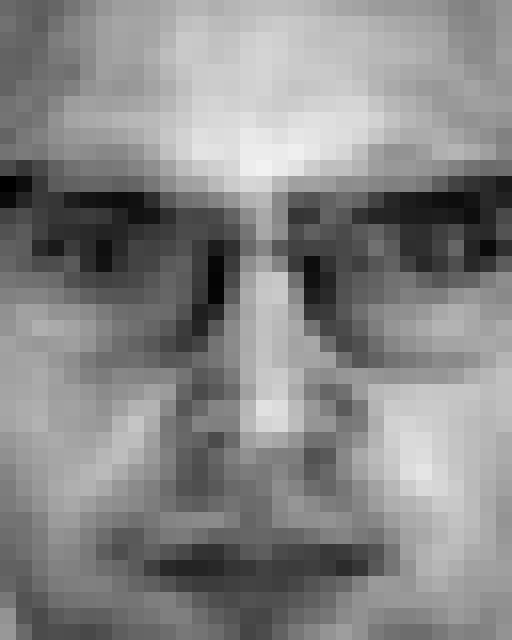}}
	\end{minipage}%
	\begin{minipage}{\hdis\textwidth} 
		\centering		
		\subfloat {\includegraphics[width=\imw cm, height=\imh cm]{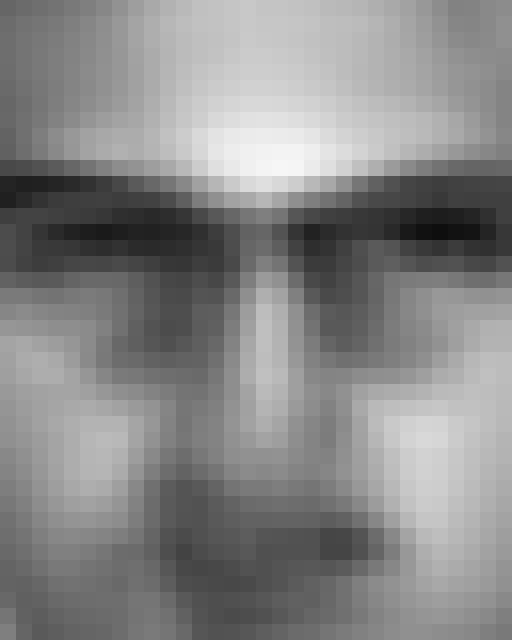}}\par\vspace{\vdiss cm}		
		\stepcounter{figure}\addtocounter{figure}{-1}
		\addtocounter{subfigure}{2}
		\subfloat[]{\includegraphics[width=\imw cm, height=\imh cm]{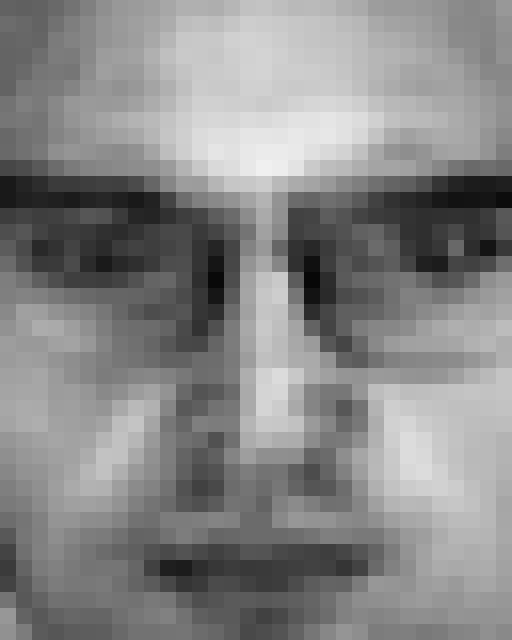}}
	\end{minipage}%
	\begin{minipage}{\hdis\textwidth} 
		\centering
		\subfloat {\includegraphics[width=\imw cm, height=\imh cm]{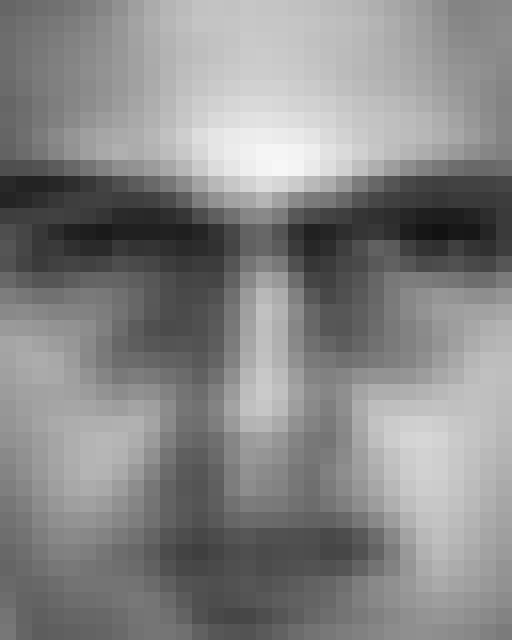}}\par\vspace{\vdiss cm}		
		\stepcounter{figure}\addtocounter{figure}{-1}
		\addtocounter{subfigure}{3}
		\subfloat[]{\includegraphics[width=\imw cm, height=\imh cm]{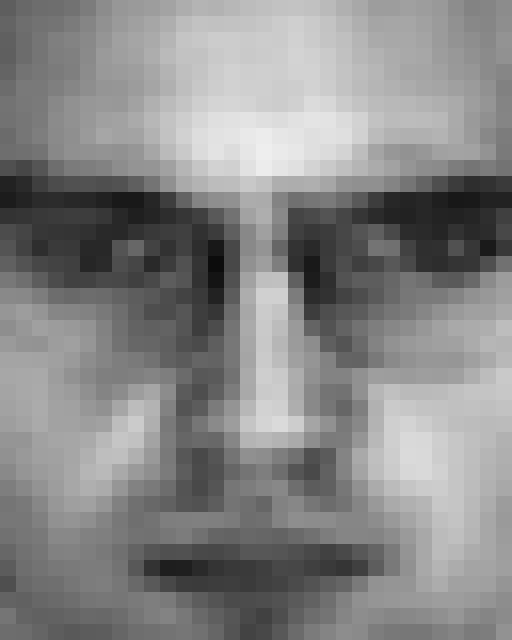}}
	\end{minipage}%
	\begin{minipage}{\hdis\textwidth} 
		\centering
		\subfloat {\includegraphics[width=\imw cm, height=\imh cm]{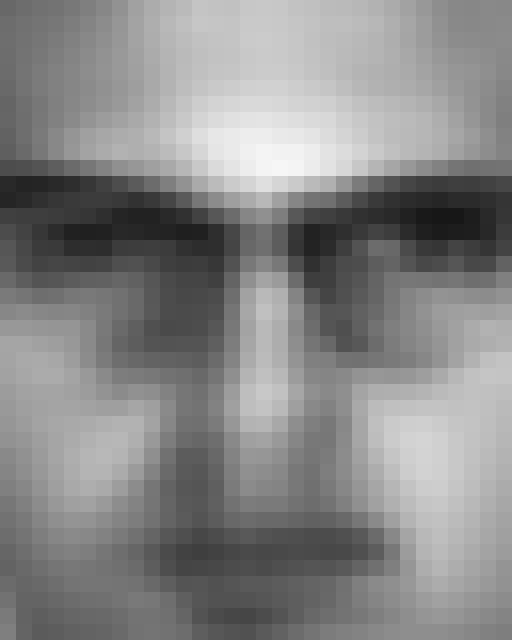}}\par\vspace{\vdiss cm}
		\stepcounter{figure}\addtocounter{figure}{-1}
		\addtocounter{subfigure}{4}
		\subfloat[]{\includegraphics[width=\imw cm, height=\imh cm]{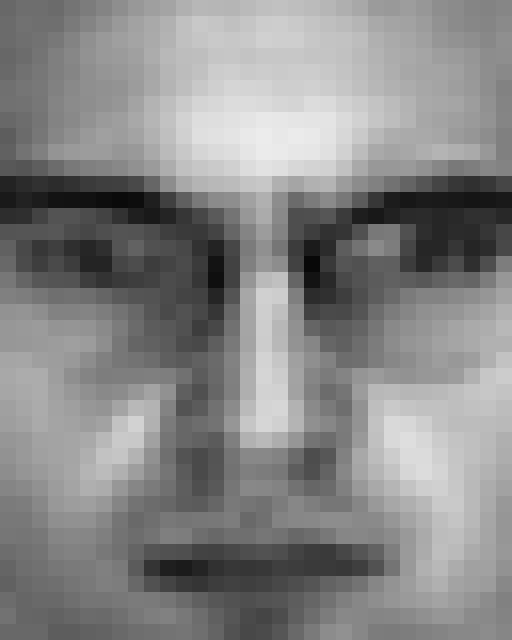}}
	\end{minipage}%
	\begin{minipage}{\hdis\textwidth} 
		\centering
		\subfloat {\includegraphics[width=\imw cm, height=\imh cm]{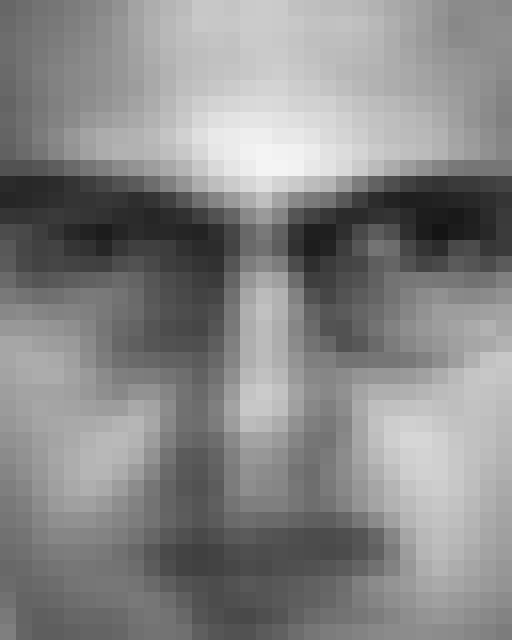}}\par\vspace{\vdiss cm}
		\stepcounter{figure}\addtocounter{figure}{-1}
		\addtocounter{subfigure}{5}
		\subfloat[]{\includegraphics[width=\imw cm, height=\imh cm]{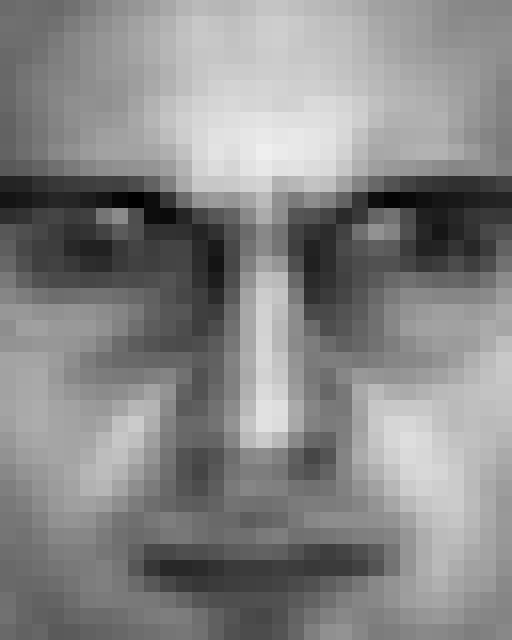}}
	\end{minipage}%
	\begin{minipage}{\hdis\textwidth} 
		\centering
		\par\vspace{1.65 cm} 
		\stepcounter{figure}\addtocounter{figure}{-1}
		\addtocounter{subfigure}{6}
		\subfloat[]{\includegraphics[width=\imw cm, height=\imh cm]{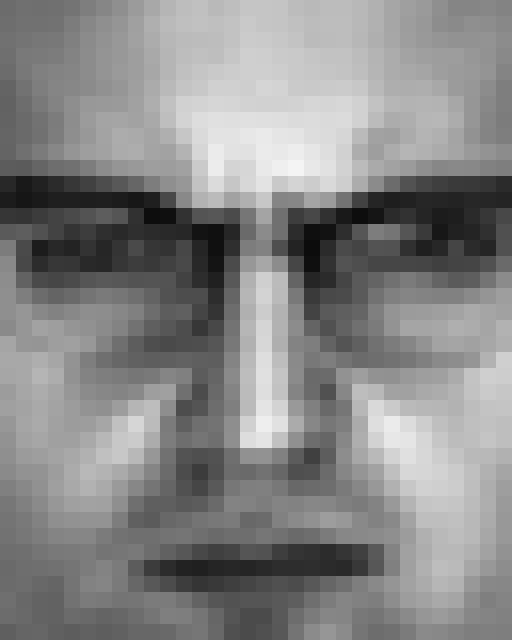}}
	\end{minipage}\par\medskip
	\caption{Qualitative influence of the number of samples per subject on the hallucinated face. (a)--(f) Reconstructed images by the proposed method (bottom) and TLcR-RL \cite{refs:Jiang2018} (top) with 1, 2, 3, 4, 7, and 13 samples per subject, respectively. (g) Ground truth.} 
	\label{fig:spsexample}
\end{figure}

\subsubsection{Number of Iterations}
\begin{figure}
	\captionsetup[subfloat]{farskip=0pt,captionskip=1pt}
	\centering
	\def\theight{0.18}
	\subfloat{\includegraphics[height=\theight\textwidth]{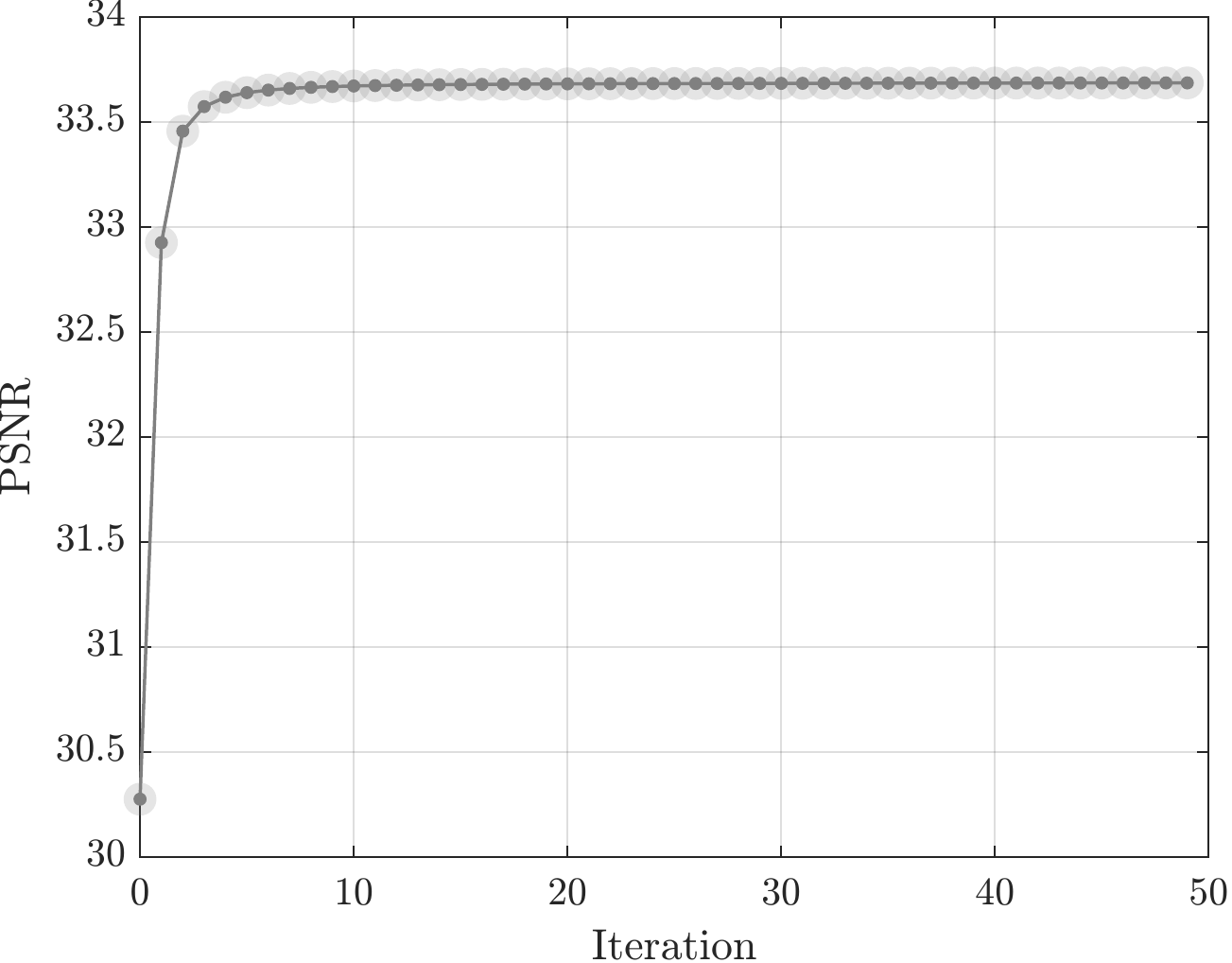}}\hfil
	\subfloat{\includegraphics[height=\theight\textwidth]{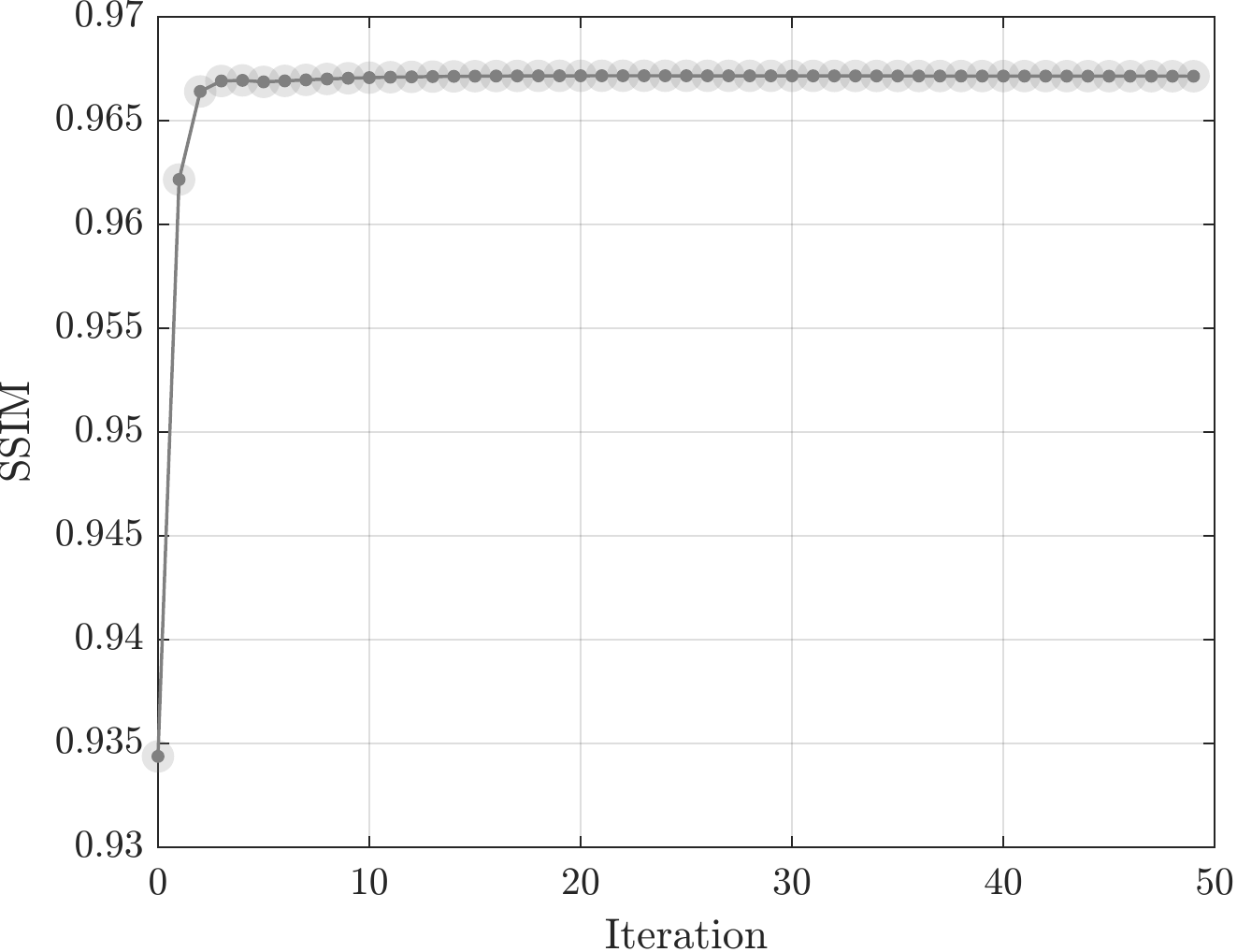}}
	\caption{Improvements of PSNR and SSIM values in different iterations.}
	\label{fig:itr}
\end{figure}

To investigate the influence of the face subspace prior, it would be worthwhile to compare the average PSNR and SSIM values across different iterations. As shown in Fig. \ref{fig:itr}, the two quantitative metrics improve dramatically in the early iterations, reaching to 32.92 dB in PSNR and 0.9622 in SSIM in a single iteration (2.65 dB and 0.0278 improvements, respectively), and surpassing 33.64 dB in PSNR and 0.9670 in SSIM after the first five iterations (3.36 dB and 0.0326 improvements, respectively), with respect to the initial state, i.e., $\hat{\textbf{x}} \approx \textbf{D}_h\boldsymbol\alpha_0$. The growth of these two indicators becomes stable roughly at iteration 30, thus, it will be considered as the number of iterations in our experiments. 

\subsubsection{Blur Kernel Size}
\begin{figure}
	\captionsetup[subfloat]{farskip=0pt,captionskip=1pt}
	\centering
	\def\theight{0.18}
	\subfloat{\includegraphics[height=\theight\textwidth]{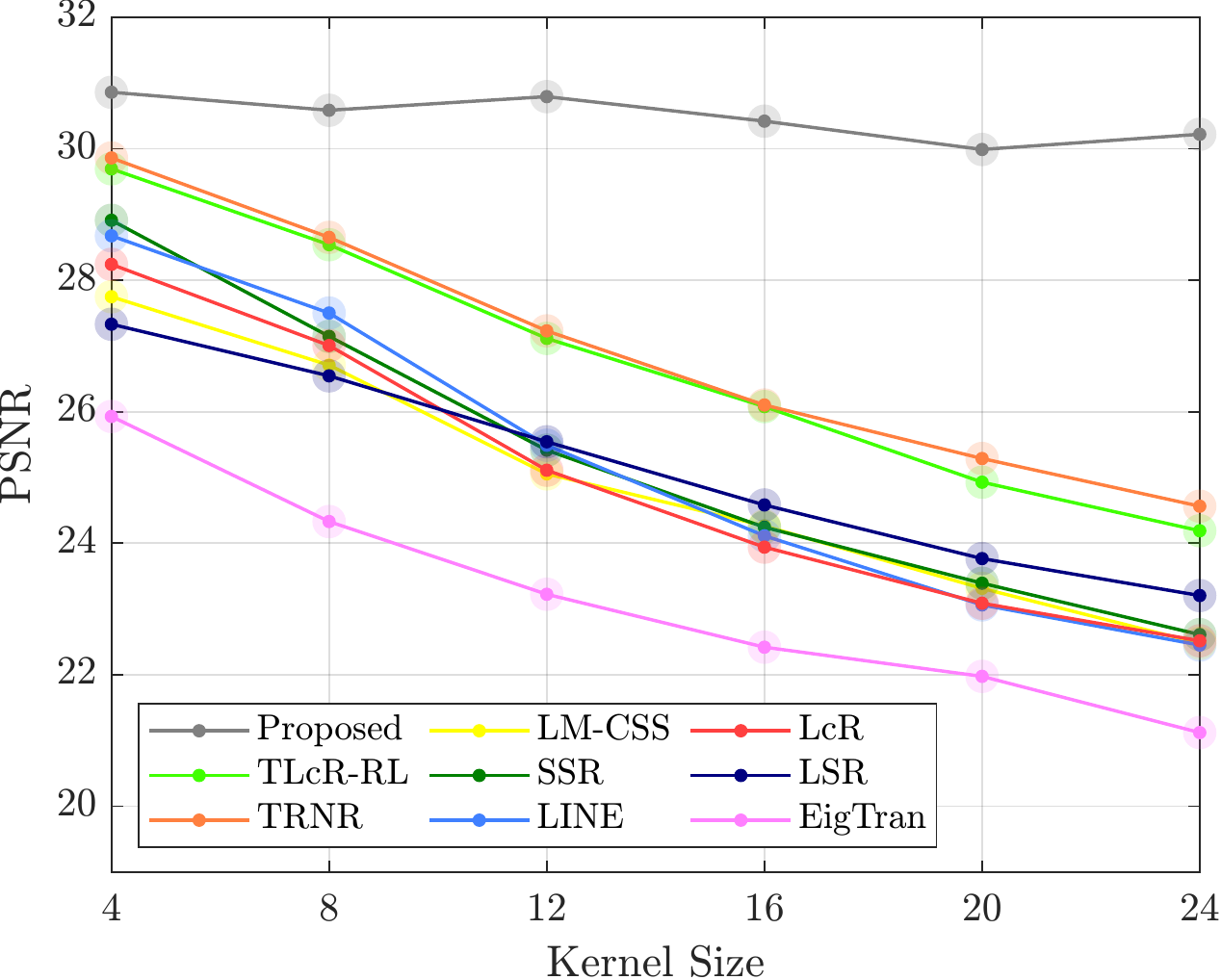}}\hfil
	\subfloat{\includegraphics[height=\theight\textwidth]{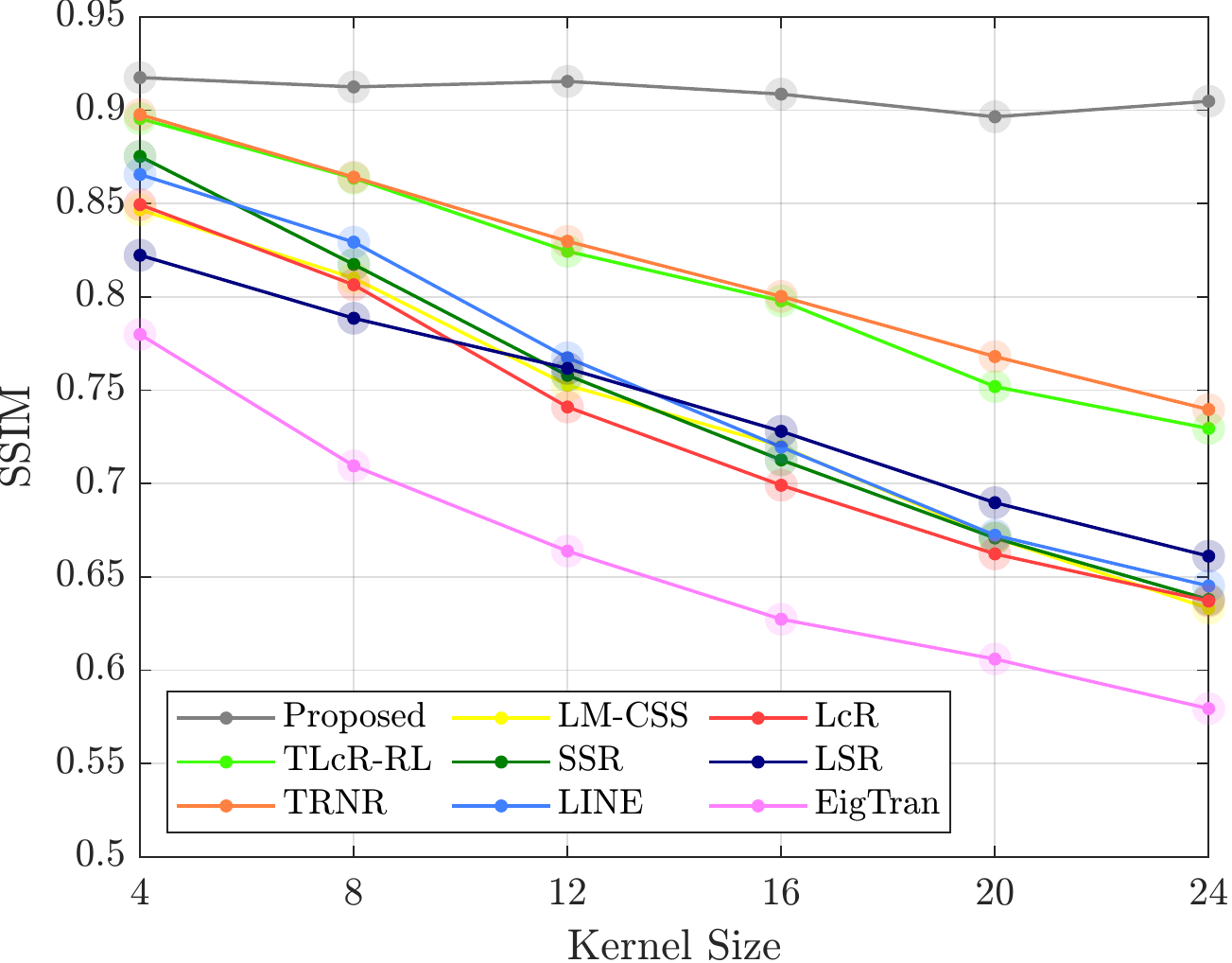}}
	\caption{Impact of blur kernel size on the performance of different algorithms.}
	\label{fig:deg}
\end{figure}

We further test the performance of our approach against various levels of degradation to measure how robust the proposed algorithm is when less facial information is available. We set the LR input size and the upsampling factor to be $12 \times 9$ and 4, respectively, and perform several experiments with average blur kernel of different sizes. The quantitative results of different approaches in this experiment are represented in Fig. \ref{fig:deg}. Despite the significant decline in the performance of the other methods, our algorithm is barely affected by the increase in blur kernel size, and even in extreme cases, its quantitative measures remain more or less the same. As discussed in section \ref{sec:intro}, as a result of selecting inappropriate neighboring patches, the performance of the patch-based face hallucination methods is prone to be seriously affected when LR input images become more degraded.

\subsection{Influence of the Face Subspace Prior}
\begin{figure}
	\captionsetup[subfloat]{farskip=0pt,captionskip=1pt}
	\centering
	\def\theight{0.18}
	\subfloat{\includegraphics[height=\theight\textwidth]{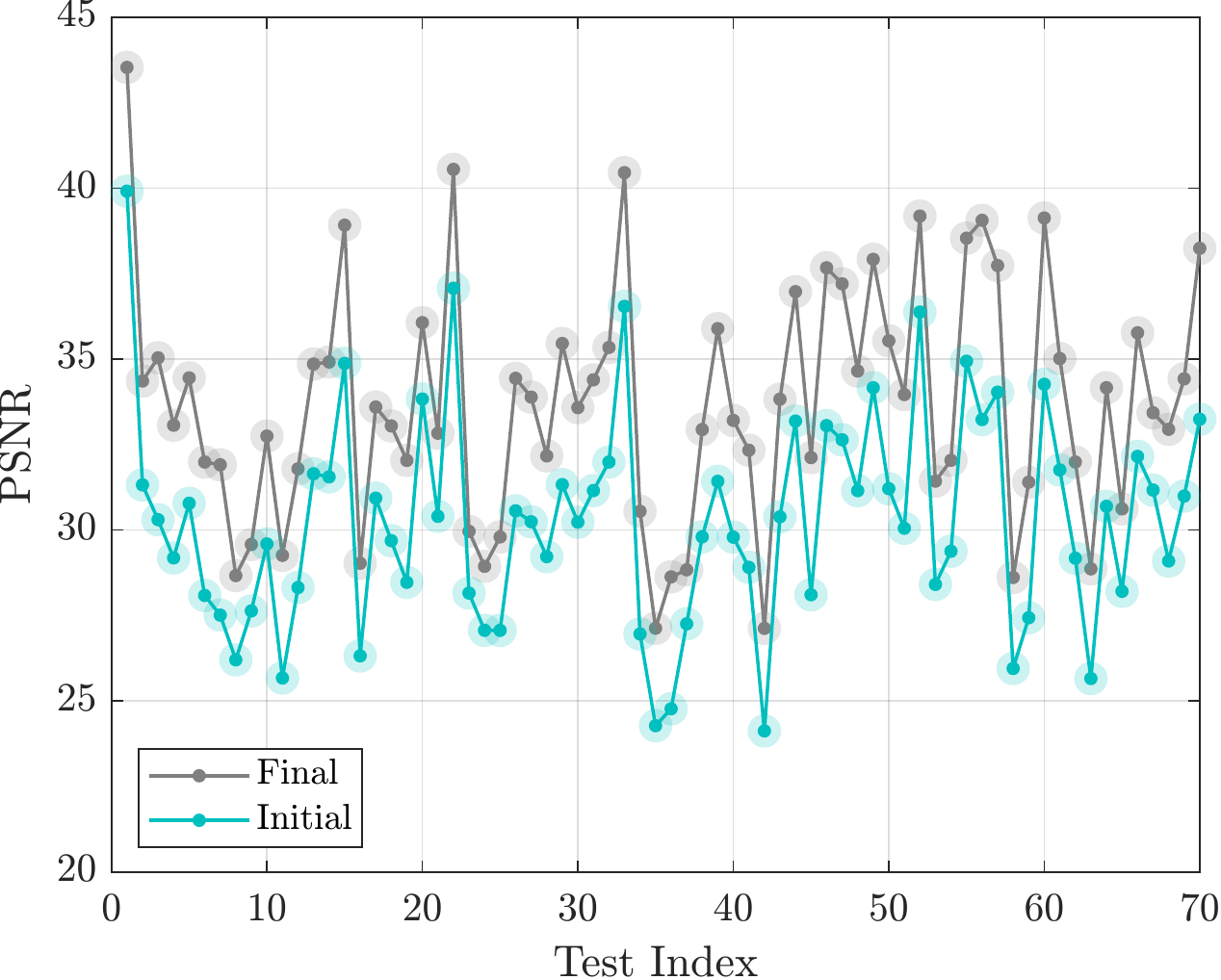}}\hfil
	\subfloat{\includegraphics[height=\theight\textwidth]{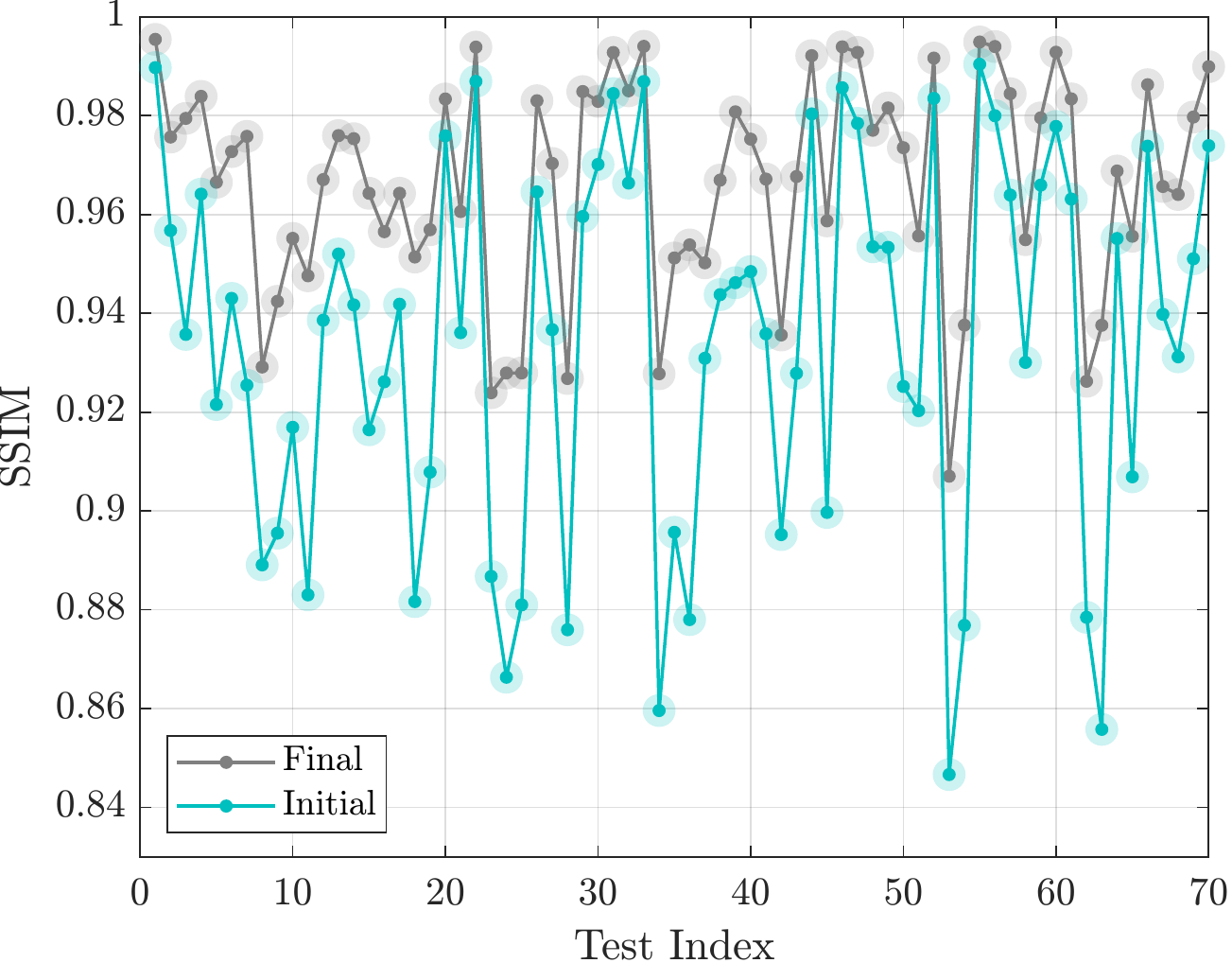}}
	\caption{Comparison of quantitative scores of the proposed algorithm in its initial and final iterations on each specific test sample in the FERET database.}
	\label{fig:fspquan}
\end{figure}

\begin{figure}
	\def\hdis{0.071}
	\def\vdiss{0.04}
	\def\caphdis{0.03}
	\def\capvdis{1.0}
	\def\imw{1.05}
	\def\imh{1.47} 
	\begin{minipage}[c]{\caphdis\textwidth}
		\subfloat[]{}\par\vspace{\capvdis cm}
		\subfloat[]{}\par\vspace{\capvdis cm}
		\subfloat[]{}
	\end{minipage}
	\begin{minipage}[c]{\hdis\textwidth}
		\includegraphics[width=\imw\textwidth]{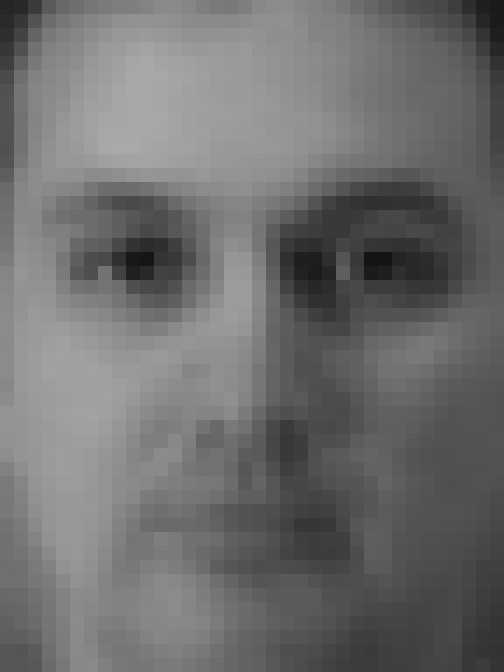}\par\vspace{\vdiss cm}
		\includegraphics[width=\imw\textwidth]{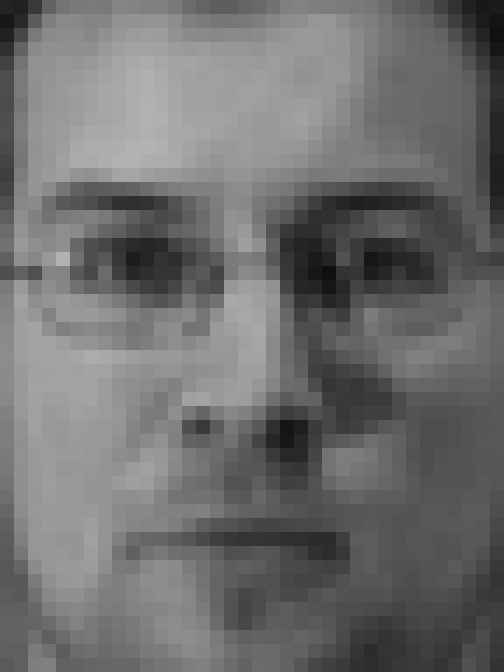}\par\vspace{\vdiss cm}
		\includegraphics[width=\imw\textwidth]{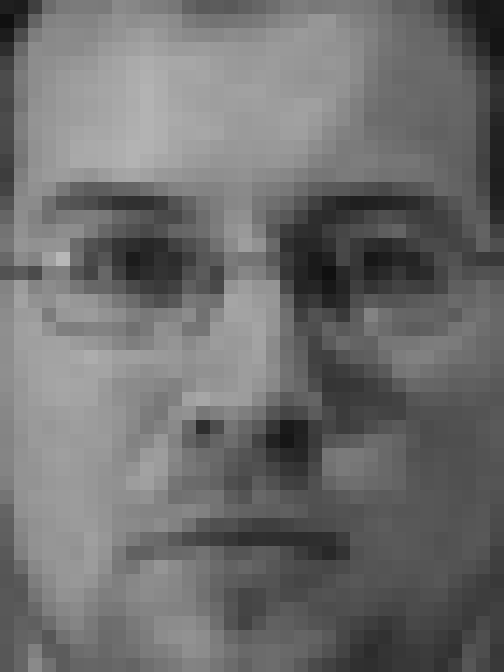}
	\end{minipage}
	\begin{minipage}[c]{\hdis\textwidth}
		\includegraphics[width=\imw\textwidth]{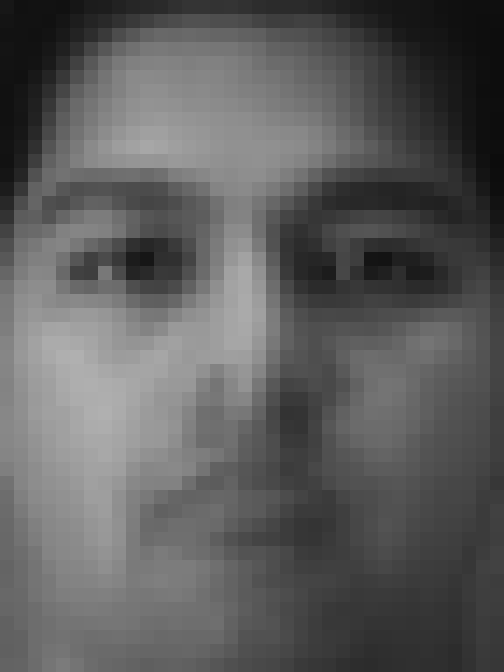}\par\vspace{\vdiss cm}
		\includegraphics[width=\imw\textwidth]{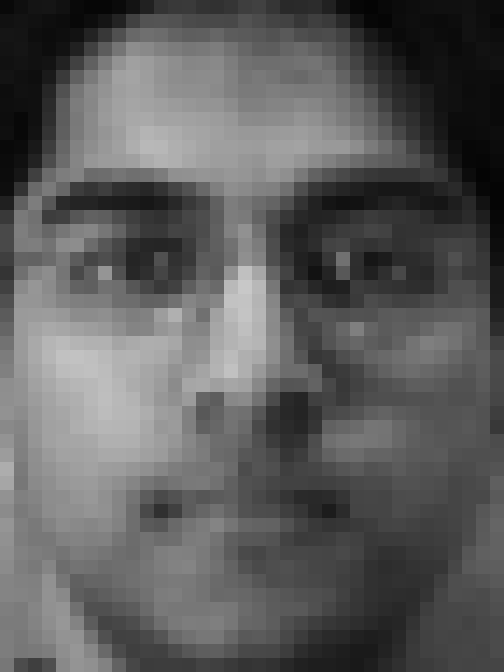}\par\vspace{\vdiss cm}
		\includegraphics[width=\imw\textwidth]{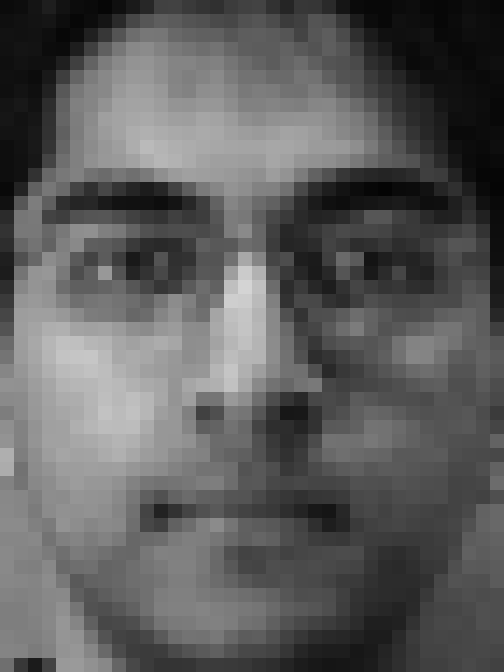}
	\end{minipage}
	\begin{minipage}[c]{\hdis\textwidth}
		\includegraphics[width=\imw\textwidth]{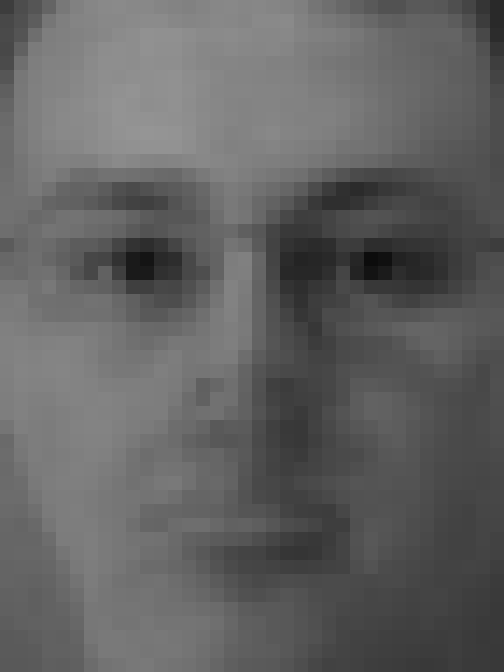}\par\vspace{\vdiss cm}
		\includegraphics[width=\imw\textwidth]{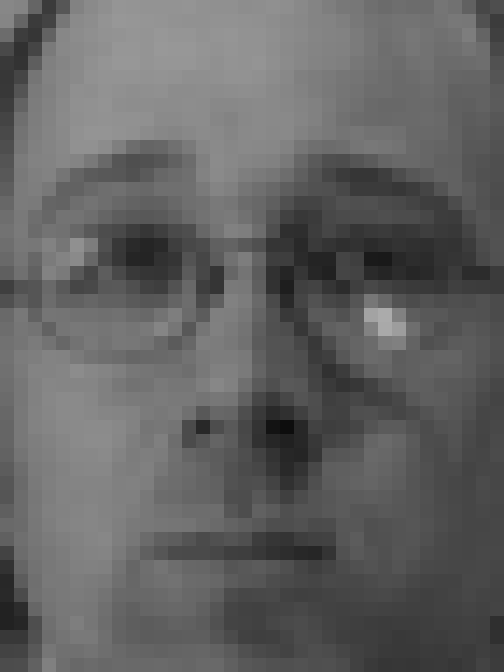}\par\vspace{\vdiss cm}
		\includegraphics[width=\imw\textwidth]{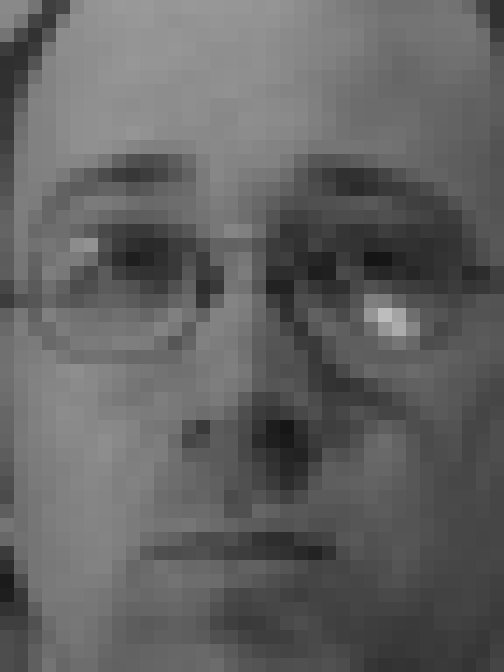}
	\end{minipage}
	\begin{minipage}[c]{\hdis\textwidth}
		\includegraphics[width=\imw\textwidth]{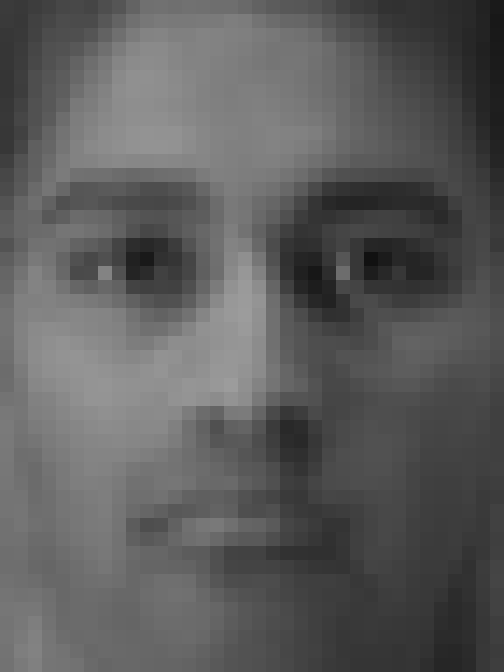}\par\vspace{\vdiss cm}
		\includegraphics[width=\imw\textwidth]{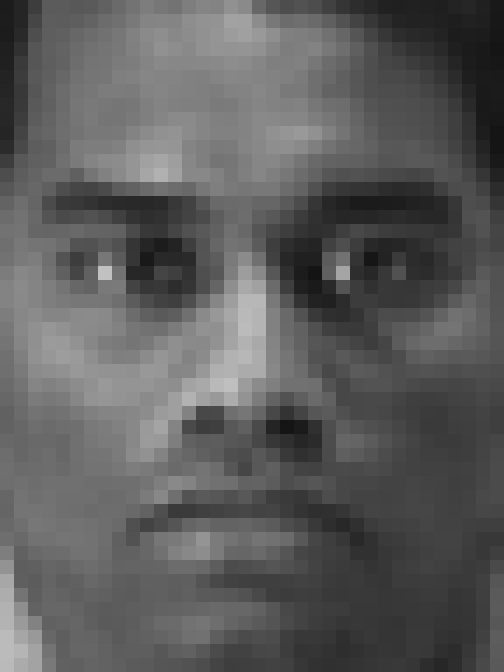}\par\vspace{\vdiss cm}
		\includegraphics[width=\imw\textwidth]{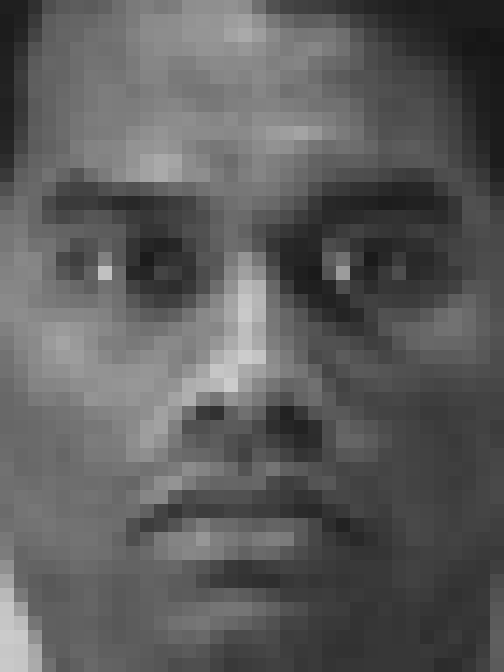}
	\end{minipage}
	\begin{minipage}[c]{\hdis\textwidth}
		\includegraphics[width=\imw\textwidth]{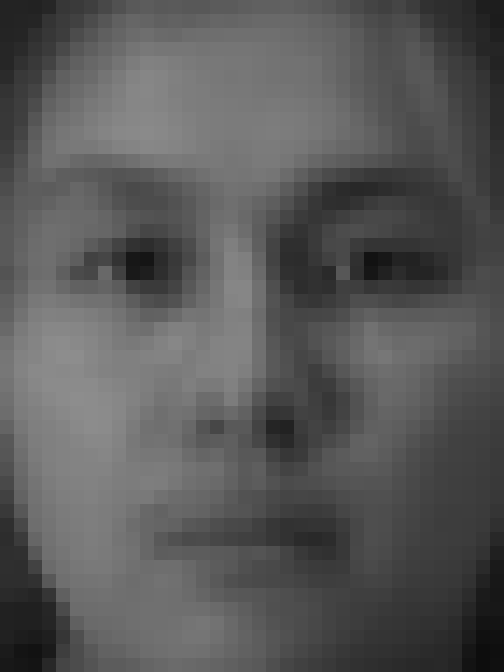}\par\vspace{\vdiss cm}
		\includegraphics[width=\imw\textwidth]{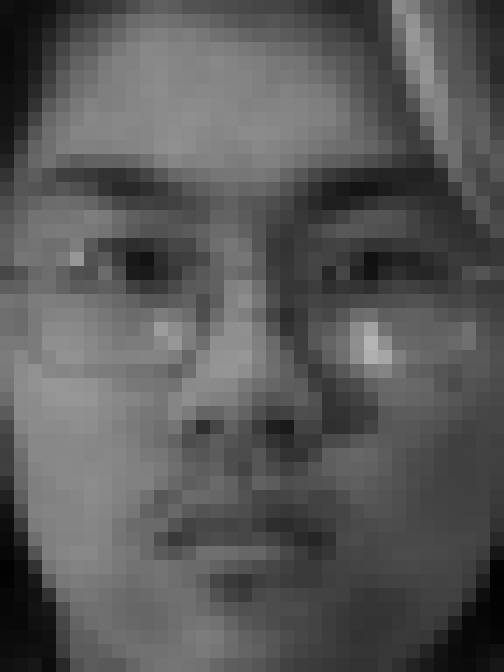}\par\vspace{\vdiss cm}
		\includegraphics[width=\imw\textwidth]{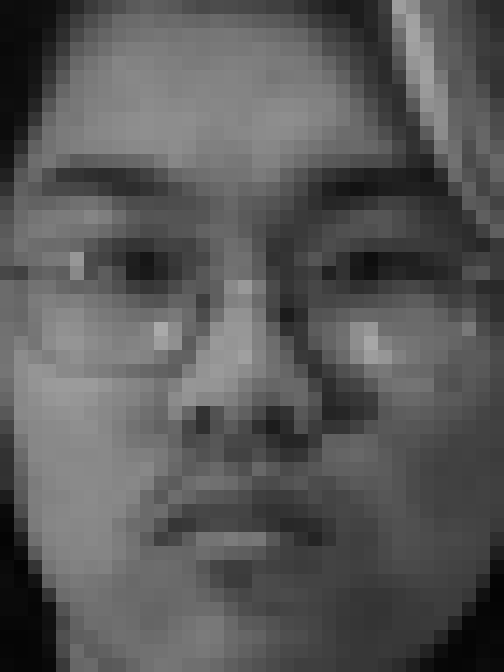}
	\end{minipage}
	\begin{minipage}[c]{\hdis\textwidth}
		\includegraphics[width=\imw\textwidth]{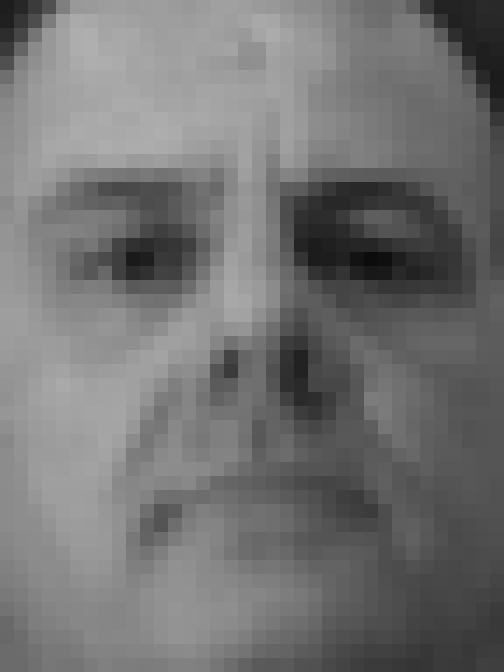}\par\vspace{\vdiss cm}
		\includegraphics[width=\imw\textwidth]{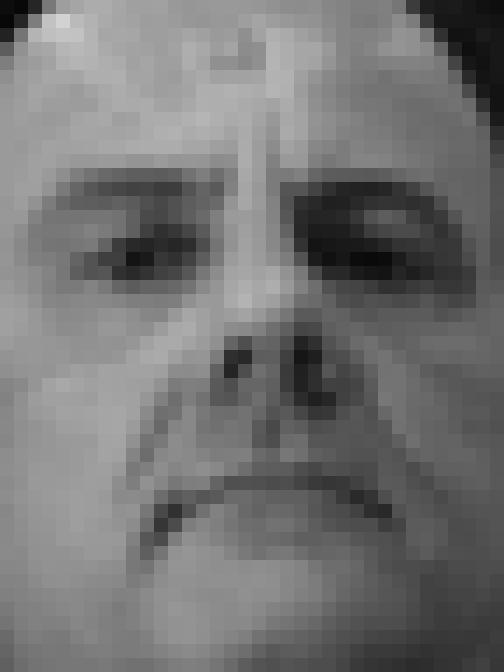}\par\vspace{\vdiss cm}
		\includegraphics[width=\imw\textwidth]{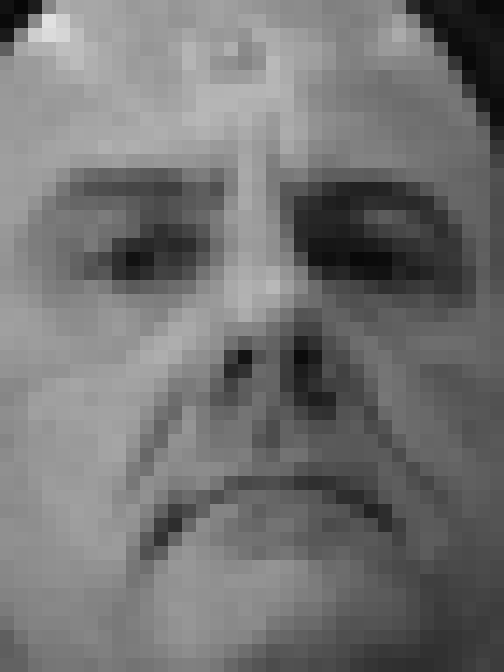}
	\end{minipage}
	\caption{Visual comparison of the initial and final hallucinated faces. (a) Initial reconstruction. (b) Final reconstruction. (c) Ground truth.}
	\label{fig:fspquali}
\end{figure}

The effectiveness of the face subspace prior in the proposed approach is further evaluated by making more quantitative and qualitative comparisons between the final reconstructed face obtained by our algorithm and its initial state, which is the case when only the concept of neighbor embedding is taken for granted and the hallucinated face is equal to $\textbf{D}_h\boldsymbol\alpha_0$. Fig. \ref{fig:fspquan} displays the quantitative indicators for the test faces in the FERET database. The linear embedding-based approach achieves the average 30.29 dB in PSNR and 0.9344 in SSIM, which are lower than the proposed algorithm by 3.42 dB and 0.0323, respectively. This substantial difference between the initial and final states of the algorithm is the result of the improvements made by the face subspace prior, which has also been clearly reflected in the facial details recovered by the two approaches. According to the examples shown in Fig. \ref{fig:fspquali}, in the initial phase of the algorithm, only basic structures of the faces are recovered, and, in contrast to the final hallucination results, fine details such as eyeglasses, nose shape, eyebrows direction, and face pose are all ignored. As a consequence of considering the faces merely as the linear combinations of the training samples, these images contain blurry regions and in some cases are by no means close to their ground truth counterparts.

\subsection{Face Recognition Accuracy}
To clarify the advantages of our algorithm in recovering person-specific facial features, we conduct experiments on the task of low-resolution face recognition. The evaluations are performed on the Multi-PIE and the AR face datasets with 130 and 100 subjects, respectively, such that each subject has only two samples in the training set. The size of the LR images in the Multi-PIE subset is $8\times6$, whereas the inputs in the AR subset are of size $10\times8$, and the scaling factor in both experiments is 4. LR faces are obtained by applying a $7\times7$ Gaussian filter with $\sigma = 2$ to the HR images, before downsampling them to the desired sizes. The resultant images of all methods are classified using the SRC classifier \cite{refs:Wright2009}. As for the proposed method, the final coefficient vector $\hat{\boldsymbol\alpha}$ is used in the classification. Table \ref{tab:recogacc} represents the recognition accuracy achieved by different approaches in both experiments, whereas Fig. \ref{fig:recogcmc} compares their cumulative recognition rates in the first five ranks. The recognition rates of the proposed method on the Multi-PIE and the AR datasets are 92.25\% and 76\%, outperforming the others by 8.53\% and 3\%, respectively. This clearly illustrates the effectiveness of the recognition-oriented aspect of our face hallucination algorithm, even when each subject has only two samples in the training set.

\begin{table}
	\caption{Evaluation of face recognition performance achieved by different algorithms on the Multi-PIE and the AR databases}
	\centering
	\begin{tabular}{c||cc} 
		\hline\hline
		Algorithm & Multi-PIE & AR \\ 
		\hline
		Wang \cite{refs:Wang2005} & 34.11 & 52.00 \\
		LSR \cite{refs:Ma2010} & 81.40 & 65.00 \\
		LcR \cite{refs:Jiang2014a} & 79.07 & 70.00 \\
		LINE \cite{refs:Jiang2014b} & 82.95 & 68.00 \\
		SSR \cite{refs:Jiang2017} & 65.89 & 73.00 \\
		LM-CSS \cite{refs:Farrugia2017} & 66.67 & 57.00 \\
		TRNR \cite{refs:Jiang2016} & 81.40 & 73.00 \\
		TLcR-RL \cite{refs:Jiang2018} & 83.72 & 70.00 \\
		Proposed & \textbf{92.25} & \textbf{76.00} \\
		HR & 98.45 & 88.00 \\		
		\hline\hline
	\end{tabular}
	\label{tab:recogacc}
\end{table}

\begin{figure}
	\captionsetup[subfloat]{farskip=0pt,captionskip=1pt}
	\centering
	\def\theight{0.18}
	\subfloat[]{\includegraphics[height=\theight\textwidth]{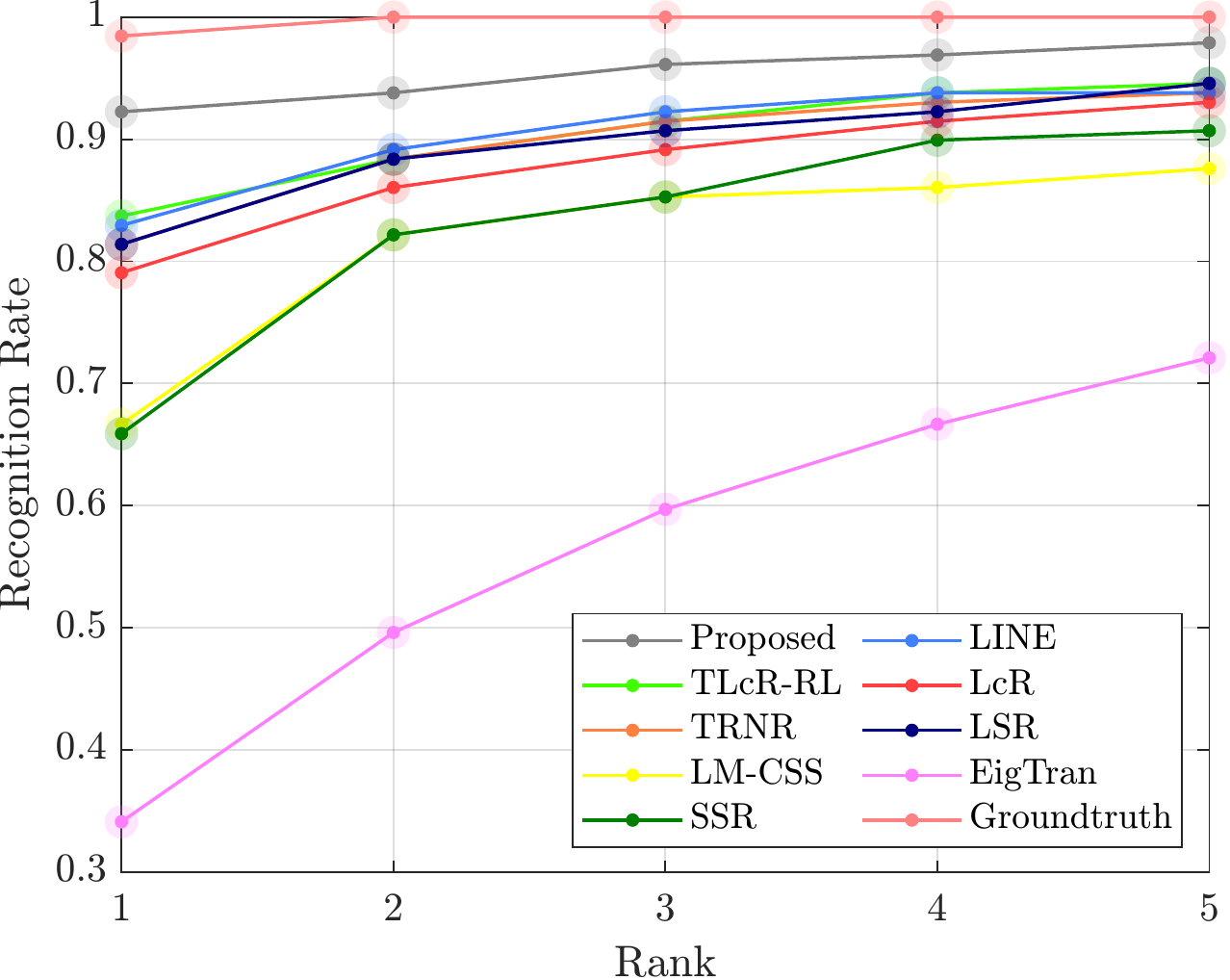}}\hfil
	\subfloat[]{\includegraphics[height=\theight\textwidth]{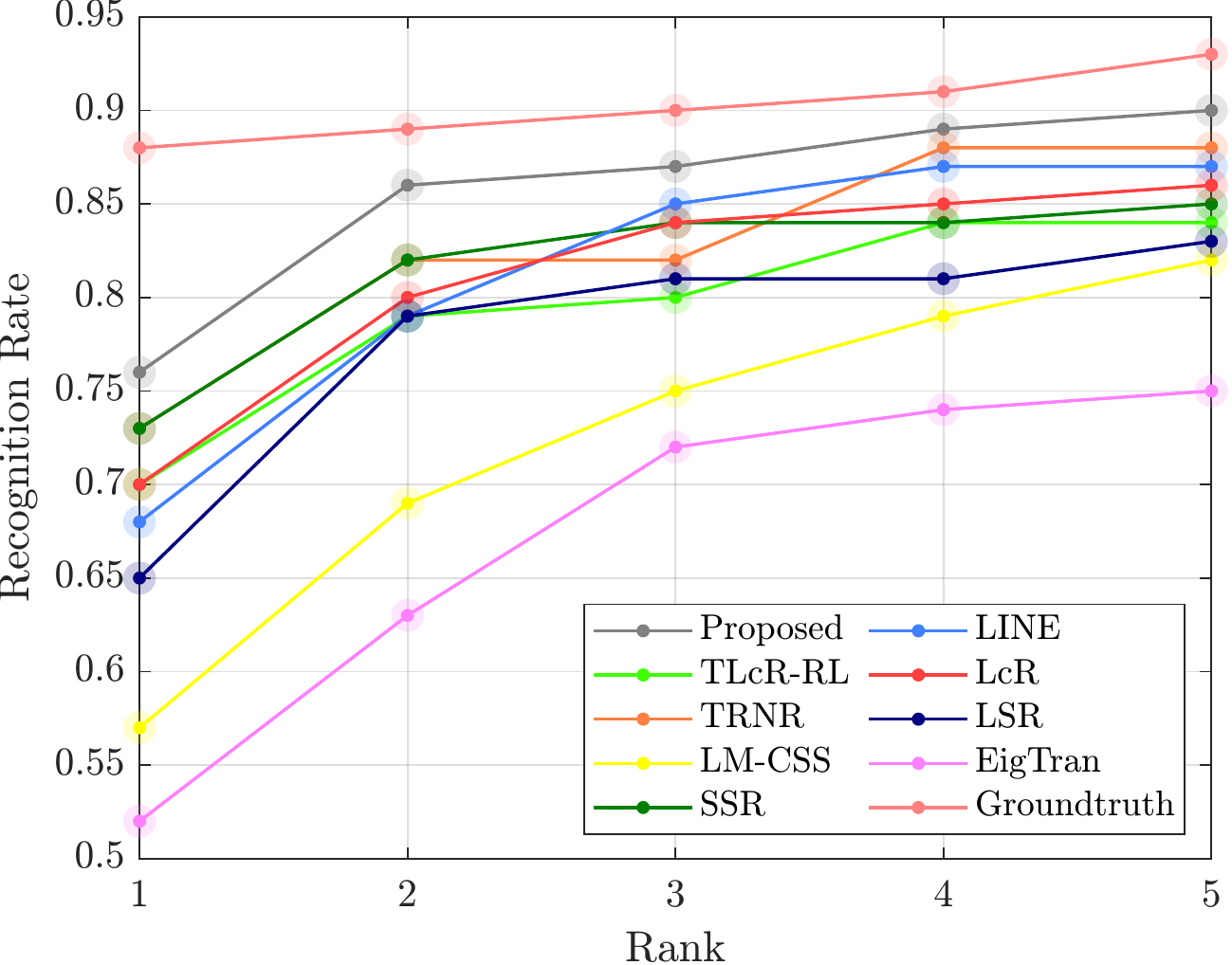}}
	\caption{Cumulative recognition rates versus rank achieved by different methods. (a) The Multi-PIE dataset. (b) The AR dataset.}
	\label{fig:recogcmc}
\end{figure}

\subsection{Comparison with Deep Learning-Based Methods}
We further evaluate our algorithm by comparing its performance on color images with several popular and/or successful deep learning-based approaches. We consider SRCNN \cite{refs:Dong2016} as a baseline algorithm along with DCSCN \cite{refs:Yamanaka2017}, DBPN \cite{refs:Haris2018}, ESRGAN \cite{refs:Wang2018}, SPSR \cite{refs:Ma2020}, SRGAN \cite{refs:Zhang2021b}, realESRGAN \cite{refs:Wang2021}, SwinIR \cite{refs:Liang2021}, and PULSE \cite{refs:Menon2020}, of which the latter is a recently developed face hallucination technique. CNN-based methods were trained using our data, whereas the remaining networks utilize their own pre-trained models. To alleviate this, the LR inputs provided for these networks are only the downsampled (and not blurred) version of the HR images. Our color results are obtained by performing the proposed method separately on each of the RGB channels. We perform the experiments on the Multi-PIE face database, and consider LR face images of size $16 \times 12$ with scaling factor 4. The PSNR and SSIM evaluation of the algorithms is reported in Table \ref{tab:deepquan}, and some hallucinated faces generated by different methods are also displayed in Fig. \ref{fig:deepqual}. It appears that SRCNN manages to enhance parts of the face images, but leaves undesired artifacts around the boundary regions. DCSCN and DBPN have done only slightly better than the bicubic interpolation, whereas the results produced by GAN-based algorithms are deformed and unclear. PULSE generates noise-free faces which hardly resemble their true identities. In general, despite achieving satisfactory results on higher resolutions, deep learning-based super-resolution algorithms seem to fail to enhance very low-resolution face images, with their results being vague and distorted.

\begin{table}
	\caption{Quantitative scores obtained by several deep learning-based algorithms and the proposed method on the Multi-PIE dataset}
	\centering
	\begin{tabular}{c||cc} 
		\hline\hline
		Algorithm & PSNR & SSIM \\ 
		\hline
		SRCNN \cite{refs:Dong2016} & 20.52 & 0.6866 \\
		DCSCN \cite{refs:Yamanaka2017} & 30.33 & 0.8398 \\
		DBPN \cite{refs:Haris2018} & 30.56 & 0.8354 \\
		ESRGAN \cite{refs:Wang2018} & 28.20 & 0.7440 \\
		SPSR \cite{refs:Ma2020} & 28.28 & 0.7535 \\
		BSRGAN \cite{refs:Zhang2021b} & 25.66 & 0.6628 \\
		realESRGAN \cite{refs:Wang2021} & 24.44 & 0.6719 \\
		SwinIR \cite{refs:Liang2021} & 22.78 & 0.5586 \\
		Proposed & \textbf{35.01} & \textbf{0.9442} \\
		\hline\hline
	\end{tabular}
	\label{tab:deepquan}
\end{table}

\begin{figure*}[!t]
	\captionsetup[subfloat]{farskip=0pt,captionskip=1pt}
	\centering
	\def\imw{1.30}
	\def\imh{1.82}
	\def\hdis{0.075}
	\def\vdiss{0.04}
	\begin{minipage}{\hdis\textwidth} 
		\centering
		\subfloat {\includegraphics[width=\imw cm, height=\imh cm]{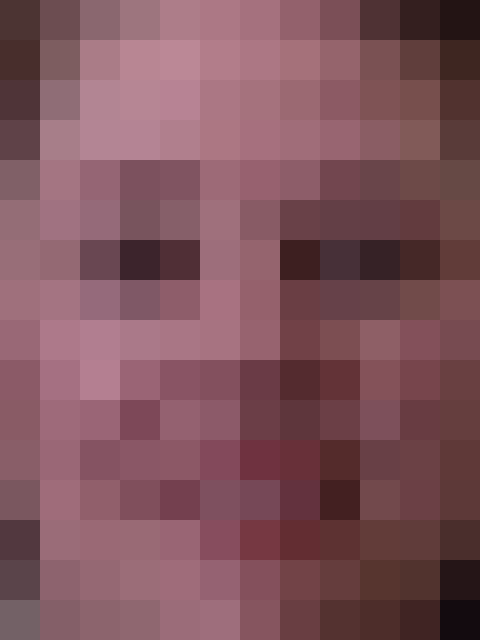}}\par\vspace{\vdiss cm}		
		\subfloat {\includegraphics[width=\imw cm, height=\imh cm]{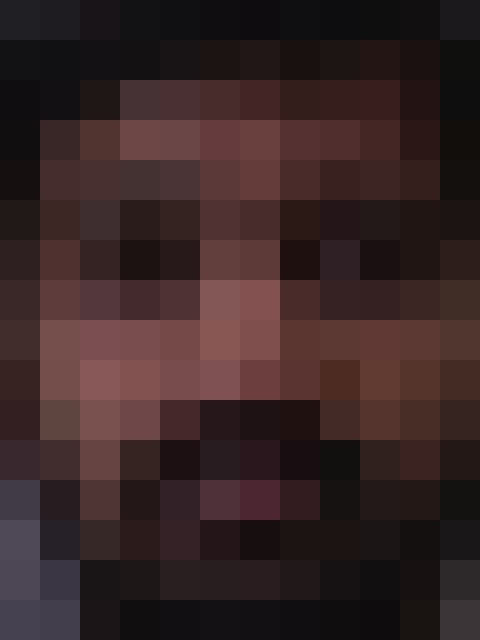}}\par\vspace{\vdiss cm}
		\subfloat {\includegraphics[width=\imw cm, height=\imh cm]{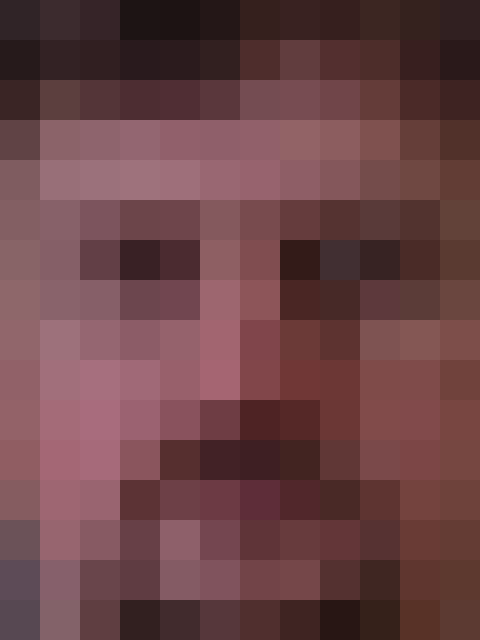}}\par\vspace{\vdiss cm}
		\subfloat {\includegraphics[width=\imw cm, height=\imh cm]{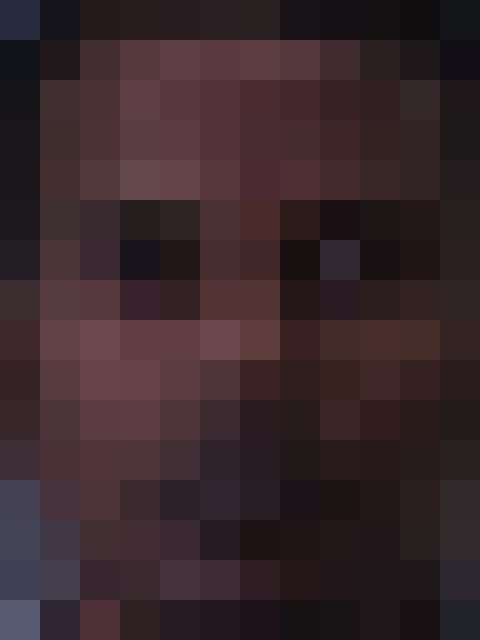}}\par\vspace{\vdiss cm}
		\stepcounter{figure}\addtocounter{figure}{-1}
		\addtocounter{subfigure}{0}
		\subfloat[]{\includegraphics[width=\imw cm, height=\imh cm]{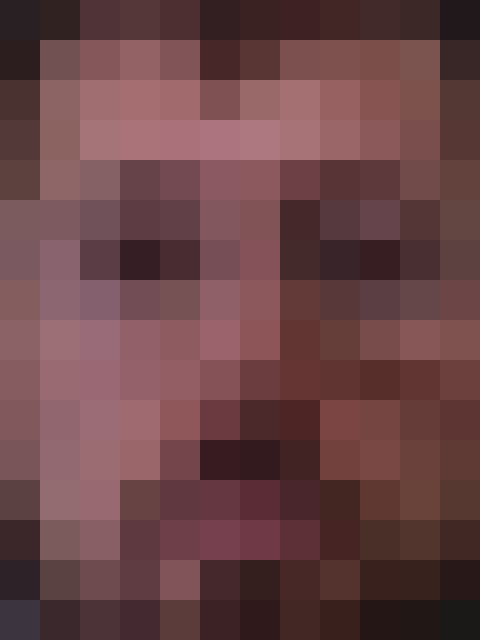}}
	\end{minipage}%
	\begin{minipage}{\hdis\textwidth} 
		\centering
		\subfloat {\includegraphics[width=\imw cm, height=\imh cm]{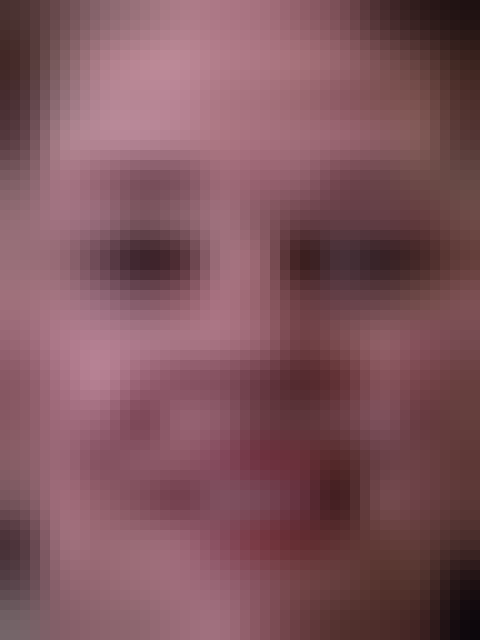}}\par\vspace{\vdiss cm}		
		\subfloat {\includegraphics[width=\imw cm, height=\imh cm]{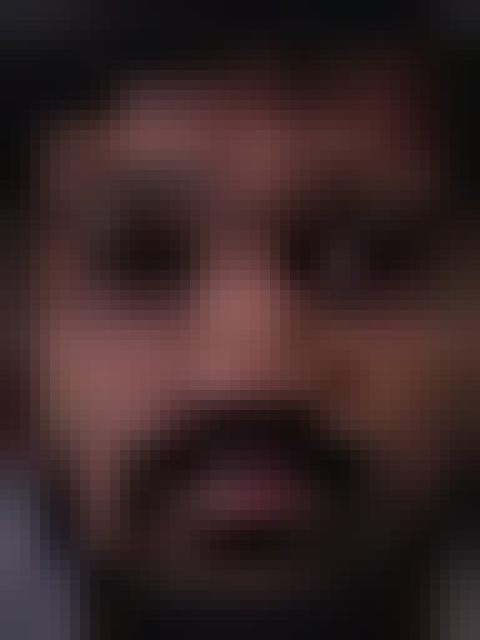}}\par\vspace{\vdiss cm}
		\subfloat {\includegraphics[width=\imw cm, height=\imh cm]{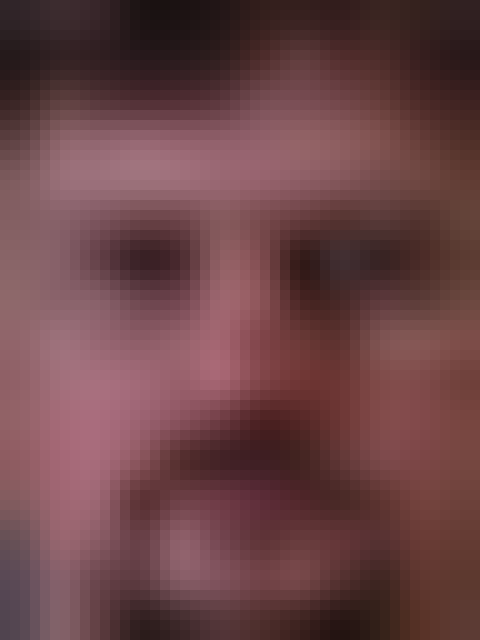}}\par\vspace{\vdiss cm}
		\subfloat {\includegraphics[width=\imw cm, height=\imh cm]{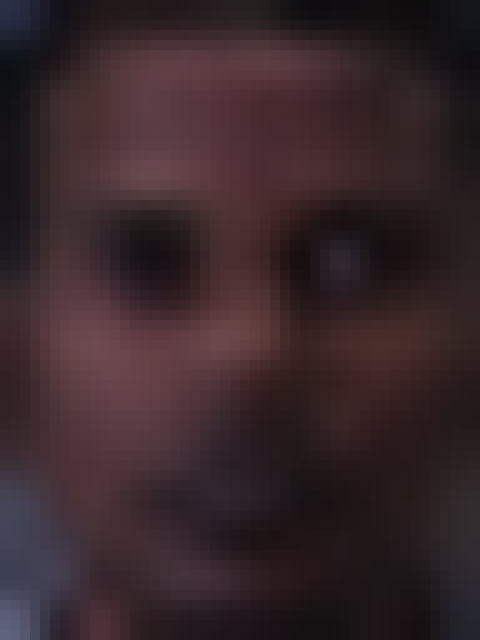}}\par\vspace{\vdiss cm}
		\stepcounter{figure}\addtocounter{figure}{-1}
		\addtocounter{subfigure}{1}
		\subfloat[]{\includegraphics[width=\imw cm, height=\imh cm]{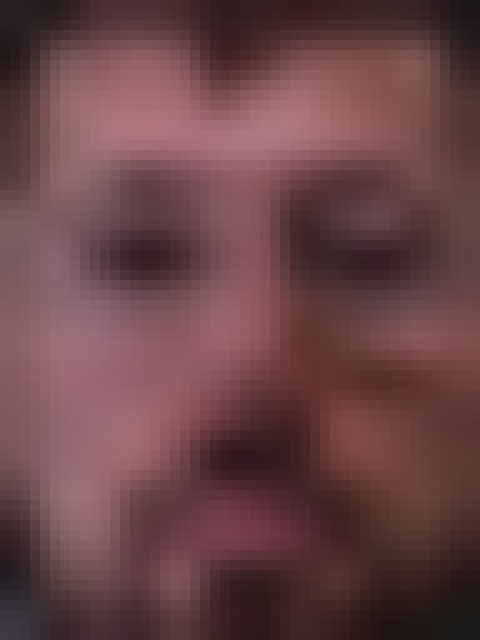}}
	\end{minipage}%
	\begin{minipage}{\hdis\textwidth} 
		\centering
		\subfloat {\includegraphics[width=\imw cm, height=\imh cm]{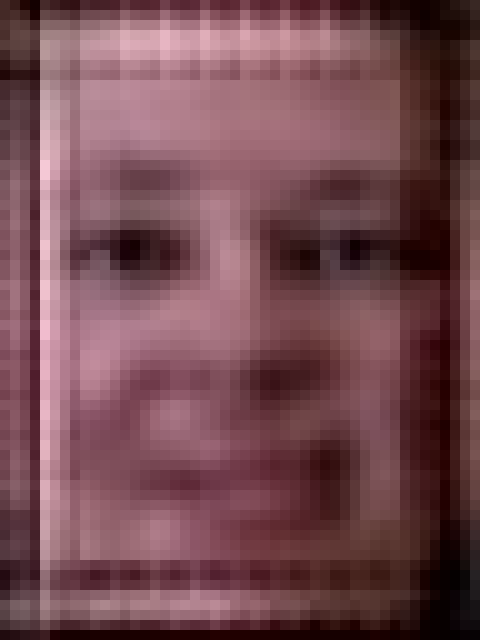}}\par\vspace{\vdiss cm}		
		\subfloat {\includegraphics[width=\imw cm, height=\imh cm]{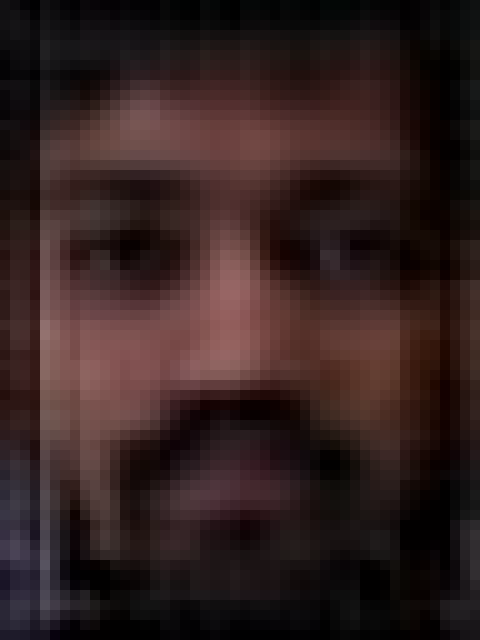}}\par\vspace{\vdiss cm}
		\subfloat {\includegraphics[width=\imw cm, height=\imh cm]{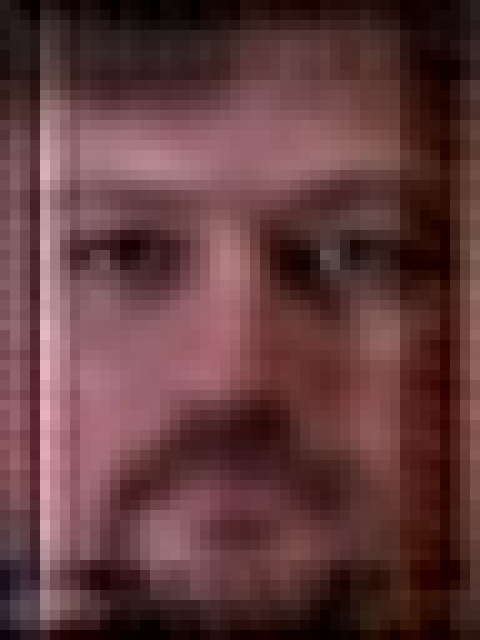}}\par\vspace{\vdiss cm}
		\subfloat {\includegraphics[width=\imw cm, height=\imh cm]{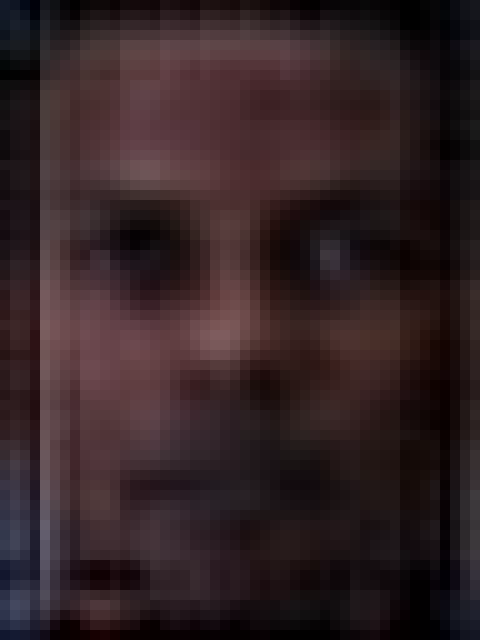}}\par\vspace{\vdiss cm}
		\stepcounter{figure}\addtocounter{figure}{-1}
		\addtocounter{subfigure}{2}
		\subfloat[]{\includegraphics[width=\imw cm, height=\imh cm]{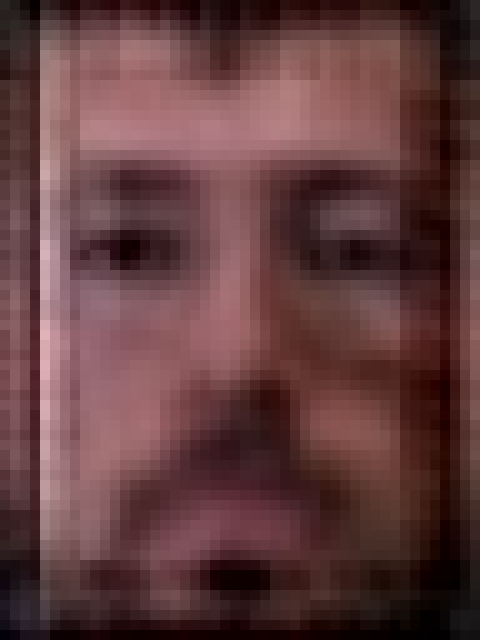}}
	\end{minipage}%
	\begin{minipage}{\hdis\textwidth} 
		\centering
		\subfloat {\includegraphics[width=\imw cm, height=\imh cm]{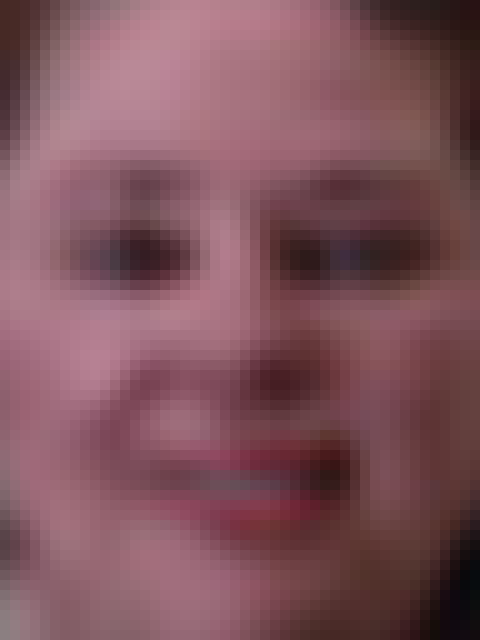}}\par\vspace{\vdiss cm}		
		\subfloat {\includegraphics[width=\imw cm, height=\imh cm]{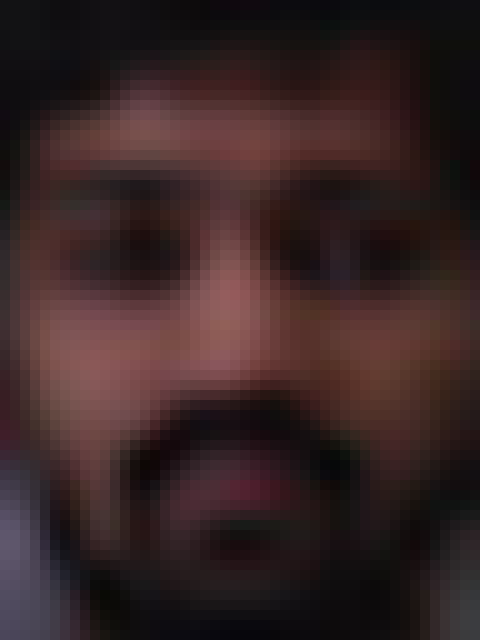}}\par\vspace{\vdiss cm}
		\subfloat {\includegraphics[width=\imw cm, height=\imh cm]{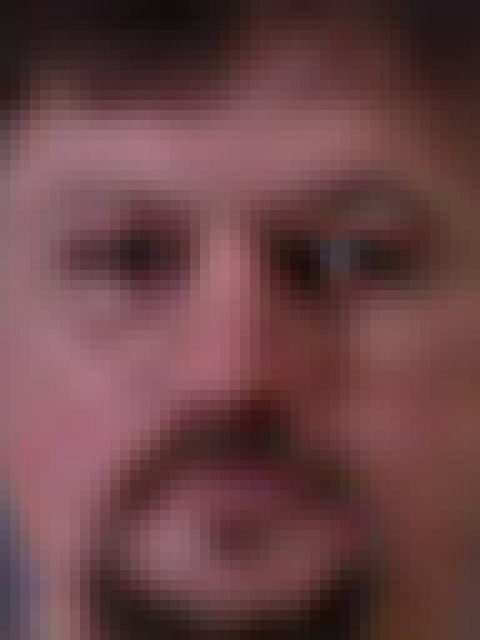}}\par\vspace{\vdiss cm}
		\subfloat {\includegraphics[width=\imw cm, height=\imh cm]{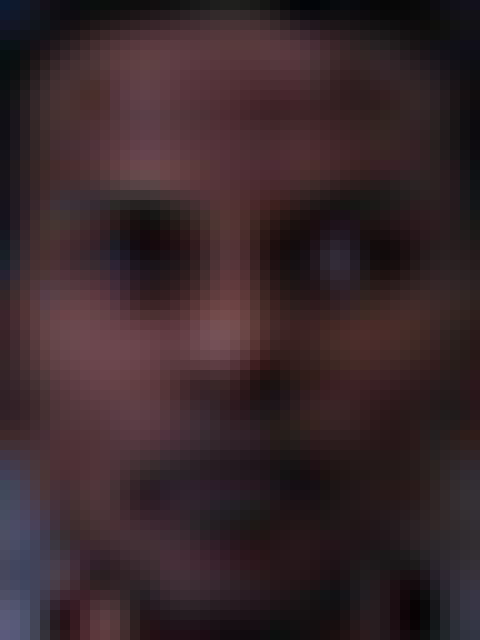}}\par\vspace{\vdiss cm}
		\stepcounter{figure}\addtocounter{figure}{-1}
		\addtocounter{subfigure}{3}
		\subfloat[]{\includegraphics[width=\imw cm, height=\imh cm]{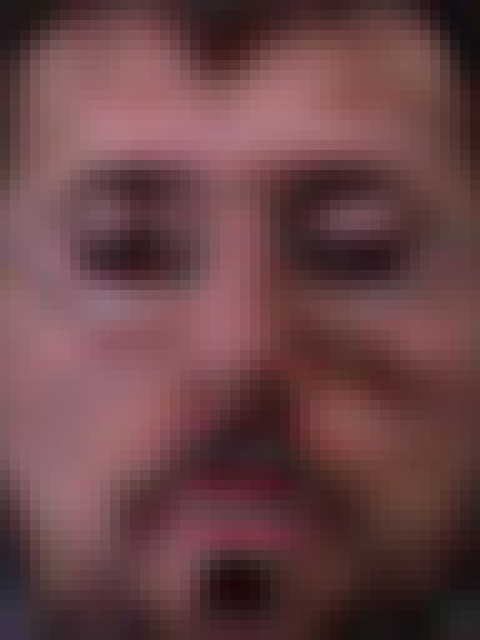}}
	\end{minipage}%
	\begin{minipage}{\hdis\textwidth} 
		\centering
		\subfloat {\includegraphics[width=\imw cm, height=\imh cm]{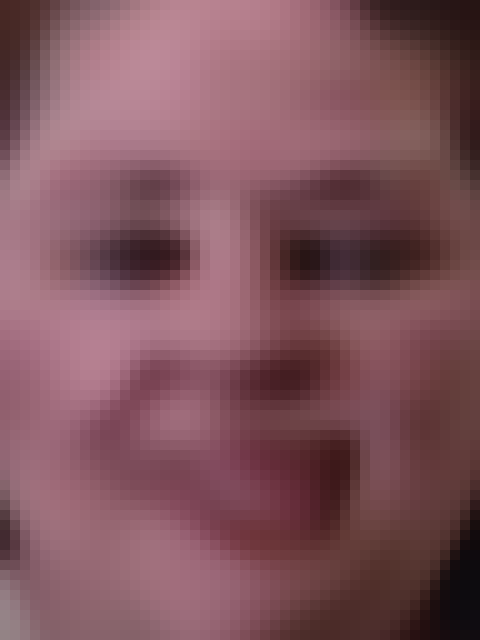}}\par\vspace{\vdiss cm}		
		\subfloat {\includegraphics[width=\imw cm, height=\imh cm]{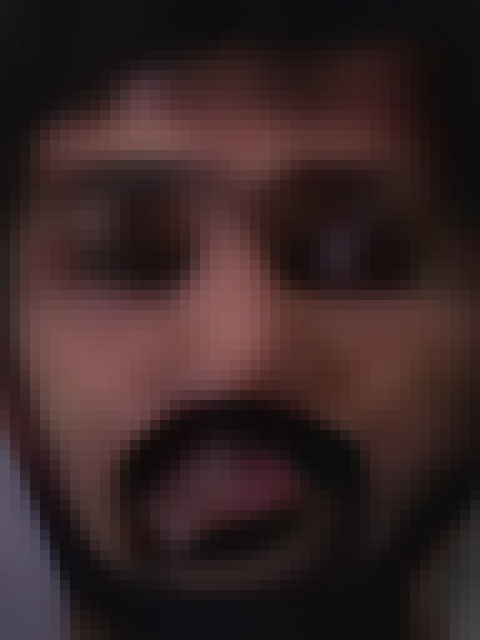}}\par\vspace{\vdiss cm}
		\subfloat {\includegraphics[width=\imw cm, height=\imh cm]{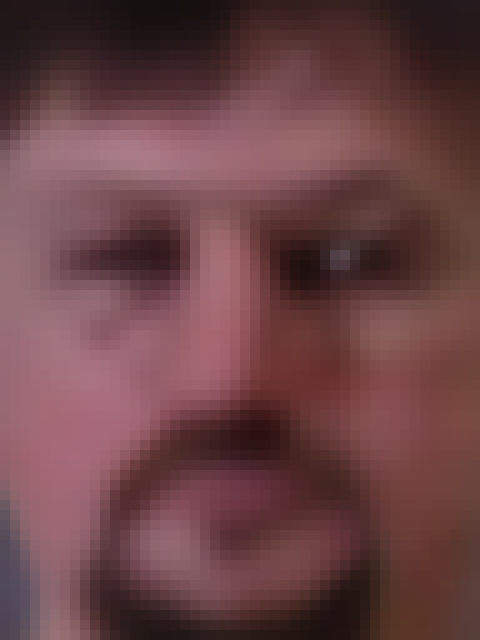}}\par\vspace{\vdiss cm}
		\subfloat {\includegraphics[width=\imw cm, height=\imh cm]{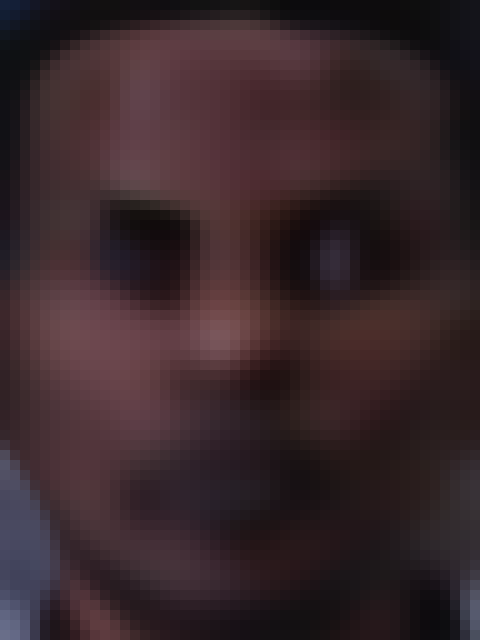}}\par\vspace{\vdiss cm}
		\stepcounter{figure}\addtocounter{figure}{-1}
		\addtocounter{subfigure}{4}
		\subfloat[]{\includegraphics[width=\imw cm, height=\imh cm]{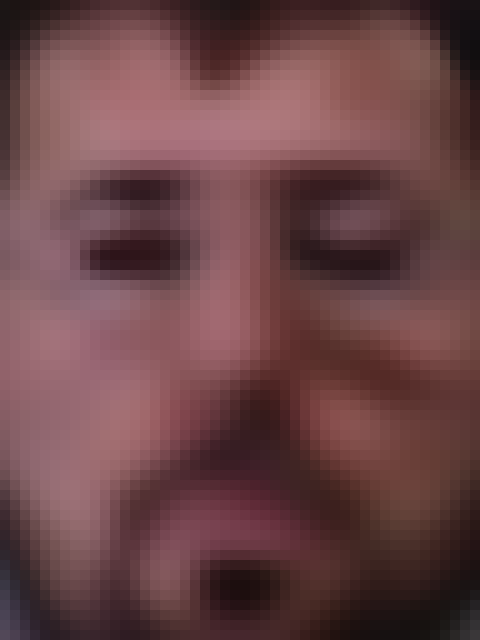}}
	\end{minipage}%
	\begin{minipage}{\hdis\textwidth} 
		\centering
		\subfloat {\includegraphics[width=\imw cm, height=\imh cm]{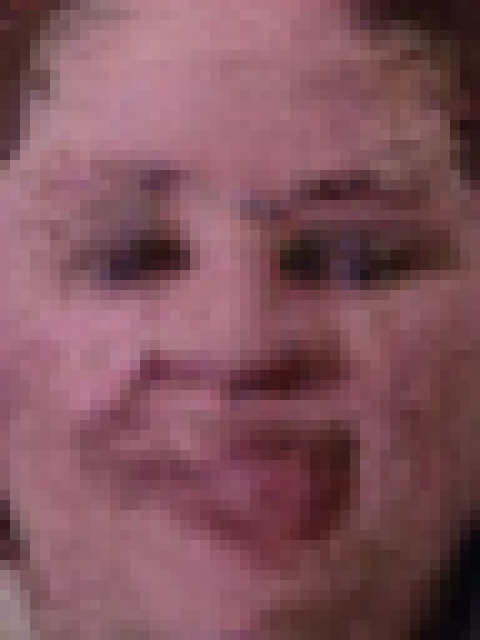}}\par\vspace{\vdiss cm}		
		\subfloat {\includegraphics[width=\imw cm, height=\imh cm]{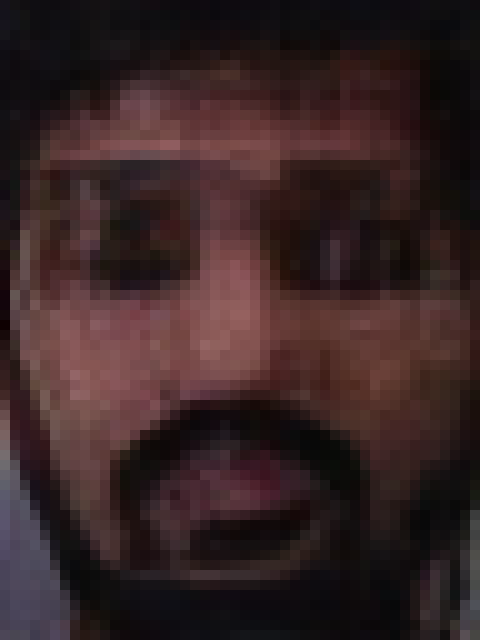}}\par\vspace{\vdiss cm}
		\subfloat {\includegraphics[width=\imw cm, height=\imh cm]{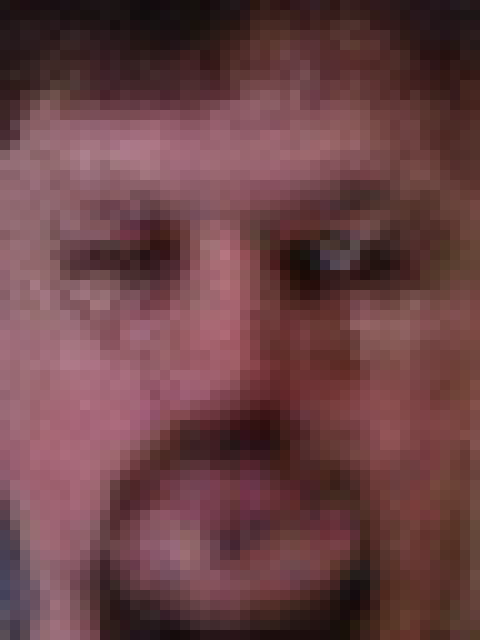}}\par\vspace{\vdiss cm}
		\subfloat {\includegraphics[width=\imw cm, height=\imh cm]{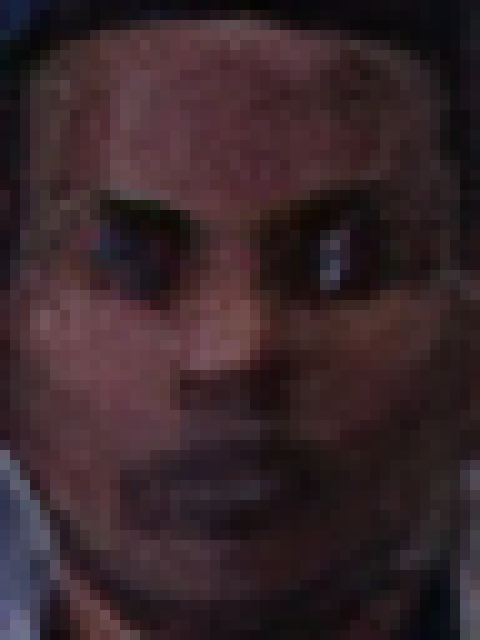}}\par\vspace{\vdiss cm}
		\stepcounter{figure}\addtocounter{figure}{-1}
		\addtocounter{subfigure}{5}
		\subfloat[]{\includegraphics[width=\imw cm, height=\imh cm]{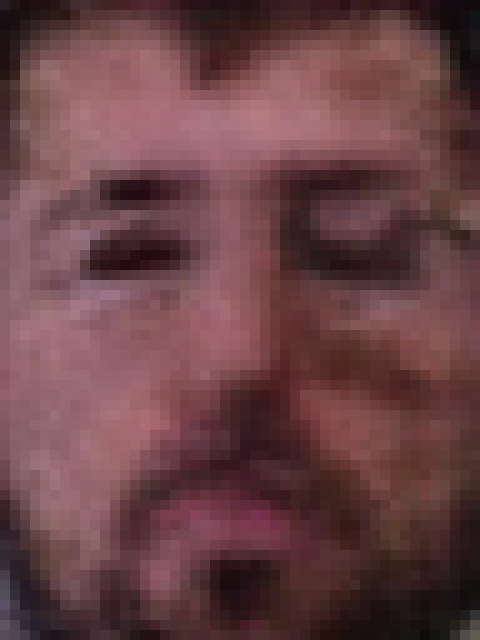}}
	\end{minipage}%
	\begin{minipage}{\hdis\textwidth} 
		\centering
		\subfloat {\includegraphics[width=\imw cm, height=\imh cm]{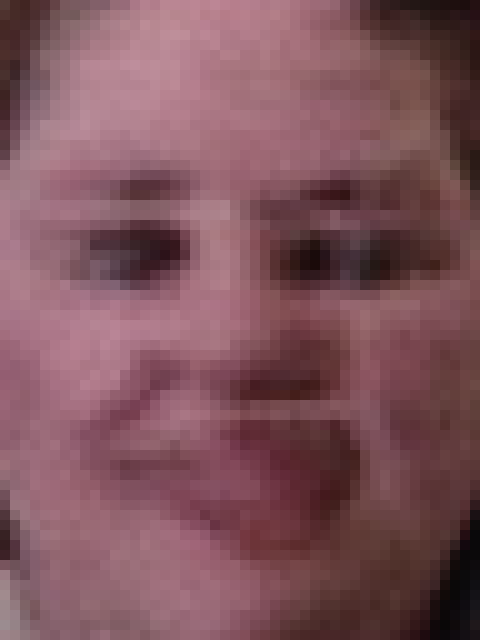}}\par\vspace{\vdiss cm}		
		\subfloat {\includegraphics[width=\imw cm, height=\imh cm]{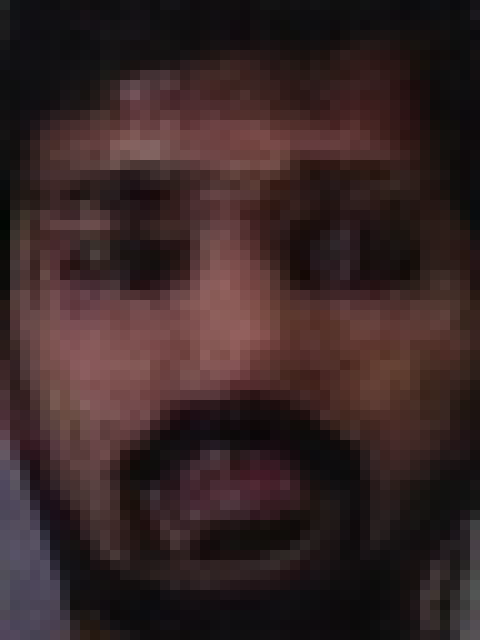}}\par\vspace{\vdiss cm}
		\subfloat {\includegraphics[width=\imw cm, height=\imh cm]{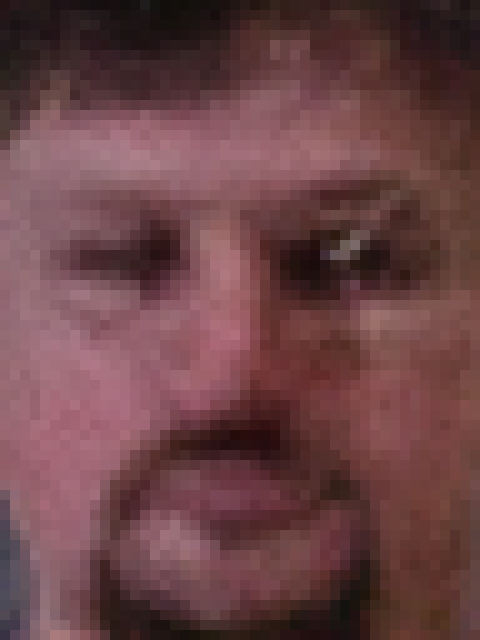}}\par\vspace{\vdiss cm}
		\subfloat {\includegraphics[width=\imw cm, height=\imh cm]{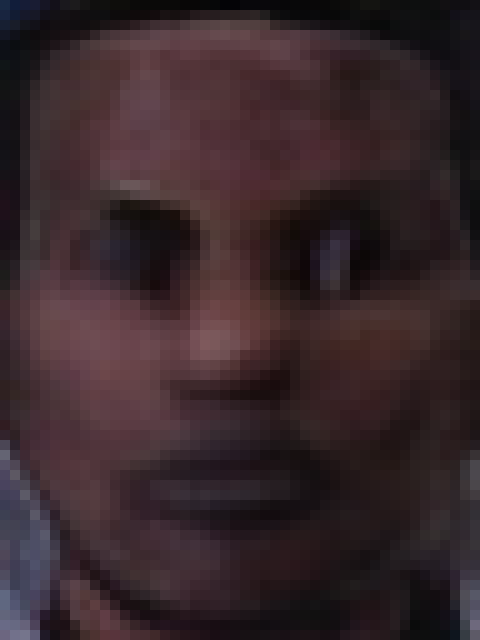}}\par\vspace{\vdiss cm}
		\stepcounter{figure}\addtocounter{figure}{-1}
		\addtocounter{subfigure}{6}
		\subfloat[]{\includegraphics[width=\imw cm, height=\imh cm]{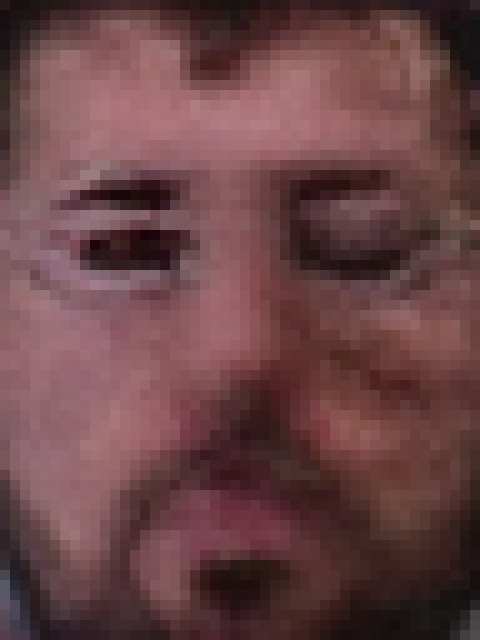}}
	\end{minipage}%
	\begin{minipage}{\hdis\textwidth} 
		\centering
		\subfloat {\includegraphics[width=\imw cm, height=\imh cm]{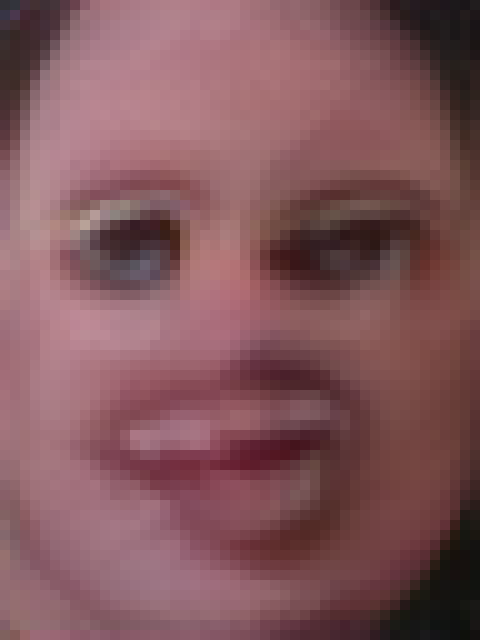}}\par\vspace{\vdiss cm}		
		\subfloat {\includegraphics[width=\imw cm, height=\imh cm]{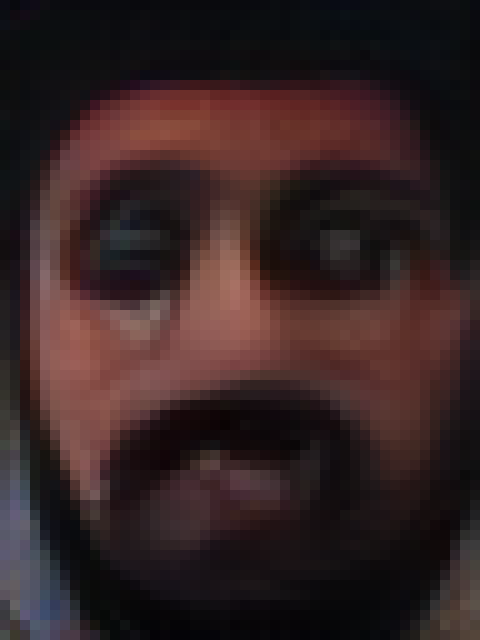}}\par\vspace{\vdiss cm}
		\subfloat {\includegraphics[width=\imw cm, height=\imh cm]{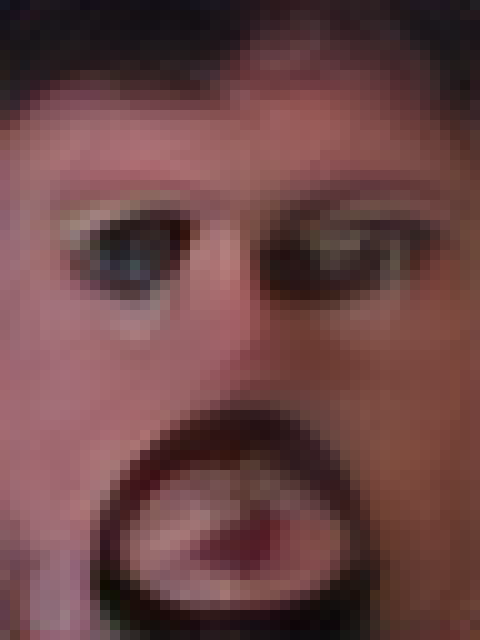}}\par\vspace{\vdiss cm}
		\subfloat {\includegraphics[width=\imw cm, height=\imh cm]{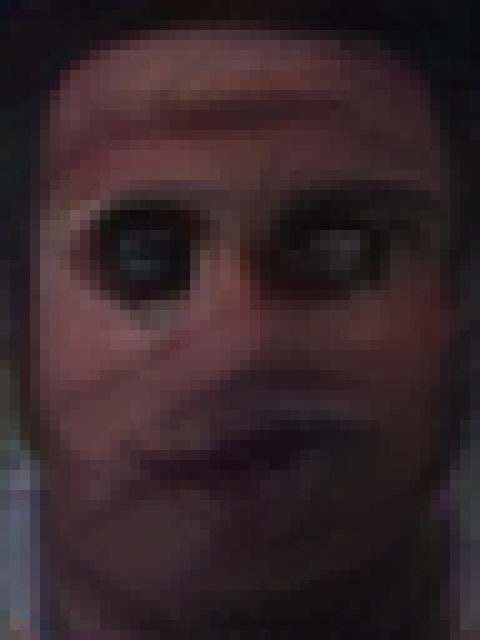}}\par\vspace{\vdiss cm}
		\stepcounter{figure}\addtocounter{figure}{-1}
		\addtocounter{subfigure}{7}
		\subfloat[]{\includegraphics[width=\imw cm, height=\imh cm]{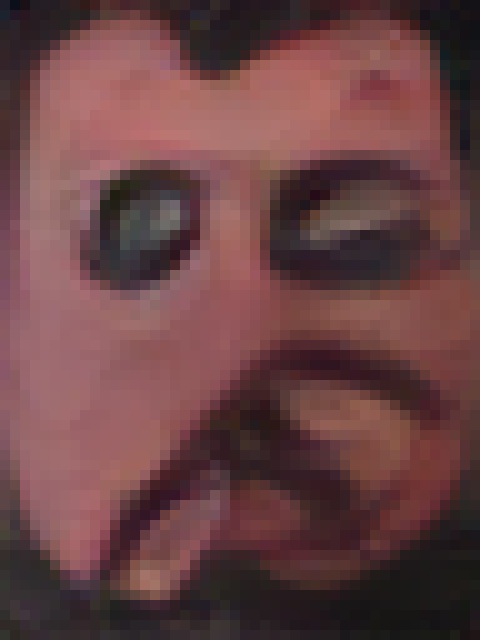}}
	\end{minipage}%
	\begin{minipage}{\hdis\textwidth} 
		\centering		
		\subfloat {\includegraphics[width=\imw cm, height=\imh cm]{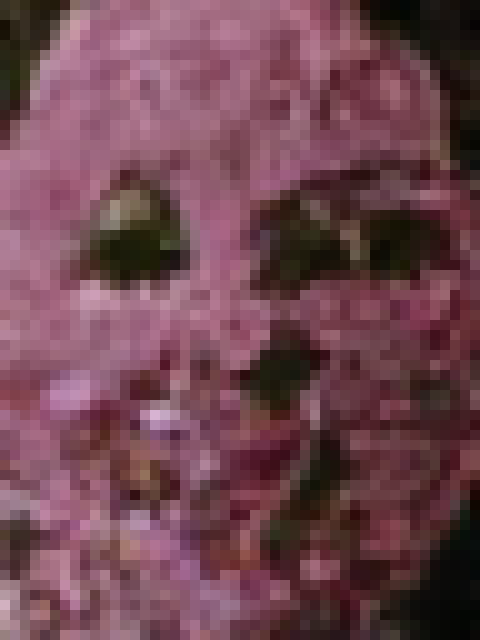}}\par\vspace{\vdiss cm}		
		\subfloat {\includegraphics[width=\imw cm, height=\imh cm]{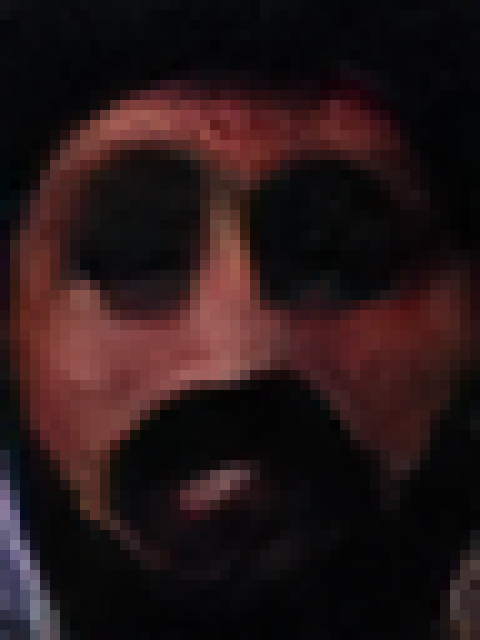}}\par\vspace{\vdiss cm}
		\subfloat {\includegraphics[width=\imw cm, height=\imh cm]{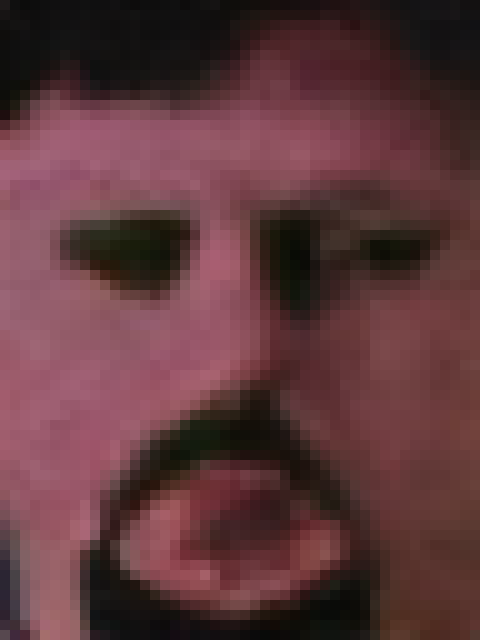}}\par\vspace{\vdiss cm}
		\subfloat {\includegraphics[width=\imw cm, height=\imh cm]{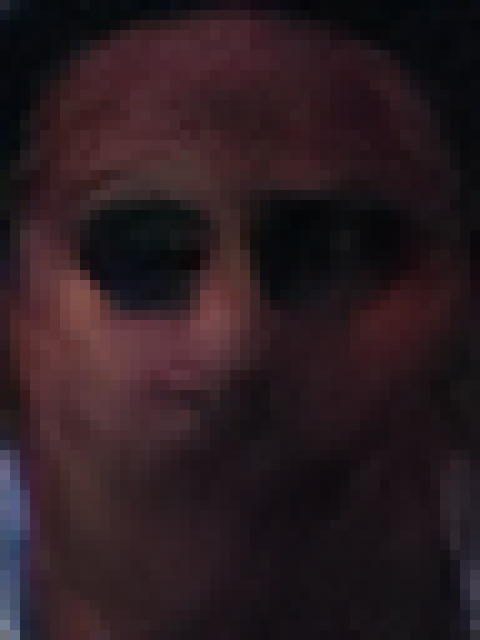}}\par\vspace{\vdiss cm}
		\stepcounter{figure}\addtocounter{figure}{-1}
		\addtocounter{subfigure}{8}
		\subfloat[]{\includegraphics[width=\imw cm, height=\imh cm]{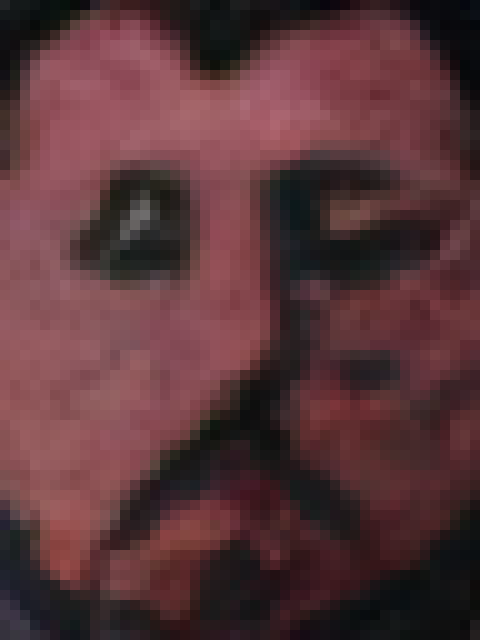}}
	\end{minipage}%
	\begin{minipage}{\hdis\textwidth} 
		\centering
		\subfloat {\includegraphics[width=\imw cm, height=\imh cm]{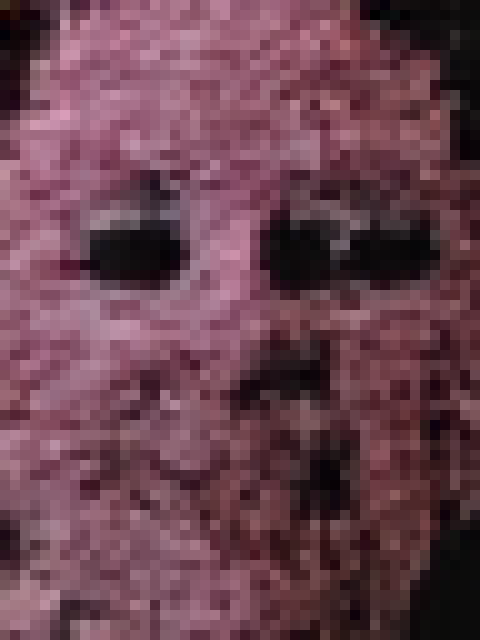}}\par\vspace{\vdiss cm}		
		\subfloat {\includegraphics[width=\imw cm, height=\imh cm]{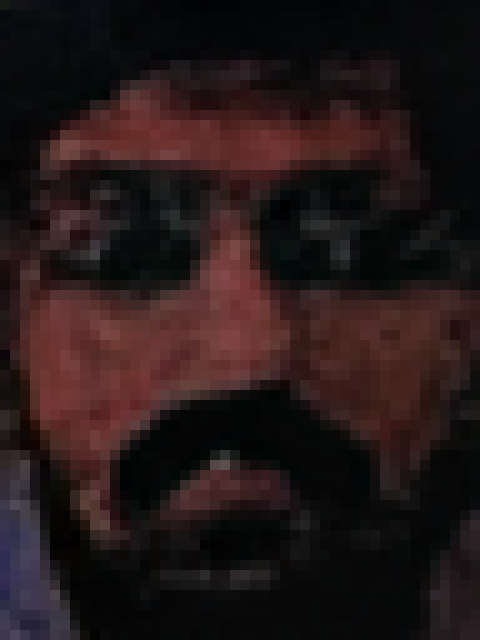}}\par\vspace{\vdiss cm}
		\subfloat {\includegraphics[width=\imw cm, height=\imh cm]{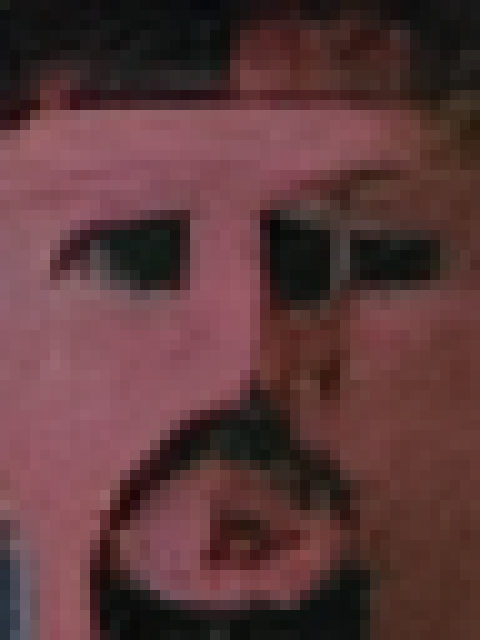}}\par\vspace{\vdiss cm}
		\subfloat {\includegraphics[width=\imw cm, height=\imh cm]{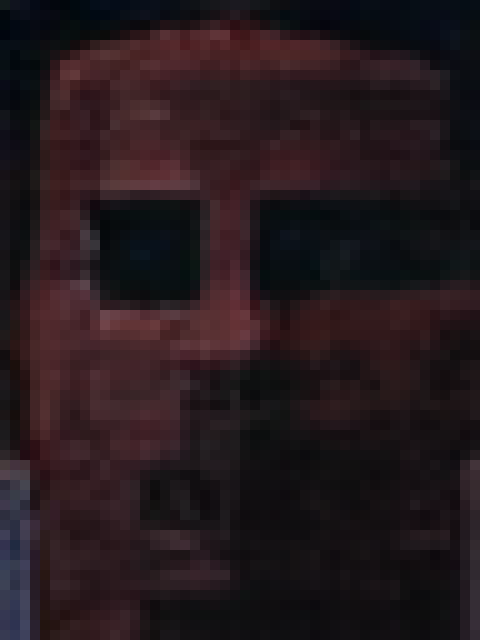}}\par\vspace{\vdiss cm}
		\stepcounter{figure}\addtocounter{figure}{-1}
		\addtocounter{subfigure}{9}
		\subfloat[]{\includegraphics[width=\imw cm, height=\imh cm]{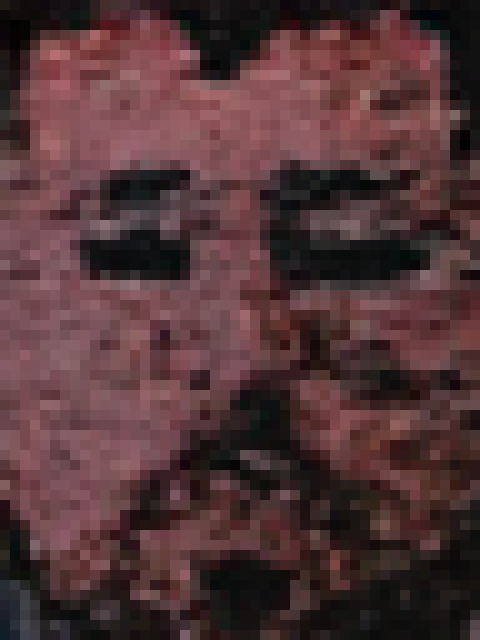}}
	\end{minipage}%
	\begin{minipage}{\hdis\textwidth} 
		\centering
		\subfloat {\includegraphics[width=\imw cm, height=\imh cm]{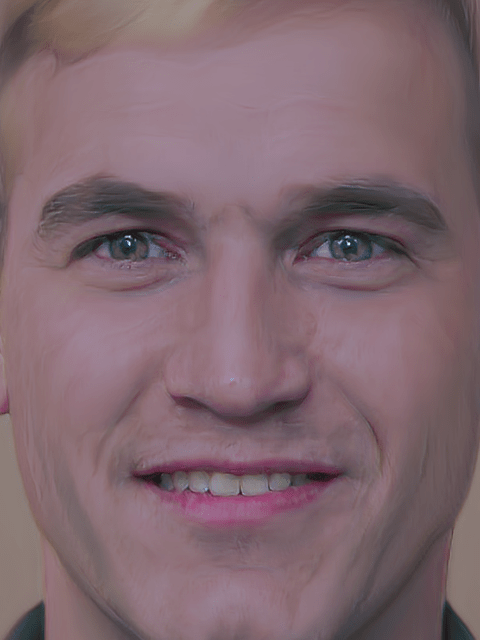}}\par\vspace{\vdiss cm}		
		\subfloat {\includegraphics[width=\imw cm, height=\imh cm]{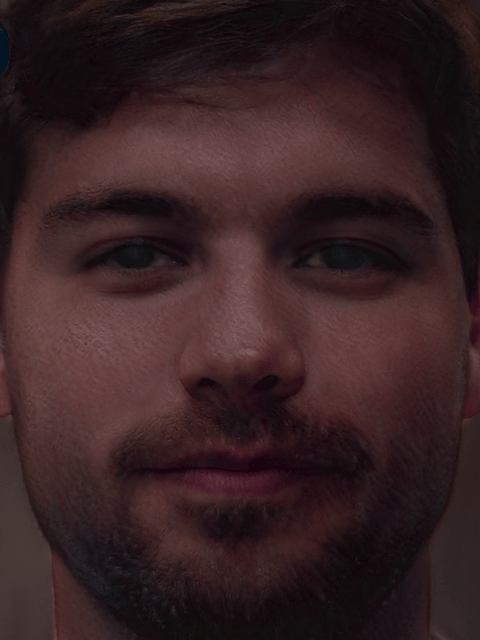}}\par\vspace{\vdiss cm}
		\subfloat {\includegraphics[width=\imw cm, height=\imh cm]{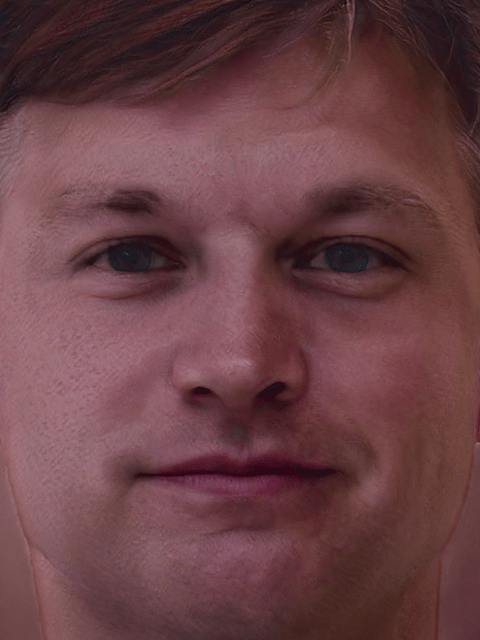}}\par\vspace{\vdiss cm}
		\subfloat {\includegraphics[width=\imw cm, height=\imh cm]{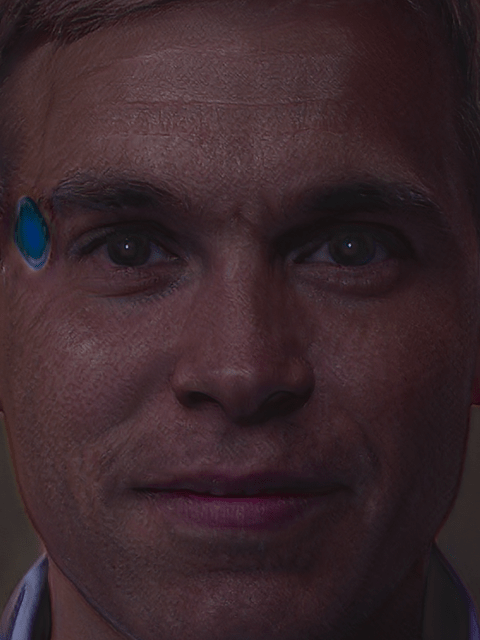}}\par\vspace{\vdiss cm}
		\stepcounter{figure}\addtocounter{figure}{-1}
		\addtocounter{subfigure}{10}
		\subfloat[]{\includegraphics[width=\imw cm, height=\imh cm]{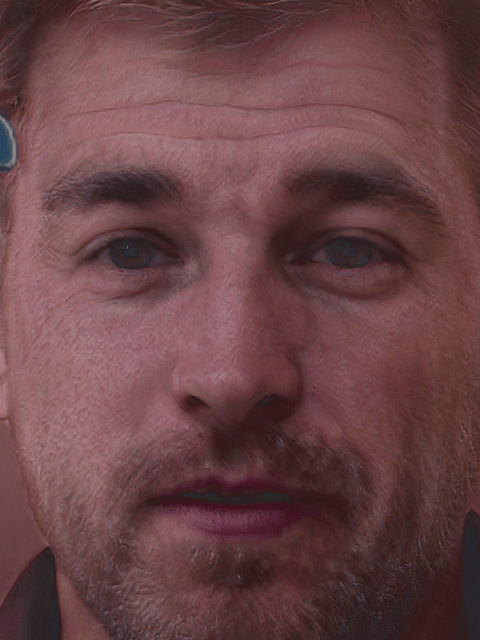}}
	\end{minipage}%
	\begin{minipage}{\hdis\textwidth} 
		\centering
		\subfloat {\includegraphics[width=\imw cm, height=\imh cm]{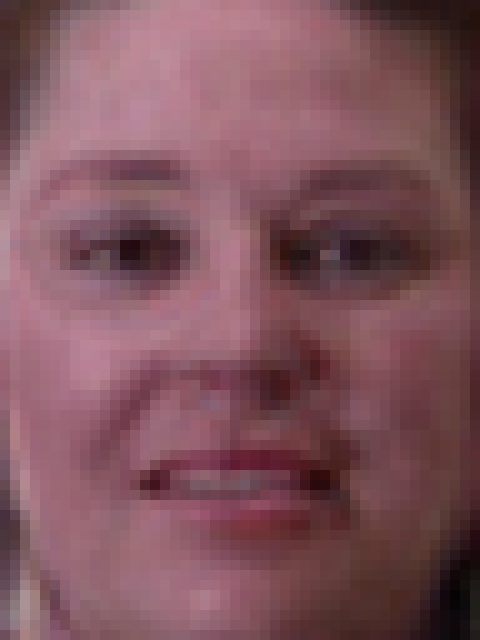}}\par\vspace{\vdiss cm}		
		\subfloat {\includegraphics[width=\imw cm, height=\imh cm]{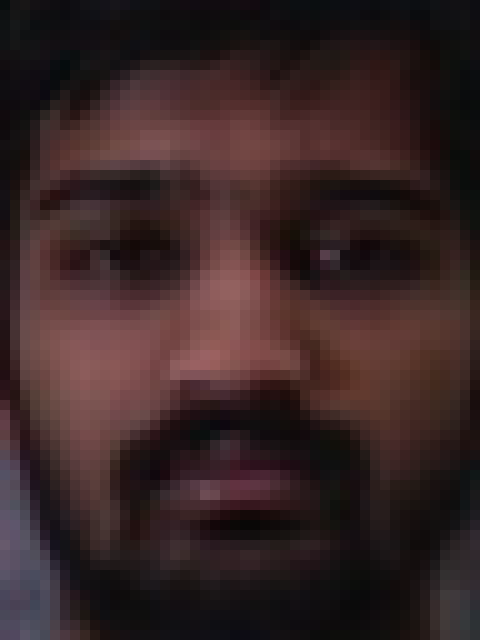}}\par\vspace{\vdiss cm}
		\subfloat {\includegraphics[width=\imw cm, height=\imh cm]{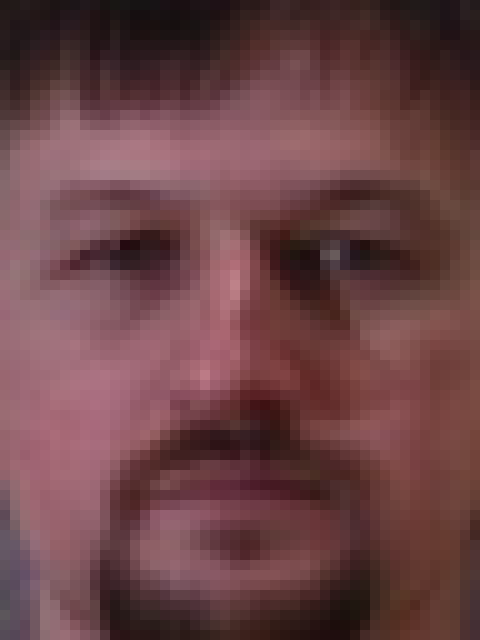}}\par\vspace{\vdiss cm}
		\subfloat {\includegraphics[width=\imw cm, height=\imh cm]{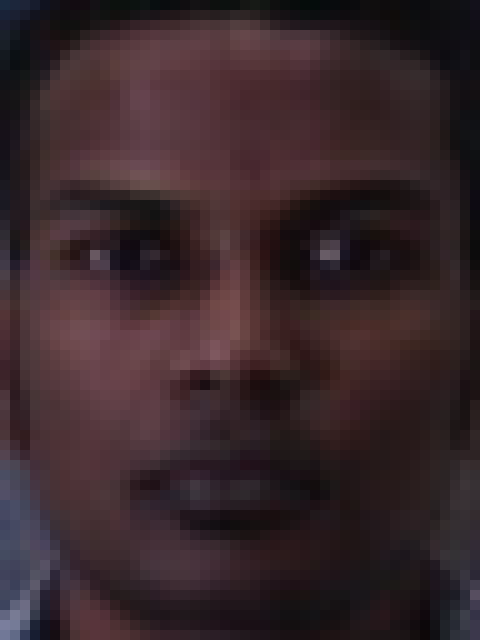}}\par\vspace{\vdiss cm}
		\stepcounter{figure}\addtocounter{figure}{-1}
		\addtocounter{subfigure}{11}
		\subfloat[]{\includegraphics[width=\imw cm, height=\imh cm]{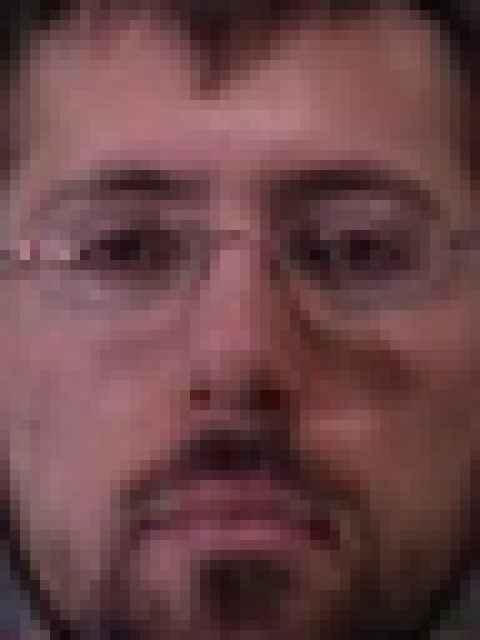}}
	\end{minipage}%
	\begin{minipage}{\hdis\textwidth} 
		\centering
		\subfloat {\includegraphics[width=\imw cm, height=\imh cm]{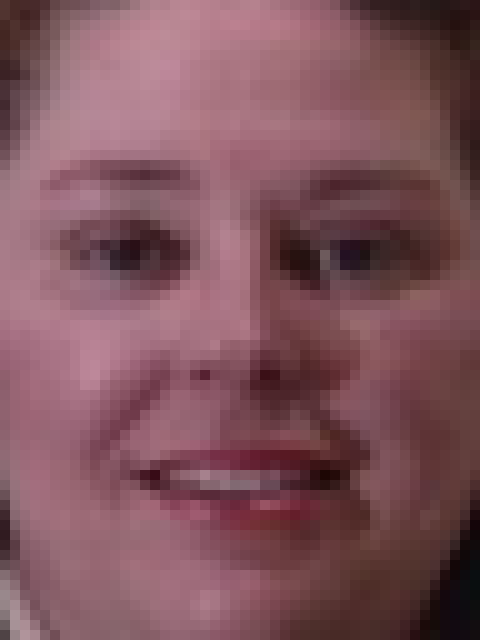}}\par\vspace{\vdiss cm}		
		\subfloat {\includegraphics[width=\imw cm, height=\imh cm]{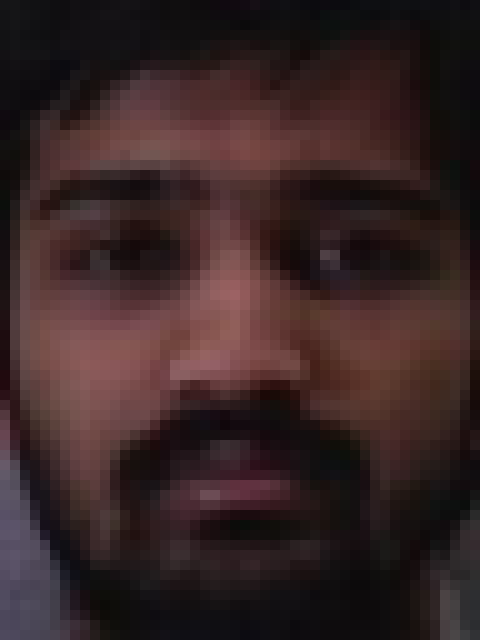}}\par\vspace{\vdiss cm}
		\subfloat {\includegraphics[width=\imw cm, height=\imh cm]{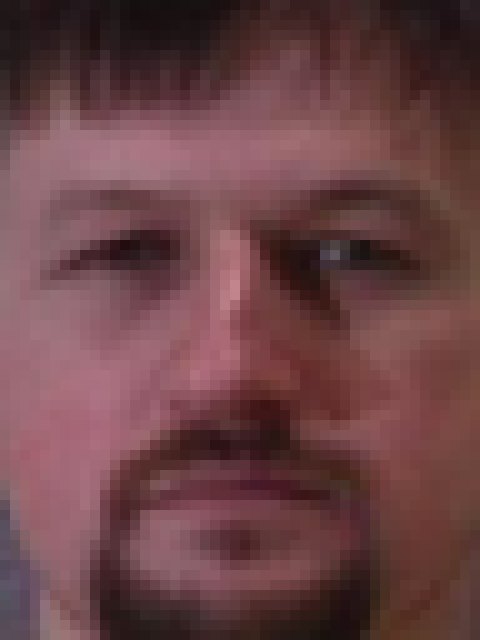}}\par\vspace{\vdiss cm}
		\subfloat {\includegraphics[width=\imw cm, height=\imh cm]{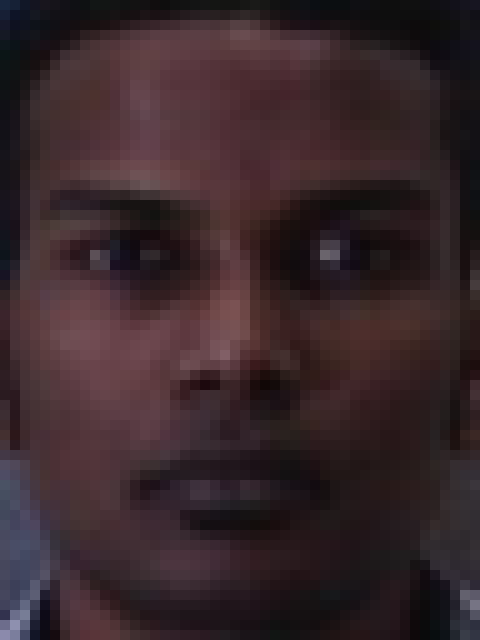}}\par\vspace{\vdiss cm}
		\stepcounter{figure}\addtocounter{figure}{-1}
		\addtocounter{subfigure}{12}
		\subfloat[]{\includegraphics[width=\imw cm, height=\imh cm]{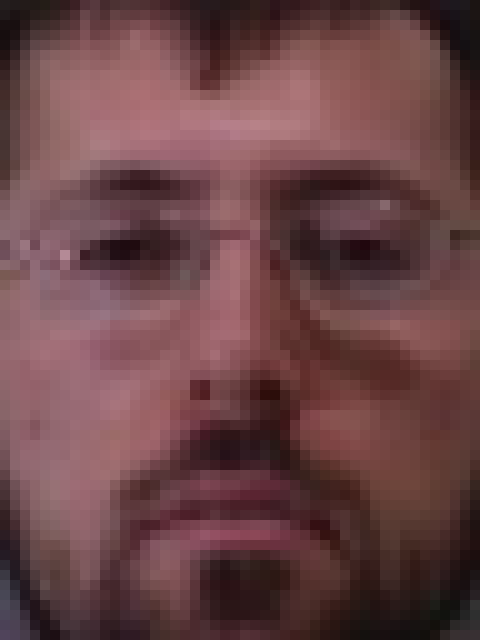}}
	\end{minipage}\par\medskip
	\caption{Qualitative comparison of the proposed method with various deep learning-based face hallucination and image super-resolution algorithms on the $16\times12$ test faces of the Multi-PIE database with scaling factor 4. (a) LR input. (b) Bicubic interpolation. (c) SRCNN \cite{refs:Dong2016}. (d) DCSCN \cite{refs:Yamanaka2017}. (e) DBPN \cite{refs:Haris2018}. (f) ESRGAN \cite{refs:Wang2018}. (g) SPSR \cite{refs:Ma2020}. (h) BSRGAN \cite{refs:Zhang2021b}. (i) realESRGAN \cite{refs:Wang2021}. (j) SwinIR \cite{refs:Liang2021}. (k) PULSE \cite{refs:Menon2020}. (l) Proposed. (m) Ground truth. }
	\label{fig:deepqual}
\end{figure*}

\subsection{Computational Complexity} 
The optimization procedure of the proposed algorithm consists of two phases. As discussed in section \ref{sec:proposed}, the $\textit{l}_2$--$\textit{l}_2$ minimization problem can be solved through a closed-form solution, hence the $\textit{l}_1$-minimization problem is basically the most time-consuming part of the reconstruction process, which can also be solved efficiently using various $\textit{l}_1$-optimization approaches. Overall, the proposed method is considered to be a very fast algorithm, with not more than 30 iterations required for it to converge. To compare the computational time of our algorithm with those of the other methods and evaluate the effects of different parameters on its runtime, we perform experiments on the AR and the FERET databases with LR input size and scaling factor of $10 \times 8$ and $4$, respectively, using MATLAB 2020b and a computer with 6GB memory and 1.8 GHz CPU. Table \ref{tab:runtime} presents the average runtime of each method on the AR database when the training set size is 100. One can notice that the proposed algorithm performs face hallucination with reasonable computational cost compared to the competitive methods. We also measure the runtime of each algorithm with respect to the dataset size and the scaling factor. The results, which are displayed in Fig. \ref{fig:runtime}, reveal that the computational time of our algorithm is not much affected by both parameters, whereas the ones for TLcR-RL \cite{refs:Jiang2018} and SSR \cite{refs:Jiang2017} grow exponentially, making them practically inefficient in the case of large images or big datasets. It should also be noted that reducing the number of iterations -- which was previously shown to be slightly significant to the quality of the hallucinated face image after the first few iterations -- will decrease the computational cost of the proposed algorithm even further. One can also think of applying collaborative representation \cite{refs:Zhang2011a} instead of sparse representation, which leads to a super-fast face hallucination procedure with both subproblems having closed-form solutions.

\begin{figure}
	\captionsetup[subfloat]{farskip=0pt,captionskip=1pt}
	\centering
	\def\theight{0.18}
	\subfloat[]{\includegraphics[height=\theight\textwidth]{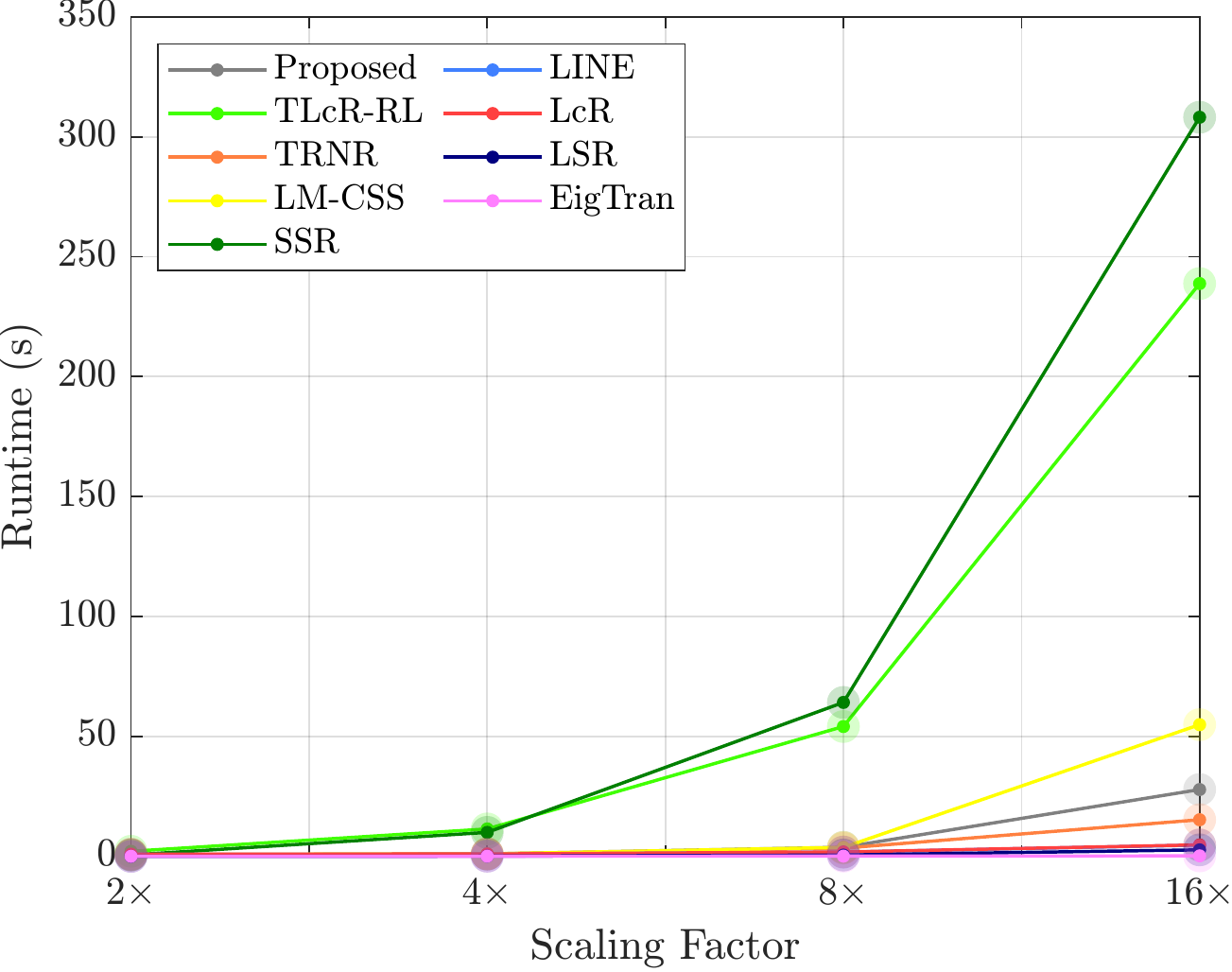}}\hfil
	\subfloat[]{\includegraphics[height=\theight\textwidth]{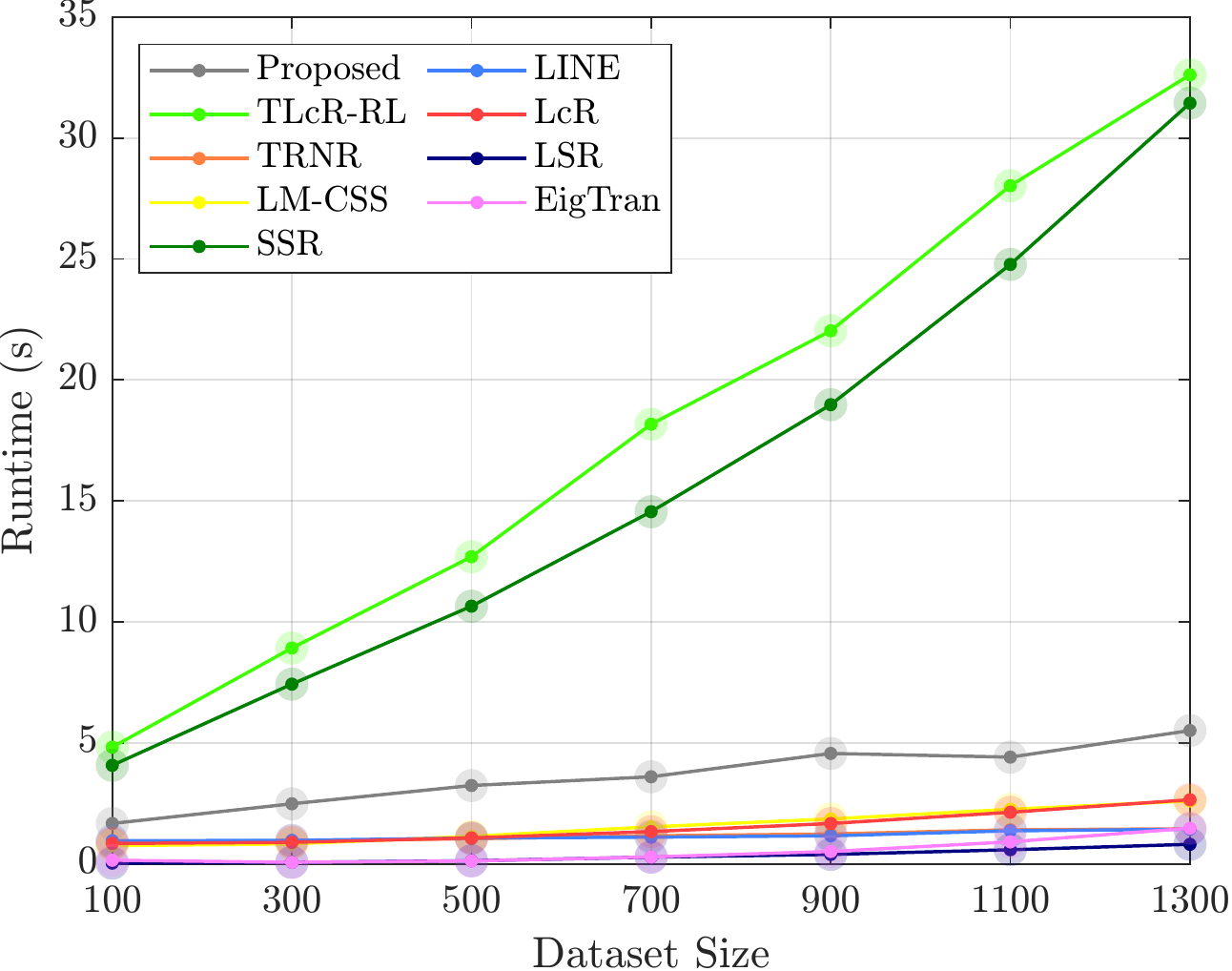}}
	\caption{Computational time variations of different approaches according to (a) Scaling factor, and (b) Dataset size.}
	\label{fig:runtime}
\end{figure}

\begin{table}
	\caption{Comparison of the average running time of different algorithms on the AR database}
	\centering
	\begin{tabular}{c||c|c|c|c|c} 
		\hline\hline
		Algorithm&Wang&LSR&LcR&LINE&SSR\\ 
		\hline
		Runtime (s)&0.11&0.13&1.04&0.93&10.03\\
		\hline\hline
		Algorithm&LM-CSS&TRNR&TLcR-RL&Proposed&\\
		\hline
		Runtime (s)&1.07&1.02&11.48&1.05&\\
		\hline\hline
	\end{tabular}
	\label{tab:runtime}
\end{table}

\subsection{Pose-Robust Face Hallucination}
In order to evaluate the efficiency of the proposed 3D dictionary alignment procedure introduced in section \ref{sec:proposed}, extensive experiments were conducted on the Multi-PIE and the LFW face databases. Since the methods used in the frontal face experiments are unable to perform pose-robust face hallucination, we integrate them with our 3D dictionary alignment scheme, hence they use the same aligned dictionary to perform face hallucination. Moreover, the parameters and settings associated with all the algorithms remain the same as in the previous experiments. We use \cite{refs:Guo2020} to extract 3D landmarks from both the LR and HR faces, and \cite{refs:Feng2018} to perform 3D face reconstruction. To apply face alignment, all 68 landmark points were taken into consideration.

\begin{table}
	\caption{Pose-robust face hallucination performance achieved by different algorithms on the Multi-PIE and the LFW datasets}
	\centering
	\begin{tabular}{c||cc|cc} 
		\hline\hline
		\multirow{2}{*}{Algorithm} & \multicolumn{2}{c|}{Multi-PIE} & \multicolumn{2}{c}{LFW}  \\ 
		\cline{2-5}
		& PSNR           & SSIM                                  & PSNR           & SSIM                                \\ 
		\hline
		Bicubic                    & 22.84          &  0.6466                                & 21.65          & 0.6876                                \\
		Wang \cite{refs:Wang2005}                  & 27.46          & 0.8038                                & 25.77          & 0.8018                                \\
		LSR \cite{refs:Ma2010}                       & 29.26          & 0.8522                                & 27.66          & 0.8469                                                     \\
		LcR \cite{refs:Jiang2014a}                       & 29.52          & 0.8534                                & 27.57          & 0.8440                                                      \\
		LINE \cite{refs:Jiang2014b}                      & 29.83          & 0.8608                                & 27.95          & 0.8543                                             \\
		SSR \cite{refs:Jiang2017}                       & 29.25          & 0.8516                                & 27.43          & 0.8413                                                     \\
		LM-CSS \cite{refs:Farrugia2017}                    & 28.91          & 0.8517                                & 27.11          & 0.8395                                                   \\
		TRNR \cite{refs:Jiang2016}                      & 30.23          & 0.8712                                & 28.47          & 0.8661                                                     \\
		TLcR-RL \cite{refs:Jiang2018}                   & 30.65          & 0.8843                                & 28.88          & 0.8757                                          \\
		Proposed                   & \textbf{32.07} & \textbf{0.8950}                       & \textbf{31.01} & \textbf{0.9149}                                 \\
		\hline\hline
	\end{tabular}
	\label{tab:poseres}
\end{table}
 
\subsubsection{The Multi-PIE Dataset}
For each of the subjects in the Multi-PIE face database, we consider illumination condition 10 and select the frontal face images (camera 05-1) taken in neutral position from all sessions as the training samples (thus, there are five samples per subject in the training set), and randomly select one sample from their images in the same settings but under different pose variations (cameras 04-1, 05-0, 13-0, and 14-0) from session one to form the test set. Original images are cropped in such a way that they contain the entire face to facilitate the procedure of 3D face reconstruction (i.e., the face regions are usually much smaller than the actual image size). The LR face images are of size $20 \times 20$ and the scaling factor is 4. Table \ref{tab:poseres} summarizes the results obtained by each method. Our algorithm outperforms the second best method by 1.42 dB in PSNR and 0.0107 in SSIM. The results shown in Fig. \ref{fig:multipiepose} also verify the effectiveness of our proposed dictionary alignment technique, which has clearly enhanced the performance of all the face super-resolution algorithms. Once again, by investigating the hallucinated faces, one can easily notice more recovered facial details in our results than those of the other approaches.

\begin{figure*}[!t]
	\captionsetup[subfloat]{farskip=0pt,captionskip=1pt}
	\centering
	\def\imw{1.42}
	\def\hdis{0.081}
	\def\vdiss{0.04}
	\begin{minipage}{\hdis\textwidth} 
		\centering
		\subfloat {\includegraphics[width=\imw cm]{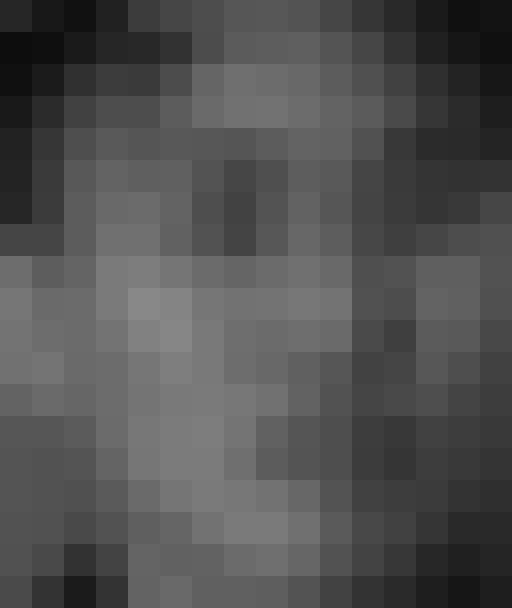}}\par\vspace{\vdiss cm}		
		\subfloat {\includegraphics[width=\imw cm]{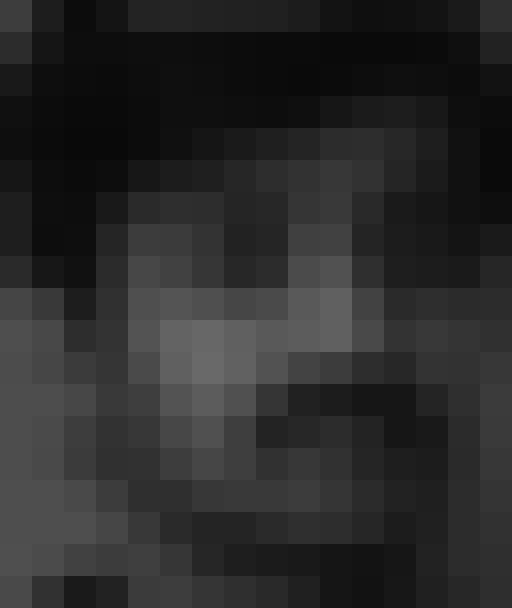}}\par\vspace{\vdiss cm}
		\subfloat {\includegraphics[width=\imw cm]{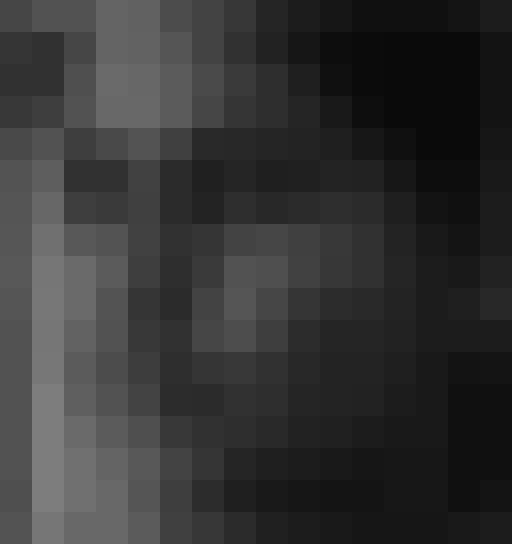}}\par\vspace{\vdiss cm}
		\subfloat {\includegraphics[width=\imw cm]{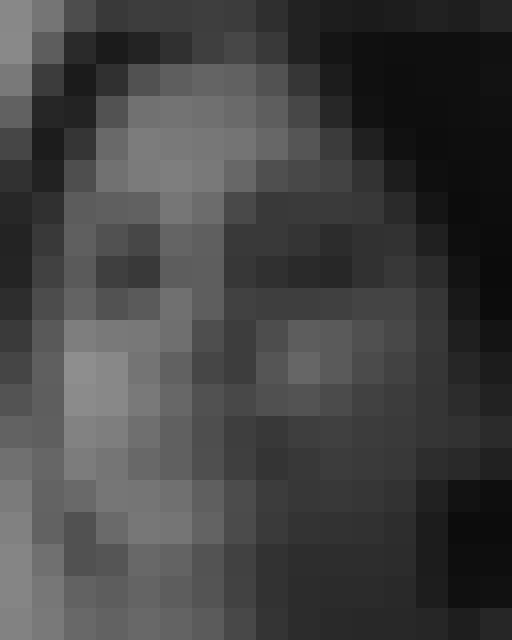}}\par\vspace{\vdiss cm}
		\stepcounter{figure}\addtocounter{figure}{-1}
		\addtocounter{subfigure}{0}
		\subfloat[]{\includegraphics[width=\imw cm]{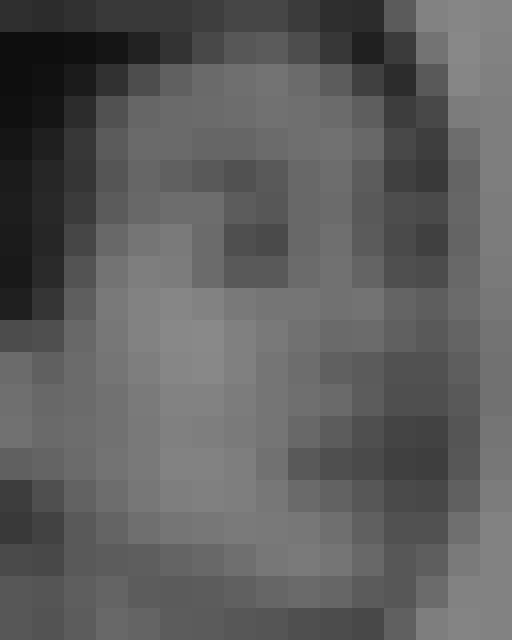}}
	\end{minipage}%
	\begin{minipage}{\hdis\textwidth} 
		\centering
		\subfloat {\includegraphics[width=\imw cm]{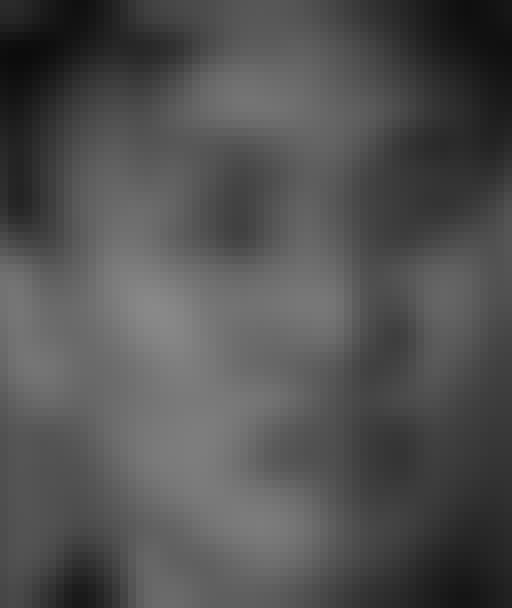}}\par\vspace{\vdiss cm}		
		\subfloat {\includegraphics[width=\imw cm]{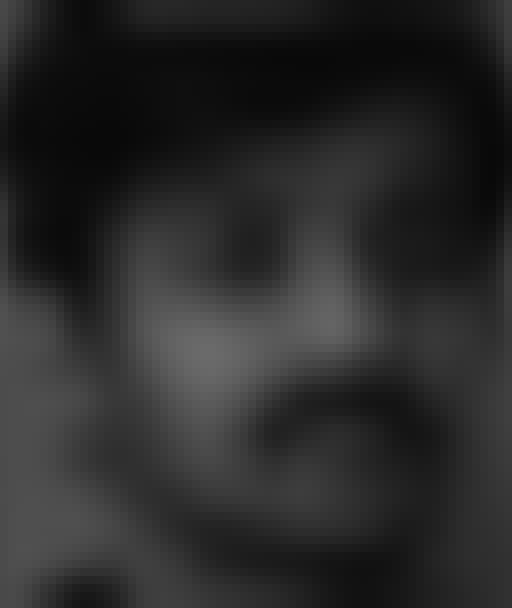}}\par\vspace{\vdiss cm}
		\subfloat {\includegraphics[width=\imw cm]{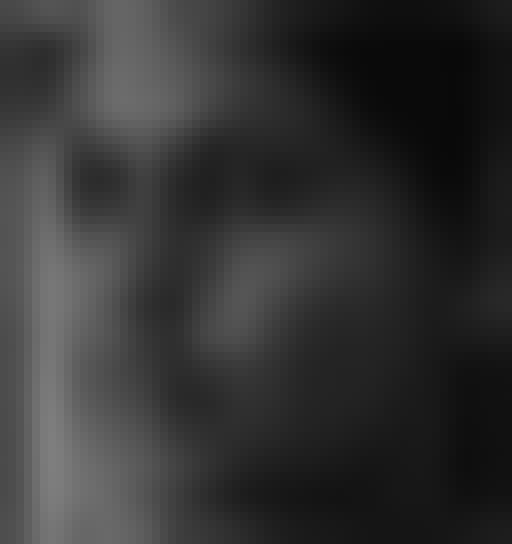}}\par\vspace{\vdiss cm}
		\subfloat {\includegraphics[width=\imw cm]{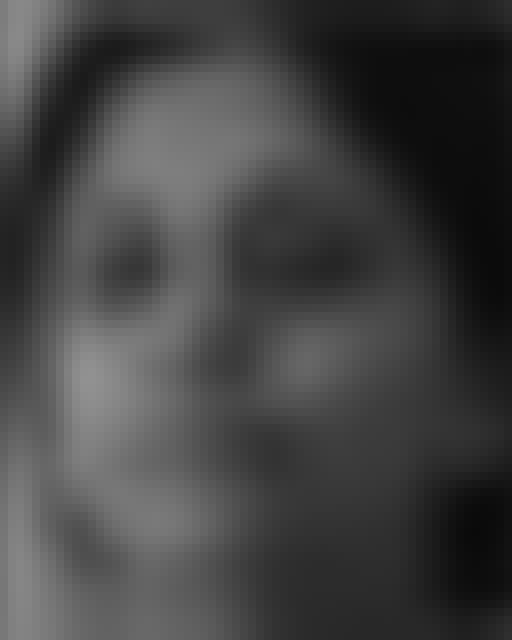}}\par\vspace{\vdiss cm}
		\stepcounter{figure}\addtocounter{figure}{-1}
		\addtocounter{subfigure}{1}
		\subfloat[]{\includegraphics[width=\imw cm]{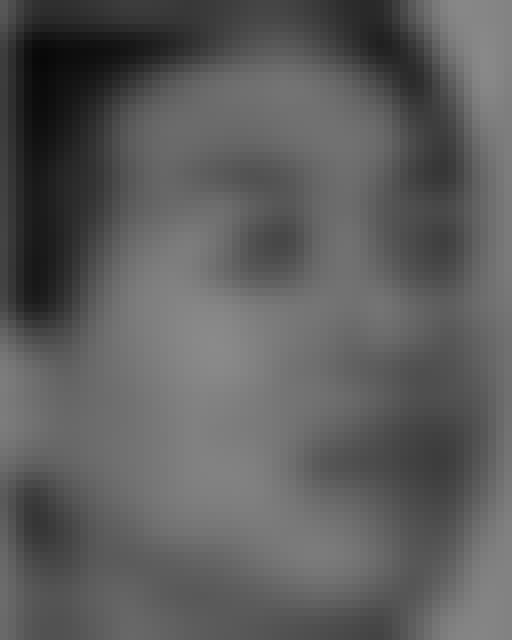}}
	\end{minipage}%
	\begin{minipage}{\hdis\textwidth} 
		\centering
		\subfloat {\includegraphics[width=\imw cm]{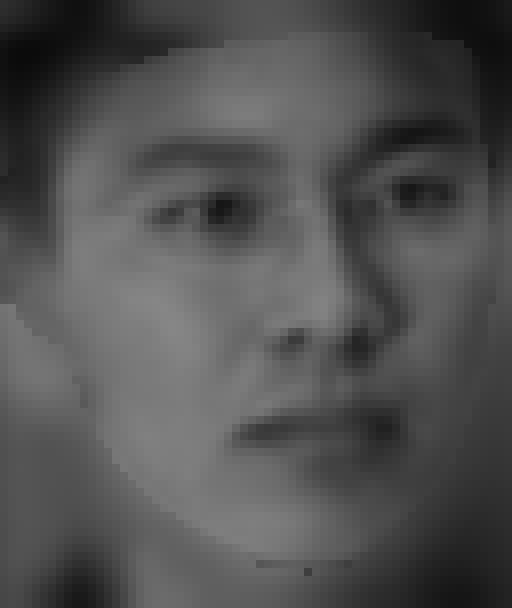}}\par\vspace{\vdiss cm}		
		\subfloat {\includegraphics[width=\imw cm]{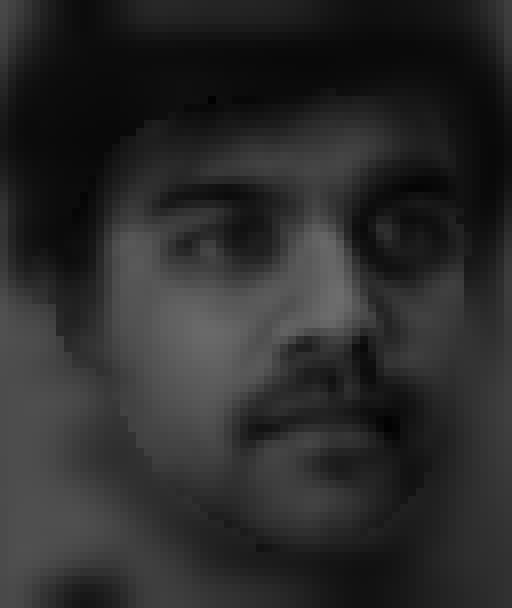}}\par\vspace{\vdiss cm}
		\subfloat {\includegraphics[width=\imw cm]{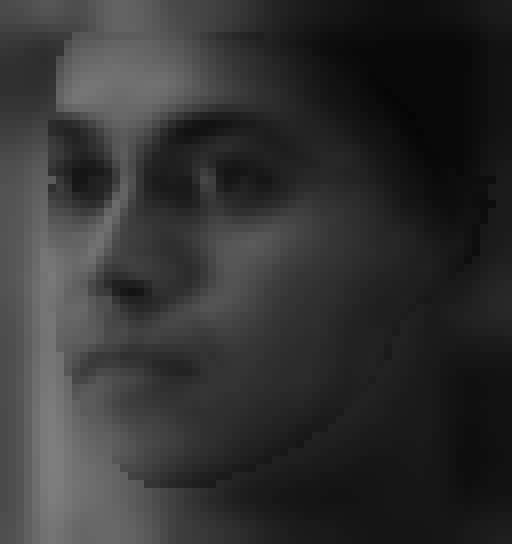}}\par\vspace{\vdiss cm}
		\subfloat {\includegraphics[width=\imw cm]{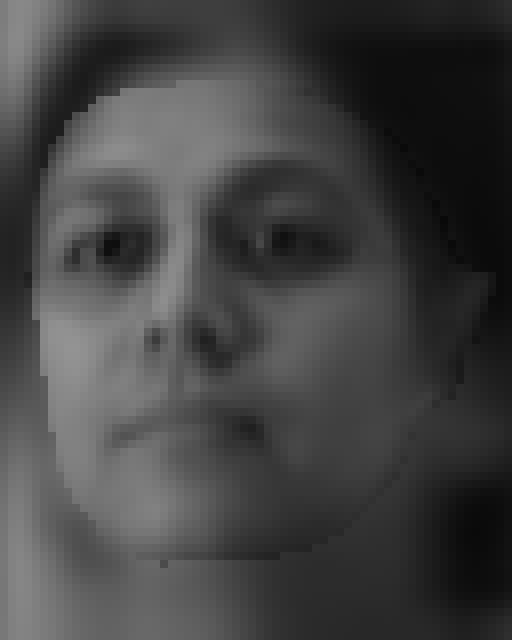}}\par\vspace{\vdiss cm}
		\stepcounter{figure}\addtocounter{figure}{-1}
		\addtocounter{subfigure}{2}
		\subfloat[]{\includegraphics[width=\imw cm]{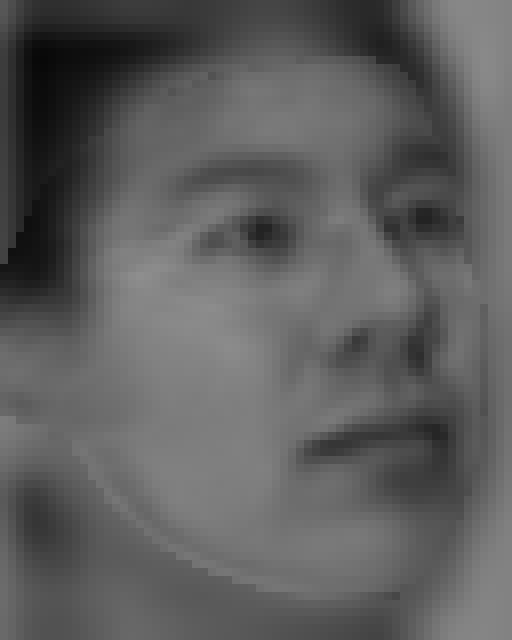}}
	\end{minipage}%
	\begin{minipage}{\hdis\textwidth} 
		\centering
		\subfloat {\includegraphics[width=\imw cm]{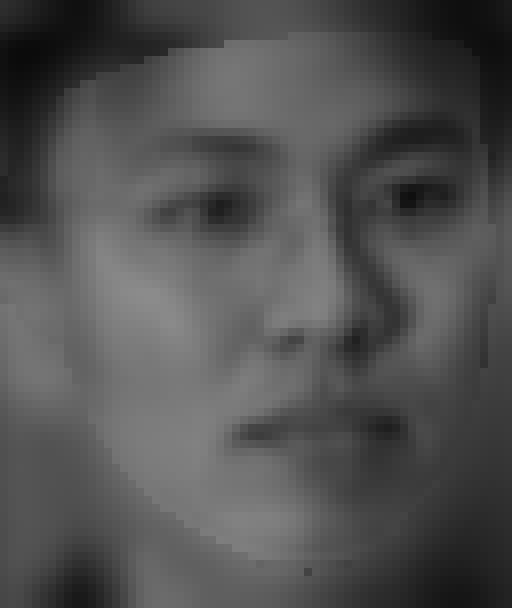}}\par\vspace{\vdiss cm}		
		\subfloat {\includegraphics[width=\imw cm]{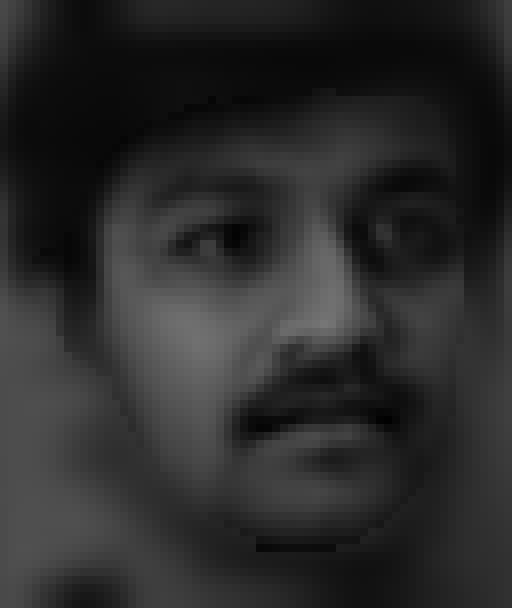}}\par\vspace{\vdiss cm}
		\subfloat {\includegraphics[width=\imw cm]{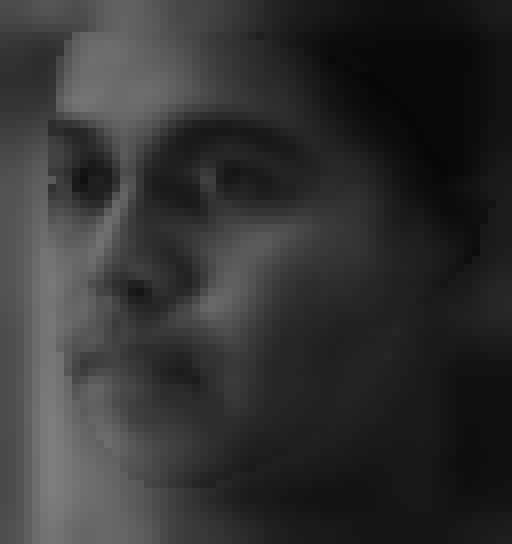}}\par\vspace{\vdiss cm}
		\subfloat {\includegraphics[width=\imw cm]{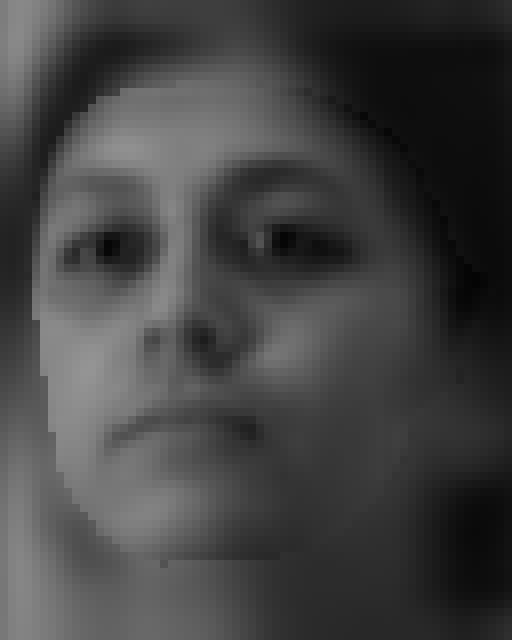}}\par\vspace{\vdiss cm}
		\stepcounter{figure}\addtocounter{figure}{-1}
		\addtocounter{subfigure}{3}
		\subfloat[]{\includegraphics[width=\imw cm]{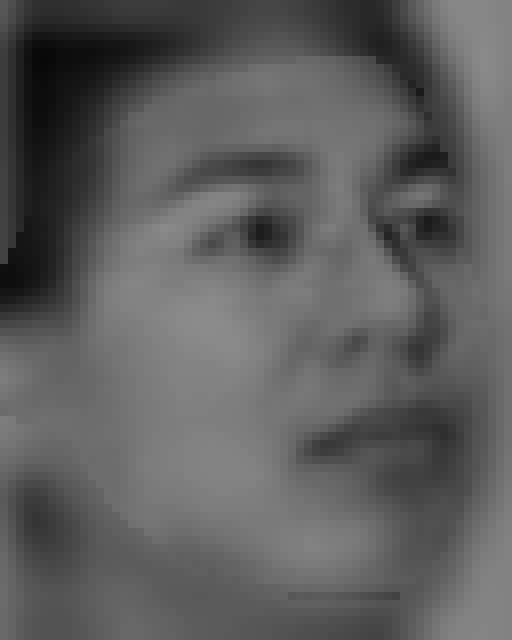}}
	\end{minipage}%
	\begin{minipage}{\hdis\textwidth} 
		\centering
		\subfloat {\includegraphics[width=\imw cm]{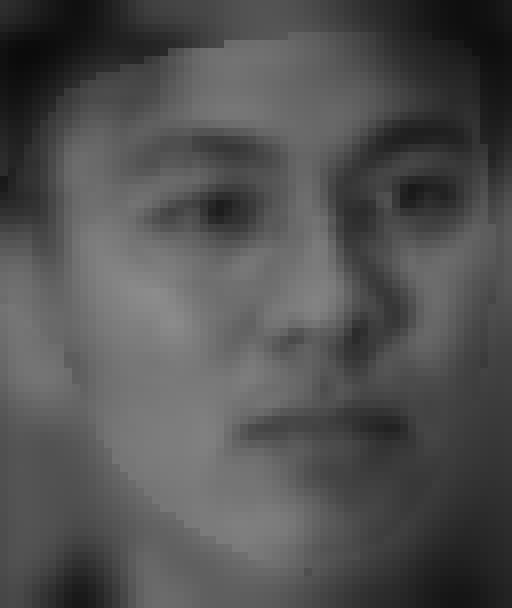}}\par\vspace{\vdiss cm}		
		\subfloat {\includegraphics[width=\imw cm]{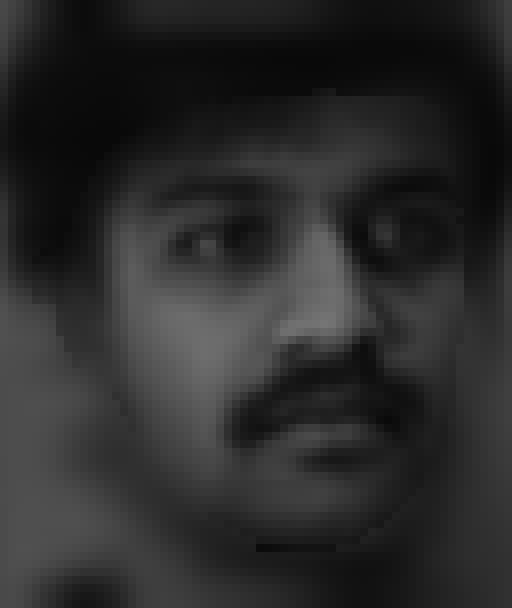}}\par\vspace{\vdiss cm}
		\subfloat {\includegraphics[width=\imw cm]{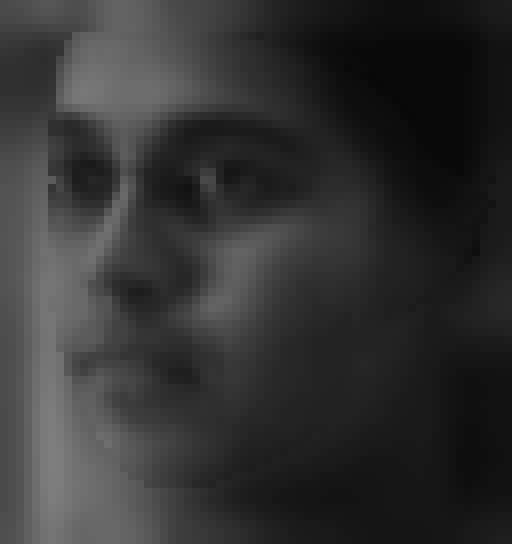}}\par\vspace{\vdiss cm}
		\subfloat {\includegraphics[width=\imw cm]{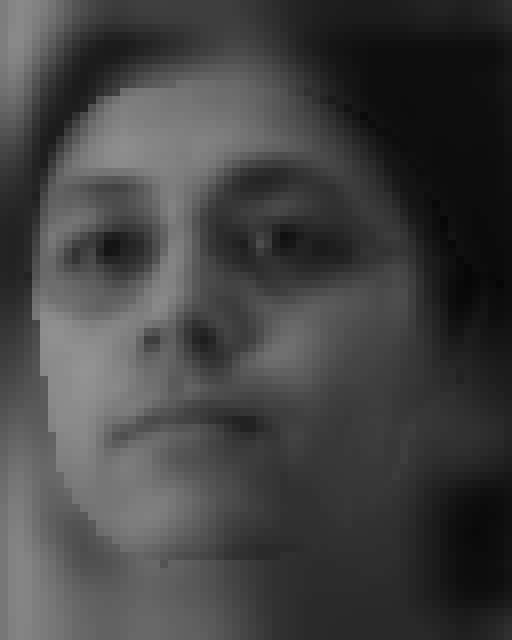}}\par\vspace{\vdiss cm}
		\stepcounter{figure}\addtocounter{figure}{-1}
		\addtocounter{subfigure}{4}
		\subfloat[]{\includegraphics[width=\imw cm]{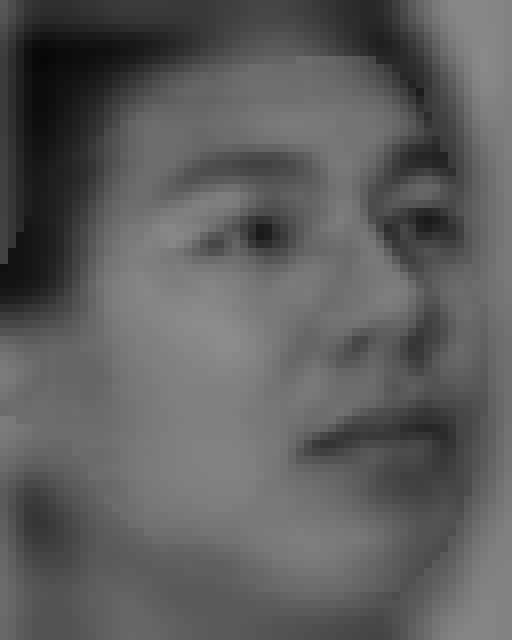}}
	\end{minipage}%
	\begin{minipage}{\hdis\textwidth} 
		\centering
		\subfloat {\includegraphics[width=\imw cm]{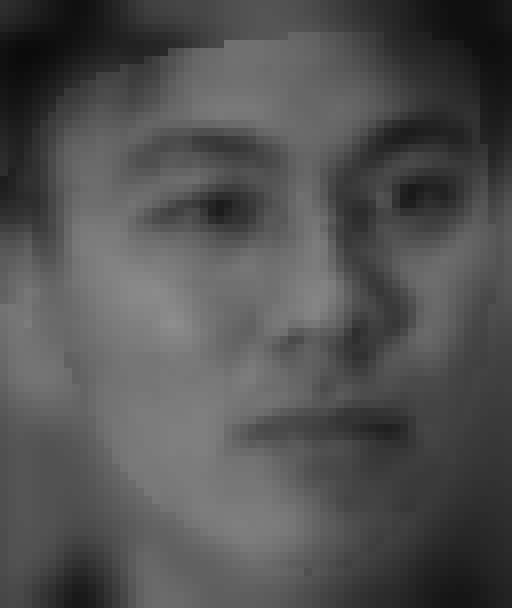}}\par\vspace{\vdiss cm}		
		\subfloat {\includegraphics[width=\imw cm]{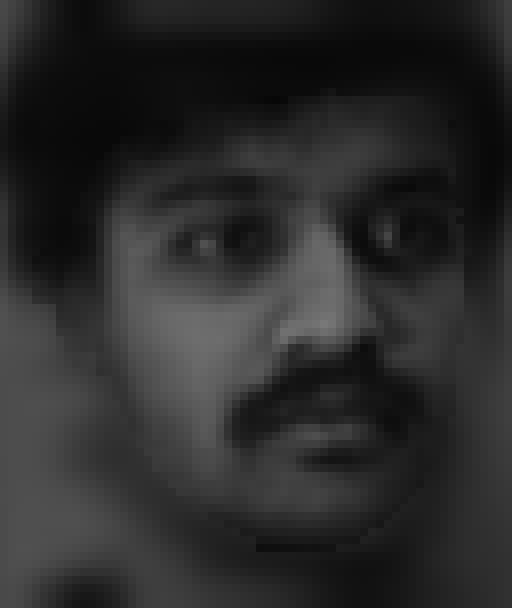}}\par\vspace{\vdiss cm}
		\subfloat {\includegraphics[width=\imw cm]{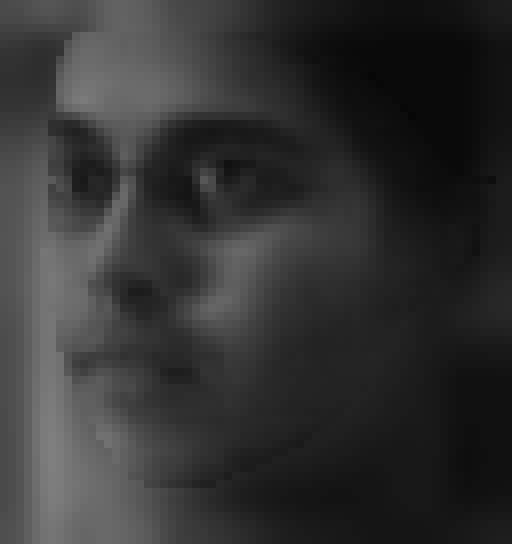}}\par\vspace{\vdiss cm}
		\subfloat {\includegraphics[width=\imw cm]{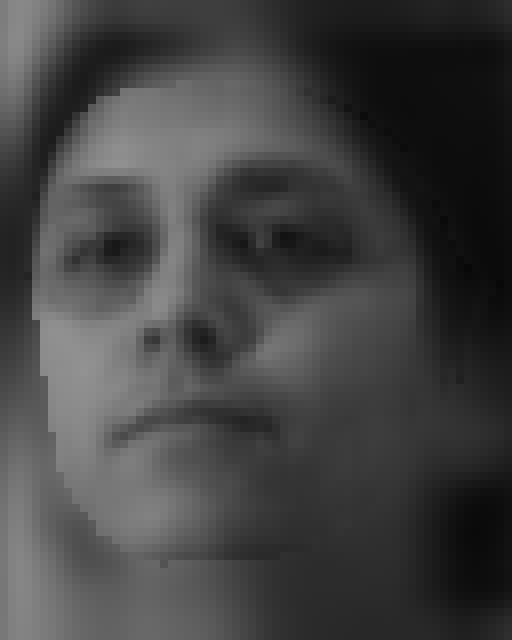}}\par\vspace{\vdiss cm}
		\stepcounter{figure}\addtocounter{figure}{-1}
		\addtocounter{subfigure}{5}
		\subfloat[]{\includegraphics[width=\imw cm]{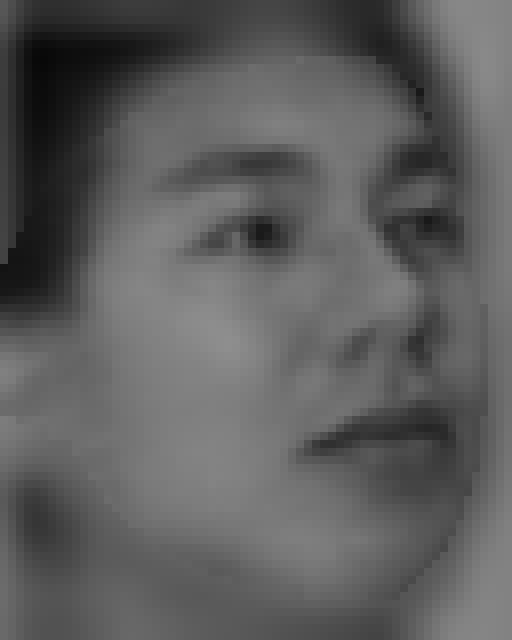}}
	\end{minipage}%
	\begin{minipage}{\hdis\textwidth} 
		\centering
		\subfloat {\includegraphics[width=\imw cm]{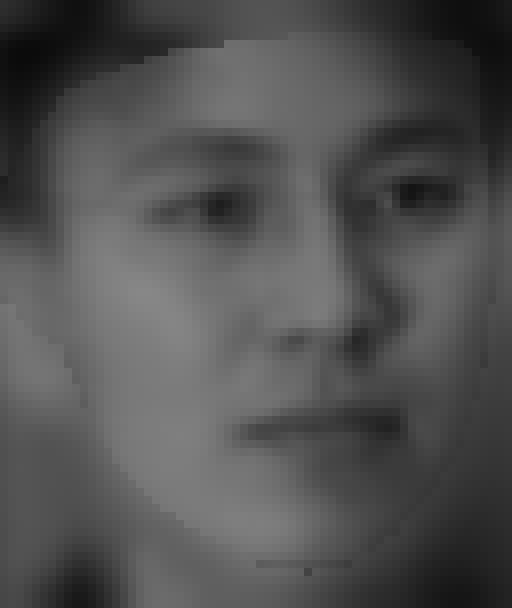}}\par\vspace{\vdiss cm}		
		\subfloat {\includegraphics[width=\imw cm]{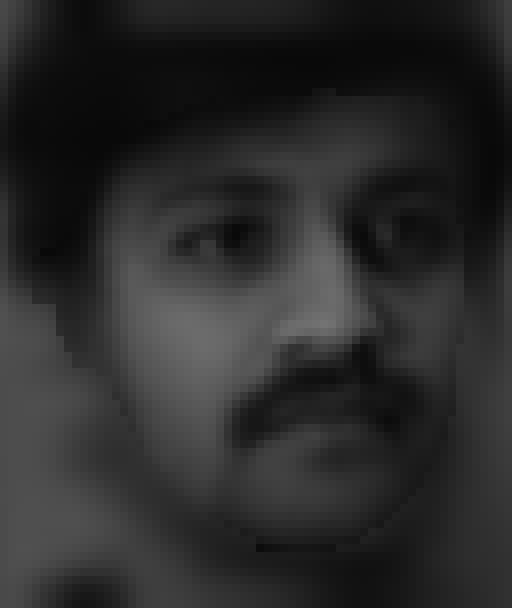}}\par\vspace{\vdiss cm}
		\subfloat {\includegraphics[width=\imw cm]{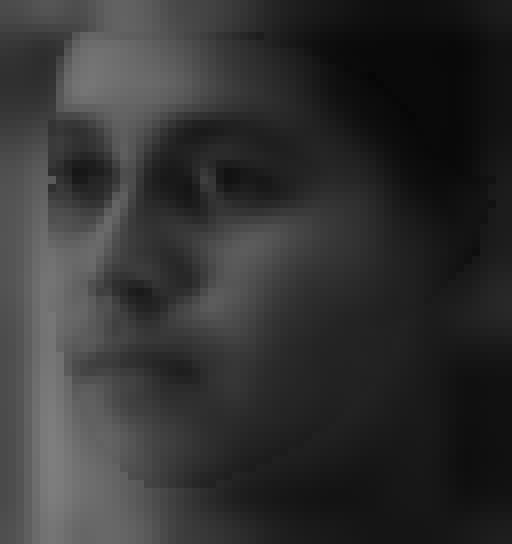}}\par\vspace{\vdiss cm}
		\subfloat {\includegraphics[width=\imw cm]{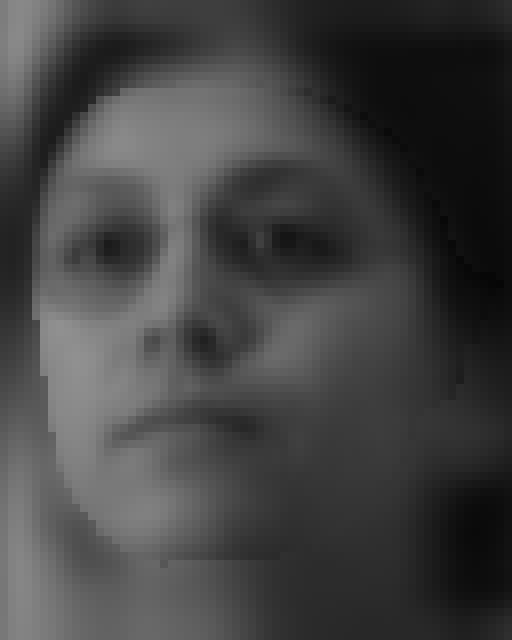}}\par\vspace{\vdiss cm}
		\stepcounter{figure}\addtocounter{figure}{-1}
		\addtocounter{subfigure}{6}
		\subfloat[]{\includegraphics[width=\imw cm]{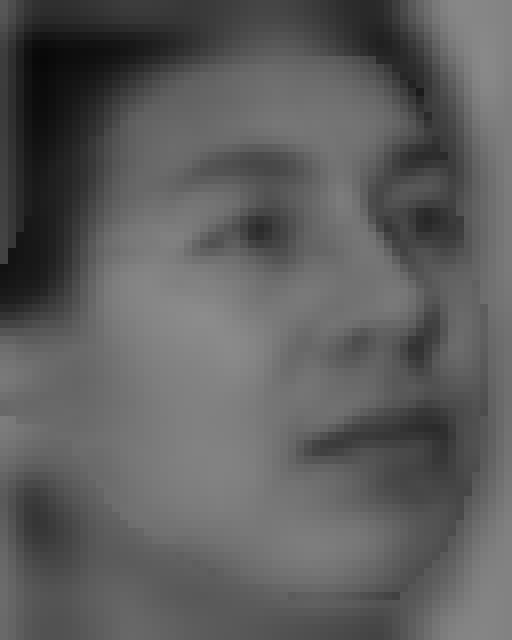}}
	\end{minipage}%
	\begin{minipage}{\hdis\textwidth} 
		\centering
		\subfloat {\includegraphics[width=\imw cm]{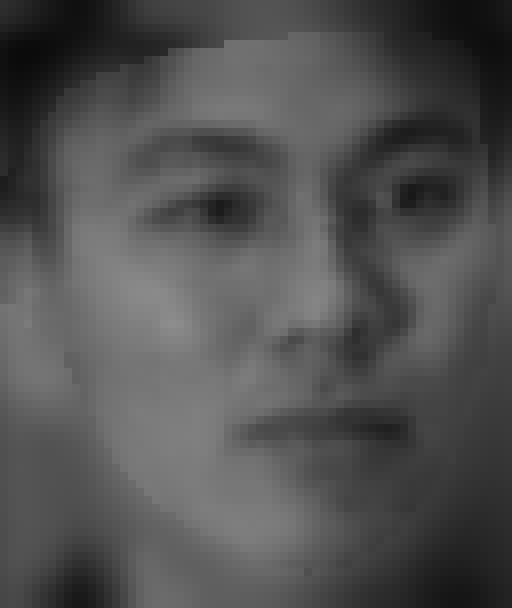}}\par\vspace{\vdiss cm}		
		\subfloat {\includegraphics[width=\imw cm]{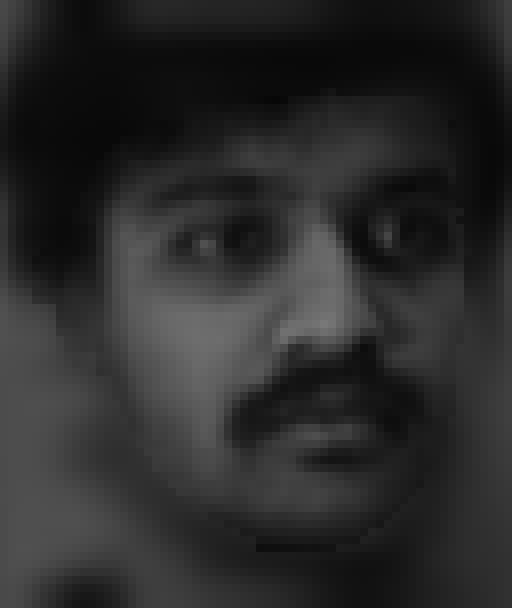}}\par\vspace{\vdiss cm}
		\subfloat {\includegraphics[width=\imw cm]{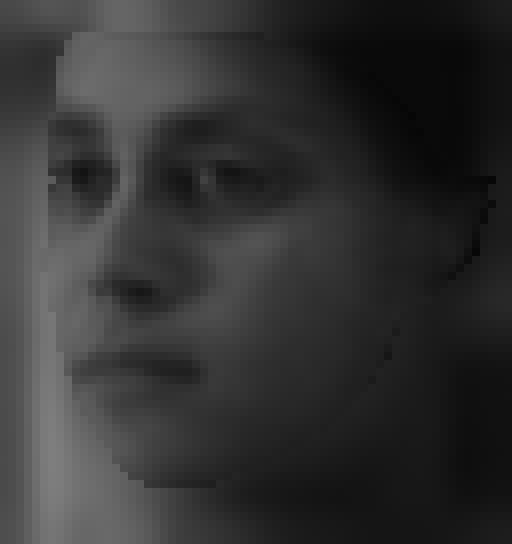}}\par\vspace{\vdiss cm}
		\subfloat {\includegraphics[width=\imw cm]{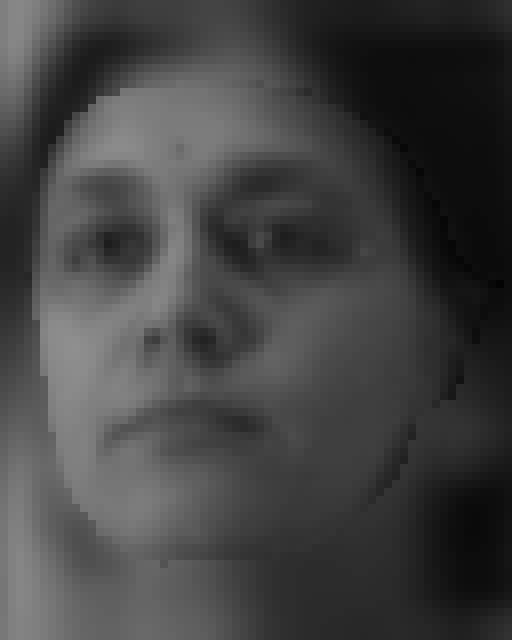}}\par\vspace{\vdiss cm}
		\stepcounter{figure}\addtocounter{figure}{-1}
		\addtocounter{subfigure}{7}
		\subfloat[]{\includegraphics[width=\imw cm]{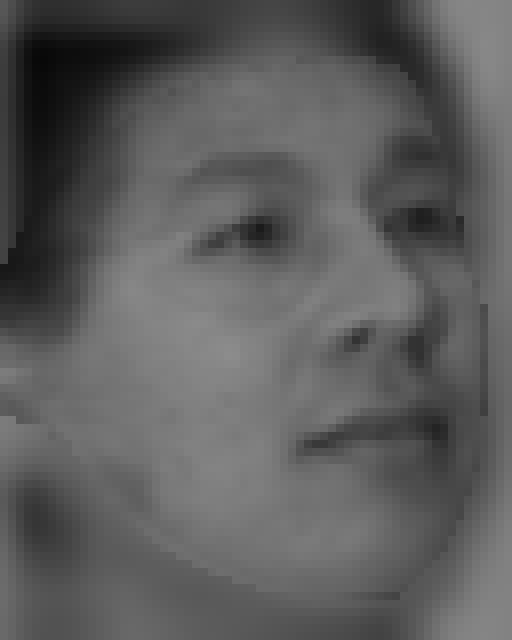}}
	\end{minipage}%
	\begin{minipage}{\hdis\textwidth} 
		\centering		
		\subfloat {\includegraphics[width=\imw cm]{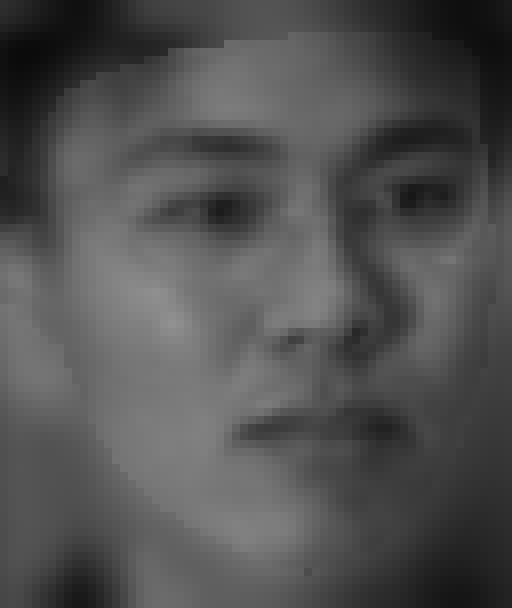}}\par\vspace{\vdiss cm}		
		\subfloat {\includegraphics[width=\imw cm]{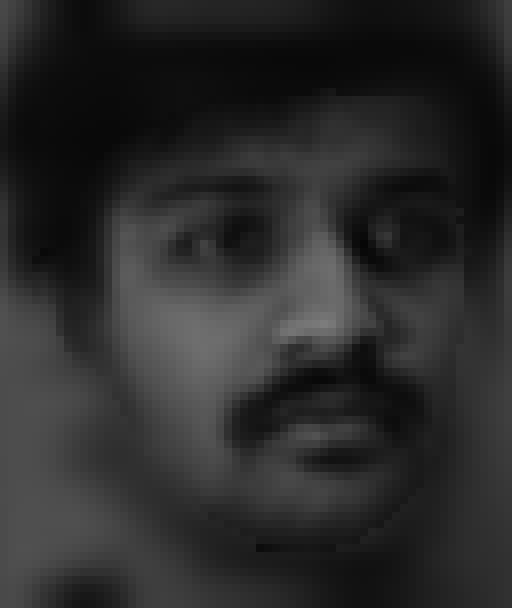}}\par\vspace{\vdiss cm}
		\subfloat {\includegraphics[width=\imw cm]{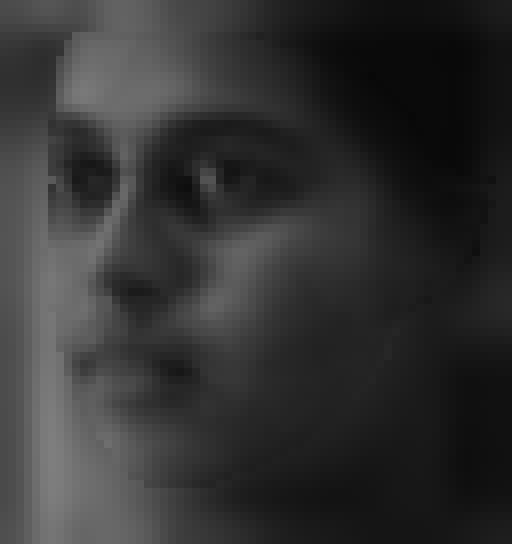}}\par\vspace{\vdiss cm}
		\subfloat {\includegraphics[width=\imw cm]{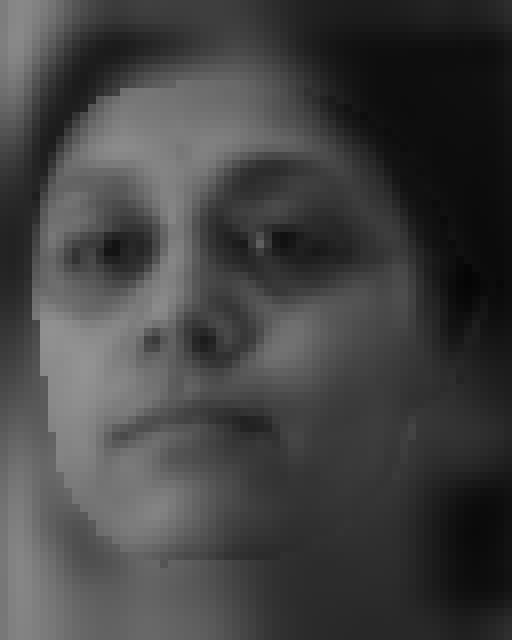}}\par\vspace{\vdiss cm}
		\stepcounter{figure}\addtocounter{figure}{-1}
		\addtocounter{subfigure}{8}
		\subfloat[]{\includegraphics[width=\imw cm]{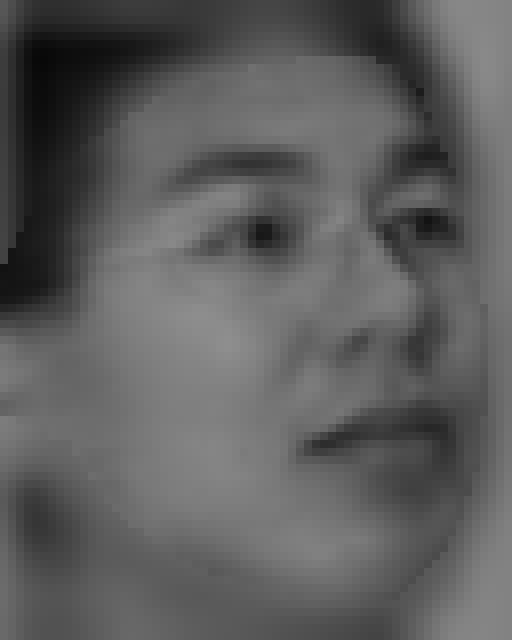}}
	\end{minipage}%
	\begin{minipage}{\hdis\textwidth} 
		\centering
		\subfloat {\includegraphics[width=\imw cm]{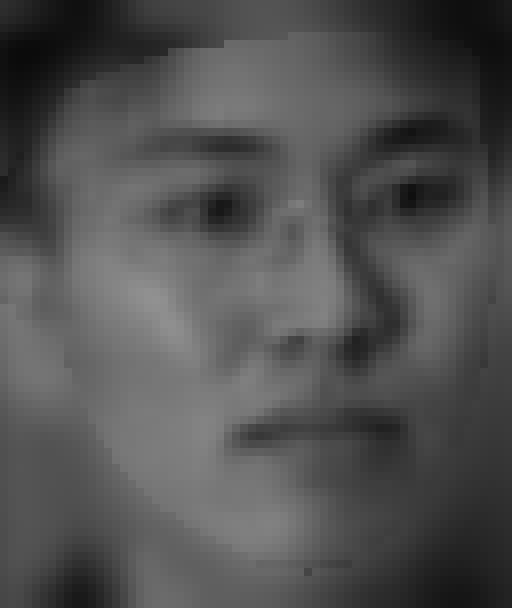}}\par\vspace{\vdiss cm}		
		\subfloat {\includegraphics[width=\imw cm]{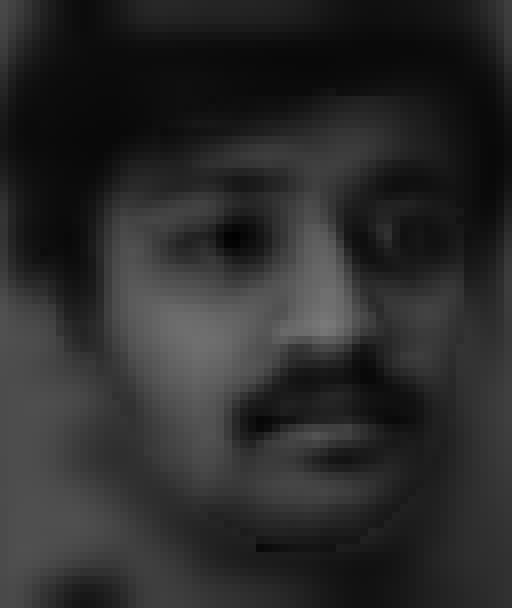}}\par\vspace{\vdiss cm}
		\subfloat {\includegraphics[width=\imw cm]{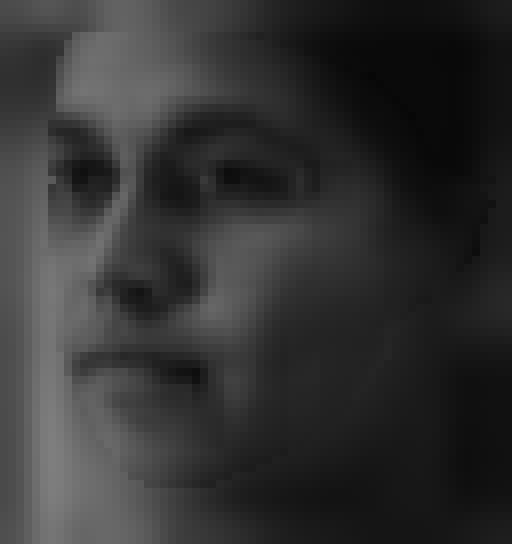}}\par\vspace{\vdiss cm}
		\subfloat {\includegraphics[width=\imw cm]{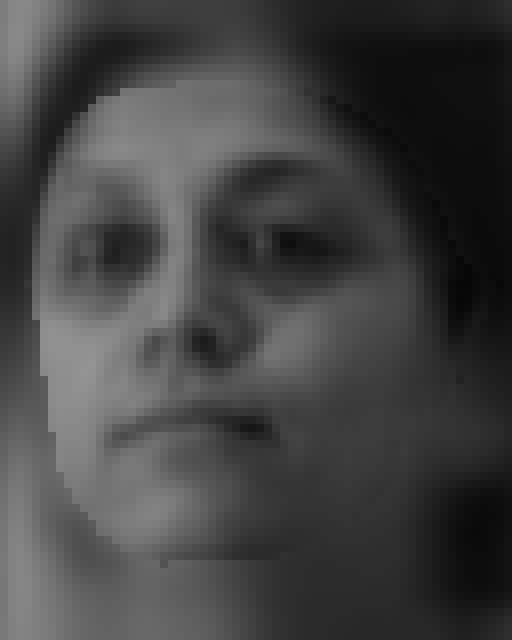}}\par\vspace{\vdiss cm}
		\stepcounter{figure}\addtocounter{figure}{-1}
		\addtocounter{subfigure}{9}
		\subfloat[]{\includegraphics[width=\imw cm]{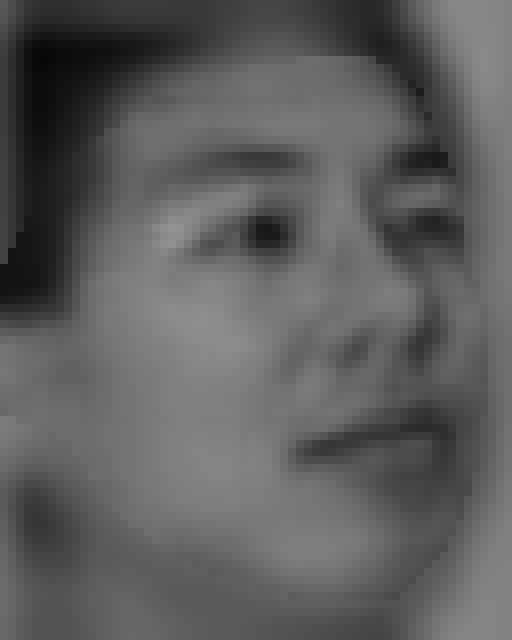}}
	\end{minipage}%
	\begin{minipage}{\hdis\textwidth} 
		\centering
		\subfloat {\includegraphics[width=\imw cm]{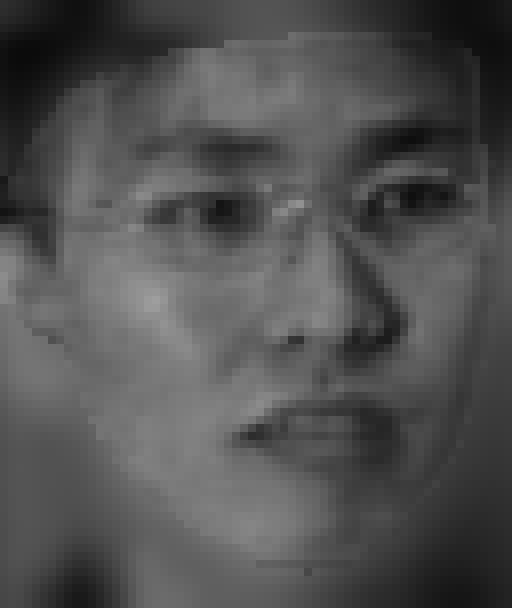}}\par\vspace{\vdiss cm}		
		\subfloat {\includegraphics[width=\imw cm]{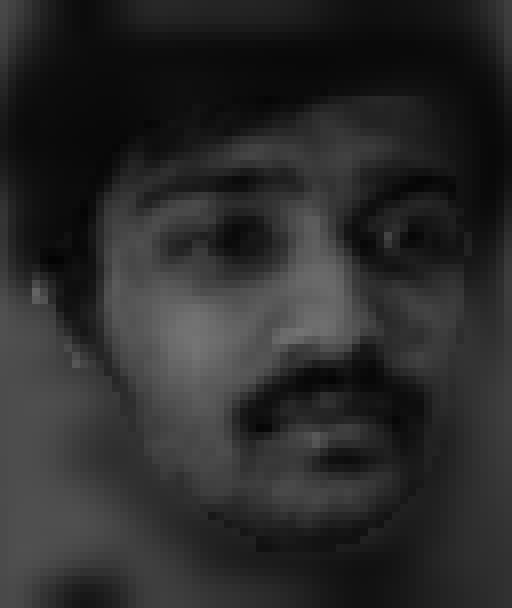}}\par\vspace{\vdiss cm}
		\subfloat {\includegraphics[width=\imw cm]{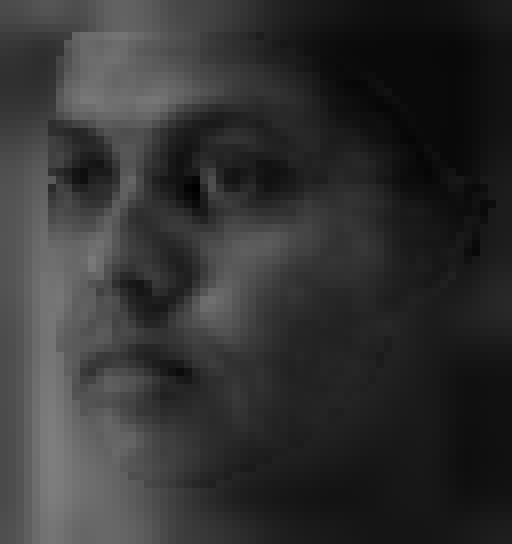}}\par\vspace{\vdiss cm}
		\subfloat {\includegraphics[width=\imw cm]{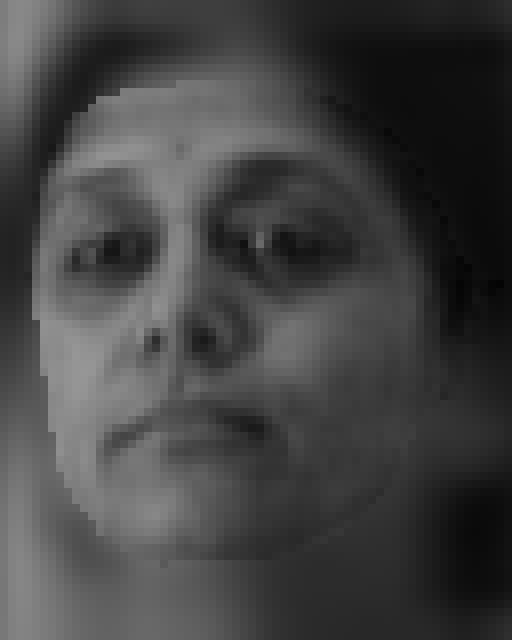}}\par\vspace{\vdiss cm}
		\stepcounter{figure}\addtocounter{figure}{-1}
		\addtocounter{subfigure}{10}
		\subfloat[]{\includegraphics[width=\imw cm]{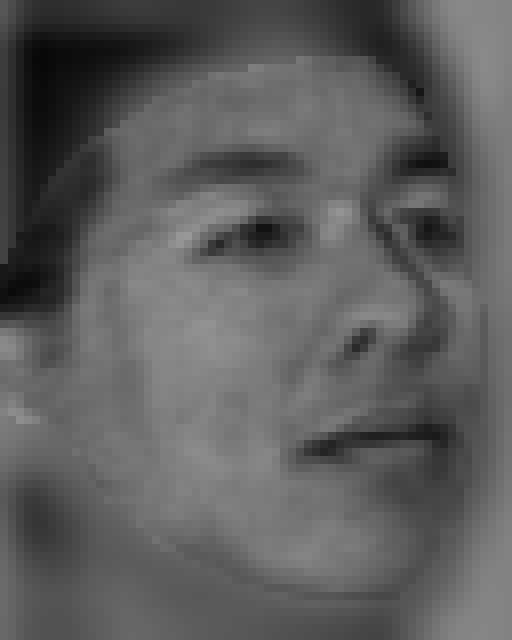}}
	\end{minipage}%
	\begin{minipage}{\hdis\textwidth} 
		\centering
		\subfloat {\includegraphics[width=\imw cm]{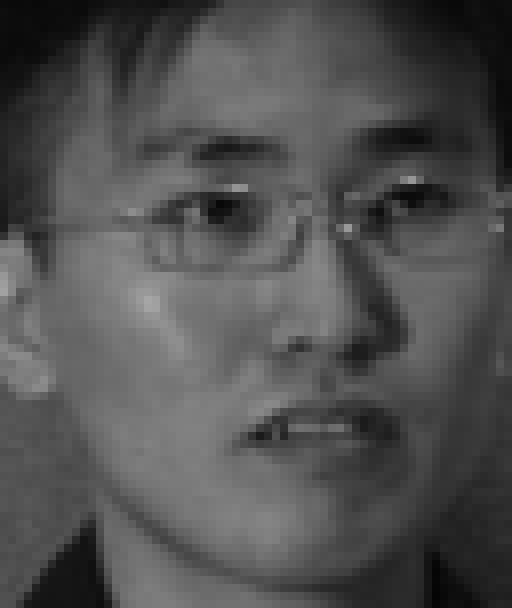}}\par\vspace{\vdiss cm}		
		\subfloat {\includegraphics[width=\imw cm]{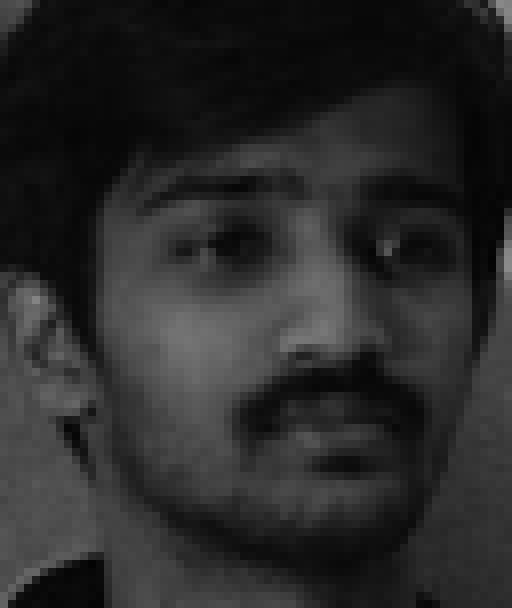}}\par\vspace{\vdiss cm}
		\subfloat {\includegraphics[width=\imw cm]{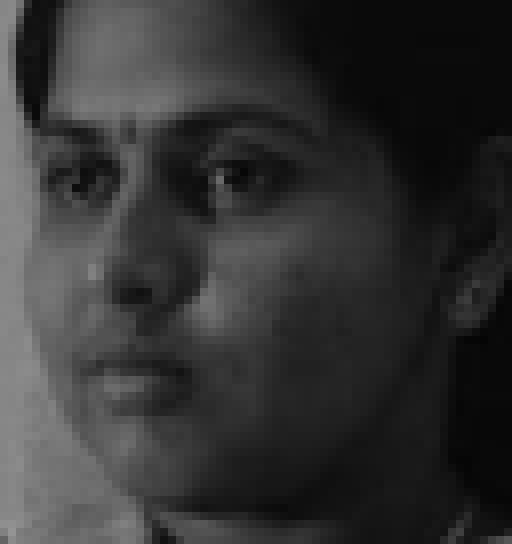}}\par\vspace{\vdiss cm}
		\subfloat {\includegraphics[width=\imw cm]{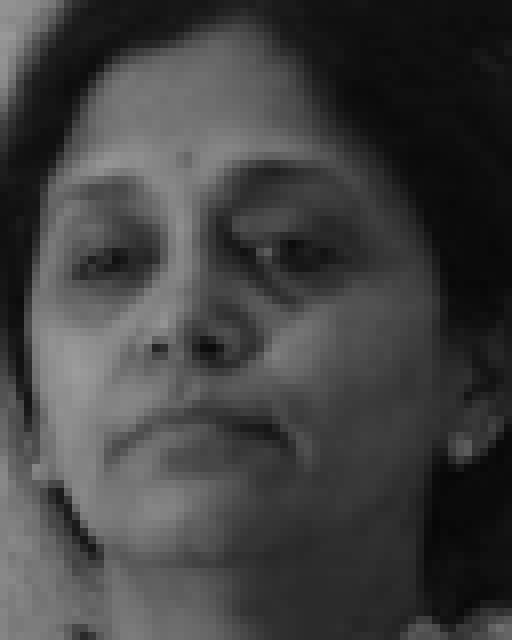}}\par\vspace{\vdiss cm}
		\stepcounter{figure}\addtocounter{figure}{-1}
		\addtocounter{subfigure}{11}
		\subfloat[]{\includegraphics[width=\imw cm]{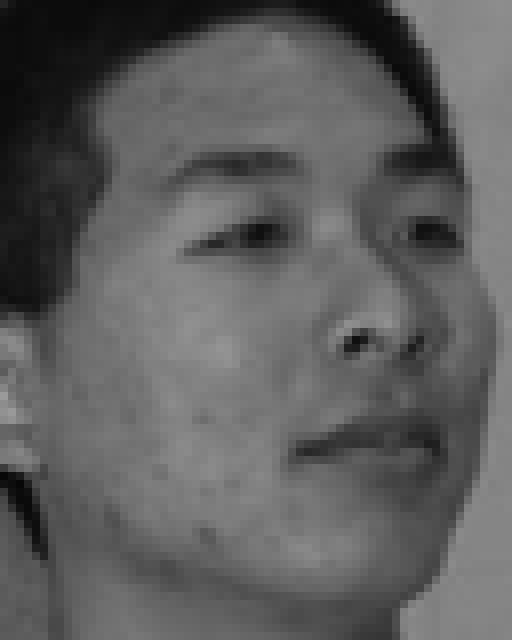}}
	\end{minipage}\par\medskip
	\caption{Face hallucination results generated by different methods using the proposed 3D dictionary alignment scheme on the $20\times20$ test samples of the Multi-PIE face database with different pose variations and scaling factor 4. (a) LR input. (b) Bicubic interpolation. (c) Wang \cite{refs:Wang2005}. (d) LSR \cite{refs:Ma2010}. (e) LcR \cite{refs:Jiang2014a}. (f) LINE \cite{refs:Jiang2014b}. (g) SSR \cite{refs:Jiang2017}. (h) LM-CSS \cite{refs:Farrugia2017}. (i) TRNR \cite{refs:Jiang2016}. (j) TLcR-RL \cite{refs:Jiang2018}. (k) Proposed. (l) Ground truth.}
	\label{fig:multipiepose}
\end{figure*}

\subsubsection{The LFW Dataset (Face Hallucination in the Wild)}
The performance of the proposed method is further evaluated in a real-world scenario, where samples in both the training and test sets are taken in uncontrolled conditions. Among the face images in the LFW database, we consider those belonging to the subjects with 10 to 14 samples in the database, which forms a subset containing 73 subjects and 894 samples. Since the images in the LFW dataset are taken in the wild, unlike the previous experiment, training samples may contain various degradations such as occlusion and illumination variations, which makes the problem even more challenging. Although the proposed alignment procedure is expected to perfectly handle the difficulties related to pose variations, in some infrequent cases, there is a considerable difference in poses of a training sample and the test face, making the registration unfavorable for the task of face hallucination. This often occurs when the two faces are significantly rotated in different directions. One intuitive way to detect these cases is to use image histogram, as an erroneous alignment causes considerable changes in the intensity values of images, Fig. \ref{fig:histreject}. We therefore define a threshold value $\theta$, and accept an alignment if
\begin{equation} \label{eq:verifytrans}
	\| hist(\textbf{I}) - hist(\textbf{I}^{a})  \| < \theta 
\end{equation}
\begin{figure}
	\captionsetup[subfloat]{farskip=0pt,captionskip=1pt}
	\centering
	\def\imw{2.8}
	\def\figw{2.7}
	\def\hdiss{0.5}
	\def\hdisb{0.16}	
	\begin{minipage}{\hdisb\textwidth} 
		\centering
		\stepcounter{figure}\addtocounter{figure}{-1}
		\addtocounter{subfigure}{0}
		\subfloat[]{\includegraphics[width=\imw cm]{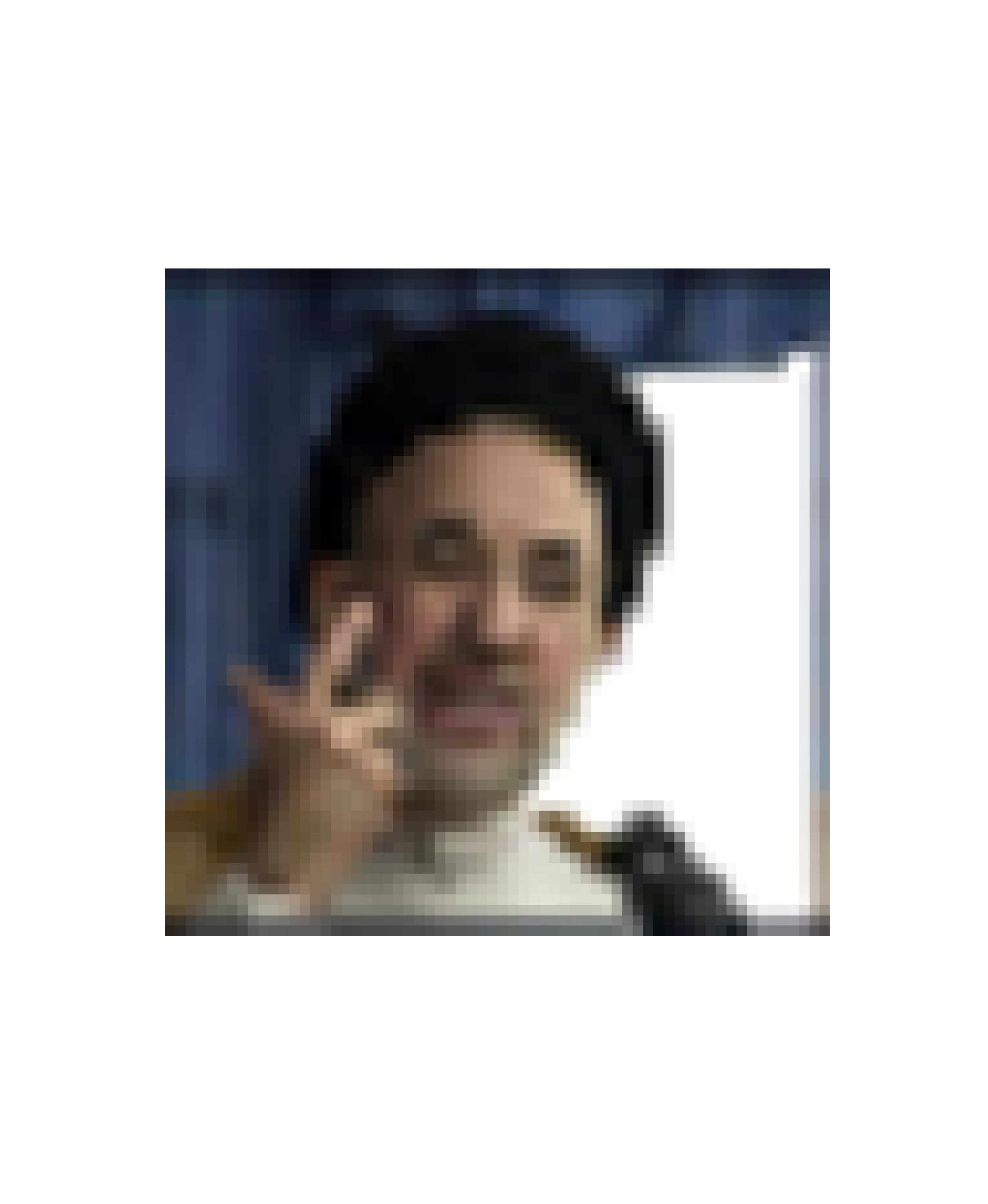}}
	\end{minipage}%
	\begin{minipage}{\hdisb\textwidth} 
		\centering
		\stepcounter{figure}\addtocounter{figure}{-1}
		\addtocounter{subfigure}{1}
		\subfloat[]{\includegraphics[width=\figw cm]{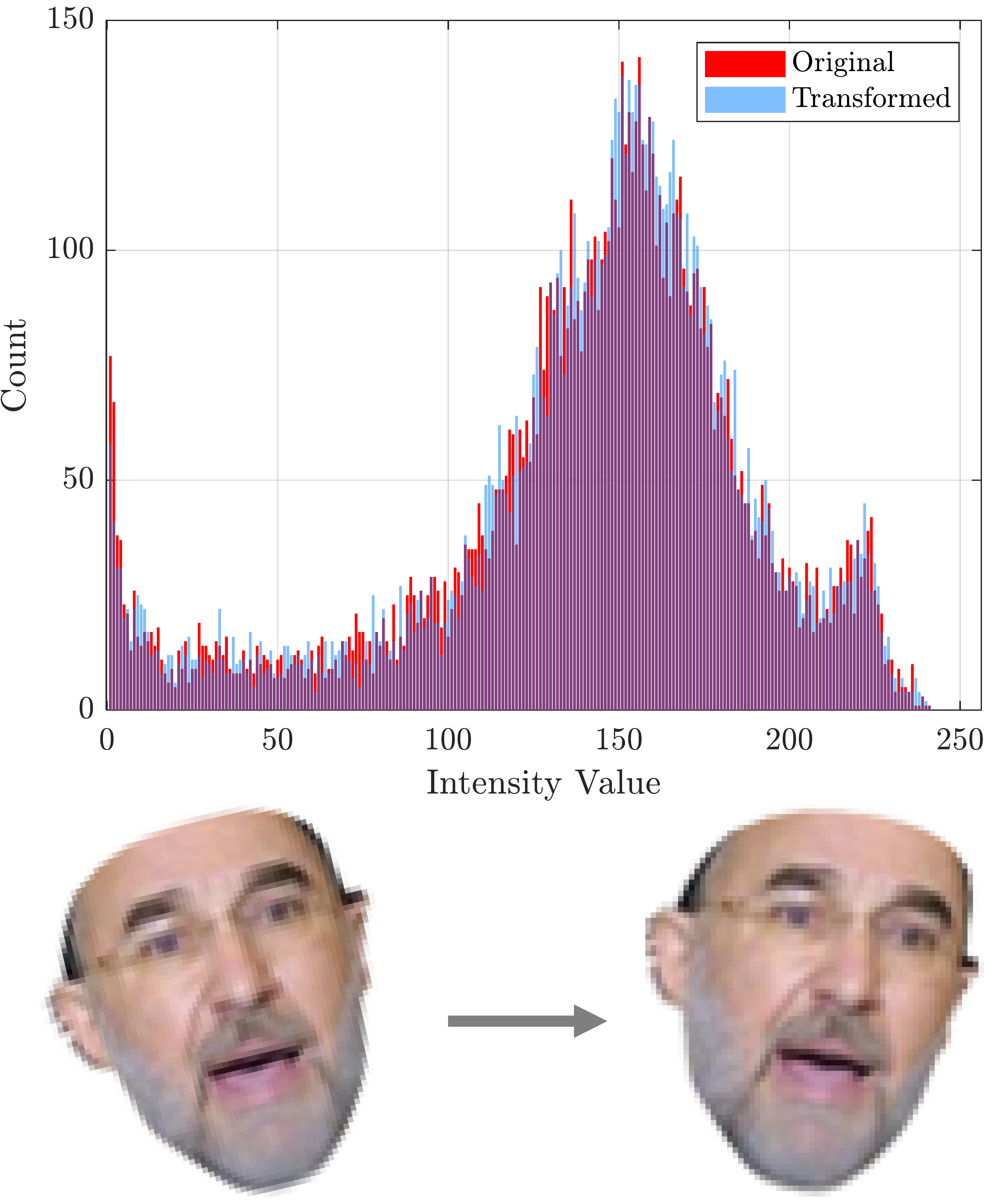}}
	\end{minipage}%
	\begin{minipage}{\hdisb\textwidth} 
		\centering
		\stepcounter{figure}\addtocounter{figure}{-1}
		\addtocounter{subfigure}{2}
		\subfloat[]{\includegraphics[width=\figw cm]{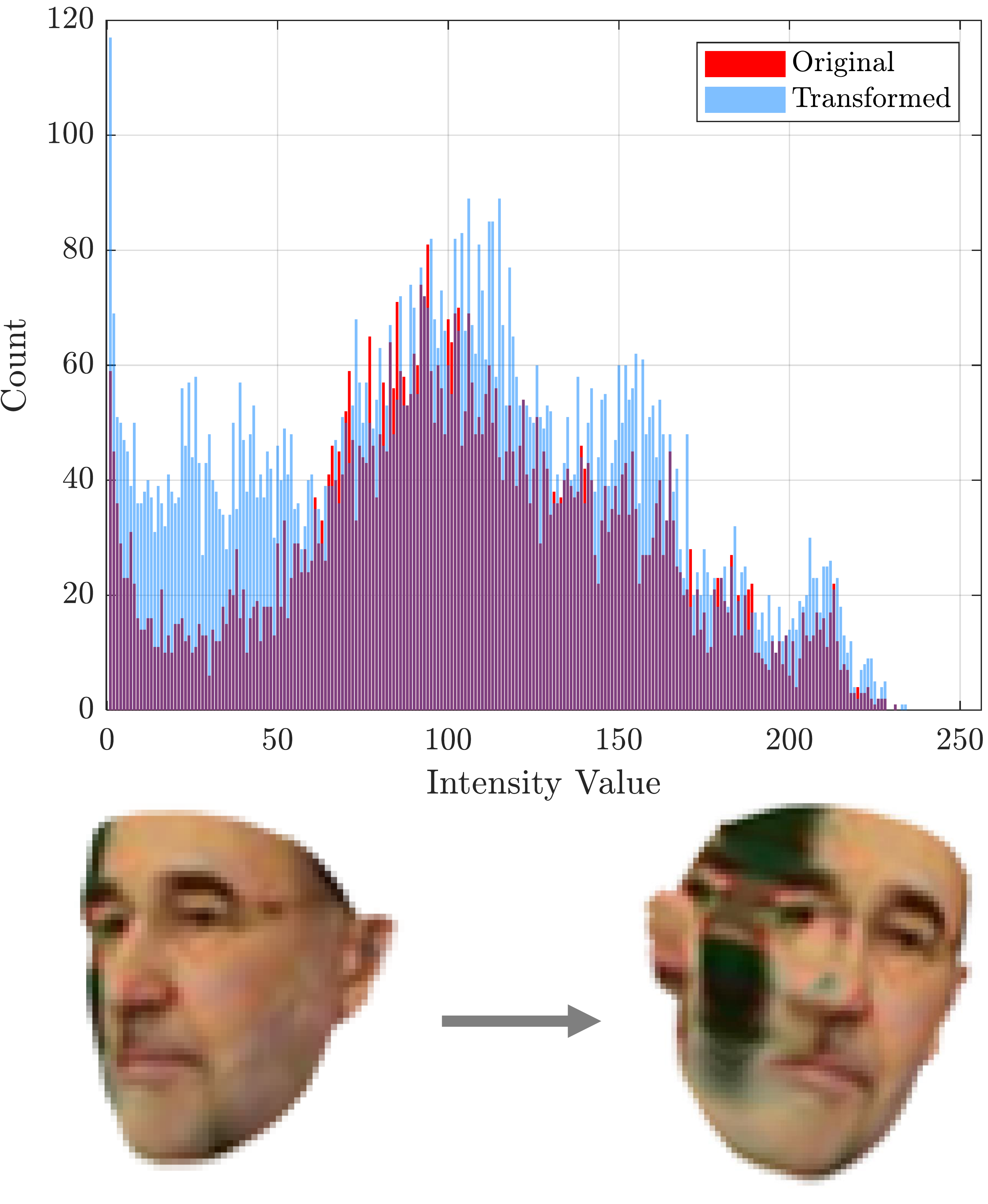}}
	\end{minipage}%
	\caption{Determination of the validity of face transformation based on image histogram. (a) An LR test example. (b) A valid transformation with respect to the test face, in which the histograms of the original and transformed faces are almost identical. (c) An invalid case, where the transformation caused a dark region in the transformed face and consequently changed its histogram.}
	\label{fig:histreject}
\end{figure}where $hist$ returns the histogram of the input image, and $\textbf{I}$ and $\textbf{I}^{a}$ are original and transformed faces, respectively. We empirically consider $\theta = 100$.  Also, the size of the LR faces is $20\times20$ (similar to the previous experiment, the actual face regions are considerably smaller), and the scaling factor is 4. The PSNR and SSIM scores obtained by different methods (Table \ref{tab:poseres}) shows differences of 2.13 dB and 0.0392 in favor of the proposed algorithm, respectively. Also from Fig. \ref{fig:lfwres}, one can observe that the proposed method has added much more information to the reconstructed faces compared to the competitive algorithms. This not only once again highlights the superiority of the proposed face hallucination algorithm over the other approaches, but also shows how efficient our 3D dictionary alignment method is, even when both the training and testing images are taken in uncontrolled conditions.

\begin{figure*}[!t]
	\captionsetup[subfloat]{farskip=0pt,captionskip=1pt}
	\centering
	\def\imw{1.42}
	\def\hdis{0.081}
	\def\vdiss{0.04}
	\begin{minipage}{\hdis\textwidth} 
		\centering
		\subfloat {\includegraphics[width=\imw cm]{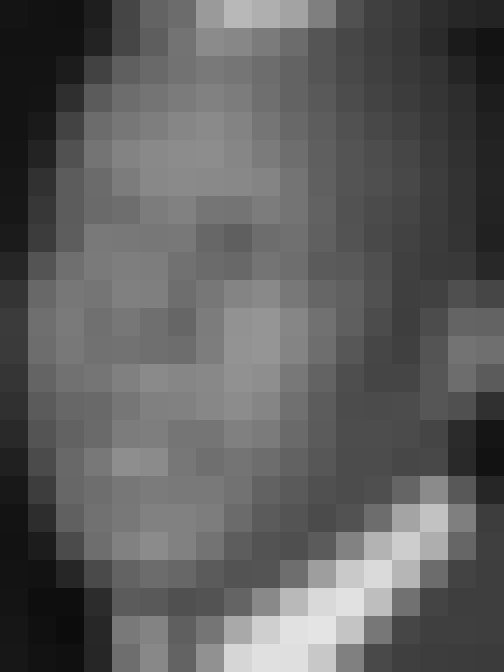}}\par\vspace{\vdiss cm}		
		\subfloat {\includegraphics[width=\imw cm]{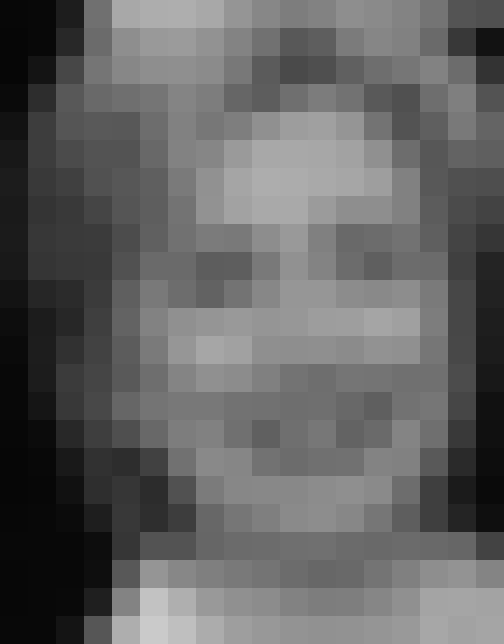}}\par\vspace{\vdiss cm}
		\subfloat {\includegraphics[width=\imw cm]{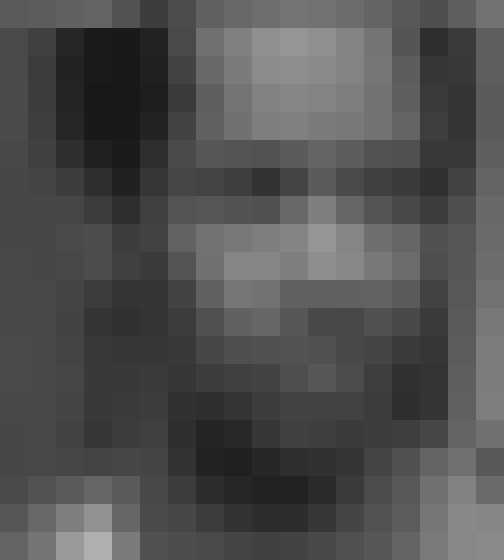}}\par\vspace{\vdiss cm}
		\subfloat {\includegraphics[width=\imw cm]{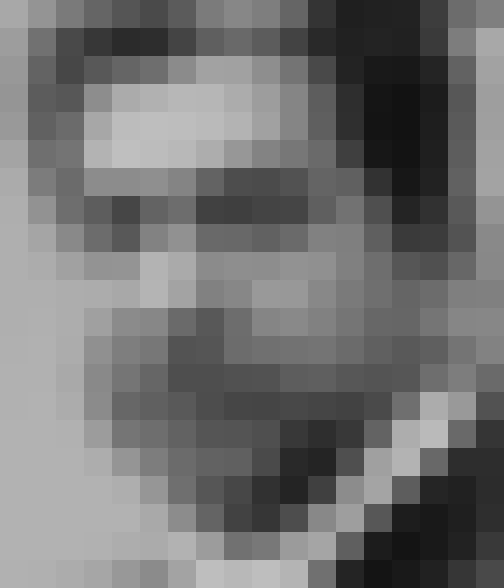}}\par\vspace{\vdiss cm}
		\stepcounter{figure}\addtocounter{figure}{-1}
		\addtocounter{subfigure}{0}
		\subfloat[]{\includegraphics[width=\imw cm]{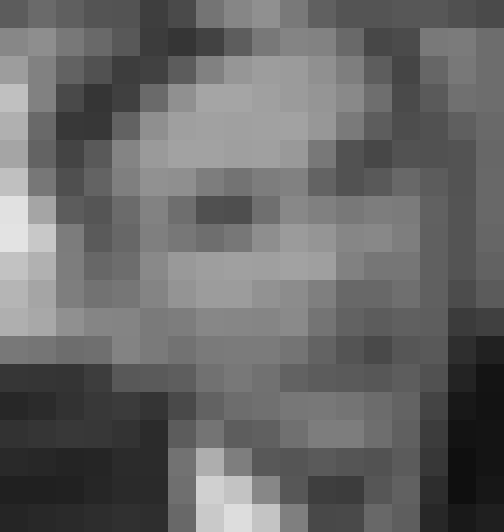}}
	\end{minipage}%
	\begin{minipage}{\hdis\textwidth} 
		\centering
		\subfloat {\includegraphics[width=\imw cm]{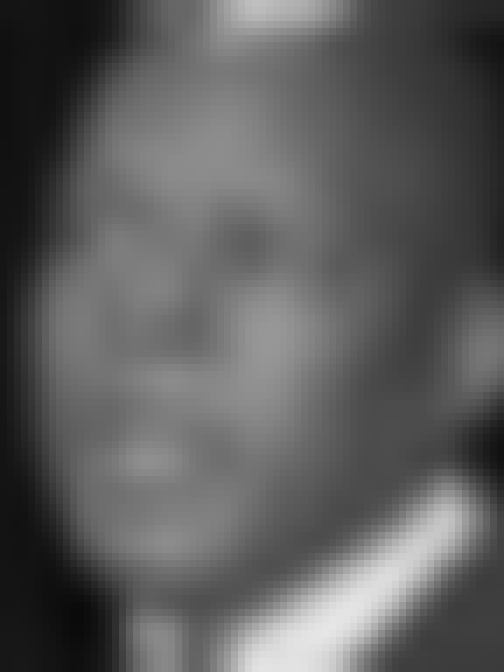}}\par\vspace{\vdiss cm}		
		\subfloat {\includegraphics[width=\imw cm]{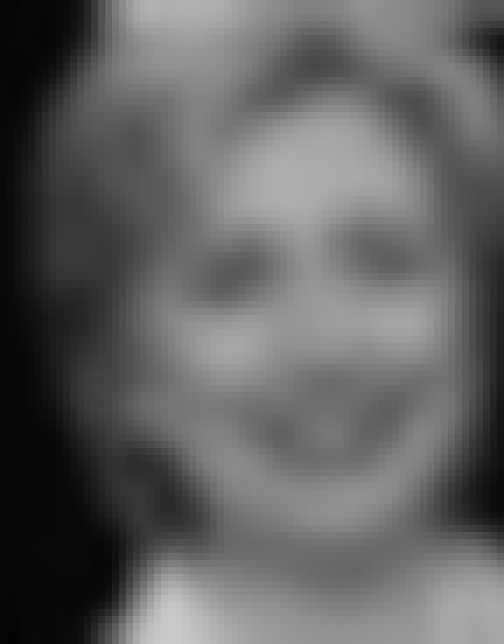}}\par\vspace{\vdiss cm}
		\subfloat {\includegraphics[width=\imw cm]{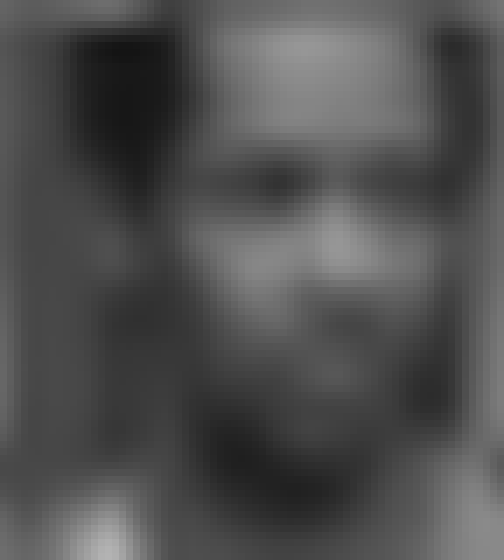}}\par\vspace{\vdiss cm}
		\subfloat {\includegraphics[width=\imw cm]{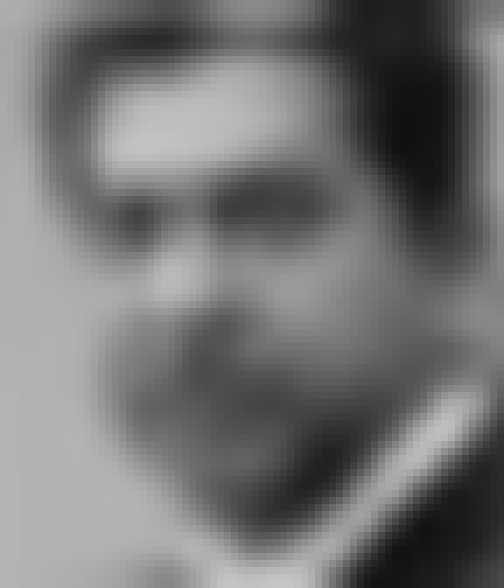}}\par\vspace{\vdiss cm}
		\stepcounter{figure}\addtocounter{figure}{-1}
		\addtocounter{subfigure}{1}
		\subfloat[]{\includegraphics[width=\imw cm]{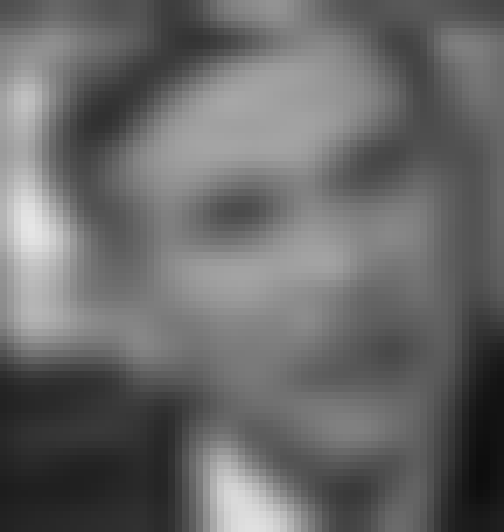}}
	\end{minipage}%
	\begin{minipage}{\hdis\textwidth} 
		\centering
		\subfloat {\includegraphics[width=\imw cm]{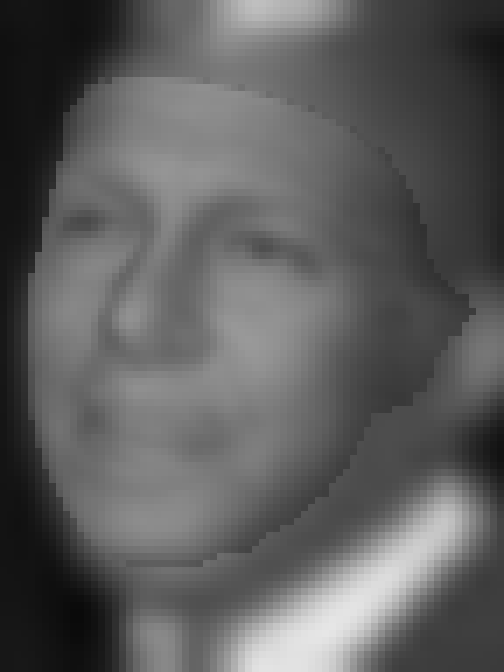}}\par\vspace{\vdiss cm}		
		\subfloat {\includegraphics[width=\imw cm]{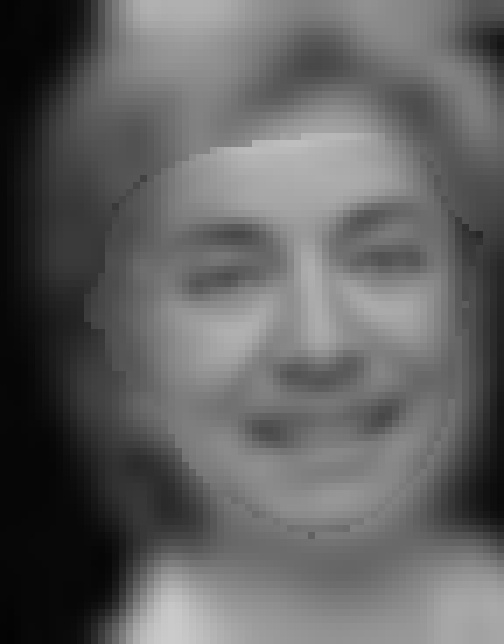}}\par\vspace{\vdiss cm}
		\subfloat {\includegraphics[width=\imw cm]{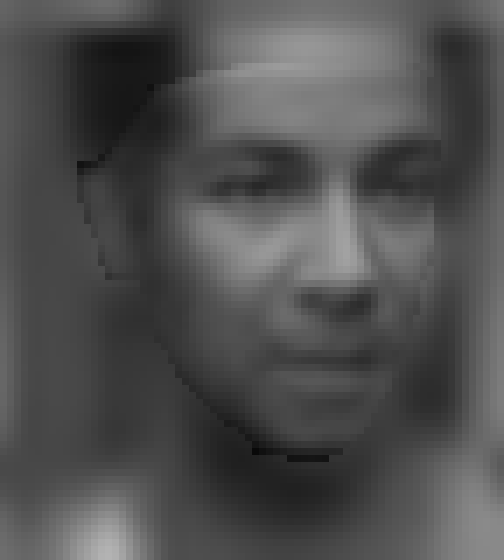}}\par\vspace{\vdiss cm}
		\subfloat {\includegraphics[width=\imw cm]{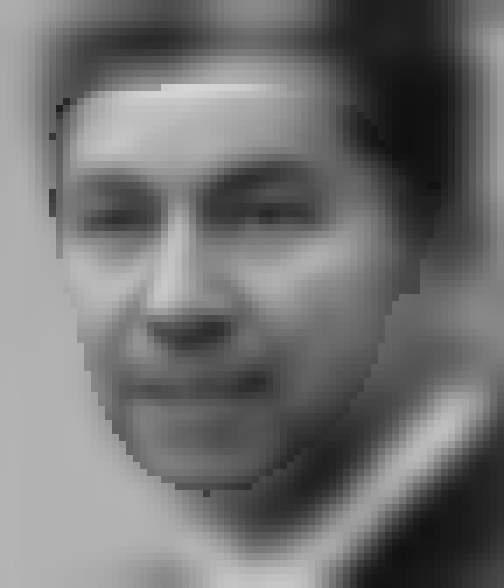}}\par\vspace{\vdiss cm}
		\stepcounter{figure}\addtocounter{figure}{-1}
		\addtocounter{subfigure}{2}
		\subfloat[]{\includegraphics[width=\imw cm]{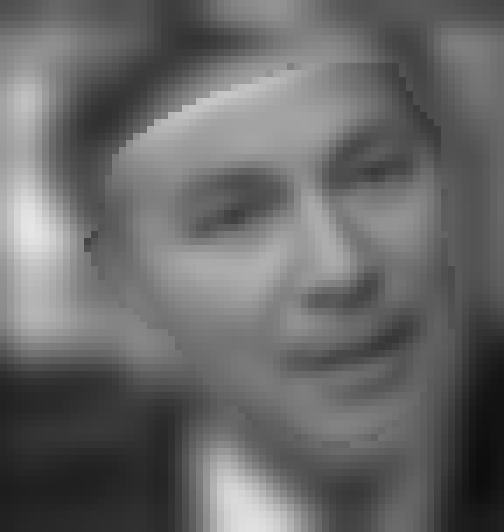}}
	\end{minipage}%
	\begin{minipage}{\hdis\textwidth} 
		\centering
		\subfloat {\includegraphics[width=\imw cm]{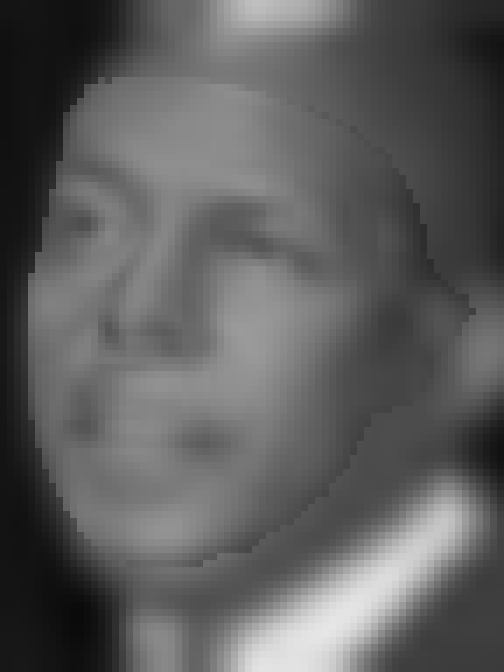}}\par\vspace{\vdiss cm}		
		\subfloat {\includegraphics[width=\imw cm]{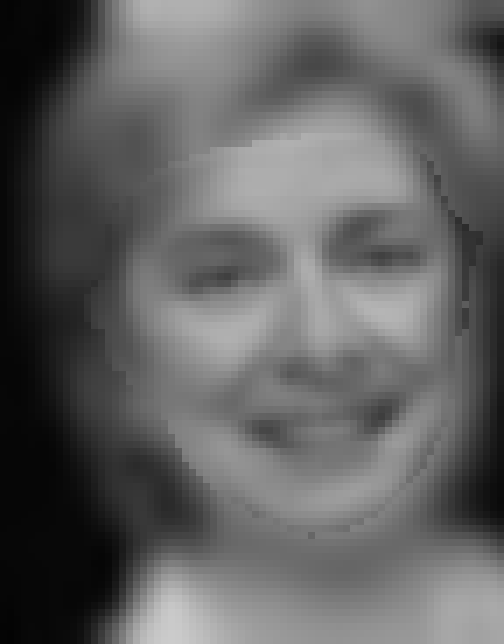}}\par\vspace{\vdiss cm}
		\subfloat {\includegraphics[width=\imw cm]{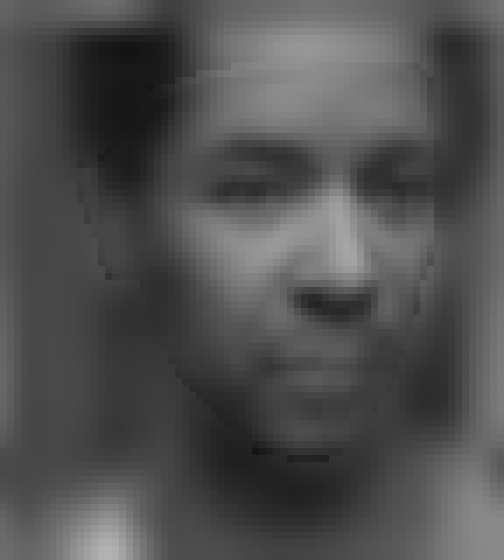}}\par\vspace{\vdiss cm}
		\subfloat {\includegraphics[width=\imw cm]{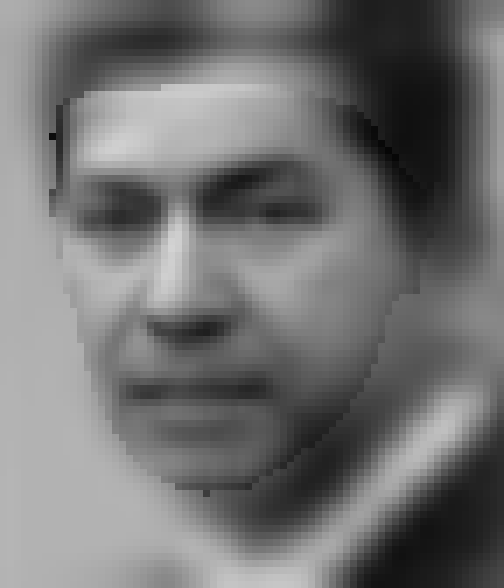}}\par\vspace{\vdiss cm}
		\stepcounter{figure}\addtocounter{figure}{-1}
		\addtocounter{subfigure}{3}
		\subfloat[]{\includegraphics[width=\imw cm]{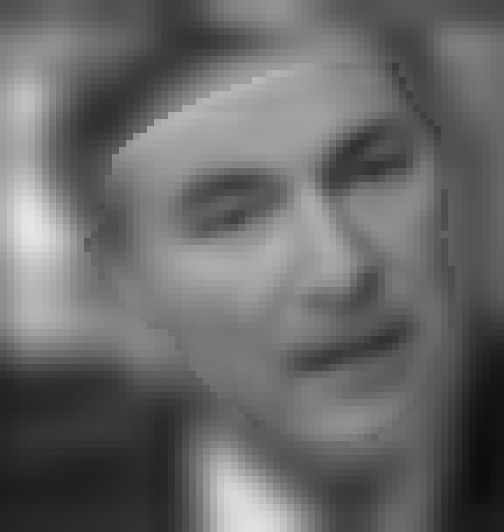}}
	\end{minipage}%
	\begin{minipage}{\hdis\textwidth} 
		\centering
		\subfloat {\includegraphics[width=\imw cm]{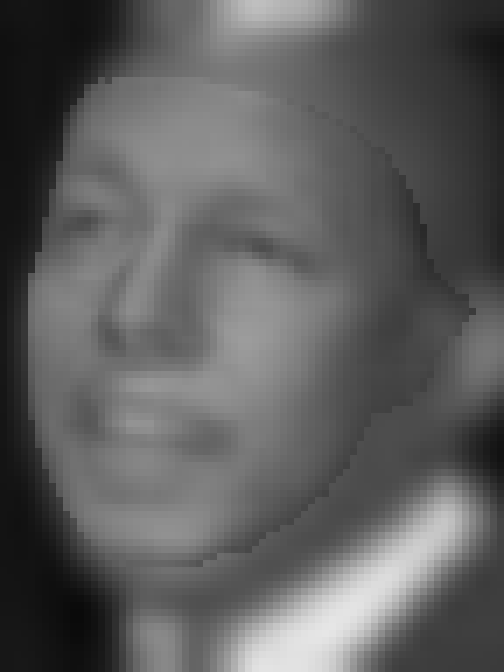}}\par\vspace{\vdiss cm}		
		\subfloat {\includegraphics[width=\imw cm]{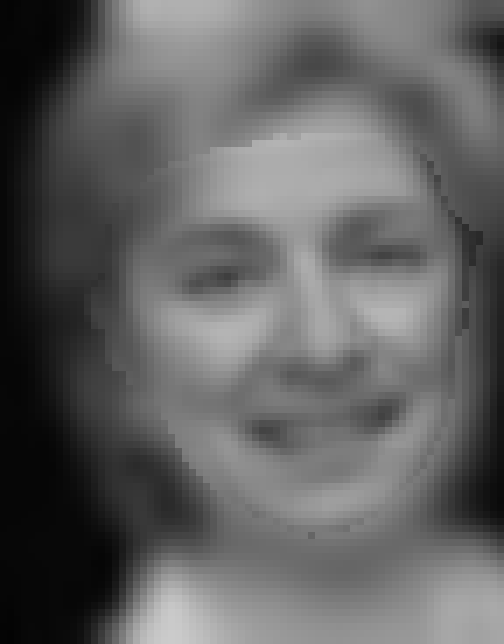}}\par\vspace{\vdiss cm}
		\subfloat {\includegraphics[width=\imw cm]{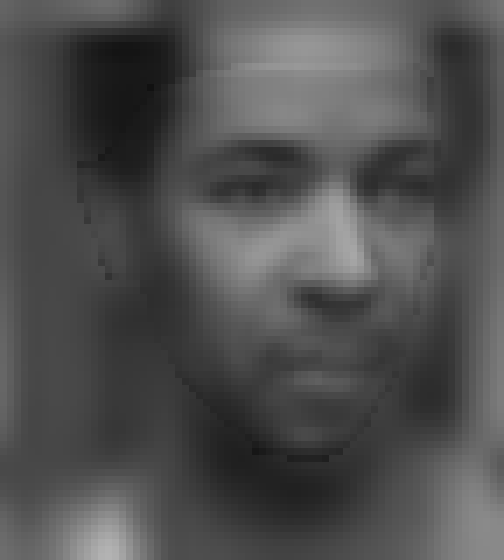}}\par\vspace{\vdiss cm}
		\subfloat {\includegraphics[width=\imw cm]{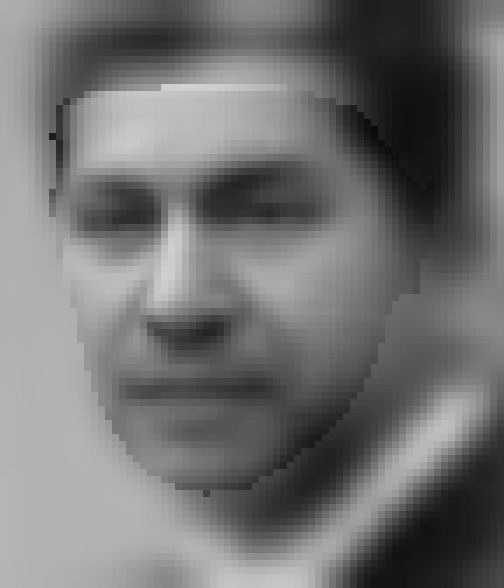}}\par\vspace{\vdiss cm}
		\stepcounter{figure}\addtocounter{figure}{-1}
		\addtocounter{subfigure}{4}
		\subfloat[]{\includegraphics[width=\imw cm]{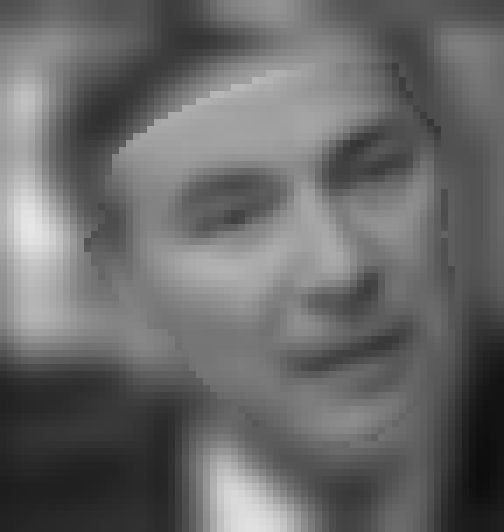}}
	\end{minipage}%
	\begin{minipage}{\hdis\textwidth} 
		\centering
		\subfloat {\includegraphics[width=\imw cm]{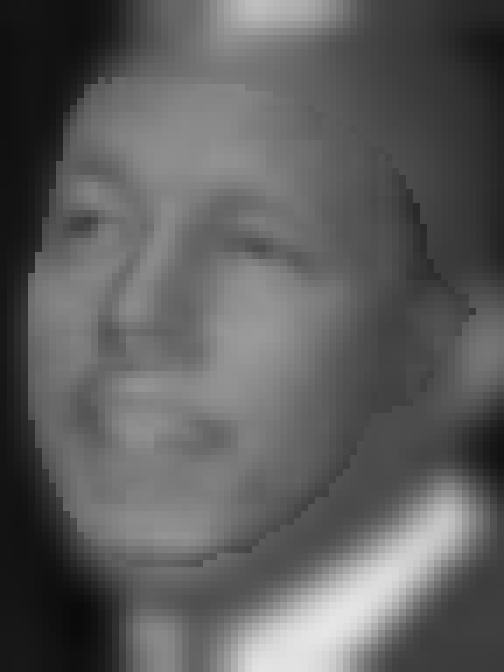}}\par\vspace{\vdiss cm}		
		\subfloat {\includegraphics[width=\imw cm]{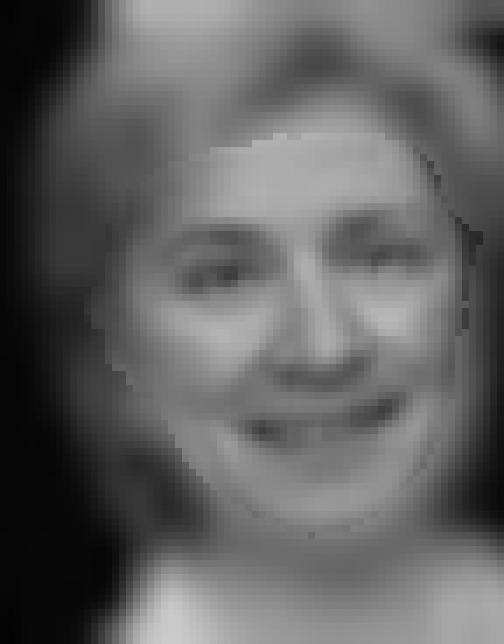}}\par\vspace{\vdiss cm}
		\subfloat {\includegraphics[width=\imw cm]{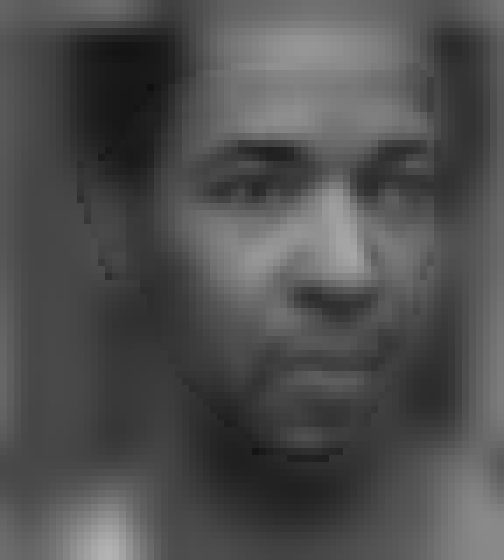}}\par\vspace{\vdiss cm}
		\subfloat {\includegraphics[width=\imw cm]{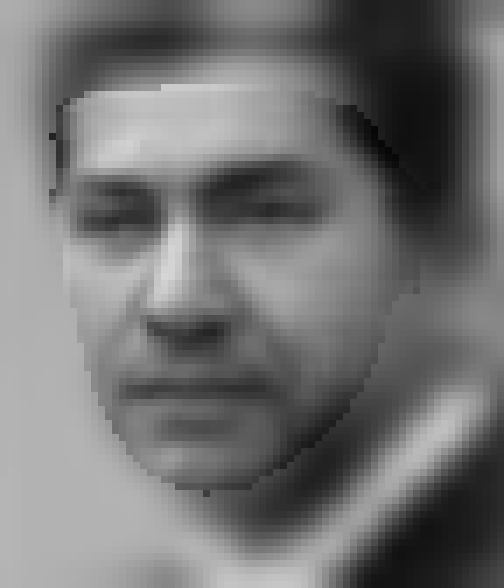}}\par\vspace{\vdiss cm}
		\stepcounter{figure}\addtocounter{figure}{-1}
		\addtocounter{subfigure}{5}
		\subfloat[]{\includegraphics[width=\imw cm]{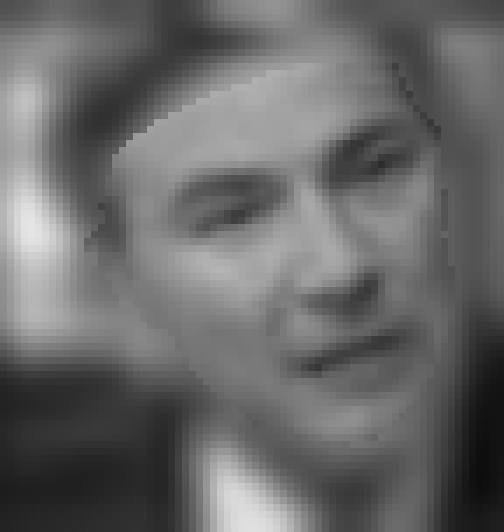}}
	\end{minipage}%
	\begin{minipage}{\hdis\textwidth} 
		\centering
		\subfloat {\includegraphics[width=\imw cm]{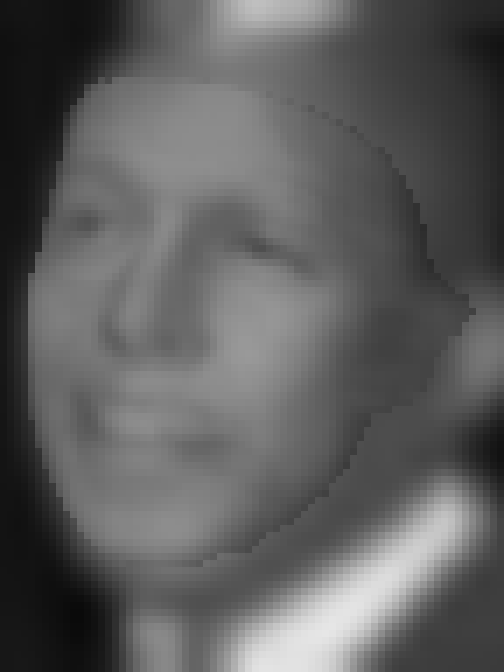}}\par\vspace{\vdiss cm}		
		\subfloat {\includegraphics[width=\imw cm]{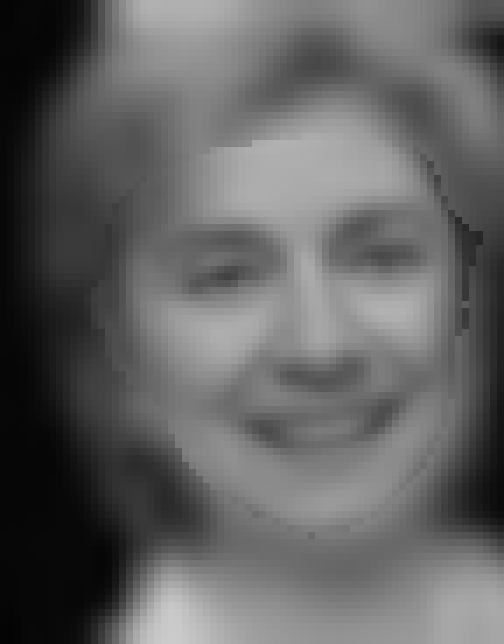}}\par\vspace{\vdiss cm}
		\subfloat {\includegraphics[width=\imw cm]{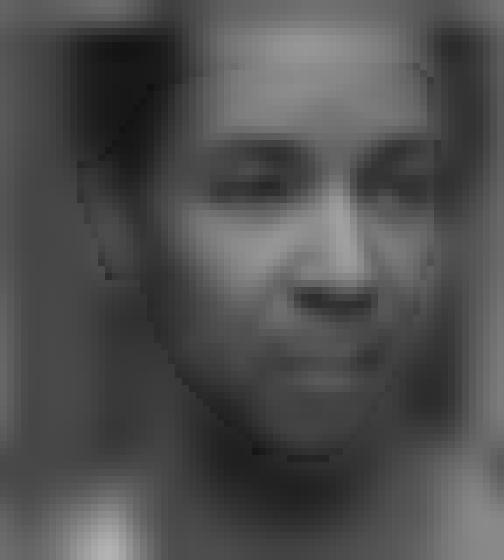}}\par\vspace{\vdiss cm}
		\subfloat {\includegraphics[width=\imw cm]{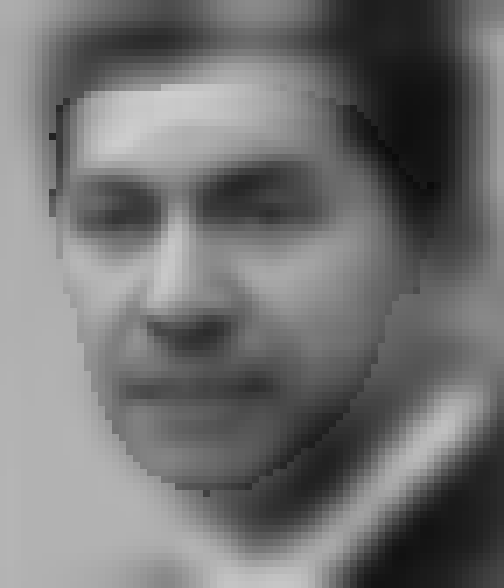}}\par\vspace{\vdiss cm}
		\stepcounter{figure}\addtocounter{figure}{-1}
		\addtocounter{subfigure}{6}
		\subfloat[]{\includegraphics[width=\imw cm]{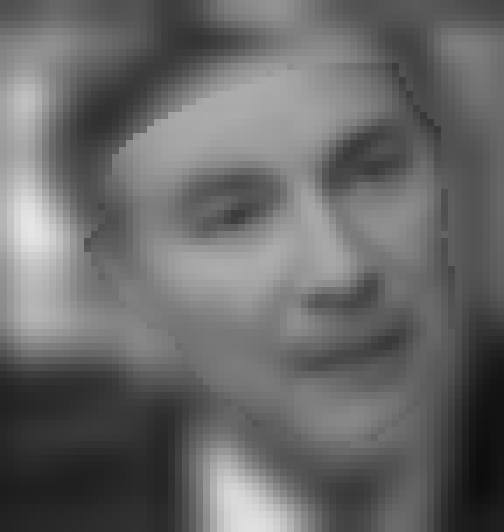}}
	\end{minipage}%
	\begin{minipage}{\hdis\textwidth} 
		\centering
		\subfloat {\includegraphics[width=\imw cm]{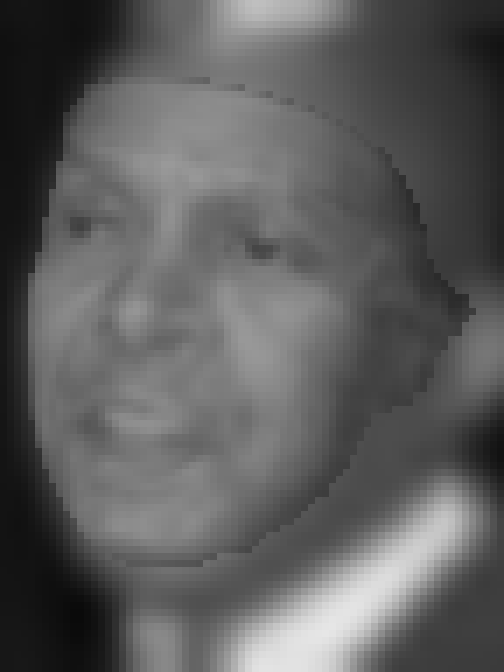}}\par\vspace{\vdiss cm}		
		\subfloat {\includegraphics[width=\imw cm]{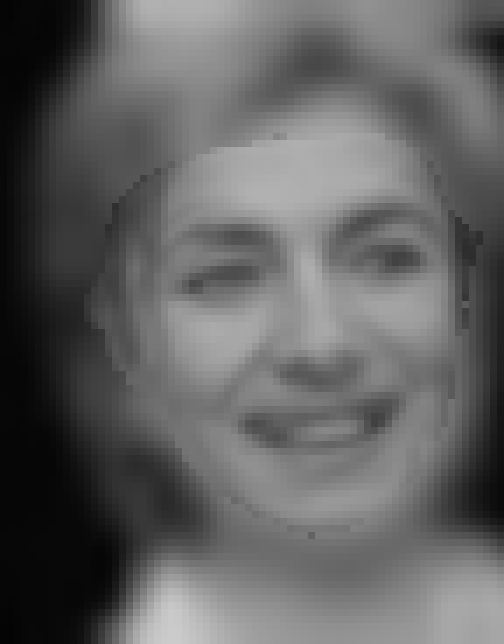}}\par\vspace{\vdiss cm}
		\subfloat {\includegraphics[width=\imw cm]{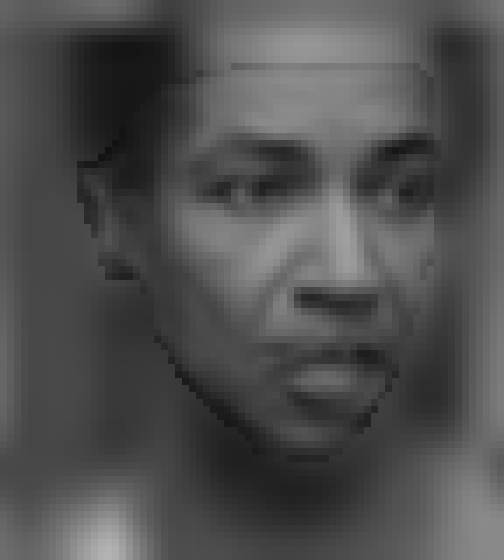}}\par\vspace{\vdiss cm}
		\subfloat {\includegraphics[width=\imw cm]{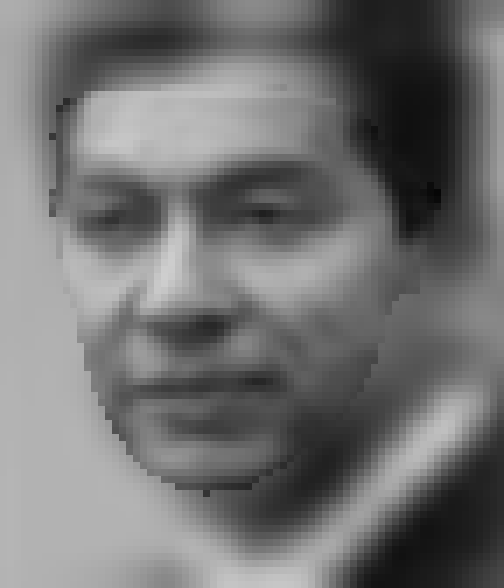}}\par\vspace{\vdiss cm}
		\stepcounter{figure}\addtocounter{figure}{-1}
		\addtocounter{subfigure}{7}
		\subfloat[]{\includegraphics[width=\imw cm]{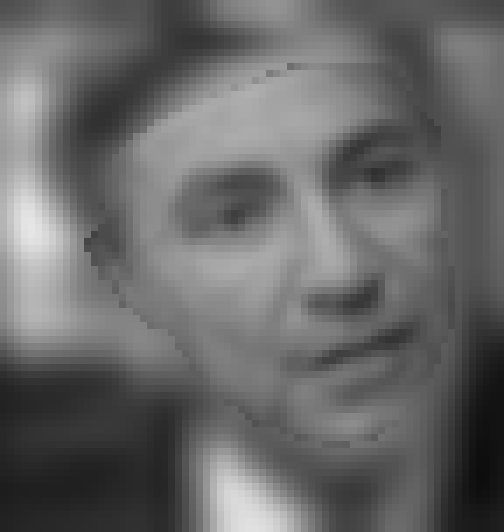}}
	\end{minipage}%
	\begin{minipage}{\hdis\textwidth} 
		\centering		
		\subfloat {\includegraphics[width=\imw cm]{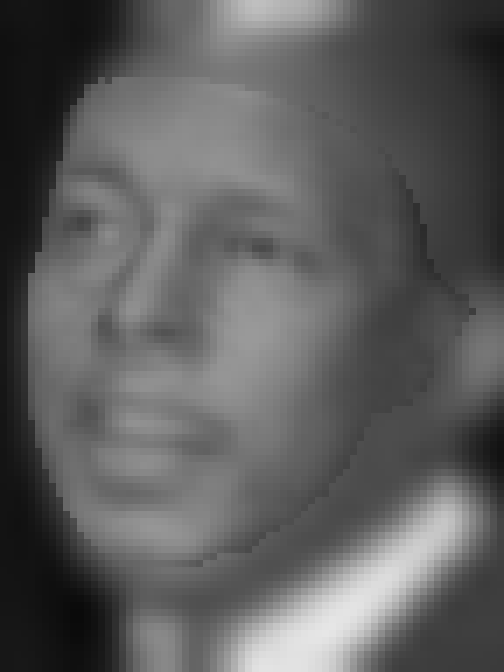}}\par\vspace{\vdiss cm}		
		\subfloat {\includegraphics[width=\imw cm]{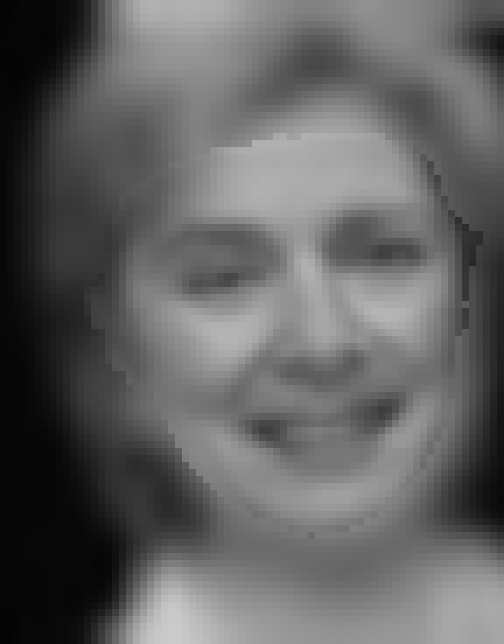}}\par\vspace{\vdiss cm}
		\subfloat {\includegraphics[width=\imw cm]{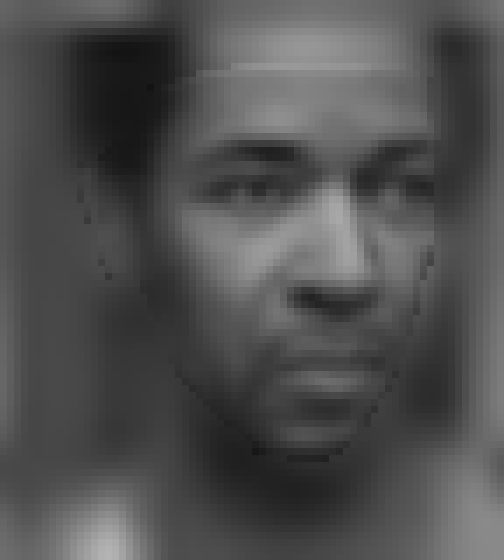}}\par\vspace{\vdiss cm}
		\subfloat {\includegraphics[width=\imw cm]{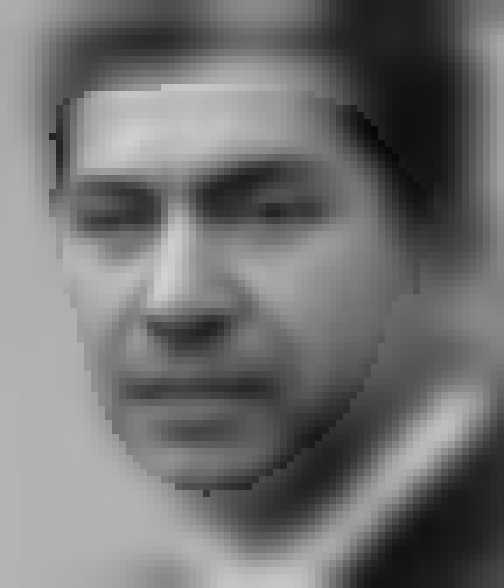}}\par\vspace{\vdiss cm}
		\stepcounter{figure}\addtocounter{figure}{-1}
		\addtocounter{subfigure}{8}
		\subfloat[]{\includegraphics[width=\imw cm]{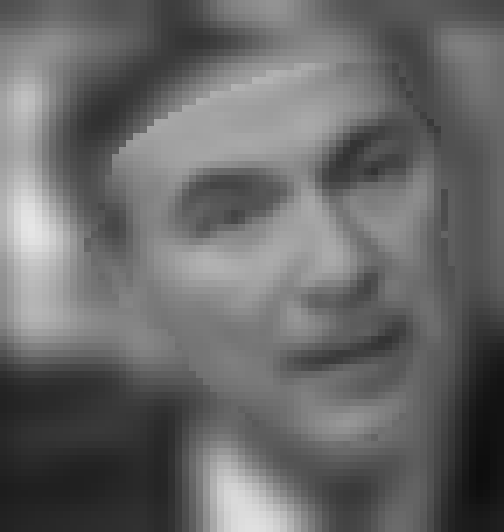}}
	\end{minipage}%
	\begin{minipage}{\hdis\textwidth} 
		\centering
		\subfloat {\includegraphics[width=\imw cm]{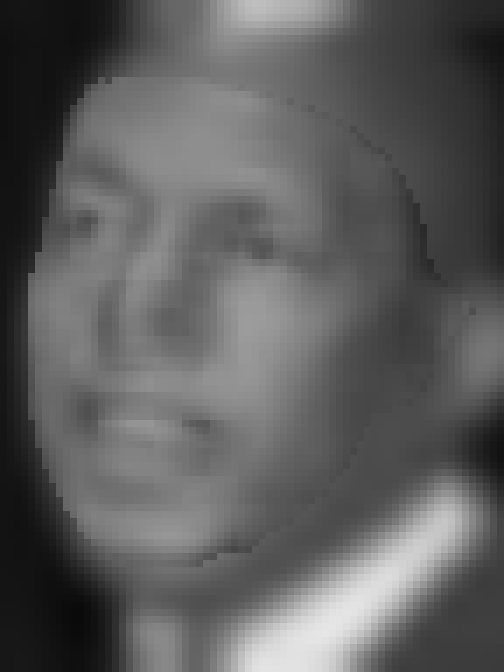}}\par\vspace{\vdiss cm}		
		\subfloat {\includegraphics[width=\imw cm]{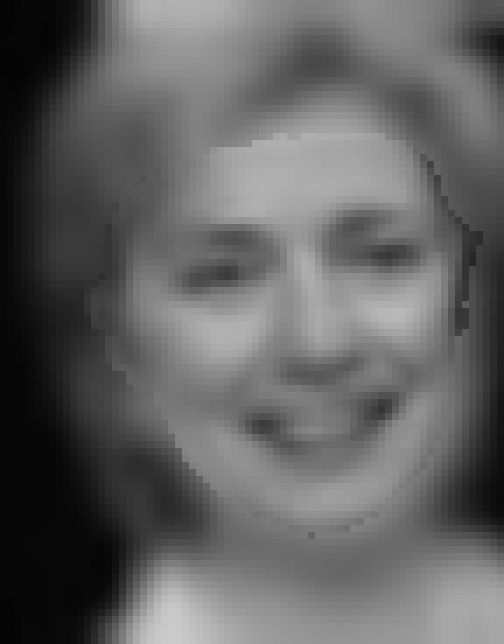}}\par\vspace{\vdiss cm}
		\subfloat {\includegraphics[width=\imw cm]{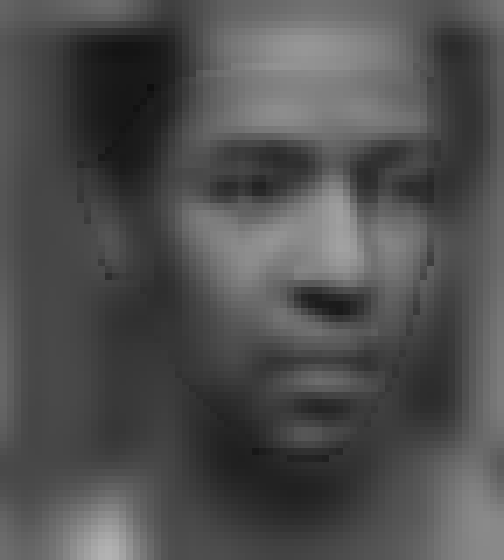}}\par\vspace{\vdiss cm}
		\subfloat {\includegraphics[width=\imw cm]{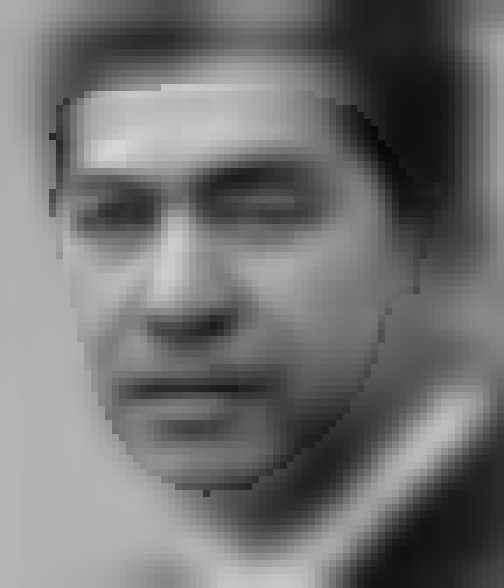}}\par\vspace{\vdiss cm}
		\stepcounter{figure}\addtocounter{figure}{-1}
		\addtocounter{subfigure}{9}
		\subfloat[]{\includegraphics[width=\imw cm]{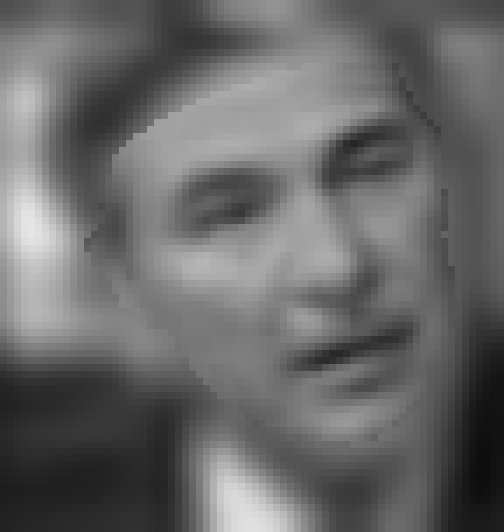}}
	\end{minipage}%
	\begin{minipage}{\hdis\textwidth} 
		\centering
		\subfloat {\includegraphics[width=\imw cm]{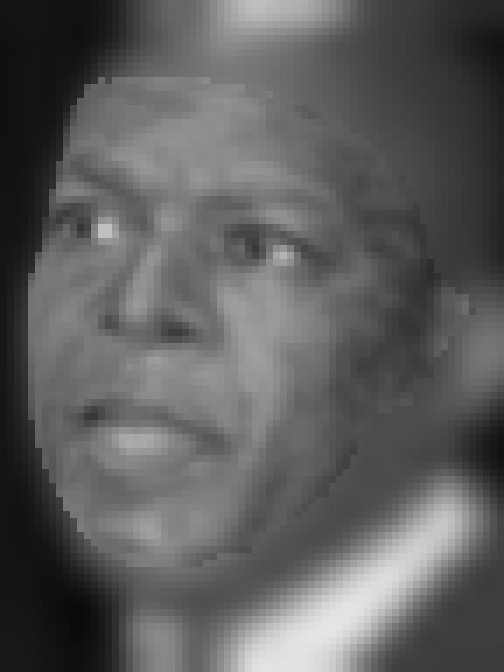}}\par\vspace{\vdiss cm}		
		\subfloat {\includegraphics[width=\imw cm]{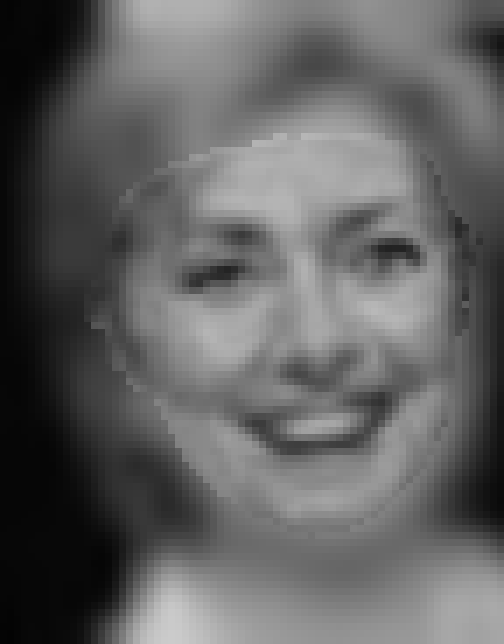}}\par\vspace{\vdiss cm}
		\subfloat {\includegraphics[width=\imw cm]{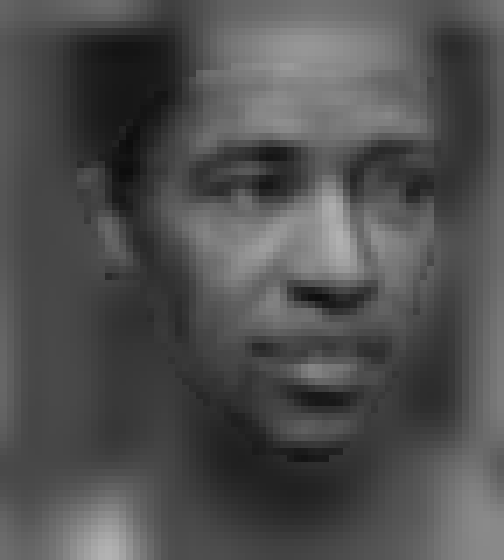}}\par\vspace{\vdiss cm}
		\subfloat {\includegraphics[width=\imw cm]{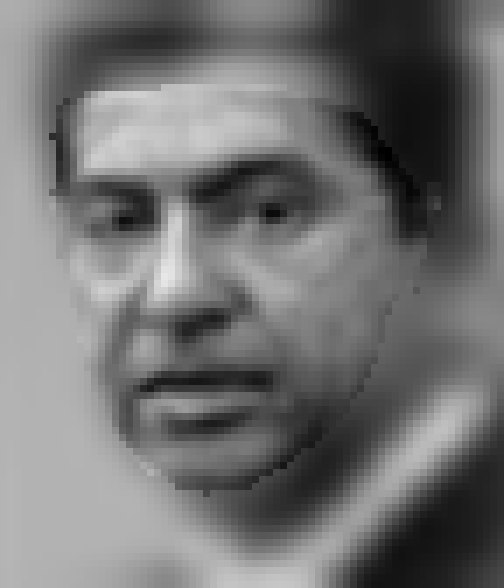}}\par\vspace{\vdiss cm}
		\stepcounter{figure}\addtocounter{figure}{-1}
		\addtocounter{subfigure}{10}
		\subfloat[]{\includegraphics[width=\imw cm]{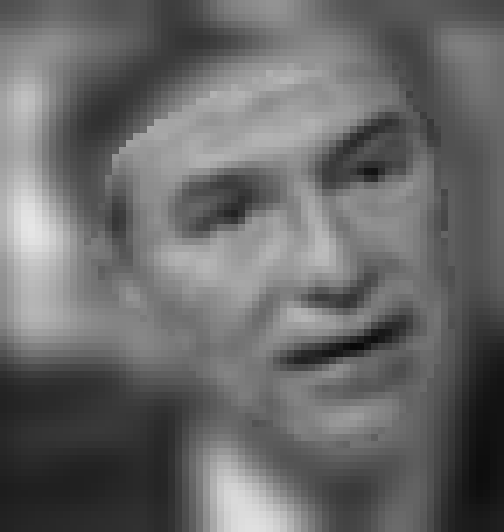}}
	\end{minipage}%
	\begin{minipage}{\hdis\textwidth} 
		\centering
		\subfloat {\includegraphics[width=\imw cm]{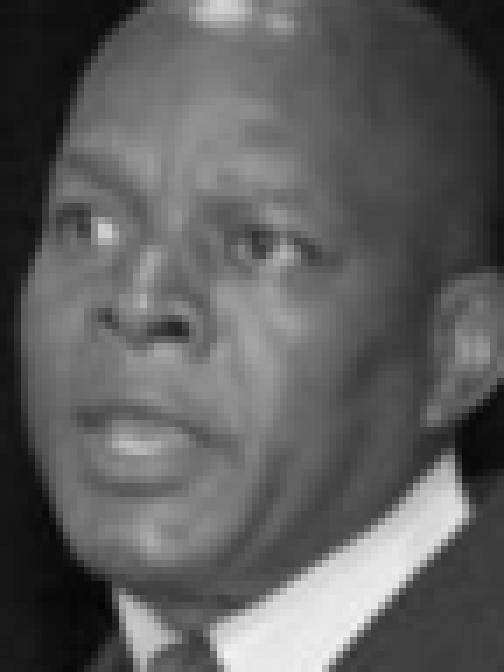}}\par\vspace{\vdiss cm}		
		\subfloat {\includegraphics[width=\imw cm]{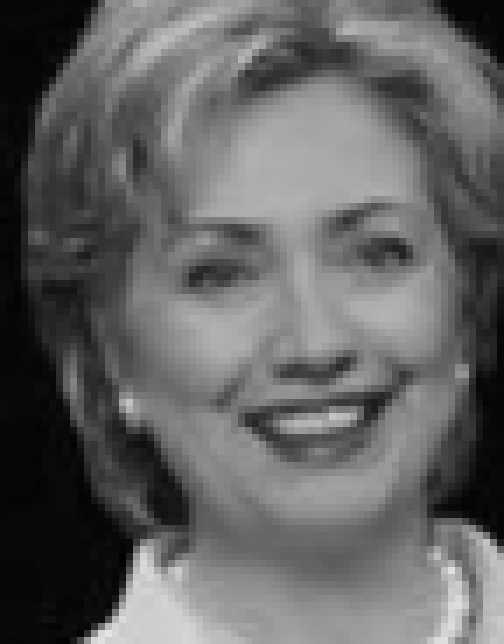}}\par\vspace{\vdiss cm}
		\subfloat {\includegraphics[width=\imw cm]{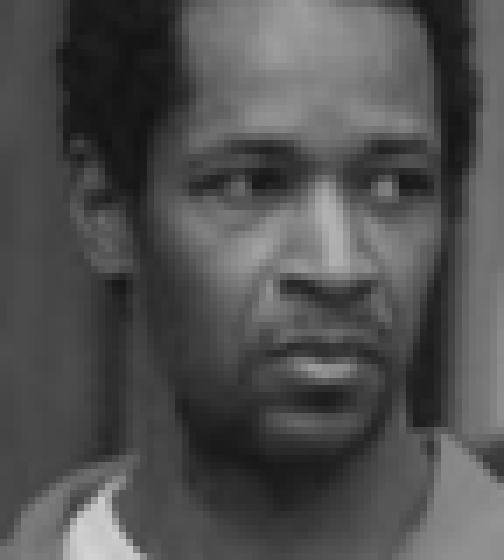}}\par\vspace{\vdiss cm}
		\subfloat {\includegraphics[width=\imw cm]{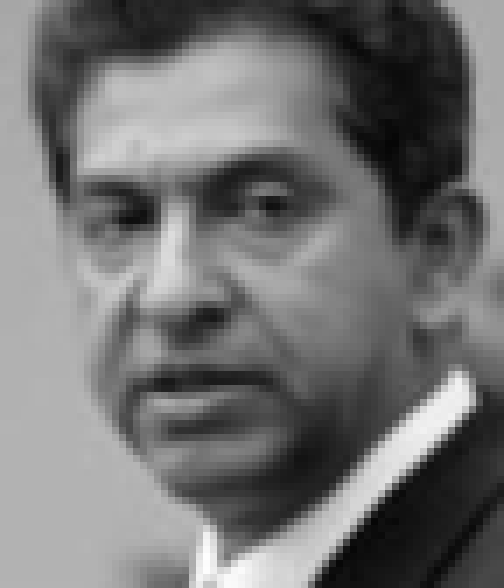}}\par\vspace{\vdiss cm}
		\stepcounter{figure}\addtocounter{figure}{-1}
		\addtocounter{subfigure}{11}
		\subfloat[]{\includegraphics[width=\imw cm]{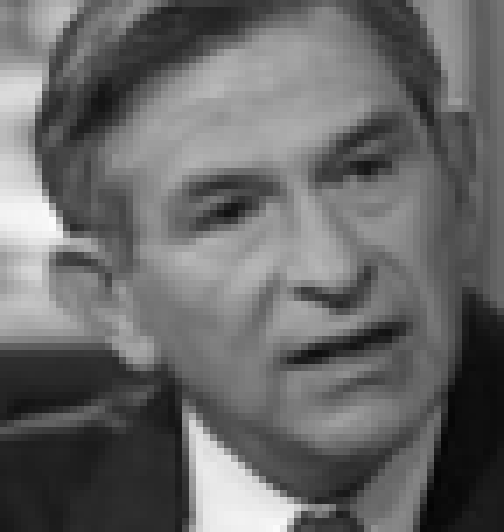}}
	\end{minipage}\par\medskip
	\caption{Evaluation of the performance achieved by the competitive methods and our algorithm in the problem of face hallucination in the wild using the proposed 3D dictionary alignment scheme on the $20\times20$ test samples of the LFW face dataset with scaling factor 4. (a) LR input. (b) Bicubic interpolation. (c) Wang \cite{refs:Wang2005}. (d) LSR \cite{refs:Ma2010}. (e) LcR \cite{refs:Jiang2014a}. (f) LINE \cite{refs:Jiang2014b}. (g) SSR \cite{refs:Jiang2017}. (h) LM-CSS \cite{refs:Farrugia2017}. (i) TRNR \cite{refs:Jiang2016}. (j) TLcR-RL \cite{refs:Jiang2018}. (k) Proposed. (l) Ground truth.}
	\label{fig:lfwres}
\end{figure*}

\section{Conclusion}
\label{sec:conclusion}
This paper presented a fast and robust face super-resolution algorithm which, different from most of the existing methods, attempts to recover true individual-specific attributes of the low-resolution input face. This is achieved by introducing a MAP estimator which encourages the reconstructed image to lie near to the subspace spanned by the training samples of the subject to which the input face belongs. Our experiments indicated that not more than three samples per subject are required to form such a subspace and obtain clear, detailed, and artifact-free outputs, even in datasets with high variations in facial expressions. We further extended our method by proposing a 3D dictionary alignment framework which proved to be vastly effective in real-world scenarios. Our evaluations revealed that, owing to its efficient closed-form based optimization procedure, the proposed algorithm is a very fast method which shows impressive robustness against face degradations. The identity-preserving aspect of our algorithm also demonstrated its significance in the task of low-resolution face recognition, even when there were only two samples of each subject available. The comparison results of the proposed method with some competitive algorithms, including both position-patch and deep learning-based methods, illustrated its superiority over the state-of-the-art face hallucination techniques in terms of both quantitative measurements and visual impressions. One possible extension of our proposed algorithm is to employ more robust sparse representation-based regularization terms (e.g., \cite{refs:Jiang2015} and \cite{refs:Xu2017}) to enhance its flexibility and robustness. Our future work will also focus on introducing the blurring operator to the main objective function as a variable and designing a blind face hallucination scheme.

\bibliographystyle{IEEEtran}
\bibliography{refs}

\vskip -25pt plus -1fil
\begin{IEEEbiography}{Ali Abbasi}
	received the B.S. degree in Computer Engineering from Ferdowsi University of Mashhad, Iran, in 2015, and the M.S. degree in artificial intelligence from the Computer Engineering Department, Amirkabir University of Technology (Tehran Polytechnic), Iran, in 2018. He is currently a research assistant at the Pattern Recognition and Image Processing lab at AUT. His main research interests are in the fields of image processing, computer vision, and pattern recognition.
\end{IEEEbiography}
\vskip -25pt plus -1fil
\begin{IEEEbiography}{Mohammad Rahmati}
	received the M.S. degree in electrical engineering from the University of New Orleans, USA in 1987 and the Ph.D. degree in electrical and computer engineering from the University of Kentucky, Lexington, KY USA in 1994. He is currently a Professor with the Computer Engineering Department, Amirkabir University of Technology (Tehran Polytechnic), and a member of IEEE Signal Processing Society. His current research interests include pattern recognition, image processing, machine learning, deep learning, and data mining.
\end{IEEEbiography}

\end{document}